%% file: thesis.tex
\documentclass[10pt,a4paper,openright]{memoir}

\usepackage[utf8]{inputenc}
\usepackage[T1]{fontenc}
\usepackage{textcomp}
\usepackage{lmodern}
\usepackage[french,english]{babel}
\usepackage{xspace}
\usepackage{url}
\usepackage[us]{datetime}
\usepackage{nth}
\usepackage{amsmath}
\usepackage{amssymb}
\usepackage{algorithm}
\usepackage[noend]{algpseudocode}
\usepackage{titletoc}
\usepackage{tikz}
\usepackage{calc}
\usepackage{pgfplots}
\usepackage{sansmath}
\usepackage{booktabs}
\usepackage{threeparttable}
\usepackage{standalone}
\usepackage{pdfpages}
\usepackage{afterpage}
\usepackage{graphicx}
\usepackage[justification=centering, caption=false]{subfig}
\usepackage{layouts}
\usepackage[
  unicode,
  pdftitle={Exploiting Modern Hardware for High-Dimensional Nearest Neighbor Search},
  pdfauthor={Fabien André}
]{hyperref}
\usepackage{hologo}

\usepackage{pq} 
\usetikzlibrary{positioning}
\usetikzlibrary{calc}
\usetikzlibrary{shapes}
\usetikzlibrary{matrix}
\usetikzlibrary{fit}
\usetikzlibrary{arrows}

\settypeblocksize{\textheight}{1.08\textwidth}{*}
\setlrmargins{\spinemargin}{*}{*}
\checkandfixthelayout

\copypagestyle{myruled}{ruled}
\makeevenhead{myruled}{\leftmark}{}{}
\makeoddhead{myruled}{}{}{\rightmark}
\makechapterstyle{mysection}{%

}
\chapterstyle{mysection}

\setsecnumdepth{subsection}

\maxtocdepth{subsection}

\newcommand\partialtocname{Contents}
\newcommand\ToCrule{\noindent\rule[5pt]{\textwidth}{1.3pt}}
\newcommand\ToCtitle{{\large\bfseries\partialtocname}\vskip2pt\ToCrule}
\makeatletter
\newcommand\ruledprintcontents{%
  \ToCtitle
  \ttl@printlist[chapters]{toc}{}{1}{}\par\nobreak
  \ToCrule}
\makeatother

\DeclareMathOperator*{\argmin}{arg\,min}

\newcommand*{\stimes}{{\times}}
\newcommand*{\pq}[2]{PQ\,${#1}\stimes{#2}$}

\pgfplotsset{compat=1.12}

\pgfplotsset{recall plot/.style={
  xtick={1,2,5,10,20,50,100,200,500,1000},
  xticklabels={1,2,5,10,20,50,100,200,500,1K},
  xlabel=$r$,
  ylabel=Recall@$r$,
  legend pos=south east,
  legend cell align=left,
}}

\pgfplotsset{time plot/.style={
  yticklabel=\empty,
  legend pos=south east,
  legend cell align=left,
  xlabel={Total query time (ms)},
  every node near coord/.append style={font=\small},
  nodes near coords align={horizontal},
  every node near coord/.append style={xshift=0.05em},
  nodes near coords={
        \pgfmathprintnumber[fixed relative, precision=2]{\pgfplotspointmeta}
  }
}}
\pgfplotsset{time bar/.style={
  xbar, point meta=explicit,
  error bars/.cd,
  x dir=both,
  x explicit
}}

\pgfplotsset{r plot/.style={
  xtick={10,20,50,100,200,500,1000},
  xticklabels={10,20,50,100,200,500,1K},
  xlabel=$r$,
  height=5cm,
  width=6.3cm,
  ticklabel style={font=\small}
}}

\input{derivedquant/fig/cmaps}

\input{simd}

\newcommand*{\eg}{e.g.,\xspace}
\newcommand*{\ie}{i.e.,\xspace}
\newcommand*{\etc}{etc.\xspace}



\begin{document}





\cleardoublepage
\input{title.tex}

\input{titleback.tex}

\cleardoublepage
\chapter*{Acknowledgements}
\input{thanks.tex}

\cleardoublepage
\input{notice.tex}

\cleardoublepage
\tableofcontents

\clearpage
\listoffigures

\clearpage
\listoftables

\frontmatter
\input{resumefr/resumefr.tex}

\mainmatter

\chapter{Introduction}
\input{intro/introduction.tex}

\chapter{State of the Art}
\input{sota/sota.tex}

\chapter{Performance Analysis}
\input{perf/perf.tex}

\chapter{PQ Fast Scan}
\input{fastpq/fastpq.tex}

\chapter{Quick ADC}
\input{quickadc/quickadc.tex}

\chapter{Derived Quantizers}
\input{derivedquant/derivedquant.tex}

\chapter{Conclusion and Perspectives}
\input{conclusion/conclusion.tex}

\bibliographystyle{abbrv}
\bibliography{references}

\cleardoublepage
\includepdf[pages=5]{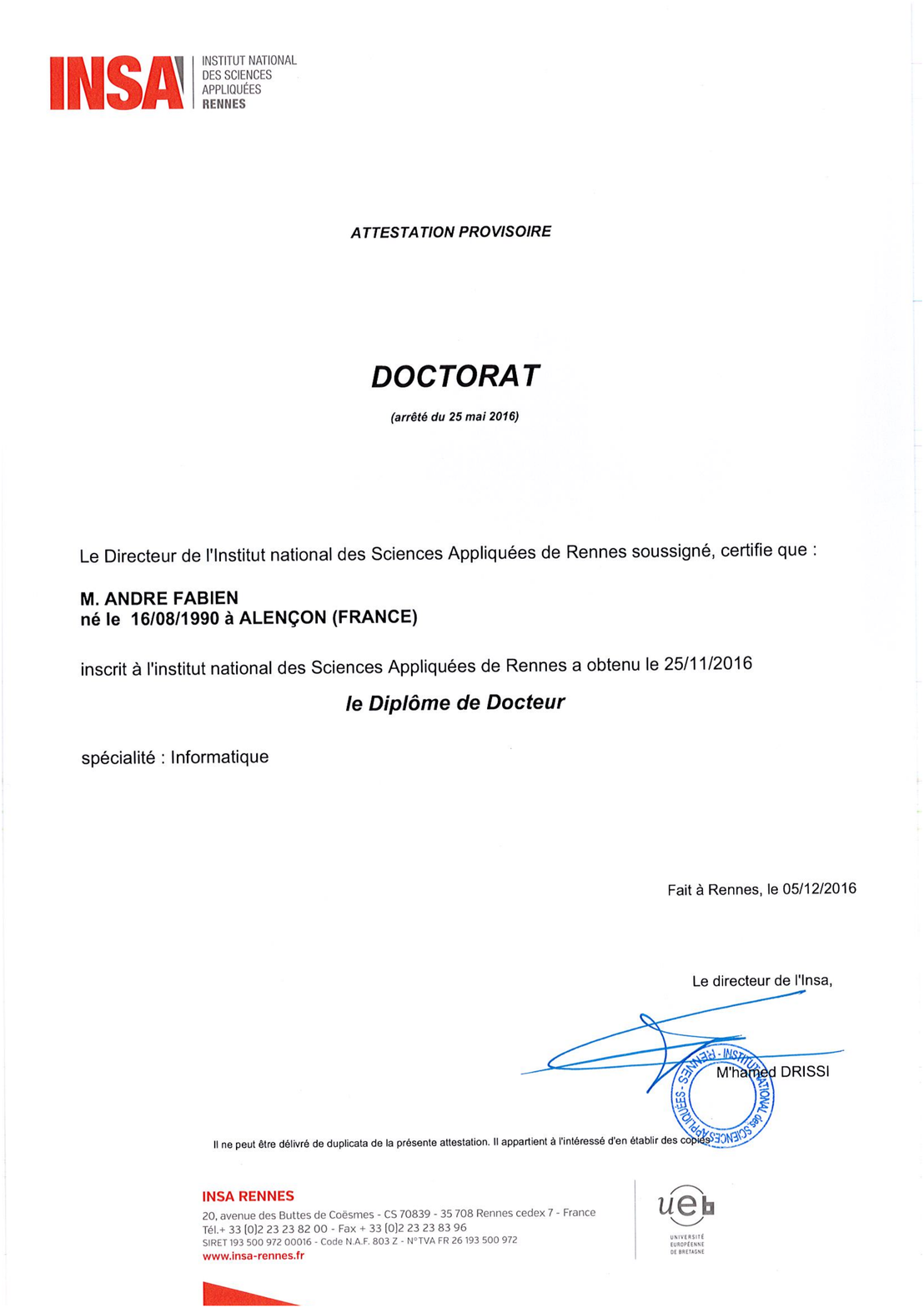}

\end{document}

%% file: derivedquant/fig/cmaps.tex
\pgfplotsset{
    /pgfplots/colormap={jet}{rgb255(0cm)=(0,0,128) rgb255(1cm)=(0,0,255)
    rgb255(3cm)=(0,255,255) rgb255(5cm)=(255,255,0) rgb255(7cm)=(255,0,0) rgb255(8cm)=(128,0,0)}
}

\pgfplotsset{
    colormap={bluered}{
        rgb255(0cm)=(0,0,180); rgb255(1cm)=(0,255,255); rgb255(2cm)=(100,255,0);
        rgb255(3cm)=(255,255,0); rgb255(4cm)=(255,0,0); rgb255(5cm)=(128,0,0)}
}

\pgfplotsset{
    colormap={cubehelix16}{
        rgb255(0cm)=(0,0,0);
        rgb255(1cm)=(22,10,34);
        rgb255(2cm)=(24,32,68);
        rgb255(3cm)=(16,62,83);
        rgb255(4cm)=(14,94,74);
        rgb255(5cm)=(35,116,51);
        rgb255(6cm)=(80,125,35);
        rgb255(7cm)=(138,122,45);
        rgb255(8cm)=(190,117,85);
        rgb255(9cm)=(218,121,145);
        rgb255(10cm)=(219,138,203);
        rgb255(11cm)=(204,167,240);
        rgb255(12cm)=(191,201,251);
        rgb255(13cm)=(195,229,244);
        rgb255(14cm)=(220,246,239);
        rgb255(15cm)=(255,255,255)}
}

\pgfplotsset{
    colormap={rainbow16}{
        rgb255(0cm)=(135,59,97);
        rgb255(1cm)=(143,64,127);
        rgb255(2cm)=(143,72,157);
        rgb255(3cm)=(135,85,185);
        rgb255(4cm)=(121,102,207);
        rgb255(5cm)=(103,123,220);
        rgb255(6cm)=(84,146,223);
        rgb255(7cm)=(69,170,215);
        rgb255(8cm)=(59,192,197);
        rgb255(9cm)=(60,210,172);
        rgb255(10cm)=(71,223,145);
        rgb255(11cm)=(93,229,120);
        rgb255(12cm)=(124,231,103);
        rgb255(13cm)=(161,227,95);
        rgb255(14cm)=(198,220,100);
        rgb255(15cm)=(233,213,117)
    }
}

\tikzset{adjust near node color/.code={%
    \pgfplotscolormapdefinemappedcolor\pgfplotspointmetatransformed%
    \definecolor{mapped node color}{rgb}{\pgfmathresult}%
    \pgfkeys{/tikz/text=mapped node color!80!black}%
    }
}

\pgfplotsset{voronoiaxis/.style={restrict x to domain*=-5:5,
    restrict y to domain*=-5:5,
    xmin=0,
    xmax=1,
    ymin=0,
    ymax=1,
    axis equal image,
    hide axis,
    width=7cm
    }}
    
\newcommand*{\nodeFont}{\sansmath\sffamily\scriptsize}

%% file: simd.tex
\newlength\eltWidth
\newlength\eltHeight
\def\horSep{0.5em}
\def\vertSep{\horSep}
\def\labelSep{2*\horSep}

\tikzstyle{elt} = [
  inner sep=0pt,
  outer sep=0pt,
  text width=\eltWidth,
  font=\small,
  align=center
]

\tikzstyle{register} = [
  rectangle split,
  minimum height=\eltHeight,
  rectangle split horizontal=true,
  rectangle split parts=6,
  rectangle split part align=base,
  draw
]

\tikzstyle{operator} = [
  chamfered rectangle,
  font=\small,
  text width=13em,
  text height=1.2ex,
  text depth=0.1ex,
  align=center,
  draw
]

\tikzstyle{label} = [
    font=\small,
    inner sep=0,
    outer sep=0
]

\newlength\memCellHeight
\newlength\memCellWidth
\tikzset{
  memory cell/.style={
    rectangle,
    draw,
    font=\small,
    text depth=0.25ex,
    text height=\memCellHeight,
    text width=\memCellWidth,
    align=center,
  },      
  memory/.style={
    matrix of math nodes,
    row sep=-\pgflinewidth,
    column sep=-\pgflinewidth,
    nodes={
        memory cell
    },
    execute at empty cell={\node[draw=none]{};}
  },
  memory multicell/.style={
    draw,
    inner ysep=\pgflinewidth/2,
    inner xsep=-\pgflinewidth/2,
    font=\small,
    align=center,
    text height=\memCellHeight,
    text depth=0.1ex,
  }
}

\tikzstyle{memory label} = [
    font=\small
]


%% file: title.tex
\begin{titlingpage}
\vspace*{2cm}
  \begin{flushright}
    \rule{0.85\textwidth}{1pt}

    {\noindent\LARGE\bfseries Exploiting Modern Hardware for\\[0.5\baselineskip] High-Dimensional Nearest Neighbor Search}

    \rule{0.85\textwidth}{1pt}
    \\[2\baselineskip]
    {\noindent \large Fabien André}
    \\[2\baselineskip]
    {\noindent \large PhD thesis defended on\\November 25, 2016}\\[0.4\baselineskip]
    {\noindent \large at INSA Rennes}
    \\[2\baselineskip]
    {\noindent \large \textbf{Thesis Advisors}\\[0.3\baselineskip]
    Anne-Marie Kermarrec\\
    {\normalsize Research Director, Inria}\\[0.4\baselineskip]
    Nicolas Le Scouarnec\\
    {\normalsize Senior Scientist, Technicolor}}
    \\[2\baselineskip]
    {\noindent \large \textbf{Thesis Committee}\\[0.3\baselineskip]
    Achour Mostefaoui\\
    {\normalsize President\\Professor, University of Nantes}\\[0.4\baselineskip]
    Gaël Thomas\\
    {\normalsize Referee\\Professor, Telecom SudParis}\\[0.4\baselineskip]
    Peter Triantafillou\\
    {\normalsize Referee\\Professor, University of Glasgow}}
  \end{flushright}
  \vfill
\end{titlingpage}

%% file: titleback.tex
\thispagestyle{empty}

\null\vspace{20ex}
\begin{center}
This work is distributed under a Creative Commons Attribution 4.0 International License (CC BY 4.0). For more information, see \url{https://creativecommons.org/licenses/by/4.0/}.\\[3ex]
\includegraphics{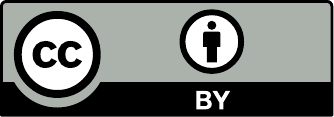}\\[3ex]
Sources are available at \url{https://github.com/Xion345/phd-thesis}.
\end{center}
\vspace{15ex}
\begin{center}
Some of the techniques presented in this thesis may be covered by patents owned by Technicolor R\&D France, Thomson Licensing and/or other parties. No rights to any party's patents are implied by the distribution of this thesis.
\end{center}

\vfill
\begin{center}
This document was compiled on \today{} by \hologo{pdfTeX} version \the\pdftexversion\pdftexrevision{} using \hologo{\fmtname} version \fmtversion{} and \textsc{PGF} version \pgfversion{}.
\end{center}

%% file: thanks.tex

%

\thispagestyle{empty}

\paragraph{}
First and foremost, I would like to express my deepest gratitude to my supervisors, Anne-Marie Kermarrec and Nicolas Le Scouarnec, for their outstanding guidance. I would like to particularly thank them for their assistance through the process of defining my thesis topic, their original ideas, their technical help, and their precious advice on scientific writing. I have a feeling that I have discovered a new world during this thesis -- the one of research -- and I owe this discovery to Nicolas and Anne-Marie. On a more personal level, I would also like to thank them for their human qualities. The completion of a PhD thesis inevitably involves moments of doubt and hesitation and I was fortunate to be able to could count on their support during these moments.

\paragraph{}
I also thank all other members of the jury, Achour Mostefaoui, Peter Triantafillou and Gaël Thomas. I am especially grateful to Peter Triantafillou and Gaël Thomas for reviewing the manuscript and providing feedback.

\paragraph{}
This thesis would not have been possible without the support of Technicolor which gave me the opportunity to fully devote myself to my research activities during these three years. I also thank Inria for providing access to computing resources without which many experiments of this thesis could not have been done. More specifically, I was fortunate to have access to Grid'5000, a highly flexible large-scale platform which enables complex experiments. I cannot stress enough how important these platforms are for the research in computer science, and especially for systems research.

\paragraph{}
I also have a thought for my colleagues and fellow PhD students at Inria and Technicolor without whom these three years would not have been the same. Thank you for creating an intellectually stimulating environment, thank you for your friendship, and thank you for the crazy discussions at the coffee break -- I will not forget them. I owe a special thank you to Clémentine, my fellow PhD student at Technicolor, for sharing with me the difficulties of doing a PhD.


\paragraph{}
Lastly, these acknowledgements would not be complete without a word of thanks for my family who gave me much-needed emotional support over these three years. They offered a sympathetic ear when I needed, and I am convinced I could not have come this far without their continuous support.

%% file: notice.tex

\thispagestyle{empty}

\null\vfill

\hfill\parbox{0.75\textwidth}{
\paragraph{}
La présente thèse comporte un résumé de onze pages rédigé en français («~Résumé en Français~») dont les pages sont numérotées en chiffres romains. Le reste de la thèse est rédigé en anglais, et les pages sont numérotées en chiffres arabes. Le résumé en Français ne contient aucune information qui n'est pas également présente dans le reste de la thèse.
}

\vspace{10ex}

\hfill\parbox{0.75\textwidth}{
\paragraph{}
The present thesis comprises a summary of eleven pages written in French («~Résumé en Français~») the pages of which are numbered with Roman numerals. The remainder of the thesis is written in English, and the pages are numbered with Arabic numerals. The summary in French contains no information that is not also present in the remainder of the thesis.
}

\vspace{40ex}

%% file: resumefr/resumefr.tex
\begin{otherlanguage}{french}
\chapter{Résumé en Français}

\section{Introduction}

\paragraph{}
Cette thèse s'intéresse à la recherche de plus proche voisin en haute dimensionnalité et à large échelle. Ce problème revêt une importance particulière dans le contexte actuel d'utilisation massive de réseaux sociaux en ligne. L'utilisation de ces réseaux sociaux par des millions d'utilisateurs a permis l'accumulation de jeux de données très volumineux, une tendance connue sous le nom de \emph{big data}. Ces jeux de données comprennent non seulement des données textuelles (messages, discussions) mais aussi multimédia (images, sons, vidéos). 

\paragraph{}
La recherche de plus proche voisin dans les espaces de haute dimensionnalité est un problème clé dans le domaine des bases de données multimédia. Les objets multimédia (images, sons, vidéos) peuvent être représentés par des vecteurs caractéristiques en haute dimensionnalité. Rechercher deux objets multimédias similaires revient alors à rechercher deux objets multimédia ayant des vecteurs caractéristiques similaires. Cependant, la recherche de plus proche voisin est un problème difficile, en particulier à large échelle. En haute dimensionnalité, la tristement célèbre « malédiction de la dimensionnalité » rend impossible la recherche de solutions exactes. Les travaux de recherche actuels s'attachent donc à trouver des solutions approchées à la recherche de plus proche voisin.

\paragraph{}
La product quantization est une des solutions approchées de recherche de plus proche voisin les plus efficaces. Son principal avantage est qu'elle permet de stocker des bases de données volumineuses entièrement en RAM, grâce à une technique de compression des vecteurs de haute dimensionnalité en codes compacts. Ceci permet alors de répondre à des requêtes de recherche de plus proche voisin sans accéder à un disque dur ou à un SSD. Ainsi, la product quantization offre des temps de réponse faibles. Dans cette thèse, nous proposons différentes solutions pour réduire davantage les temps de réponse offerts par la product quantization en exploitant les fonctionnalités des processeurs modernes.

\section{Quantification Produit}

\paragraph{Quantifieur vectoriel} Pour compresser les vecteurs de haute dimensionnalité en codes compacts, la quantification produit s'appuie sur des quantifieurs vectoriels. Un quantifieur vectoriel est une fonction $q$ qui associe un vecteur $x \in \mathbb{R}^d$ a un vecteur $c_i \in \mathbb{R}^d$ appartenant à un ensemble prédéfini de vecteurs $\mathcal{C}=\{c_0,\dotsc,c_{k-1}\}$. Les vecteurs $c_i$ sont appelés centroides, l'ensemble de centroides $\mathcal{C}$ est le dictionnaire. Un quantifieur optimal associe un vecteur $x$ à son centroide le plus proche : $\operatorname{q}(x) = \argmin_{c_i \in \mathcal{C}}{||x-c_i||}$.

\paragraph{Quantification produit} La quantification produit, ou \emph{Product Quantization} (PQ), divise un vecteur $x \in \mathbb{R}^d$ en $m$ sous-vecteurs $x=(x^0,\dotsc,x^{m-1})$. Chaque sous-vecteur $x^j$ est ensuite quantifié en utilisant un quantifieur $\operatorname{q^j}$ distinct. Chaque quantifieur $\operatorname{q^j}$ dispose d'un dictionnaire $\mathcal{C}^j$ distinct. Un quantifieur produit $\operatorname{pq}$ quantifie un vecteur $x$ comme suit :
\begin{align*}
\operatorname{pq}(x) &= \left ( \operatorname{q}^0(x^0),\dots,\operatorname{q}^{m-1}(x^{m-1}) \right )\\
&= (\mathcal{C}^0[i_0],\dotsc,\mathcal{C}^{m-1}[i_{m-1}]).
\end{align*}
Un quantifieur produit peut être utilisé pour encoder un vecteur $x$ en un code compact défini par la concaténation des indices des $m$ centroides retournés par les $m$ quantifieurs $\operatorname{q^j}$ : $\operatorname{code}(x)=(i_0,\dotsc,i_{m-1})$. Un code compact $\operatorname{code}(x)$ utilise $m{\cdot}b$ bits de mémoire, avec $b=\log_2(k)$ où $k$ est le nombre de centroides de chaque quantifieur $\operatorname{q^j}$. Un code compact $\operatorname{code}(x)$ utilise beaucoup moins de mémoire qu'un vecteur de haute dimensionnalité $x$. Ainsi, un vecteur en 128 dimensions peut être représenté par un code de 64 bits ($m=8, b=8$, $m{\stimes}b=64$), alors qu'il occupe $d\cdot 32 = 4096$ bits stocké sous forme d'un tableau de flottants. 

\paragraph{Recherche de plus proche voisin} Lorsque la quantification produit est utilisée, les vecteurs de haute-dimensionalité de la base de données sont compressés en codes compacts. Le codes compacts sont ensuite stockés dans un tableau contigu en RAM. Pour trouver le plus proche voisin d'un vecteur requête $y$, il est nécessaire de calculer la distance entre le vecteur $y$ et chacun des codes stockés en RAM. Pour ce faire, la quantification produit s'appuie sur une procédure nommée Asymmetric Distance Computation (ADC), ou calcul de distance asymmétrique, qui permet de calculer la distance entre un vecteur de haute dimensionnalité $y$ et n'importe quel code $c$ de la base de données. Plus précisément, la recherche de plus proche voisin se déroule en deux étapes, nommées Tables et Scan: 
\begin{itemize}
\item{\emph{Tables.}} Le vecteur $y$ est divisé en $m$ sous-vecteurs $y=(y^0,\dotsc,y^{m-1})$. On construit ensuite $m$ tables de correspondance $\{D_0,\dotsc,D_{m-1}\}$. Chaque table de correspondance $D_j$ contient la distance entre le sous-vecteur $y^j$ et chaque centroide du dictionnaire $\mathcal{C}^j$
\item{\emph{Scan.}} Ensuite, on calcule la distance entre le vecteur $y$ et chaque code $c$ de la base de données grace à la procédure ADC, comme suit:
\begin{equation}
\operatorname{adc}(y,c) = \sum_{j=0}^{m-1} D^{j}[c[j]]
\end{equation}
\end{itemize}
Cette thèse s'attache a améliorer les performances de la procédure ADC, et en particulier de l'étape Scan, car celle-ci consomme la majorité des cycles CPU.

\paragraph{Index inversés} La quantification produit est généralement associée à des systèmes d'index inversés. Ces systèmes divisent la base de données en un ensemble de listes inversées. Pour répondre à une requête, les listes inversées les plus appropriées sont sélectionnées. Ces listes sont ensuite scannées en utilisant la procédure ADC. Les index inversés permettent de réduire les temps de réponse en évitant de scanner la totalité de la base de données. Ils permettent également d'augmenter la précision de la recherche de plus proche voisin, grâce à un système connu sous le nom d'encodage de résidus.

\paragraph{Quantification produit optimisée} La quantification produit divise un vecteur $x$ en sous-vecteurs $x^j$ et encode chaque sous-vecteur en un indice de $b$ bits, indépendamment de la quantité d'information contenue dans chaque sous-vecteur. Ce procédé donne des résultats sous-optimaux lorsque les sous-vecteurs ne contiennent pas la même quantité d'information ou que certaines dimensions sont corrélées avec d'autres. La quantification produit optimisée, ou \emph{Optimized Product Quantization} (OPQ), remédie a ce problème en multipliant les vecteurs $x$ par une matrice orthonormale $R \in \mathbb{R}^{d\times d}$. Cette matrice permet un rotation et une permutation arbitraire des composantes du vecteur $x$, et est apprise de façon à minimiser l'erreur de quantification. Un quantifieur produit optimisé $\operatorname{opq}$ quantifie un vecteur $x$ comme suit :
\[
\operatorname{opq}(x) = \operatorname{pq}(Rx), \text{ tel que } R^TR=I,
\]
où $\operatorname{pq}$ est quantifieur produit. Un quantifieur produit optimisé peut encoder des vecteurs en codes compacts de la même manière qu'un quantifieur produit. 


\section{Analyse de Performance}

%
\paragraph{Analyse des opérations} Afin d'améliorer la performance de la procédure ADC, et en particulier de l'étape Scan, on commence par identifier les goulots d'étranglement qui la limitent. Lors de l'étape Scan, chaque calcul de distance asymétrique nécessite:
\begin{itemize}
\item $m$ accès mémoire pour charger les indices des centroides $c[j]$ (mem1)
\item $m$ accès mémoire pour charger les valeurs $D_j[.]$ depuis les tables de correspondance (mem2)
\item $m$ additions ($\sum_{j=0}^{m-1}$)
\end{itemize}
Parmi ces opérations, les accès mémoires sont les plus coûteuses. En effet, un accès mémoire prend entre 4 et 40 cycles CPU en fonction du niveau de cache atteint, alors qu'une addition ne prend qu'un seul cycle. Ainsi, la performance de PQ Fast Scan est principalement limitée par le grand nombre d'accès cache qu'elle effectue.

\begin{table}
  \centering
  \caption{Temps de réponse et précision (codes 64 bits, SIFT1M)\label{tbl:fr-resume-taille}}
  \begin{tabular}{llllll}
    \toprule
    $m \stimes b$ & Tables size & Cache & Recall@100 & Tables time & Scan time\\
    \midrule
    $16 \stimes 4$ & 1 KiB & L1 & 83.1\% & 0.001 ms & 6 ms\\
    $8 \stimes 8$  & 8 KiB & L1 & 92.2\% & 0.011 ms & 2.6 ms\\
    $4 \stimes 16$ & 1 MiB & L3 & 96.5\% & 0.82 ms & 7.9 ms\\
    \bottomrule
  \end{tabular}
\end{table}

\paragraph{Accès mémoire}
Les accès mémoire mem1 atteignent toujours le cache le plus rapide (cache L1), grâce aux \emph{hardware prefetchers} inclus dans les processeurs modernes. On accède aux codes $c$ \emph{séquentiellement}: premier code, puis second code \etc ce qui permet aux \emph{hardware prefetchers} de précharger les codes $c$ dans le cache L1. En revanche, le niveau de cache atteint par les accès mem2 dépend du paramètre $b$ du quantifieur produit utilisé. Un quantifieur produit est complètement défini par deux paramètres $m$, le nombre de quantifieurs, et $b$ le nombre de bits par quantifieur (qui détermine le nombre $k=2^b$ de centroides par quantifier).

\paragraph{Compromis précision-vitesse}
La précision d'un quantifieur produit dépend principalement du produit $m{\cdot}b$. Ce produit définit l'occupation mémoire de chaque code, et donc l'occupation mémoire de la base de données. Ainsi un quantifieur produisant des codes de 128 bits ($m{\cdot}b=128$) sera plus précis qu'un quantifieur produisant des codes de 64 bits ($m{\cdot}b=64$). Le quantifieur produisant des codes de 128 bits engendrera cependant un doublement de la taille de la base de données, puisque chaque code occupera 128 bits au lieu de 64 bits. Pour limiter l'occupation mémoire, le produit $m{\stimes}b$ est souvent fixé à 64. Pour un produit $m{\cdot}b$ fixé (par exemple $m{\cdot}b=64$), les paramètres $m$ et $b$ peuvent varier. Plus $b$ est grand (et donc plus $m$ est petit), meilleure est la précision du quantifieur, et donc meilleure est la précision de la recherche de plus proche voisin. Ainsi, un quantifieur produit $4{\stimes}16$ ($m=4, b=16$) et un quantifieur produit $8{\stimes}8$ ($m=8, b=8$) produisent tous deux des codes de 64 bits, mais le quantifieur $4{\stimes}16$ offre une meilleure précision. Pour mesurer la précision des techniques de recherche de plus proche voisin approchées, on utilise le \emph{Recall@R}. Le Recall@R est défini comme la proportion de requêtes pour lesquelles le plus proche voisin \emph{exact} du vecteur requête se trouve parmi les R plus proches voisins renvoyés par la technique de recherche approchée. En général, on paramètre les techniques de recherche approchées pour qu'elles retournent R=100 voisins, donc on utilise le Recall@100 comme mesure de précision (Table~\ref{tbl:fr-resume-taille}). On constate alors qu'un quantifieur produit $4{\stimes}16$ ($b=16$) offre un meilleur Recall@100 qu'un quantifieur $8{\stimes}8$ ($b=8$). Cependant, plus $b$ est grand, plus les tables de correspondance $D_j$ sont volumineuses, et doivent alors être stockés dans des niveaux de cache plus lents. Pour $b=16$, les tables doivent être stockées dans le cache L3 (latence de 40 cycles), au lieu du cache L1 (latence de 4-5 cycles). Ceci se traduit par un triplement du temps de réponse (Table~\ref{tbl:fr-resume-taille}). Pour $b=8$ et $b=4$, les tables sont stockées dans le cache L1. Le paramètre $b=8$ offre alors un meilleur de temps de réponse car il nécessite moins d'opérations ($m=8$ contre $m=16$). Le paramétrage $m{\stimes}b=8{\stimes}8$ offre un bon compromis entre précision et temps de réponse, raison pour laquelle il est utilisé dans la quasi-totalité de la littérature sur la quantification produit. Cependant, même lorsqu'ils atteignent le cache L1 (niveau de cache le plus rapide), les accès mémoire restent un facteur limitant la performance de la procédure ADC.

\paragraph{SIMD} Les instructions Single Instruction Multiple Data (SIMD) sont couramment utilisées pour améliorer la performance des algorithmes, en particulier lorsqu'ils réalisent un grand nombre d'opérations arithmétiques. Les instructions SIMD appliquent la même opération (par exemple une addition) à plusieurs données simultanément, pour produire plusieurs résultats simultanément. Pour se faire, les instructions SIMD opèrent sur des registres larges, en général 128 bits. Ces registres sont divisés en plusieurs \emph{voies}, par exemple 4 voies flottantes ($4$ voies de $32$ bits, soit 128 bits). Les instructions SIMD sont donc susceptibles d'être utilisées pour réduire le nombre de cycles CPUs dédiés aux $m$ additions nécessaires pour chaque calcul de distance. Cependant, la structure de la procédure ADC empêche une utilisation efficace des instructions SIMD. En effet, les valeurs $D_j[c[j]]$ chargées depuis les tables de correspondance $D_j$ ne sont pas contiguës en mémoire, ce qui rend impossible le chargement d'un registre SIMD de 128 bits en un seul accès mémoire. Au contraire, les valeurs doivent être insérées une à une dans chacune des voies des registres SIMD, ce qui nécessite un grand nombre d'instructions. Au final, ceci annule le gain offert par l'utilisation d'additions SIMD.

\paragraph{}
En définitive, nous pouvons tirer trois conclusions de cette analyse de performance~:
\begin{itemize}
\item Les accès mémoire sont le principal facteur limitant la performance de la procédure ADC, bien qu'ils atteignent le cache L1 (niveau de cache le plus rapide) dans la majorité des cas.
\item Les quantifieurs produit utilisant des quantifieurs 16 bits ($b=16$) offrent une meilleure précision que les quantifieurs produit utilisant des quantifieurs 8 bits ($b=8$) mais ils engendrent un triplement du temps de réponse lors de la recherche de plus proche voisin. Ceci est du au fait que les quantifieurs 16 bits nécessitent de stocker les tables de correspondance utilisées par la procédure ADC dans le cache L3, alors que ces tables peuvent être stockées dans le cache L1 lorsque des quantifieurs 8 bits sont utilisés.
\item La structure de la procédure ADC empêche l'utilisation efficace des instructions SIMD. En effet, la procédure ADC récupère des valeurs dans le cache avant des les insérer une à une dans les registres SIMD. Ces opérations sont coûteuses et annulent le gain offert par les instructions SIMD.
\end{itemize}

\section{PQ Fast Scan}

\paragraph{} Grâce au stockage des bases de données en RAM, la quantification produit permet de scanner un grand nombre de codes en peu de temps. Cependant, la procédure ADC reste consommatrice de temps CPU. En particulier, nous avons montré que la performance de la procédure ADC est principalement limitée par le grand nombre d'accès cache qu'elle occasionne. De plus, la structure de cette procédure empêche une utilisation efficace des instructions SIMD pour augmenter sa performance. Ces limitations appellent donc une modification de la procédure ADC.

\paragraph{} Pour ces raisons, nous avons conçu PQ Fast Scan, une procédure de scan de listes de codes compacts novatrice. PQ Fast Scan offre une performance 4 à 6 fois supérieure à la procédure ADC conventionnelle, tout en retournant exactement les mêmes résultats. PQ Fast Scan est exclusivement compatible avec les quantifieurs produits $8{\stimes}8$. Il ne s'agit pas en pratique d'une limitation forte, étant donné que ce type de quantifieur est utilisé dans presque la totalité des cas.  L'idée maîtresse de PQ Fast Scan consiste à remplacer les accès au cache L1 par des permutation SIMD (instruction \texttt{pshufb}). Cette modification permet également une utilisation efficace des additions SIMD. Utiliser des permutations SIMD nécessite de stocker les tables de correspondance dans les registres SIMD. Le principal défi que nous avons eu à résoudre pour la conception de PQ Fast Scan est que les tables de correspondances $D_j$ sont beaucoup plus volumineuses (1 KiB chacune) que les registres SIMD (128 bits chacun).

\paragraph{}
PQ Fast Scan dépasse cette difficulté en utilisant des \emph{petites tables} dimensionnées pour tenir dans les registres SIMD à la place des tables de correspondance stockées en cache. Ces petites tables sont utilisées pour calculer des bornes basses sur les distances, sans accéder au cache L1. Par conséquent les calculs de bornes basses sont rapides. De plus, ils sont implémentés en utilisant des additions SIMD, ce qui améliore encore la performance. Nous utilisons les calculs de bornes basses pour élaguer les calculs de distances, qui nécessitent des accès au cache L1 et ne peuvent pas tirer profit des instructions SIMD. Ainsi, pour chaque code $c$ dans la liste à scanner, on calcule la borne basse sur sa distance au vecteur requête $y$. Si la borne basse est supérieure à la distance entre le vecteur $y$ et le plus proche voisin actuel, alors le code $c$ est écarté et on passe au code suivant. Si la borne basse est inférieure ou égale à la distance entre le vecteur $y$ et le plus proche voisin, on calcule la distance entre le code $c$ et $y$ pour déterminer si $c$ peut devenir le plus proche voisin courant. Nos résultats expérimentaux sur des vecteurs SIFT montrent que les calculs de borne basses permettent d'élaguer plus de 95\% des calculs de distance.

\paragraph{}
Chaque petite table $S^j$ est calculée à partir de la table $D^j$ correspondante, $j \in \{0,\dotsc,7\}$. Pour constuire les petites tables, nous combinons trois techniques: (1) le groupage des codes, (2) le calcul de tables de minimums et (3) la quantification des distances flottantes en entiers 8 bits. Les deux premières techniques, le groupage des codes et le calcul de tables de minimums, sont utilisées pour transformer les tables $D_j$ de 256 flottants en tables de 16 flottants (256$\times$32 bits $\rightarrow$ 16$\times$32 bits). La troisième technique, la quantification des distance flottantes en entier 8 bits, est utilise pour réduire chaque élément de 32 à 8 bits (16$\times$32 bits $\rightarrow$ 16$\times$8 bits).

\begin{figure}
  \centering
  \begin{tikzpicture} [baseline, trim axis left]
  \begin{axis} [
  ybar,
  height=5cm,
  width=3.5cm,
  bar width=15pt,
  ymin=0,
  symbolic x coords={time},
  ylabel= Temps de scan (ms),
  xtick=data,
  nodes near coords={\pgfmathprintnumber[fixed, precision=1]{\pgfplotspointmeta}},
  every node near coord/.append style={font=\small},
  enlarge y limits = {abs=3ex, upper}
  ]
  \addplot+ coordinates {
    (time, 73.8)
  };
  \addplot+ coordinates {
    (time, 12.9)
  };
  \end{axis}
  \end{tikzpicture}
  \begin{tikzpicture} [baseline, trim axis right]
  \begin{axis} [
  ybar,
  height=5cm,
  width=7cm,
  bar width=15pt,
  ymin=0,
  enlarge x limits=0.3,
  symbolic x coords={l1_loads, instructions, cycles},
  ylabel = Compteurs,
  xtick=data,
  xticklabels={cycles, instructions, cache},
  nodes near coords={\pgfmathprintnumber[fixed, precision=1]{\pgfplotspointmeta}},
  every node near coord/.append style={font=\small},
  enlarge y limits = {abs=3ex, upper},
  legend to name=fr-legende-timeperf,
  legend columns=2
  ]
  \addplot+ table[x=counter, y=pq_median] {fastpq/plots/perf_counters.gnuplot};
  \addplot+ table[x=counter, y=fast_pq_median] {fastpq/plots/perf_counters.gnuplot};
  \legend{ADC Scan, PQ Fast Scan};
  \end{axis}
  \end{tikzpicture}
  \\
  \pgfplotslegendfromname{fr-legende-timeperf}
  \caption{Comparaison de ADC Scan et PQ Fast Scan (25M vectors)\label{fig:frfastpq}}
\end{figure}
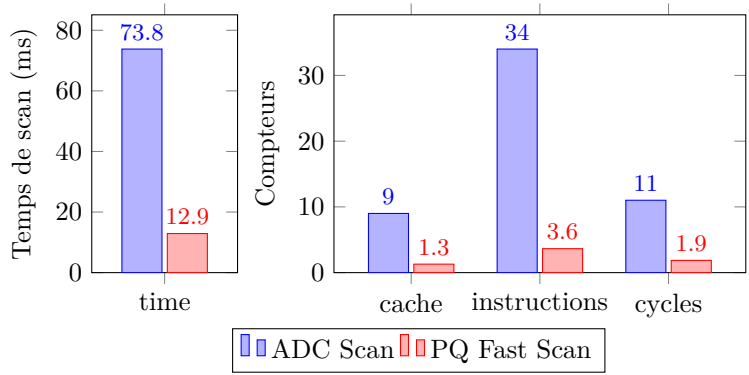

\paragraph{}
Pour valider notre approche, on mesure la performance de PQ Fast Scan et de la procédure ADC Scan conventionnelle sur une liste de 25 millions de codes (Figure~\ref{fig:frfastpq}). PQ Fast Scan atteint un temps de scan de 5.72 fois plus faible qu'ADC Scan à 12.9 ms contre 73.8 ms. Grace au remplacement des accès cache par des permutations SIMD, PQ Fast Scan ne nécessite que 1.3 accès cache par code scanné contre 9 pour ADC Scan. Le nombre d'instructions nécessaires pour scanner un code passe de 34 pour ADC Scan à 3.6 pour PQ Fast Scan, à la fois grace à l'utilisation de permutations SIMD et d'additions SIMD. Au global, ceci se traduit par une baisse du nombre de cycles CPU par code scanné de 11 pour ADC Scan à 1.9 pour PQ Fast Scan.

\begin{figure}
  \centering
  \begin{tikzpicture}[trim axis left, trim axis right]
    \begin{axis} [
      height=5cm,
      width=8cm,
      bar width=5pt,
      ybar,
      symbolic x coords={0,7,2,4,5,3,6,1},
      xtick=data,
      xticklabels={25M, 23M, 11M, 11M, 11M, 11M, 4M, 3.4M},
      every tick label/.append style={font=\small},
      xlabel=Inverted List,
      ylabel style={align=center},
      ylabel=Vitesse de scan\\(M codes/s),
      error bars/y dir=both,
      error bars/y explicit,
      legend columns=2,
      legend to name=fr-legende-fastpqtaille,
    ]
  \addplot+ table[
      x=cluster_id,
      y=pq_speed_median,
      y error plus expr=\thisrow{pq_speed_quart3}-\thisrow{pq_speed_median},
      y error minus expr=\thisrow{pq_speed_median}-\thisrow{pq_speed_quart1}
  ] {fastpq/plots/database_pruned_speed.gnuplot};
  \addplot+ table[
      x=cluster_id,
      y=fast_pq_speed_median,
      y error plus expr=\thisrow{fast_pq_speed_quart3}-\thisrow{fast_pq_speed_median},
      y error minus expr=\thisrow{fast_pq_speed_median}-\thisrow{fast_pq_speed_quart1}
  ] {fastpq/plots/database_pruned_speed.gnuplot};
  \legend{ADC Scan, PQ Fast Scan}
  \end{axis}
  \end{tikzpicture}
  \\
  \pgfplotslegendfromname{fr-legende-fastpqtaille}
  \caption{Impact de la taille des listes inversées sur PQ Fast Scan\label{fig:frfastpqinv}}
\end{figure}
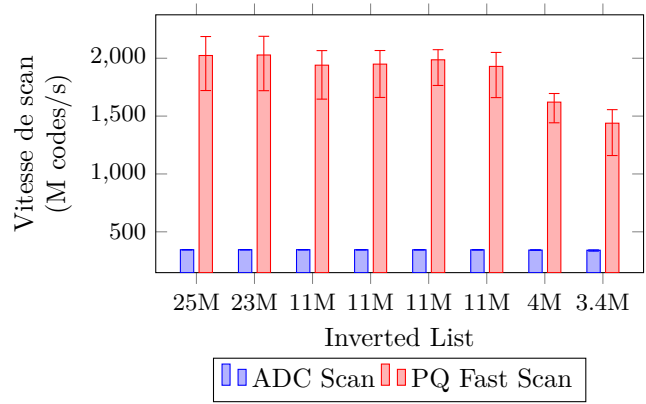

\paragraph{} PQ Fast Scan utilise trois techniques pour obtenir des petites tables qui tiennent dans les registres SIMD: (1) le groupage des codes, (2) le calcul de tables de minimums et (3) la quantification des distances flottantes en entiers 8 bits. Parmi ces trois techniques, le groupage des codes implique une réogarnisation des listes de codes, et impose une taille minimale sur ces listes de codes. En effet, les codes sont répartis dans un nombre fixe de groupes, et la taille moyenne d'un groupe ne doit pas descendre en dessous d'un certain seuil, au risque d'impacter la performance. On peut observer cet effet en mesurant la vitesse de PQ Fast Scan sur 8 listes dont les tailles varient de 3.4 million de codes à 25 millions de codes (Figure \ref{fig:frfastpqinv}). La taille de chaque liste est indiquée en abscisse. PQ Fast Scan a une vitesse constante pour les listes de 11 à 25 million de vecteurs. Cepedant, la vitesse diminue fortement avec la taille de la liste à partir de 4 millions de vecteurs (Figure \ref{fig:frfastpqinv}). Ainsi, PQ Fast Scan n'est intéressant que pour les listes dont la taille dépasse 2 à 3 millions de vecteurs. Ceci rend PQ Fast Scan incompatible avec les configurations d'index inversés les plus performantes qui divisent la base de données en un grand nombre de listes inversées de petite taille (10000 vecteurs ou moins). En pratique, il est donc difficile de combiner l'accélération offerte par PQ Fast Scan et celle offerte par les index inversés.

\section{Quick ADC}

\begin{table}
\centering
\begin{threeparttable}
  \caption{Comparaison ADC - Quick ADC (SIFT1M, 64 bit)\label{tbl:frqadcsift1m}}
  \begin{tabular}{lllllll}
    \toprule
    PQ \tnote{1} & ADC \tnote{2} & Recall@100 & Index & Tables & Scan & Total \\
    \midrule
    \multicolumn{7}{c}{\textbf{Sans index inversé}} \\
    \midrule
    PQ & ADC & 0.916 & - & 0.005 & 2.8 & 2.8 \\
       & QADC & 0.826 & - & 0.001 & 0.38 & 0.38 \\
       &      & \emph{-9.8\%} & & \emph{-80\%} & \emph{-86\%} & \emph{-86\%} \\
    \midrule
    \multicolumn{7}{c}{\textbf{Avec index inversé (K=256, ma=24)}} \\
    \midrule
    \input{quickadc/tables/SIFT1M.256.24.booktabs.tex}
    \bottomrule
  \end{tabular}
  \begin{tablenotes}
    \item[1] PQ: Quantification Produit, OPQ: Quantification Produit Optimisée
    \item[2] ADC: ADC ($8{\stimes}8$), QADC: Quick ADC ($16{\stimes}4$)
  \end{tablenotes}
\end{threeparttable}
\end{table}

\paragraph{}
Nous avons montré que le principal inconvénient de PQ Fast Scan est qu'il n'est pas compatible avec les index inversés les plus performants. Pour cette raison, nous proposons Quick ADC, un autre procédure de scan rapide, mais pouvant être combinée avec n'importe quel type d'index inversé, y compris les plus performants. Comme PQ Fast Scan, Quick ADC, offre des performances 4 à 6 fois supérieures à la procédure ADC conventionnelle. Quick ADC s'appuie également sur un remplacement des accès cache par des permutations SIMD, et utilise des additions SIMD. Cependant, Quick ADC utilise des techniques différentes de PQ Fast Scan pour obtenir les petites tables stockées registres SIMD. Quick ADC s'appuie sur : (1) l'utilisation de quantifieurs 4 bits et (2) la quantification des distances flottantes en entiers 8 bits. L'utilisation de quantifieurs 4 bits permet d'obtenir des tables des correspondance $D_j$ de $2^4=16$ flottants ($16{\stimes}32$ bits). En quantifiant les flottants en entiers 8 bits, on obtient des tables de 16 entiers 8 bits ($16{\stimes}8$ bits) pouvant être stockées dans les registres SIMD.

\paragraph{}
Le principal inconvénient des quantifieurs 4 bits est qu'ils entraînent une perte de précision en comparaison aux quantifieurs 8 bits. Nous montrons cependant que cette perte de précision est faible, voire négligeable, surtout lorsque Quick ADC est combiné avec des index inversés et la quantification produit optimisée. La quantification produit optimisée est un dérivé de la quantification produit offrant une meilleure précision et de plus en plus utilisé dans les publications récentes sur la recherche de plus proche voisin. Sur le jeu de données SIFT1M (1 million de vecteurs SIFT), Quick ADC est plus de 7 fois plus rapide (-86\% sur le temps de réponse) que la procédure ADC conventionnelle (Table~\ref{tbl:frqadcsift1m}). Quick ADC entraine cependant une baisse faible, mais non-négligeable de 9.8\% du rappel. En combinant Quick ADC avec un index inversé, cette baisse est ramenée à 4.4\%. On constate par ailleurs qu'utiliser un index inversé offre à la fois un meilleur temps de réponse (0.48 ms contre 2.8 ms) et un meilleur rappel (0.949 contre 0.916), avec ou sans Quick ADC. Pour cette raison, les index inversés sont utilisés dans la majorité des cas. Enfin, en combinant Quick ADC avec un index inversé et la quantification produit optimisée, la baisse de rappel devient négligeable (-1.5\%). Quick ADC offre alors un gain de temps de réponse de 67\%. Nous obtenons des résultats similaires sur d'autres types de vecteurs (vecteurs GIST \etc) lorsque l'on utilise OPQ. Sur les jeux de données volumineux, par exemple le jeu de données SIFT1B (1 milliard de vecteurs SIFT), Quick ADC offre également un gain en temps de réponse proche de 70\% (Table~\ref{tbl:frqadcsift1b}). La perte de recall est cependant un peu plus importante, -7.3\% sur le jeu de données SIFT1B. Quick ADC propose donc de troquer une perte de recall faible ou négligeable (-1.5\% à -7.3\%), pour un fort gain en temps de réponse (proche de 70\%).

\begin{table}
  \centering
  \begin{threeparttable}
    \caption{Comparaison ADC - Quick ADC (SIFT1B, 64 bit)\label{tbl:frqadcsift1b}}
    \begin{tabular}{lllllll}
      \toprule
      PQ \tnote{1} & ADC \tnote{2} & Recall@100 & Index & Tables & Scan & Total \\
      \midrule
      \multicolumn{7}{c}{\textbf{Avec index inversé (K=65536, ma=64)}} \\
      \midrule
      \input{quickadc/tables/SIFT1B.65536.64.booktabs.tex}
      \bottomrule
    \end{tabular}
    \begin{tablenotes}
      \item[1] OPQ: Quantification Produit Optimisée
      \item[2] ADC: ADC ($8{\stimes}8$), QADC: Quick ADC ($16{\stimes}4$)
    \end{tablenotes}
  \end{threeparttable}
\end{table}

\section{Derived Quantizers}

\paragraph{}
Pour la plupart des cas d'utilisation actuels de la recherche de plus proche voisin, la quantification produit avec des quantifieurs 8 bits offre une précision suffisante. À travers notre solution Quick ADC, nous avons montré qu'il était même possible d'utiliser des quantifieurs 4 bits. Pour améliorer la performance dans ces cas d'utilisation, nous avons proposé deux procédures de scan hautement optimisées, PQ Fast Scan et Quick ADC. Cependant, de nouveaux cas d'utilisations émergents nécessitent une meilleure précision. Par exemple, les descripteurs générés par des réseaux de neurones profonds sont de plus en plus populaires dans les applications multimédia. Ces vecteurs comprennent plusieurs milliers de dimensions, et sont donc plus difficiles à quantifier. Pour ces cas d'utilisation, il est intéressant d'utiliser des quantifieurs produits utilisant de quantifieurs 16 bits, étant donné qu'ils offrent une meilleur précision que les quantifieurs produit utilisant des quantifieurs 8 bits. Par ailleurs, l'utilisation de codes de 32 bits, à la place des codes de 64 bits usuels, soulève un intérêt grandissant. Les codes de 32 bits rendent l'utilisation de quantifieurs de 16 bits nécessaire parce que les quantifieurs 8 bits offrent une précision trop basse dans ce cas. Les quantifieurs 16 bits souffrent cependant d'un inconvénient majeur: ils entraînent un triplement du temps de réponse (Table~\ref{tbl:fr-resume-taille}), si bien qu'ils ne sont pas utilisés en pratique.

\paragraph{}
Pour résoudre ce problème, nous proposons une approche novatrice, les quantifieurs dérivés. Les quantifieurs dérivés rendent les quantifieurs 16 bits aussi rapides que les quantifieurs 8 bits, tout en conservant leur précision. L'idée principale de notre approche consiste à associer un quantifieur dérivé de 8 bits avec chaque quantifieur 16 bits utilisé dans le quantifieur produit.  Les quantifieurs dérivés (8 bits) sont utilisés pour calculer des \emph{tables de correspondance compactes}, stockées dans le cache L1. Ces tables de distances compactes sont utilisées pour calculer des distances \emph{approchées} entre le vecteur requête et les codes de la liste à scanner. Ces calculs de distance approchés servent à sélectionner les $r2$ codes les plus proches du vecteur requête et construire ainsi un ensemble de candidats. Un calcul de distance précis est ensuite effectué pour chacun des $r2$ codes de l'ensemble de candidats. Cette évaluation précise de distance s'appuie sur des tables de correspondance calculées à partir des quantifieurs 16 bits. Étant donné qu'elles ont été calculées à partir de quantifieurs 16 bits, ces tables de distances sont volumineuses, et donc stockées dans le cache L3. En conséquence, les évaluation de distance précises sont lentes, car elles nécessitent d'accéder au cache L3. Cependant, ces évaluations précises de distances sont réalisées uniquement pour les $r2$ codes de l'ensemble de candidats précédemment généré. Au global, ceci permet d'obtenir des temps de réponse bas, très proches du temps de réponse obtenu avec des quantifieurs 8 bits.

\begin{figure}
  \centering
  \begin{tikzpicture} [baseline, trim axis left]
  \begin{semilogxaxis} [
  r plot,
  cycle list={black},
  ylabel=Recall,
  legend to name=fr-legend-derived-64-1m,
  legend columns=3,
  legend cell align=left,
  ]
  \addplot+ [solid, color=red, mark=*]
  table[x=r, y=pq8_recall] {derivedquant/plots/sift1m_r_recall.dat};
  \addplot+ [solid, color=blue, mark=square*]
  table[x=r, y=pq16_recall] {derivedquant/plots/sift1m_r_recall.dat};
  \addplot+ [solid, color=brown, mark=diamond*]
  table[x=r, y=pq816_recall] {derivedquant/plots/sift1m_r_recall.dat};
  
  \addplot+ [dashed, color=red, mark=*]
  table[x=r, y=opq8_recall] {derivedquant/plots/sift1m_r_recall.dat};
  \addplot+ [dashed, color=blue, mark=square*]
  table[x=r, y=opq16_recall] {derivedquant/plots/sift1m_r_recall.dat};
  \addplot+ [dashed, color=brown, mark=diamond*]
  table[x=r, y=opq816_recall] {derivedquant/plots/sift1m_r_recall.dat};
  \legend{PQ $8{\stimes}8$\\
    PQ $4{\stimes}16$\\
    PQ $4{\stimes}8,16$\\
    OPQ $8{\stimes}8$\\
    OPQ $4{\stimes}16$\\
    OPQ $4{\stimes}8,16$\\}
  \end{semilogxaxis}
  \end{tikzpicture}
  \begin{tikzpicture} [baseline, trim axis right]
  \begin{semilogxaxis} [
  r plot,
  cycle list={black},
  legend pos=north west,
  ylabel=Temps par requête (ms)
  ]
  \addplot+ [solid, color=red, mark=*]
  table[x=r, y=pq8_query_us] {derivedquant/plots/sift1m_r_time.dat};
  \addplot+ [solid, color=blue, mark=square*]
  table[x=r, y=pq16_query_us] {derivedquant/plots/sift1m_r_time.dat};
  \addplot+ [solid, color=brown, mark=diamond*]
  table[x=r, y=pq816_query_us] {derivedquant/plots/sift1m_r_time.dat};
  
  \addplot+ [dashed, color=red, mark=*]
  table[x=r, y=opq8_query_us] {derivedquant/plots/sift1m_r_time.dat};
  \addplot+ [dashed, color=blue, mark=square*]
  table[x=r, y=opq16_query_us] {derivedquant/plots/sift1m_r_time.dat};
  \addplot+ [dashed, color=brown, mark=diamond*]
  table[x=r, y=opq816_query_us] {derivedquant/plots/sift1m_r_time.dat};
  \end{semilogxaxis}
  \end{tikzpicture}
  \\
  \pgfplotslegendfromname{fr-legend-derived-64-1m}
  \caption{Recall et temps de réponse (SIFT1M, codes 64 bits)\label{fig:fr-derived-quantizers}}
\end{figure}
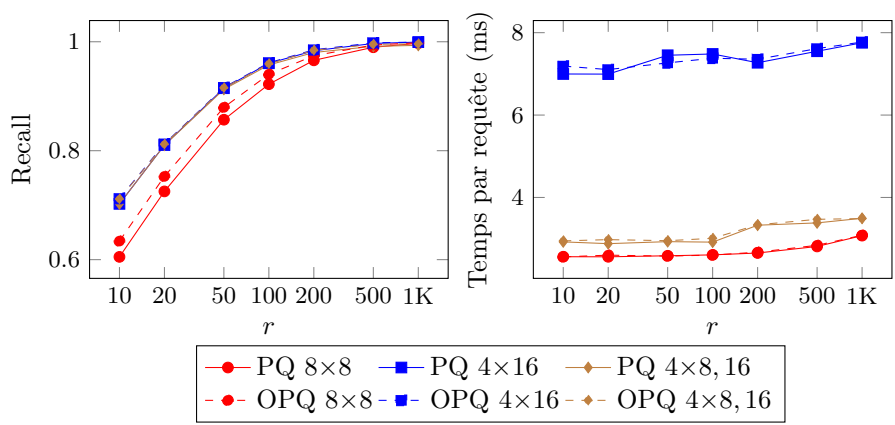

\paragraph{}
Une solution naïve pour obtenir des quantifieurs 8 bits et des quantifieurs 16 bits consisterait à apprendre indépendamment ces deux types de quantifieurs. Chaque vecteur serait alors compressé en code compact deux fois: une première fois en utilisant les quantifieurs 8 bits, et une seconde fois en utilisant les quantifieurs 16 bits. Nous obtiendrions alors deux codes par vecteur, ce qui doublerait l'utilisation mémoire de la base de données. Un tel inconvénient serait rédhibitoire: au lieu d'utiliser des quantifieurs 16 bits pour améliorer la précision, on pourrait simplement utiliser des codes deux fois plus grands avec des quantifieurs 8 bits. Au lieu d'apprendre indépendamment les quantifieurs 8 bits et les quantifieurs 16 bits, on \emph{dérive} le dictionnaire des quantifieurs 8 bits à partir du dictionnaire des quantifieurs 16 bits de tel sorte qu'ils puissent partager un même code. Les quantifieurs 16 bits sont utilisés pour encoder les vecteurs en code compacts et les quantifieurs 8 bits dérivés sont utilisés uniquement pour la recherche de plus proche voisin.

\paragraph{}
On évalue les quantifieurs dérivés sur le jeu de données SIFT1M (Figure~\ref{fig:fr-derived-quantizers}). On observe que les quantifieurs 16 bits ($4{\stimes}16$) offrent un meilleur recall que les quantifieurs 8 bits, que ce soit avec la quantification produit (PQ) ou la quantification produit optimisée (OPQ). Cependant les quantifieurs 16 bits engendrent un triplement du temps de réponse en comparaison aux quantifieurs 8 bits. Notre solution, les quantifieurs dérivés, permet de bénéficier du meilleur des deux mondes. En effet, les quantifieurs dérivés (notés $4{\stimes}8,16$) offrent le même recall que les quantifieurs 16 bits, tout en offrant un temps de réponse très proche des quantifieurs 8 bits (Figure~\ref{fig:fr-derived-quantizers}).

\section{Conclusion et Perspectives}

\paragraph{}
Dans ce résumé, nous avons présenté quatre contributions sur le problème de la recherche de plus proche voisin dans les espaces de haute dimensionnalité, et à large échelle. L'exploitation des bases de données multimédia repose largement sur la recherche de plus proche voisin. Ainsi, ce problème revêt une importance particulière dans le contexte actuel de collecte massive de données multimédia. Les contributions de cette thèse s'appuient sur la quantification produit, une des solutions actuelles de recherche de plus proche voisin les plus efficaces. La quantification produit compresse les vecteurs de haute dimensionnalité en codes compacts, ce qui permet de stocker des bases de données volumineuses en RAM. Pour trouver le plus proche voisin d'un vecteur requête, la quantification produit calcule la distance entre ce vecteur requête et les codes compacts stockés en RAM. Pour ce faire, la quantification produit s'appuie sur une procédure de scan nommée Asymmetric Distance Computation (ADC), ou calcul de distance asymmétrique. Cette procédure utilise des tables de correspondances stockées en cache pour les calculs de distances. Elle consomme un grand nombre de cycles CPU, et est le goulot d'étranglement des systèmes de recherche de plus proche voisin utilisant la quantification produit. 

\paragraph{}
Cette thèse s'intéresse à l'amélioration de la performance de la procédure ADC. Notre première contribution consiste en une analyse des facteurs limitant la performance de la procédure ADC. Nous montrons que la performance de la procédure ADC est principalement limitée par le grand nombre d'accès cache qu'elle effectue. De plus, la structure de la procédure ADC empêche une implémentation efficace utilisant les instructions SIMD, couramment utilisées pour améliorer les performances. Suite à cette analyse, nous proposons deux procédures de scan optimisées: PQ Fast Scan et Quick ADC. Ces deux procédures s'appuient sur un remplacement des accès caches par des permutations SIMD. PQ Fast Scan n'entraine aucune perte de précision, mais n'est pas compatible avec les index inversés, un méthode d'accélération de la quantification produit largement répandue. Quick ADC entraine une faible perte de précision, mais est compatible avec les index inversés. Enfin, notre dernière contribution, nommée quantifieurs dérivés améliore la performance de la procédure ADC dans le cas ou une très haute précision est nécessaire. 

\paragraph{}
Ces contributions ouvrent des perspectives de recherche, à la fois à court terme, et à plus long terme. À court terme, on trouve l'adaptation de PQ Fast Scan et Quick ADC au jeu d'instructions SIMD AVX-512 qui sera introduit dans la prochaine génération de processeurs Intel Xeon (Skylake Purley, sortie prévue en 2017). Le jeu d'instructions AVX-512 offrira des instructions pour effectuer des permutations SIMD sur des tables de 512 bits (contre 128 bits actuellement). Ceci permettra à PQ Fast Scan d'offrir une meilleure compatibilité avec les index inversés, et à Quick ADC d'offrir une meilleure précision. En outre, AVX-512 permettra de traiter plus données par instruction, offrant ainsi des gains de performance. Parmi les perspectives à plus long terme, les techniques utilisées dans PQ Fast Scan pour réduire les tables de correspondance afin qu'elles tiennent dans des registres SIMD peuvent être adaptés pour accélérer d'autre algorithmes utilisant des tables de correspondances, ou certains algorithmes de base de données (requêtes topk \etc).

\end{otherlanguage}{french}

%% file: quickadc/tables/SIFT1M.256.24.booktabs.tex
PQ & ADC & 0.949 & 0.008 & 0.18 & 0.3 & 0.48 \\ 
    & QADC & 0.907 & 0.008 & 0.055 & 0.072 & 0.14 \\ 
    &      & \emph{-4.4\%} &    & \emph{-69\%} & \emph{-76\%} & \emph{-72\%} \\ 
\midrule
OPQ & ADC & 0.963 & 0.008 & 0.21 & 0.29 & 0.52 \\ 
    & QADC & 0.949 & 0.008 & 0.089 & 0.073 & 0.17 \\ 
    &      & \emph{-1.5\%} &    & \emph{-59\%} & \emph{-75\%} & \emph{-67\%} \\ 

%% file: quickadc/tables/SIFT1B.65536.64.booktabs.tex
OPQ & ADC & 0.806 & 0.52 & 0.51 & 4.2 & 5.2 \\ 
    & QADC & 0.747 & 0.53 & 0.22 & 0.92 & 1.7 \\ 
    &      & \textbf{\emph{-7.3\%}} &    & \emph{-57\%} & \emph{-78\%} & \textbf{\emph{-68\%}} \\ 

%% file: intro/introduction.tex
\startcontents[chapters]
\ruledprintcontents

\section{The Big Data Trend}

\paragraph{}
It is indisputable that humanity is producing an increasingly important amount of digital data, which is collected, stored and processed by computer systems in large data centers. A combination of factors arguably explain this surge in the amount of data we produce. First, the wide availability and popularity of capture devices such as digital cameras and smartphones have enabled mass acquisition of digital data. Second, the development of fast internet access and online social networks has allowed corporations to collect the data produced by massive amounts of users. Thus, over 1.6 billion users communicate via the Facebook social network every month~\cite{FacebookQ1}, over 80 million photos are uploaded every day to the Instagram social network~\cite{Instagram400M} and over 400 hours of videos are uploaded every minute to the Youtube~\cite{Youtube400hours} streaming service.
This trend of pervasive data production and collection, often referred to as \emph{big data}, impacts many aspects of society: economy, marketing, politics etc. The term "big data" is no longer a jargon word: between June 2015 and June 2016, the newspaper \emph{The New York Times} published 83 articles including this term. Science is no exception to this trend with the advent of data-intensive science~\cite{FourthParadigm}. For instance, the LHC particle accelerator generates about 30 petabytes of data each year~\cite{CERNCompute}.

\paragraph{}
These large datasets collected by corporations are highly valuable assets, that generate a lot of revenue. Amazon uses the purchase history of its millions of users to automatically generate product recommendations, which allowed it to increase its sales by 29\%~\cite{FortuneAmazon}. In the same vein, Youtube massive video database attracts millions of users, which in turn generates advertising revenue. Generating value requires such data to be processed, analyzed or searched. Thus, Amazon uses \emph{machine learning} algorithms on the purchase history of its users to learn a product recommendation model. Similarly, Youtube indexes its videos by title and keywords so that its database can be searched. Processing and extracting information from these large datasets is a challenging task, not only because of the high \emph{volume} of data and the high \emph{velocity} of acquisition, but also because of the \emph{variety} of data types. Modern distributed processing frameworks such Hadoop MapReduce~\cite{HadoopWebsite, Dean2004} or Spark~\cite{SparkWebsite, Zaharia2012} allow handling large volumes of data, potentially generated at a high velocity by taking advantage of computer clusters. This thesis focuses on the variety aspect of big data. Current datasets contain not only text data (web pages, messages, chats \etc), but also other types of data (images, videos, music, purchase histories \etc), for which conventional text processing approaches do not apply.

\begin{figure}
  \centering
  \input{intro/fig/cbir.tex}
  \caption{Overview of Content Based Image Retrieval (CBIR) systems\label{fig:cbiroverview}}
\end{figure}

\section{Applications of Nearest Neighbor Search}
\label{sec:annapplications}

\paragraph{}
This thesis deals with high-dimensional data, a specific type of data that is found in many applications. In a scalar dataset, each record is a single integer or floating-point number. For instance, datasets of temperature records, power records, or any scalar physical quantity are scalar datasets. Datasets can also be composed of 2-dimensional or 3-dimensional points. For instance, in a dataset of post office positions or train station positions, each record would be a 2-dimensional point. More generally, a dataset can be seen as a collection of $d$-dimensional points, or vectors. Thus, a scalar dataset can be seen as a dataset of 1-dimensional vectors. By contrast, a high-dimensional dataset comprise high-dimensional vectors, \ie $d$-dimensional vectors with a high dimensionality $d$. We consider that vectors with more than 100 dimensions, \ie $d > 100$, are high-dimensional.

\paragraph{}
High-dimensional datasets span many different domains: genetics, recommender systems, multimedia data processing \etc In genetics, DNA microarrays are used to measure the expression of genes in different individuals. A DNA microarray is able to measure the expression of thousands of genes, resulting in a vector with thousands of dimensions. Gene expression measurements are used to study the impact of drugs and diseases on multiple individuals, producing a dataset of multiple high-dimensional vectors. A recommender system recommends items \eg books, movies or products, based on how a given user rated other items. In a recommender system, the profile of a user consists of the ratings she gave to the items she bought. This profile is represented by a high-dimensional vector, where each dimension corresponds to an item in the recommender system. Typical recommender systems obviously include the profile of multiple users, thus resulting in a large dataset of high-dimensional vectors. Multimedia objects (such as pictures or movies), can be represented by high-dimensional feature vectors, that capture their contents. Therefore, a multimedia dataset, including multiple pictures or movies, can be represented as a datasets of high-dimensional vectors. 

\paragraph{}
In this thesis, we focus on nearest neighbor search in high-dimensional datasets. Simply put, this problem consists in finding the closest vector to a query vector among a database of high-dimensional vectors. Despite its apparent simplicity, nearest neighbor search has many applications in machine learning and multimedia data processing. In machine learning, nearest neighbor search can be used as a standalone algorithm, but it is also at the basis of many algorithms. Examples of such algorithms include collaborative filtering algorithms, used in recommender systems. A classifier is a machine learning system which determines the category a new observation belongs to based on a training set of prior observations, the categories of which are known. A nearest neighbor classifier determines the category of a new observation by finding its nearest neighbor in the training set, and returning the category of the found nearest neighbor.

\paragraph{}
Above all, nearest neighbor search is of particular importance for multimedia information retrieval problems. Content Based Image Retrieval (CBIR), consists in finding the images in a database that are the most similar to a query image provided by the user. As the user essentially provides an example image of what he wants to retrieve, content-based image retrieval is also known as query by example. CBIR systems rely heavily on nearest neighbor search in high-dimensional spaces, and are the guiding theme of this thesis. In particular, the nearest neighbor search systems presented in this thesis were evaluated in the context of CBIR systems, but may be used for other purposes. In the remainder of this section, we give an overview of how CBIR systems work. CBIR systems rely on the extraction of feature vectors, or descriptors, that capture the visual contents of pictures. Finding two similar pictures then comes down finding two pictures that have a similar set of descriptors. CBIR systems typically operate in two distinct phases: an offline phase and an online phase. During the offline phase, descriptors are extracted from every image in the dataset, therefore creating a database of high-dimensional descriptors. During the online phase, an user submits a query image to the system. High-dimensional query descriptors are extracted from the query image. The nearest neighbors of the query descriptors are searched in the database of high-dimensional descriptors created during the offline phase, which in turn allows to identify the most similar images to the query image (Figure~\ref{fig:cbiroverview}). Because nearest neighbor search is used during the online phase, it is essential to use fast nearest neighbor search solutions in order to answer queries in a timely manner.

\paragraph{}
As the nearest neighbor search systems presented in this thesis have been tested on databases of high-dimensional descriptors extracted from pictures, we present these descriptors into more detail. The dimensionality of currently used descriptors ranges from a hundred dimensions to a few thousand dimensions. We distinguish between global descriptors and local descriptors, depending on the number of descriptors extracted per picture. Global descriptors describe a complete image, therefore an image can be described by a single global descriptor. On the contrary, local descriptors describe a small region of the picture, named patch. Therefore, an image can be described by a set of local descriptors. The number of local descriptors required to describe an image usually ranges from a hundred to a thousand. One of the most popular global descriptor is the GIST descriptor~\cite{Oliva2001} the dimensionality of which ranges 384 and 960 dimensions. The most popular local descriptor is arguably the Scale Invariant Feature Transform (SIFT) descriptor~\cite{Lowe1999}, with 128 dimensions. An alternative to the SIFT descriptor with similar properties is the SURF descriptor~\cite{Bay2006}, implemented in the open-source library OpenCV. Descriptors like SIFT and SURF descriptors were designed manually to have a strong descriptive power while being resistant to image variations. It has recently been proposed to use deep neural networks to learn robust descriptors with strong descriptive power without human intervention. These deep feature vectors tend to very high-dimensional (usually 4096 dimensions), but have been found to have a higher descriptive power than other descriptors, which is why they are gaining in popularity.

\paragraph{}
For CBIR, local descriptors generally offer a better accuracy than their global counterparts. In addition, local descriptors enable object detection, \ie identifying objets inside pictures, which is not possible with global descriptors. For these reasons, local descriptors are often preferred to global descriptors. The drawback of local descriptors is that they lead to the creation of very large high-dimensional databases, as a single image is described by hundreds of local descriptors instead of a single global descriptor. This in turn increases the pressure on the nearest neighbor search system, which has to deal with a database hundreds of times larger.

\section{Classification of Nearest Neighbor Search Problems}

\paragraph{}
Most practical applications of nearest neighbor search require a set of near neighbors of the query vectors instead of the nearest neighbor. For instance a CBIR system returns multiple images similar to the query image, and not a single one. Similarly, recommender systems recommend items to an user based on the preferences of a set its nearest neighbors. We generally distinguish between two different nearest neighbor search problems: k-nearest neighbor search (k-NN) and $\epsilon$-nearest neighbor search ($\epsilon$-NN). Given a dataset $\mathcal{X}$ of $n$ $d$-dimensional vectors, $\mathcal{X} = \{x_0, \ldots, x_{n-1}\}$ and $d$-dimensional query vector $q$, these two problem are defined as follows:
\begin{description}
  \item[k-NN] The k-NN set of $q$ is the set of $k$ vectors of $\mathcal{X}$ that are the closest to $q$.
  \item[$\epsilon$-NN] The $\epsilon$-NN set of $q$ is the set of vectors of $\mathcal{X}$ whose distance to $q$ is less than $\epsilon$. As $\epsilon$-NN search returns all neighbors in the hypersphere centered on $q$ of radius $\epsilon$, it is also known as radius search.
\end{description}
For many applications, it is generally more intuitive to pick a number $k$ of nearest neighbors than a threshold $\epsilon$, which is harder to determine. For instance, it is easier to determine that a CBIR system should return $k=100$ similar images, rather than to determine a similarity threshold $\epsilon$. Therefore, k-NN is often preferred to $\epsilon$-NN. In this thesis, we focus on the k-NN problem. An important observation is that these two problems are related, the $\epsilon$-NN set of $q$ is the same as the $k$-NN set of $q$ if the radius $\epsilon$ is the distance between $q$ and its $k$th nearest neighbor. Therefore, methods designed to solve the $\epsilon$-NN problem can be adapted to solve the $k$-NN problem.

\paragraph{}
The notion of nearest neighbor implies a similary measure or distance metric. Depending on applications, different distance metric are used: $l_1$ norm, $l_2$ norm or even cosine similarity. In this thesis, we focus on the $l_2$ norm, as it is one of the most widely used distance metric. More specifically, it is used in CBIR systems as well as many machine learning algorithms.

\section{The Curse of Dimensionality}

\paragraph{}
The naive approach to solve the k-NN problem is to compute the distance between the query vector $q$ and each vector of the dataset $\mathcal{X}$. This solution, known as linear scan or bruteforce scan, is obviously not tractable for large datasets. A dataset of $n=10^{9}$ of SIFT descriptors ($d=128$ dimensions) stored as a contiguous array of floats (4 bytes/float) uses $S = n \cdot d \cdot 4\text{ bytes} = 512\text{ GB}$ of memory. If this dataset is stored on an SSD with a read throughput of 512 MB/s, scanning the whole dataset would require 1000 seconds, or 17 minutes. Thus, answering a single k-NN query would require 17 minutes, assuming that there are no other bottlenecks. This is of course not practical for an interactive CBIR system. Offline machine learning algorithms, such as recommender systems, often require to compute thousands to millions k-NN sets so linear scan is also not tractable for many non-interactive applications.

\paragraph{}
In computer science, divide-and-conquer is a common approach that often allows solving problems with a better efficiency than bruteforce algorithms. In a 1-dimensional space \ie when the dataset $\mathcal{X}$ contains scalar values, nearest neighbor search admits a simple divide-and-conquer solution, binary search. The dataset $\mathcal{X}$ should first be stored in an array and sorted in ascending order. The binary search algorithm then compares the scalar $q$ to the middle element of $\mathcal{X}$. If the $q$ is less than the middle element, then its nearest neighbor is in the first half of $\mathcal{X}$. Conversely, if $q$ is greater than the middle, its nearest neighbor is in the second half. This procedure is recursively repeated until the nearest neighbor of $q$ is found. Binary search has a logarithmic time complexity, while a bruteforce scan has a linear complexity.

\paragraph{}
For spaces with 2 or more dimensions, the dataset $\mathcal{X}$ obviously cannot be sorted, but divider-and-conquer approaches based on space partitioning can still be designed. A popular structure for nearest neighbor search in $d$-dimensional spaces is the KD-tree~\cite{Bentley1975,Friedman1977}. 
Each non-terminal node of KD-tree consists of one high-dimensional vector and an hyperplane orthogonal to one dimension of the hyper-space. Each terminal node only consists of one high-dimensional vector. The hyperplane of the root of the KD-tree is positioned at the median of the dimension where the data exhibits the greatest variance, thus halving the dataset. To build the other nodes, the dataset is recursively split in the same way until partitions contain only a single point. These single points from the terminal nodes of the tree. Thus, each node of a KD-tree defines a cell in the high-dimensional space. To answer a nearest neighbor query, a tree descent is performed by determining on which side of the hyperplanes the query vector $q$ lies. This yields a first nearest neighbor candidate. This candidate is not necessarily the true nearest neighbor of $q$. Cells intersecting with the ball whose radius is the distance between $q$ and the first candidate can contain a closer neighbor than this first candidate. These cells are therefore searched for better candidates. Nearest neighbor search stops when there are no cells intersecting with the hypersphere whose radius is the distance of $q$ to the best candidate found so far. 

\paragraph{}
Unfortunately, it has been shown that the number of cells that must be explored to be sure to find the true nearest neighbor of $q$ grows exponentially with the dimensionality $d$~\cite{Friedman1977}. Thus, nearest neighbor search with a KD-tree degrades to linear scan as the dimensionality increases. Worse, in~\cite{Weber1998}, the authors show that it is not possible to design a space partitioning method that does not degrade to linear scan as the dimensionality increases. This phenomenon is known as the curse of dimensionality. As exact nearest neighbor search is inherently hard, the research community has focused on approximate nearest neighbor search. Approximate Nearest Neighbor (ANN) search aims at returning close enough neighbors instead of the exact closest ones. The key idea is to trade exactness for efficiency. Moreover, approximate nearest neighbors are sufficient in many applications, including CBIR systems. An obvious tradeoff in the case of KD-trees is to stop exploring cells after a fixed number have been explored. This reduces search time, but does not guarantee that the exact nearest neighbor will be found. More sophisticated approaches have been designed; we review the most prominent ones in the state of the art section.

\section{Product Quantization}

\paragraph{}
Product Quantization is a widespread ANN search solution that compresses high-dimensional vectors into short codes. In most cases, high-dimensional vectors can be represented by codes of 8 bytes to 16 bytes (64 bits to 128 bits). This allows very large datasets to be stored in RAM, therefore enabling nearest neighbor queries without accessing the SSD of HDD. RAM typically has a throughput in excess of 40GB/s and a latency of around 100ns, while the highest-performing SSDs have a throughput of 2-3GB/s and a latency of 100{\textmu}s. Because it entirely relies on RAM, product quantization allows answering nearest neighbor queries faster than approaches that rely on SSDs. To compress a high-dimensional vector into a short code, product quantization splits it into $m$ sub-vectors and encodes each sub-vector using a distinct sub-quantizer. Each sub-quantizer has a codebook of $2^b$ entries, and produces a sub-code of $b$ bits. For fixed memory budget of $m \times b$ bits per short code, a large $b$ (and therefore a small $m$) offers a better accuracy, but makes nearest neighbor search slower.

\paragraph{}
In practice, product quantization in used in combination with a form of inverted index. Different types of inverted indexes have been proposed, but they all split the database into multiple inverted lists so that only a fraction of the database has to be scanned at query time. However, because of the curse of dimensionality, it is difficult to build efficient indexes. Thus, it is still necessary to scan a large part of the database to answer queries (typically 1-20\%). Generally, database preparation takes two steps. First, the database is split into several parts using the index. Second, the high-dimensional vectors composing each part are compressed into short codes using product quantization, as described in the previous paragraph. Both the index and the short codes are stored in RAM, to enable fast answers to ANN queries.

\paragraph{}
One of the unique features of product quantization is that it is able to compute the distance between an uncompressed high-dimensional query vector $y$ and compressed database vectors. Therefore, it is not necessary to decompress the database to answer ANN queries. To answer an ANN query, the relevant partition is first selected using the index, and then scanned for nearest neighbors. Scanning the partition consists in computing the distance between the query vector $y$ and all short codes of the partition. To do so, a set of $m$ lookup tables, derived from the quantizer codebooks, are computed. Then, the distance between the query vector $y$ and any short code can be computed using a technique named Asymmetric Distance Computation (ADC).
ADC consists in (1) performing $m$ lookups in the previously computed tables (2) adding the looked up values, thus performing $m-1$ additions. To date much work has been dedicated to improving the indexes used with product quantization. \emph{On the contrary, this thesis focuses on improving the performance of ADC, by making it more hardware friendly.}

\section{Thesis Outline}

\paragraph{}
In the next chapter of this thesis, we review the state of the art of ANN search in high-dimensional spaces. Each contribution (Performance Analysis, PQ Fast Scan, Quick ADC, Derived Quantizers) is then detailed in a dedicated chapter. The last chapter draws conclusions from the contributions presented in this thesis, and elaborates on some perspectives that they open.

\paragraph{State of the Art} We present three families of nearest neighbor search approaches: (1) Space and data partitioning trees, (2) Locality Sensitive Hashing (LSH) and (3) Product Quantization. We show that partitioning trees and LSH use large amounts of memory when dealing with large databases. This makes it necessary to use the SSDs or HDDs, causing high response times. We then present product quantization and show that it can store large databases entirely in RAM, which allows low response times. All our contributions build on product quantization, therefore the last section of the State of the Art chapter is also the background of our work.

\paragraph{Performance Analysis} Using hardware performance counters, a mechanism included in CPUs that allows monitoring the operations they perform, we analyze the performance of ADC. We show that the table lookups performed by ADC account for the majority of CPU time, even if these lookup tables are cache resident. We show that number $b$ of bits per sub-quantizer strongly impacts response time, because its impacts the cache level in which lookup tables are stored. In almost all publications on product quantization, 8-bit sub-quantizers ($b=8$) are used because it has been experimentally observed that they offer a good tradeoff between speed and accuracy. Our analysis explains why, but also suggests means of improving this tradeoff. Lastly we introduce SIMD instructions (Single Instruction Multiple Data) and show that they are a candidate for accelerating the additions performed by ADC. We however demonstrate that ADC cannot be easily implemented in SIMD, which prevents obtaining a large speedup. All systems presented in this thesis (PQ Fast Scan, Quick ADC, Derived Quantizers) stem from the conclusions of our performance analysis.

\paragraph{PQ Fast Scan} PQ Fast Scan achieves a 4-6 times speedup over the conventional ADC algorithm. This is achieved by replacing cache accesses to lookup tables by SIMD in-register shuffles. SIMD in-register shuffles are much faster that cache accesses, but they require lookup tables to be stored in SIMD registers, which are much smaller than cache. PQ Fast Scan shrink lookup tables by (1) reorganizing short codes, (2) computing tables of minimums and (3) quantizing floating-point distances to 8-bit integers. PQ Fast Scan uses 8-bit sub-quantizers, as it is common practice in the literature, and provides the exact same results as the conventional ADC algorithm. The drawback of PQ Fast Scan is that because it re-organizes short codes, it requires the database to be split in relatively coarse parts. Therefore, it is not compatible with the most efficient types of inverted indexes usually used with product quantization.

\paragraph{Quick ADC.} Like PQ Fast Scan achieves a 4-6 times speedup over the conventional ADC algorithm using SIMD in-register shuffles. However, Quick ADC uses a different approach to shrink lookup tables. Quick ADC shrinks lookup tables by (1) using 4-bit sub-quantizers instead of the common 8-bit sub-quantizers and (2) quantizing floating point distances to 8-bit integers. Using 4-bit sub-quantizers instead of 8-bit sub-quantizers decreases accuracy but contrary to PQ Fast Scan, Quick ADC imposes no constraints on inverted lists, and may be combined with any type of inverted index. Moreover, we show that by using optimized product quantization, a derivative of product quantization, the loss of accuracy caused by the use of 4-bit sub-quantizers can be made negligible in most cases.

\paragraph{Derived Quantizers.} While PQ Fast Scan and Quick ADC aim at slashing response time without impacting accuracy, derived quantizers aim at increasing accuracy without increasing response time. To increase accuracy, it is possible to use 16-bit sub-quantizer instead of the 8-bit sub-quantizer used in most publications. However, this strategy causes a threefold increase in response time, because lookup tables have to be stored in slower cache levels. Derived quantizers allow combining the accuracy of 16-bit sub-quantizers with the speed of 8-bit sub-quantizers. This is achieved by deriving 8-bit sub-quantizers from 16-bit sub-quantizers, so that they generate the same short codes. At query time, 8-bit (fast) sub-quantizers are used to build a small candidate set, which is then reranked using the 16-bit (precise) sub-quantizers.

\paragraph{Conclusion and Perspectives} We conclude by reviewing the key features of our contributions, and we discuss the research opportunities that stem from them. More specifically, we discuss how our contributions can be adapted to recent derivatives of product quantization that offer a higher accuracy. We also show that our contributions can be implemented on other CPUs than Intel CPUs (and in particular ARM CPUs). Lastly, we discuss the applicability of the techniques we developed in the context of product quantization to other algorithms that rely on lookup tables.

\stopcontents[chapters]

%% file: intro/fig/cbir.tex
\newcommand{\imWidth}{4.5cm}

\tikzset{
  vector array cell/.style = {
    rectangle,
    align=center,
  },
  vector array/.style = {
    matrix of math nodes,
    row sep = -\pgflinewidth,
    column sep = -\pgflinewidth,
    nodes={
      vector array cell
    },
  }
}

\begin{tikzpicture}

\node [
] (db1) {\includegraphics[width=\imWidth]{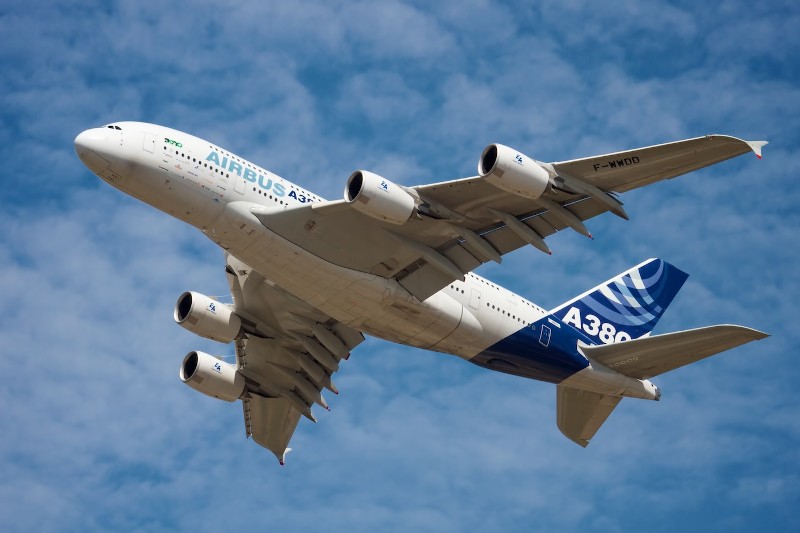}};

\draw [decoration={brace,raise=5pt},
  decorate] (db1.north west) --  node (dblbl) [above=10pt, align=center, text depth=0.25ex] {Database of Pictures} (db1.north east);

\coordinate[draw, color=red] (a1) at ($ (db1.west) + (0.3, 0.1 cm) $);
\coordinate (vp1) at ($ (db1.north east) + (0,-1cm) $);

\node [
  below = 0 cm of a1.south,
  anchor = north west,
] (db2) {\includegraphics[width=\imWidth]{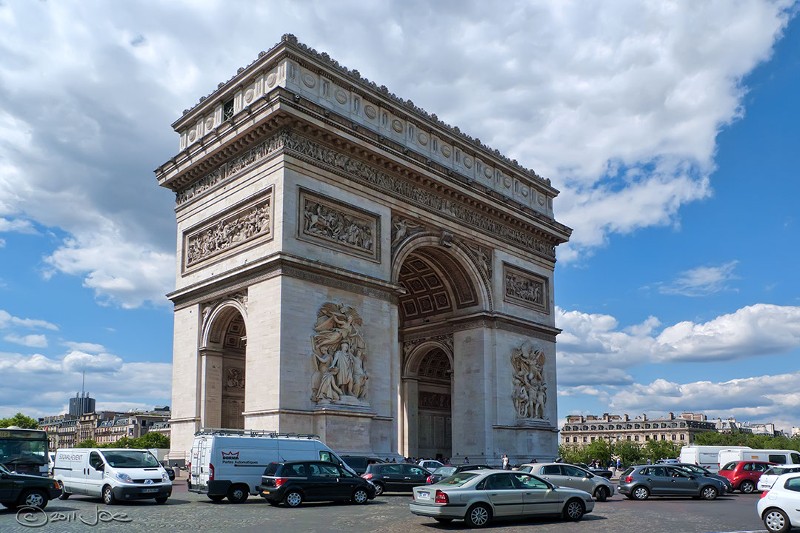}};

\coordinate[draw, color=red] (a2) at ($ (db2.west) + (0.3,0.1 cm) $);

\node [
  right = 1 cm of db2.west,
  below = 0 cm of a2.south,
  anchor = north west,
] (db3) {\includegraphics[width=\imWidth]{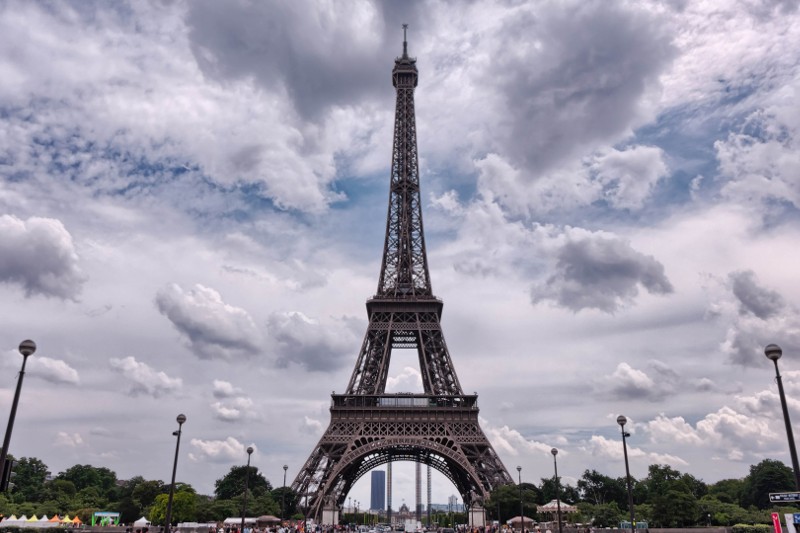}};

\coordinate (vp3) at ($ (db3.south east) + (0,1cm) $);

\node[vector array,
  right=1 cm of vp3,
  anchor=west,
] (vecs3)
{%
  $(0.568, 0.967, 0.188, \dotsc, 0.988)$\\
  $(0.081, 0.691, 0.695, \dotsc, 0.778)$\\
  $(0.465, 0.381, 0.640, \dotsc, 0.991)$\\
};

\coordinate (v2) at (db2.east -| vecs3.west);

\node[vector array,
  right=0 cm of v2.east,
  anchor=west,
] (vecs2)
{%
  $(0.423, 0.374, 0.431, \dotsc, 0.053)$\\
  $(0.026, 0.845, 0.800, \dotsc, 0.760)$\\
  $(0.868, 0.211, 0.013, \dotsc, 0.475)$\\
};

\coordinate (v1) at (vp1 -| vecs3.west);

\node[vector array,
  right=0 cm of v1.east,
  anchor=west,
] (vecs1)
{%
  $(0.137, 0.576, 0.369, \dotsc, 0.886)$\\
  $(0.489, 0.741, 0.504, \dotsc, 0.599)$\\
  $(0.656, 0.335, 0.512, \dotsc, 0.942)$\\
};

\coordinate (bracew) at (db1.north east -| vecs1.north west);
\coordinate (bracee) at (db1.north east -| vecs1.north east);

\draw [decoration={brace,raise=5pt},
  decorate] (bracew) --  node (dblbl) [above=10pt, align=center, text depth=0.25ex] {Database of Descriptors} (bracee);

\draw[->] (vp1) -- (vecs1.west);
\draw[->] (db2.east) -- (vecs2.west);
\draw[->] (vp3) -- (vecs3.west);


\node[
  below = 1.8 cm of db3.south,
  anchor=north
] (imq) {\includegraphics[width=\imWidth]{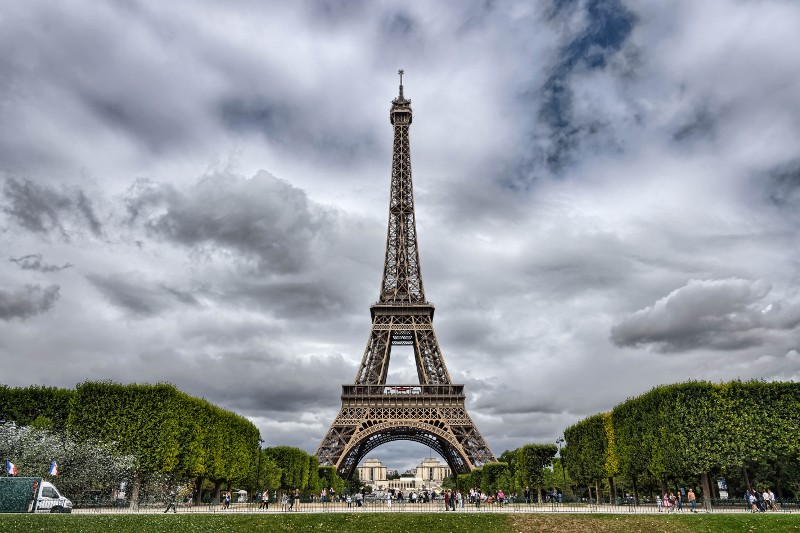}};

\coordinate (q) at (imq.east -| vecs3.west);

\node[vector array,
  right=0 cm of q.east,
  anchor=west,
] (vecsq)
{%
  $(0.362, 0.259, 0.249, \dotsc, 0.187)$\\
  $(0.223, 0.279, 0.460, \dotsc, 0.324)$\\
  $(0.604, 0.194, 0.548, \dotsc, 0.558)$\\
};

\draw[->] (imq.east) -- (vecsq.west);

\draw[<->, thick] (vecsq.north) -- (vecs3.south) node[pos=0.5, anchor=west, right=0.2cm, align=center] {\emph{Nearest Neighbor}\\\emph{Search}};

\draw [decoration={brace,raise=5pt, mirror},
  decorate] (imq.south west) --  node (piclbl) [below=10pt, align=center, text depth=0.25ex] {Query Picture} (imq.south east);

\coordinate (brace2w) at (imq.south east -| vecsq.south west);
\coordinate (brace2e) at (imq.south east -| vecsq.south east);

\draw [decoration={brace,raise=5pt,mirror},
  decorate] (brace2w) --  node (dblbl) [below=10pt, align=center, text depth=0.25ex] {Query Vector(s)} (brace2e);
  
\node[
  below=9.4cm of db1.south west,
  anchor=north west,
  align=left,
  node font=\scriptsize]{\textbf{Photo Credits:} (Top to bottom)\\[0.3em]
  --\hspace{0.5em}\href{https://www.flickr.com/photos/davidgsteadman/4829995791/}{"Farnborough Air Show - A380"} by \href{https://www.flickr.com/photos/davidgsteadman/}{davidgsteadman}, public domain\\[0.15em]
  --\hspace{0.5em}\href{https://www.flickr.com/photos/mustangjoe/6356230029}{"Arc de Triomphe de l'Étoile"} by \href{https://www.flickr.com/photos/mustangjoe/}{Joe deSousa}, public domain\\[0.15em]
  --\hspace{0.5em}\href{https://www.flickr.com/photos/mustangjoe/9257105067/}{"Eiffel Tower"} by \href{https://www.flickr.com/photos/mustangjoe/}{Joe deSousa}, public domain\\[0.15em]
  --\hspace{0.5em}\href{https://www.flickr.com/photos/mustangjoe/20520637385/}{"The Eiffel Tower, Paris"} by \href{https://www.flickr.com/photos/mustangjoe/}{Joe deSousa}, public domain\\[0.3em]
    Photos downloaded from Flickr};

\end{tikzpicture}

%% file: sota/sota.tex
\startcontents[chapters]
\ruledprintcontents

\paragraph{}
In this chapter, we present three families of ANN search approaches: (1) Space and data partitioning trees, (2) Locality Sensitive Hashing (LSH) and (3) Product Quantization (PQ). We first present Space and data partitioning trees and LSH. We show that these two families of approaches use large amounts of memory, mandating the use of SSDs or HDDs to store large databases. We then present product quantization, an approach that compresses high-dimensional into short codes, making it possible to store large databases in RAM. Because RAM is much faster than SSDs or HDDs, product quantization offers low response times, especially for large databases. We explain how product quantization can be combined with inverted indexes to further decrease response times. Some of the inverted indexes used in combination with product quantization were inspired by the space and data partitioning trees presented in the beginning of this chapter.

\paragraph{}
All the contributions of this thesis are based on product quantization, thus the last section of this chapter (Section~\ref{sec:sota-pq}) also describes the background of our work.

\paragraph{} 

\section{Space and Data Partitioning Trees}
\label{sec:spacetree}

\paragraph{}
The first data structures proposed for efficient nearest neighbor search are space and data partitioning trees. Space-partitioning trees (KD-trees) divide the high-dimensional space along predetermined lines. On the other hand, data partitioning trees (Vocabulary trees, k-means trees) divide the high-dimensional space according to data clusters.

\subsection{KD-trees}
\label{sec:kdtree}

\paragraph{Tree structure}
KD-trees~\cite{Bentley1975, Friedman1977} can be seen as a generalization of binary search trees. At each non-terminal node of a KD-tree, the dataset is divided into two halves, by a hyperplane orthogonal to a chosen dimension at a threshold value. In the literature, the chosen dimension is sometimes known as the discriminator and the threshold is known as the partition value. Generally, the dimension where the data exhibits the greatest variance, or the greatest spread, is chosen as the discriminator. The median of the values in the chosen dimension is chosen as the partition value. Each non-terminal node of a KD-tree contains two pointers to two sons, or successors nodes. These two successor nodes start two sub-tree. The left sub-tree contains the high-dimensional vectors for which the value in the dimension chosen as discriminator is less than the partition value. Correspondingly, the right sub-tree contains the high-dimensional vectors for which the value in the dimension chosen as discriminator is greater or equal to the partition value. Each terminal node generally contains a single high-dimensional vector of the dataset, although in some implementations, terminal nodes may contain more than one vector. Each node in a KD-tree defines a cell in $\mathbb{R}^d$. Moreover, The cells defined by terminal nodes are mutually exclusive, and thus form a partition of $\mathbb{R}^d$.

\paragraph{Build process}
KD-trees are built in a top-down fashion, following a simple recursive process. At the root, the whole dataset is split by a hyperplane at the median of the dimension where the data has the greatest variance. The resulting two halves of the data are then recursively split until the tree is fully built. The resulting height of the tree is $\operatorname{O}(\log_2(n))$, where $n$ is the number of points in the dataset. Figure~\ref{fig:searchtree} shows a KD-tree with 3 levels. The hyperplane of the root is shown by a solid line, the hyperplanes of the second level nodes are shown by dashed lines, while the hyperplanes of the third level nodes are shown by dotted lines.

\paragraph{Nearest Neighbor Search} To answer a nearest neighbor query, a tree descent is first performed. This tree descent consists in determining to which side of the hyperplane the query vector falls at each node of the tree. The descent involves $\log_2(n)$ comparisons and yields a terminal node, and thus a cell in $\mathbb{R}^d$. The vectors located in this cell are examined to find a first nearest neighbor candidate. However this first candidate may not be the nearest neighbor of the query vector. For instance, if the query vector is close to a boundary of the cell the tree descent landed in, the first candidate will be a vector in this cell. There may however be a closest neighbor in a adjacent cell, close to the boundary. In the remainder of this section, we denote $B_{query}$, the ball centered at the query vector with a radius equal to the distance between the query vector and the current nearest neighbor. To find the exact nearest neighbor, it is therefore necessary to explore all cells the boundaries of which overlap the ball $B_{query}$  (Figure~\ref{fig:searchtree}). In~\cite{Weber1998}, the authors show that the number of cells that need to be explored grows rapidly with the dimensionality, to the point where almost the whole dataset needs to be explored when the dimensionality exceeds 15. This phenomenon is a consequence of the curse of dimensionality. This issue can be overcome by stopping search when a fixed number $n_{\mathit{max}}$ of cells have been explored. However, the guarantee of finding the exact nearest neighbors is lost, and this is therefore an approximate solution. 

\paragraph{Priority Search} In the original search algorithm, the cells to explore are determined by backtracking in the tree, and checking if hyperplanes overlap with the ball centered at the query vector. The order in which cells are explored thus depends on the structure of the tree, and not on the position of the cells relative to the query vector. However, cells that are close to the query vectors are more likely to contain good neighbors than cells that are far away. Therefore, an intuitive idea to find better approximate neighbors is to explore cells based on their proximity with the query vector. This can be achieved by maintaining a priority queue during the initial tree descent. At each node, the position in the tree, and the distance of the query vector to the unexplored cell (\ie the cell at the other side of the hyperplane) are stored in the priority queue. When backtracking, the top entry of the priority queue is removed, and the search continues by exploring the corresponding cell. This approach is known as Best Bin First (BBF)~\cite{Beis1997} or priority search.

\paragraph{KD-forests} Exploring more cells to improve the quality of approximate neighbors leads to diminishing returns~\cite{Silpa2008}: the more cells have been explored, the less the benefit of exploring an additional cell is. This is because searches in different cells are not independent: the more explored cells, the further away the cells are from the query point. In~\cite{Silpa2008}, the authors propose building $l$ KD-trees with a different structure, such that they allow independent searches. In total, $n_{\mathit{max}}$ cells are explored, so $n_{\mathit{max}}/l$ cells are searched in each tree. Three techniques are proposed to build forests of independent trees: NKD-trees, RKD-trees, and PKD-trees. To build an NKD-tree, vectors are randomly rotated before building the tree. To build an RKD-tree, instead of using the one dimension with the greatest variance to position the hyperplane, a random dimension among the few with the greatest variance is chosen. Lastly, PKD-trees rely on Principal Component Analysis and random rotations. The authors show that NKD-trees and RKD-trees perform equally. PKD-trees perform slightly better but are much more costly to build.

\subsection{Bag of Features and Vocabulary Trees}

\begin{figure}
  \centering
  \input{sota/fig/invertedindex.tex}
\end{figure}

\paragraph{Bag of features} The Bag of Features (BoF) approach~\cite{Sivic2003} was not original presented as generic nearest neighbor search system, but as a Content Based Image Retrieval (CBIR) system. However, it was one of the first image retrieval system that scaled well, and can be adapted into a generic nearest neighbor search system, which is why we present it. We first present the original approach, and then show how to generalize it into a generic nearest neighbor search system. In a CBIR systems, an image is described by a set of high-dimensional feature vectors, or descriptors, where each descriptor captures the contents of a patch of the image (Section~\ref{sec:annapplications}). The key idea of the BoF approach is to partition the vector space into $K$ mutually exclusive Voronoi cells using the k-means algorithm. A subset of the descriptors extracted from the database of images is used to learn the partition. This partition is used to build and \emph{inverted index}. The inverted index takes the form of an array of $K$ elements, where each element corresponds to one Voronoi cell of the partition. Each element contains a pointer to a list of image IDs. When an image is added to the database, its descriptors are extracted. Then the cell in which each descriptor falls is determined, and the image ID is added to the corresponding lists of the inverted index. For instance, if an image has descriptors that fall in cells 5, 768 and 1023, the ID of this image is added to the lists 5, 768 and 1023 of the inverted index. To retrieve the most similar image to a query image, the descriptors of the query image are first extracted. The cells in which they fall are determined, and the corresponding lists of image IDs are retrieved from the inverted index. A voting scheme is then applied to determine the most similar image. The simplest voting scheme consists in finding the image with the most frequent image ID among the retrieved lists, but more elaborate schemes have been proposed~\cite{Sivic2003}. As this system was inspired by text search engines, where documents are indexed based on the words they contain, the Voronoi cells are sometimes named \emph{visual words}.

\paragraph{} The idea of using an inverted index, based on a Voronoi partition of the vector space, can be adapted to build a generic nearest neighbor search system. Instead of storing images IDs in the lists of the inverted index, high-dimensional vectors can be stored. Adding a vector to the database then consists in determining in each cell it falls, and adding it to the corresponding inverted list (Figure~\ref{fig:invindex}). To answer a nearest neighbor query, the cell in which the query vector falls is first determined. The corresponding list is retrieved, and the distance of the query vector to all vectors in the retrieved list is computed to determine the nearest neighbor. To improve the accuracy, the $\mathit{ma}$ inverted lists corresponding to the \emph{$\mathit{ma}$} closest cells to the query vector can be scanned. 

\begin{figure}
  \centering
  \subfloat[][KD-tree]{
    \input{sota/fig/kdtree.tex}
  }%
  \hfil
  \subfloat[][Vocabulary tree]{
    \input{sota/fig/voctree.tex}
  }
  \caption{ANN search in KD-trees and Vocabulary trees\label{fig:searchtree}}
\end{figure}
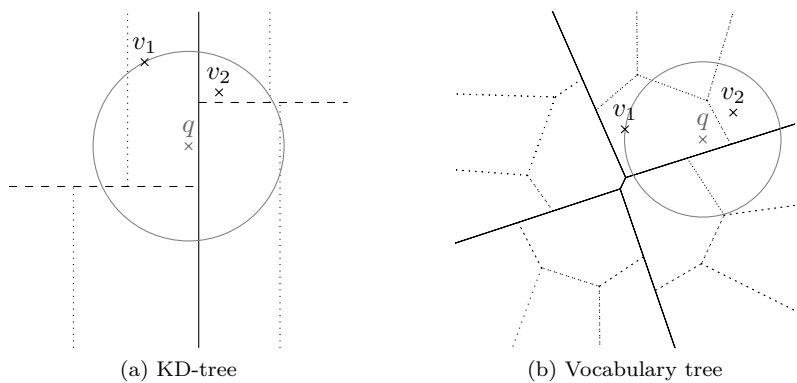

\paragraph{Vocabulary trees} The main issue with the bag of features approach, especially when used as an image retrieval system, is that the size of the visual vocabulary, \ie the number $K$ of Voronoi cells, is limited. Dividing the space into more than $K=2^{16}$ using k-means is not tractable. However, using a large vocabulary is desirable, as it better discriminates between similar images patches, and thus provides a better accuracy. The vocabulary tree~\cite{Nister2006} allows using a larger visual vocabulary (\ie millions of visual words), while maintaining an acceptable computational cost. In~\cite{Nister2006}, vocabulary trees store image IDs, like in the bag of features approach described in~\cite{Sivic2003}. The vocabulary tree is built by the means of a hierarchical k-means clustering. Like in the bag of features approach, the vector space is first partitioned into $k$ cells using the k-means algorithm. This partitioning is used to build the root of the vocabulary tree: the root node has $k$ child nodes, each one corresponding to a cell. Each cell is then divided into $k$ cells using the k-means algorithm to build the second level of the tree. This process is recursively applied to build the other levels of the tree, until the tree reaches a depth $l$. The resulting vocabulary has a branching factor of $k$, and a depth of $l$, therefore the number of terminal nodes and the total number of cells is $k^{l}$. Figure~\ref{fig:searchtree} shows a vocabulary tree with a branching factor $k=4$ and a depth $l=2$. The root divides the space in $k=4$ cells (solid lines). At the second level of the tree, each cell is again divided into $k=4$ cells (dotted lines). A similar idea (hierarchical k-means tree) had been proposed in~\cite{Fukunaga1975}.

\subsection{FLANN Library}
\label{sec:flann}

\paragraph{}
FLANN~\cite{Muja2014} (Fast Library for Approximate Nearest Neighbors) is an open-source ANN search library, included in the widely used computer vision framework OpenCV. Both FLANN and OpenCV are distributed under the BSD license. Because it is open-source, and because it is one of the only production ready ANN search libraries that can be used by non-specialists, FLANN is highly popular and used in many computer vision projects. FLANN mainly relies on (1) forests of randomized KD-trees (RKD-trees, Section~\ref{sec:kdtree}) and (2) the priority search k-means tree. The k-means trees used in FLANN are similar to the vocabulary trees used in~\cite{Nister2006}, except that they store high-dimensional vectors, instead of image IDs. Therefore, high-dimensional vectors of the database are added to the tree by perform a tree descent. The vectors are then stored in the obtained terminal nodes. To search these trees, the authors of~\cite{Muja2014} propose using the same priority search method as the one used to search KD-trees. Thus, to answer a nearest neighbor search query, a first tree descent is performed ot generate a first candidate. It is followed by a backtracking process where the cells that overlap $B_{query}$ (ball centered at the query vector with a radius equal to the distance between the query vector and the current nearest neighbor) are explored (Figure~\ref{fig:searchtree}). The key advantage of FLANN is that it chooses the best algorithm, and determines the optimum parameters automatically. This selection is based on the type of high-dimensional vectors, and user-specified constraints: desired accuracy, search time, maximum memory overhead, maximum tree build time. This feature makes the FLANN library easy to use for non-specialists. 

\section{Locality Sensitive Hashing (LSH)}

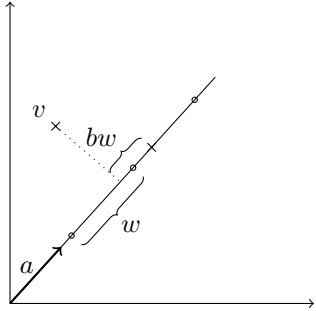
\begin{figure}
  \centering
  \input{sota/fig/lsh.tex}
  \caption{Geometrical interpretation of LSH functions\label{fig:lshgeo}}
\end{figure}

\paragraph{}
Locality Senstive Hashing (LSH) is another prominent nearest neighbor search approach. The key difference between LSH and other approaches mentioned in this thesis is that LSH offers theoretical guarantees on the quality of the returned nearest neighbors. LSH tackles a variant of ANN search , the $c$-approximate nearest neighbor search problem. Formally, a vector $v$ is a $c$-approximate neighbor nearest neighbor of a query vector $q$ if the distance between $v$ and $q$ is at most $c$ times the distance between $q$ and its exact nearest neighbor $v^{*}$, \ie $\lVert v - q \rVert \leq c \cdot \lVert v^{*} - q\rVert$. Accordingly, the $c$-approximate $k$ nearest neighbors problem consists in finding $k$ vectors that are respectively the $c$-approximation of the exact $k$ nearest neighbors of $q$.

\subsection{Euclidean Distance LSH (E2LSH)}
\paragraph{}
LSH was originally proposed for the Hamming space ($l_1$ norm)~\cite{Gionis1999}, and later adapted to Euclidean spaces ($l_2$ norm), with the E2LSH method~\cite{Datar2004}. As this thesis focuses on nearest neighbor search methods for the $l_2$ norm, we only describe the E2LSH method, and not the original Hamming space LSH method.

\paragraph{}
E2LSH does not solve the $c$-approximate nearest neighbor problem directly but rather focuses on the $(R,c)$-approximate nearest neighbor problem, a decision version of the $c$-approximate nearest neighbor problem. In the remainder of this section, we denote $\operatorname{B}(q, r)$, the ball centered at the query vector $q$ with the radius $r$. With this notation, the $(R,c)$-approximate nearest neighbor problem is defined as follows:
\begin{enumerate}
  \item If there is at least one vector in $\operatorname{B}(q, R)$, a vector in $\operatorname{B}(q, cR)$ is returned.
  \item If there is no vector in $\operatorname{B}(q, cR)$, nothing is returned
\end{enumerate}
The $(R,c)$-approximate nearest neighbor problem is sometimes known as "ball-cover" search in the literature.

\paragraph{}
E2LSH uses $(R, cR,p_1,p_2)$-sensitive LSH functions to solve the $(R,c)$-approximate nearest neighbor problem. A $(R, cR,p_1,p_2)$-sensitive LSH function $h$ has the following properties, for any $v \in \mathbb{R}^d$:
\begin{enumerate}
  \item If $v \in \operatorname{B}(q,R)$, then $\operatorname{Pr}[\operatorname{h}(v)=\operatorname{h}(q)] \geq p_1$
  \item If $v \notin \operatorname{B}(q,cR)$, then $\operatorname{Pr}[\operatorname{h}(v)=\operatorname{h}(q)] \leq p_2$
\end{enumerate}
The LSH functions used in E2LSH have the following form:
\begin{equation}
\operatorname{h}(v) = \left \lfloor \frac{a\cdot v + bw}{w} \right \rfloor
\end{equation}
In these functions, $a$ is a vector $a \in \mathbb{R}^d$, where each dimension has been drawn from a normal distribution (mean 0, variance 1). The parameter $w$ is a scalar constant, and $b$ is a real number uniformly drawn from $[0,1)$. These LSH functions can be interpreted geometrically: the vector $v$ is first projected onto a line, the direction of which is given by the vector $a$ (Figure~\ref{fig:lshgeo}). The projection of $v$ is then shifted by the constant $bw$. The line is divided into intervals of size $w$, and the hash of $v$ is the number of the interval in which its shifted projection falls. It has been shown~\cite{Datar2004} that the probability $\operatorname{Pr}[\operatorname{h}(v)=\operatorname{h}(q)]$ that two vectors $q$ and $v$ collide under $h$ decreases when $r = \lVert q - v \rVert$ increases, but increases when $w$ increases. Intuitively, we see that two far away points may still fall in the same interval, for instance, if they are far away in a direction orthogonal to $a$. To increase the discriminating power, E2LSH therefore uses compound hash functions $\operatorname{G}$, which consist in the concatenation of $m$ LSH functions $h$: $\operatorname{G}(v) = (\operatorname{h_1}(v),\dotsc,\operatorname{h_m}(v))$. E2LSH uses $L$ distinct compound hash functions, $\operatorname{G_1},\dotsc,\operatorname{G_L}$ to build $L$ hash tables, $T_1,\dotsc,T_m$. Each database vector $v$ is hashed using the $L$ compound hash functions and stored in the corresponding bucket of each of the $L$ hash tables. To answer nearest neighbor queries, E2LSH hashes the query vector $q$ with the $L$ compound hash functions $\operatorname{G_1}(v),\dotsc,\operatorname{G_m}(v)$ and retrieves the vectors stored in the corresponding buckets of $T_1,\dotsc,T_m$. The number of vectors retrieved is limited to $3L$ to avoid checking a too large number of candidates. For each one of the $3L$ retrieved vectors, E2LSH checks if it is in $\operatorname{B}(q,cR)$. As soon as a vectors in $\operatorname{B}(q,cR)$ is found, it is returned and search stops. If no vector in $\operatorname{B}(q,cR)$ can be found, E2LSH returns nothing. The $m$ and $L$ parameters are chosen such that the two following properties hold with a constant probability: (1) if there is a database vector $v$ in $v \in \operatorname{B}(q,cR)$, then $\operatorname{G_j}(v) = \operatorname{G_j}(q)$ must be true for at least one compound hash function, and (2) the total number of vectors $v$ that collide with $q$ under one $G_j$ and which are not in $\operatorname{B}(q,cR)$ is less than $3L$.

\paragraph{}
The scheme described so far solves the $(R,c)$-approximate nearest neighbor problem. The $c$-approximate nearest neighbor problem can be solved by issuing several $(R,c)$ nearest neighbor queries with an increasing radius $R=\{1,c,c^2,c^3,\dotsc\}$. As the $m$ and $w$ parameters used in compound hash functions $\operatorname{G_1}(v),\dotsc,\operatorname{G_L}(v)$ depend on the radius $R$, different compound hash functions must be used for different radius $R$. Therefore, a set of $L$ hash tables has to be built for each radius $R=\{1,c,c^2,c^3,\dotsc\}$. This approach is known as \emph{rigorous-LSH}, and offers theoretical guarantees on the quality of nearest neighbors. However, the duplication of hash tables leads to a very high memory use, and makes rigorous-LSH intractable for anything but very small datasets (up to a few tens of thousands of vectors). Another approach was proposed, \emph{adhoc-LSH}, which consists in building hash tables for a single "magic" radius $r_m$. This greatly reduces the memory cost, but theoretical guarantees are lost. Furthermore, it has been shown that this approach offers poor empirical performance.

\subsection{Virtual Rehashing Schemes}

\paragraph{}
The main challenge in designing an efficient LSH scheme is to offer theoretical guarantees or good empirical performance while maintaining an acceptable memory use. Modern LSH schemes rely on a form a \emph{virtual rehashing}, although the details vary widely. The key idea of virtual rehashing is to allows queries with different radii, without physically building a set of hash tables for each radius $R=\{1,c,c^2,c^3,\dotsc\}$. We present three LSH schemes that rely on virtual rehashing: LSB-forest~\cite{Tao2009}, Collision-counting LSH (C2LSH)~\cite{Gan2012}, and Sorting-Keys LSH (SK-LSH)~\cite{Liu2014}. Both LSB-forest and C2LSH offer theoretical guarantees while SK-LSH does not. However, SK-LSH has been shown to offer better empirical performance than LSB-forest or C2LSH.

\paragraph{LSB-forest~\cite{Tao2009}} Like E2LSH, LSB-forest hashes database vectors $v$ using compound hash functions $\operatorname{G_j}(v)$. The $m$-dimensional hash values $\operatorname{G_j}(v)$ are converted to Z-order values $\operatorname{z_j}(v)$. Z-order values are bit strings, obtained by grid partitioning the $m$-dimensional hash value space. An LSB-tree is then created by building a conventional B-tree over the Z-order values. Lastly, an LSB-forest is created by building $L$ LSB-trees using $L$ distinct compound hash functions $\operatorname{G_1},\dotsc,\operatorname{G_L}$ (converted to Z-order values $\operatorname{z_1(v),\dotsc,\operatorname{z_L(v)}}$). To answer a nearest neighbor search query, the $L$ LSB-trees are explored. To explore an LSB-tree, the Z-order value of the query vector $q$ is computed $\operatorname{z_j}(q)$. The entries in LSB-trees are explored by decreasing Length of the Longest Common Prefix (LLCP) with $\operatorname{z_j}(q)$. Exploring Z-order values $\operatorname{z_j}(v)$ in decreasing order of their LLCP with $\operatorname{z_j}(q)$ simulates the process of issuing several $(R,c)$ nearest neighbor queries with an increasing radius.

\paragraph{Collision-counting LSH (C2LSH)~\cite{Gan2012}} Unlike E2LSH or LSB-forest, C2LSH does not use predetermined compound hash functions $\operatorname{G_j}$. Instead, C2LSH builds $m$ hash tables $\operatorname{T_1},\dotsc,\operatorname{T_m}$ using $m$ $(1, c,p_1,p_2)$-sensitive LSH functions ($R=1$) $\operatorname{h_1},\dotsc,\operatorname{h_m}$, and stores database vectors into these hash tables. To answer a nearest neighbor search query, C2LSH hashes the query vector $q$ with all hash functions $\operatorname{h_1},\dotsc,\operatorname{h_m}$, and retrieves the vectors in the corresponding buckets. The list of retrieved vectors is likely to contain duplicates, as it is likely that some databases vector collide with $q$ under multiple hash functions. C2LSH builds candidate set $C$ of database vectors that collide with $q$ under $l$ or more hash functions, \ie that are duplicated at least $l$ times in the list of retrieved vectors. This collision counting procedure can be seen as can be seen as the dynamic creation of compound hash functions, tailored to the query vector. This allows C2LSH to answer $(R,c)$ nearest neighbor queries for $R=1$. In~\cite{Gan2012}, the authors show that if a function $\operatorname{h}(\cdot)$ is $(1,c,p_1,p_2)$-sensitive, then the function $\operatorname{H^R}(\cdot)=\operatorname{h}(\cdot)/R$ is $(R,cR,p_1,p_2)$-sensitive. If $R$ is an integer, the $\operatorname{T_1},\dotsc,\operatorname{T_m}$ hash tables can therefore be used to answer $(R,c)$ nearest neighbor queries for $R > 1$ but retrieving $R$ contiguous buckets.

\paragraph{Sorting-Keys LSH (SK-LSH)~\cite{Liu2014}} Like E2LSH and LSB-forest, SK-LSH uses compound hash functions $\operatorname{G_j}$. In~\cite{Liu2014}, the authors build a linear order on the $m$-dimensional hash keys generated by compound hash functions. This linear order allows building B-trees on compound hash keys. Database vectors are hashed using a small number $L$ (usually $L=3$) of compound hash functions $\operatorname{G_1},\dotsc,\operatorname{G_L}$, and stored in corresponding B-trees $\operatorname{T_1},\dotsc,\operatorname{T_L}$. To answer nearest neighbor search queries, the query vector is hashed using compound hash functions, and a tree descents are performed. Entries are explored by increasing distance with the compound hash key of the query vector, until a fixed number of entries has been explored.

\paragraph{} While E2LSH is only suitable for very small datasets (up to a few tens of thousands of vectors), the recent improvements presented in this section make LSH suitable for small and medium datasets (up to a few million vectors). Table~\ref{tbl:lshmem} summarizes the memory consumption of the LSH schemes presented in this section for a medium dataset of 1 million 128-dimensional SIFT vectors (SIFT1M). The size of the data structures used by recent LSH approaches makes it possible to store the database in RAM. However, the memory use of these approaches remains high. The size of the SIFT1M dataset is 128MB. Therefore, even SK-LSH causes a threefold increase in memory use. FLANN (Section~\ref{sec:flann}) typically uses less than 250MB of RAM for this dataset. For large datasets (\eg 1 billion vectors), LSH requires storing the dataset on an SSD or HDD, which causes high response times.

\begin{table}
  \centering
  \caption{Memory use of different LSH approaches (SIFT1M)\label{tbl:lshmem}}
  \begin{tabular}{llll}
    \toprule
    Approach & LSB & C2LSH & SK-LSH \\
    Memory use & 81 GB & 1.1 GB & 384 MB\\
    \bottomrule
  \end{tabular}
\end{table}

\section{Product Quantization}
\label{sec:sota-pq}

\paragraph{}
So far, we presented two families of ANN search approaches : (1) Space and data partitioning trees, and (2) LSH approaches. As these approaches do not compress the database, large databases do not fit in RAM and must stored in SSDs or HDDs.
Unlike the previously presented approaches, product quantization allows large databases to be stored in RAM, by compressing high-dimensional vectors into short codes. To answer a query, short codes are scanned for nearest neighbors, without decompressing the database. Thus, for the medium SIFT1M dataset (1 million 128-dimensional SIFT vectors), product quantization only uses 8 to 20 MB of RAM. For a large database of 1 billion SIFT vectors, product quantization uses only 8 GB to 20 GB or RAM. Such as database would be impossible to store in RAM with partitioning trees or LSH.

\paragraph{}
Moreover, product quantization can be combined with inverted indexes to decrease response times. Inverted index makes it possible to answer nearest neighbor queries by only scanning a part of the database, while product quantization makes it possible to store the database in RAM. Some inverted indexes used with product quantization were inspired by the space partitioning trees presented in Section~\ref{sec:spacetree}. Because RAM has much higher performance than SSDs, combining product quantization with inverted indexes allows for shorter response times than approaches that solely rely on inverted indexes.

\paragraph{}
In this section, we first describe how product quantization encodes high-dimensional vectors into short codes. We then present the different types of inverted indexes that can be combined with product quantization. We show how a database of short codes can be searched for nearest neighbors, without decompressing it. In particular, we introduce the Asymmetric Distance Computation (ADC) procedure, which is the focus of this thesis. Lastly, we review some derivatives of product quantization, that also compress high-dimensional vectors into short codes. These recently introduced derivatives achieve a lower quantization error but sometimes involve a more costly search process. We fully discuss their advantages and drawbacks.

\subsection{Vector Encoding}
\label{sec:vectorencoding}

\paragraph{Vector Quantizer} To compress high dimensional vectors as short codes, product quantization builds on vector quantizers~\cite{Gray1984}. A vector quantizer, or quantizer, is a function $\operatorname{q}$ which maps a vector $x \in \mathbb{R}^d$, to a vector $c_i \in \mathbb{R}^d$ belonging to a predefined set of vectors $\mathcal{C}=\{c_0, \ldots, c_{k-1}\}$. Vectors $c_i$ are called \emph{centroids}, and the set of centroids $\mathcal{C}$, of cardinality $k$, is the \emph{codebook}. For a given codebook $\mathcal{C}$, a quantizer which minimizes the quantization error must satisfy Lloyd's condition~\cite{Lloyd1982}. The Lloyd condition requires that the vector $x$ is mapped to its closest centroid $c_i$:
\[
\operatorname{q}(x) = \argmin_{c_i \in \mathcal{C}}{||x-c_i||}.
\]
At the implementation level, the set of centroids $\mathcal{C}=\{c_0, \ldots, c_{k-1}\}$, or codebook, is stored as an array of vectors. Therefore, $\mathcal{C}[0]$ denotes the first centroid, $\mathcal{C}[1]$ the second centroid \etc Storing the codebook $\mathcal{C}$ as an array amounts to \emph{ordering} it, but the ordering is arbitrary. Thus, any two centroids $\mathcal{C}[i], \mathcal{C}[j], i\neq j$ may be swapped without affecting the properties of the quantizer. The index of each centroid \ie its position in the array, is used as a unique identifier for this centroid. 

\paragraph{}
From a vector quantizer $\operatorname{q}$, it is possible to define an encoder, a function which encodes a vector $x \in \mathbb{R}^d$ as a short code $i \in \{0, \ldots, k-1\}$. For given quantizer $q$, the corresponding encoder $\operatorname{enc}$ maps a vector $x \in \mathbb{R}^d$ to the index of its closest centroid, as follows:
\[
\operatorname{enc}(x) = i, \text{ such that }\operatorname{q}(x) = \mathcal{C}[i].
\]
The short code $i$ is an integer using $b = \lceil \log_2(k) \rceil$ bits, which is typically much less than the $d\cdot 32$ bits used to store a vector $x \in \mathbb{R}^d$ as an array of floats. For instance, using a quantizer with $2^{8}$ centroids, a 16-dimensional vector can be encoded into a 8 bits (1 byte) short code. Stored as an array of floats, a 16-dimensional vector would use $16 \cdot 32 = 512$ bits (64 bytes) of memory. Vector quantizers are not specific to high-dimensional nearest neighbor search and are used in many fields of computer science, especially in telecommunications systems and data compression. However, encoding high-dimensional vectors ($> 100$ dimensions) while maintaining the quantization error low requires quantizers with a very large number of centroids \eg $2^{64}$ or $2^{128}$ centroids. Training such quantizers is obviously intractable: storing the codebook of a quantizer with $2^{64}$ centroids would require more memory than the aggregated HDD or SSD capacity of all computers on earth.

\paragraph{Product Quantizer} Product quantizers~\cite{Jegou2011} overcome this issue by dividing a vector $x \in \mathbb{R}^d$ into $m$ sub-vectors $(x^0,\dotsc,x^{m-1})$, assuming that $d$ is a multiple of $m$:
\[
x = (\underbrace{x_0,\dotsc, x_{d/m-1}}_{x^0}, \dotsc, \underbrace{x_{d-d/m},\dotsc, x_{d-1}}_{x^{m-1}}).
\]
Each sub-vector $x^j \in \mathbb{R}^{d/m}$ is quantized using a distinct sub-quantizer $\operatorname{q^j}$. Each sub-quantizer $\operatorname{q}^j$ has a distinct codebook $\mathcal{C}^j=\{c_{0}^j,\dotsc,c_{k-1}^j\}$ of cardinality $k$. Overall, a product quantizer $\operatorname{pq}$ maps a vector $x \in \mathbb{R}^d$ as follows:
\begin{align*}
\operatorname{pq}(x) &= \left ( \operatorname{q}^0(x^0),\dots,\operatorname{q}^{m-1}(x^{m-1}) \right )\\
&= (\mathcal{C}^0[i_0],\dotsc,\mathcal{C}^{m-1}[i_{m-1}]).
\end{align*}

By definition, a product quantizer has an implicit codebook $\mathcal{C}$ which is the Cartesian product of the sub-quantizers codebooks \ie $\mathcal{C} = \mathcal{C}^0 \times \dots \times \mathcal{C}^{m-1}$. The codebook $\mathcal{C}$ of the product quantizer has a cardinality of $k^m$, but it is never stored explicitly. Here lies the key feature of product quantizers: they are able to mimic a quantizer with $k^m$ centroids, while only requiring to train and store $m$ codebooks of $k$ centroids. For instance, a product quantizer with $8$ sub-quantizers of $256$ centroids each has an implicit codebook of $256^8 = 2^{64}$ centroids, while only requiring to store 8 codebooks of 256 centroids.

\paragraph{}
Like simple vector quantizers, product quantizers can be used to encode vectors as short codes. For a given product quantizer $\operatorname{pq}$, the corresponding encoder $\operatorname{enc}$ maps a vector $x \in \mathbb{R}^d$ to the concatenation of indexes of the closest centroids of the sub-vectors $x^j$ in each sub-quantizer codebook:
\[
\operatorname{enc}(x) = (i_0,\dots,i_{m-1}),\text{ such that }\operatorname{q}(x) = (\mathcal{C}^0[i_0],\dotsc,\mathcal{C}^{m-1}[i_{m-1}])
\]
The short code $(i_0,\dots,i_{m-1})$ is the concatenation of $m$ integers of $b = \lceil \log(k) \rceil $ bits each, for a total size of $m \cdot b$ bits. A common practice is to use product quantizers with 8 sub-quantizers of 256 centroids each to encode 128-dimensional SIFT vectors into short codes. In this case, 128-dimensional vectors use $128\cdot32 = 4096$ bits (512 bytes) while their short code is composed of $8$ integers of $8$ bits each, or 64 bits (8 bytes) in total. Storing the database as short codes therefore achieves a 64 times decrease in memory use.

\paragraph{Product Quantizer Parameters}
When a product quantizer is used for approximate nearest neighbor search, its parameters $m$, number of sub-quantizers, $k$, number of centroids per sub-quantizer impact: (1) the memory use of the database, (2) the accuracy of nearest neighbor search and (3) the speed of ANN search. We only consider sub-quantizers whose number of centroids is a power of 2 \ie $k=2^{b}$. We name $b$-bit sub-quantizer a sub-quantizer with $k = 2^{b}$ centroids. We denote \pq{m}{b} a product quantizers with $m$ $b$-bit sub-quantizers ($k=2^{b}$ centroids per sub-quantizer). An $m{\stimes}b$ product quantizer has total number of $2^{m\cdot b}$ centroids, produces short codes of $m\cdot b$ bits. The total number of centroids of a product quantizer is the parameter that has the most impact on quantization error and search accuracy~\cite{Jegou2011}. This creates a trade-off between accuracy and memory use: a high $m \cdot b$ product increases both accuracy and memory use. In practice, product quantizers such that $m\cdot b = 64$ or $m \cdot b = 128$ are selected, depending on vector type. For a fixed $m\cdot b$ product (\eg $m \cdot b=64$), the relative values of $m$ and $b$ create a trade-off between accuracy and speed. For a fixed $m \cdot b$ product, it has been observed that the higher the $b$, the higher the accuracy but also the slower the search speed \cite{Jegou2011}. If search speed generally decreases when $b$ increases, this decrease is not continuous; we fully discuss and explain the impact of $b$ on search speed in Chapter~\ref{sec:perf}.

\subsection{Inverted Indexes}
\label{sec:invindex}

\paragraph{Exhaustive search} Product quantization encodes high-dimensional vectors into short codes, and is able to compute the distance between a high-dimensional query vector and any short code. The simplest search strategy is to encode all database vectors into short codes, and store the short codes in a contiguous array in RAM. Answering a nearest neighbor query then comes down to computing the distance between the query vector and all short codes in the array, \ie scanning the whole database. This strategy, known as \emph{exhaustive search}~\cite{Jegou2011} is suitable for databases up to a few million vectors. Even if product quantization offers an efficient way to compute distances between a high-dimensional query vector and short codes, scanning the whole database becomes too costly as its size increases. 

\paragraph{Inverted index, non-exhaustive search}
To overcome this issue, \emph{inverted indexes}, also known as \emph{inverted files} are combined with product quantization. The database is first partitioned into several parts. Each part is then stored as an \emph{inverted list}, a contiguous array of short codes. The inverted index structure holds pointers to all inverted lists, and provides a fast access to any one of them. At query time, the most relevant inverted lists to the query are scanned for nearest neighbors. Inverted indexes allow scanning only a fraction of the database to answer a query; this strategy is known as \emph{non-exhaustive} search.

\paragraph{}
In general, making small inverted lists \ie finely partitioning the database and scanning several inverted lists at query time gives better results than making large inverted lists and scanning only one. Therefore, in practice, multiple inverted lists are scanned to answer a query. This method is known as \emph{multiple assignement}. There are two variants of this approach: either (1) a fixed number ($\mathit{ma}$) of inverted lists are scanned or (2) inverted lists are scanned until the total number of short codes scanned reaches a given threshold. Note that multiple assignement is for queries only, database vectors are always stored in a single inverted list.

\paragraph{}
Different indexing techniques have been proposed to partition the database and retrieve the relevant inverted lists at query time. We detail the two most popular indexing approaches (IVFADC~\cite{Jegou2011}, Multi-D-ADC~\cite{Babenko2015}), and quickly review less popular approaches. Because they avoid scanning the whole database, inverted indexes combined with product quantization offer shorter response times than product quantization alone. More surprisingly, some indexing techniques (in particular IVFADC and Multi-D-ADC) offer better accuracy than product quantization alone, despite the fact that only a fraction of the database is scanned. For theses reasons, these techniques are often preferred to exhaustive search in practical scenarios.

\paragraph{Simple inverted index (IVFADC)~\cite{Jegou2011}} Vector quantizers can be used to encode vectors as short codes, but they can also be used to partition the database. IVFADC (Inverted File with Asymmetric Distance Computation) uses a vector quantizer with $K$ centroids to partition the vector space into $K$ Voronoi cells. Each centroid and its corresponding Voronoi cell are assigned an integer identifier between $0$ and $K-1$. This approach is similar to the Bag of Features presented in Section~\ref{sec:spacetree}, except that instead of storing high-dimensional vectors in inverted lists, short codes produced by a product quantizer are stored. The partitioning quantizer is named index quantizer, and denoted $\operatorname{q_i}$. Before they are encoded into short codes, the residual $r(x)$ of each database vector $x$ is computed:
\[
r(x) = x - \operatorname{q_i}(x)
\]
The residual $r(x)$ of an high-dimensional vector can therefore be interpreted at is relative position to its closest centroid in the index quantizer $\operatorname{q_i}$. Each residual $r(x)$ is then encoded into a short code using a product quantizer $\operatorname{pq}$, and added to the proper inverted list. In this manner, if vector $x$ lies in the Voronoi cell the identifier of which is 3, its residual will be encoded and added to the inverted list the identifier of which is 3. IVFADC uses two quantizers for two different purposes: (1) a vector quantizer to partition the database and (2) a product quantizer to encode high-dimensional vectors as short codes. Residuals $r(x)$ typically have a lower entropy than the original vectors $x$. Therefore, encoding residuals instead of the original vectors leads to a lower quantization error, which in turn increases accuracy. To answer a query, the $\mathit{ma}$ closest centroids to the query vector are searched in the index quantizer $q_i$ codebook, and the corresponding inverted lists are scanned. For a dataset of 1 billion SIFT vectors, two typical configurations are: (1) using an index quantizer $q_i$ with $K=2^{13}$ centroids, and scanning $ma=64$ inverted lists at query time and (2) using an index quantizer $q_i$ with $K=2^{13}$ centroids, and scanning $ma=64$ inverted lists at query time. In configuration 1, about $1/128^{\text{th}}$ of the database is scanned to answer a query while in configuration 2, $1/1024^{\text{th}}$ of the database is scanned. Both configurations offer similar accuracy, but configuration 2 is faster as less short codes are scanned. In general, finer inverted indexes allow scanning a smaller fraction of the dataset while retaining the same accuracy, and thus allow faster response times. However, retrieving the $ma$ most relevant inverted lists requires computing the distance of the query vector to each of the $K$ centroids of the index quantizer in order to find the $ma$ closest ones. This operations becomes costly as $K$ increases. Above $K=2^{16}$ centroids, the increased cost of matching the query vector against the index quantizer codebook starts to outweigh the benefit of scanning less short codes.

\paragraph{Multi-index (Multi-D-ADC)~\cite{Babenko2015}} Multi-D-ADC overcomes this issue. More specifically, Multi-D-ADC makes it possible to build fine indexes, \eg with more than $K=2^{16}$ inverted lists, without making the retrieval of inverted lists costly during query processing. Multi-D-ADC therefore offers faster response times than IVFADC by scanning an even smaller fraction of the dataset. Like IVFADC, Multi-D-ADC uses two quantizers: one for the inverted index, and one to encode vectors as short codes. However, instead of using a vector quantizer $q_i$ for the index, Multi-D-ADC uses a product quantizer for the index. Multi-D-ADC thus uses two product quantizers: (1) an index product quantizer $\operatorname{pq_i}$ to partition the database, (2) a product quantizer $\operatorname{pq}$ to encode high-dimensional vector as short codes. These two product quantizers typically have different parameters. The product quantizer $\operatorname{pq}$ used for vector encoding typically has $m=4$ to $m=16$ sub-quantizers (with $k=2^{b}$ centroids each, $b=4$ to $b=16$). By contrast, in~\cite{Babenko2015}, the authors show that the index product quantizer should have only $m=2$ sub-quantizers ($K \approx 2^{14}$ centroids each). The $m=2$ sub-quantizers, $\operatorname{q_1}$ and $\operatorname{q_2}$, of the index product quantizer have two distinct codebooks $\mathcal{C}_1 = \{c^1_i\}_{i=0}^K$ and $\mathcal{C}_2 = \{c^2_i\}_{i=0}^K$. The index product quantizer $\operatorname{pq_i}$ partitions the vector space in $K^2$ cells (\eg $K^2=2^{28}$ cells for $K=2^{14}$). Each cell is defined by the concatenation of two centroids ($c^1_i$,$c^2_j$) $i \in \{0,\dotsc,k-1\}$, $j \in \{0,\dotsc,k-1\}$, and the corresponding inverted list is identified by the $(i,j)$ tuple. Like IVFADC, Multi-D-ADC allows encoding the residuals of high-dimensional vectors instead of the original vectors, as follows: $r(x) = x - (c^1_i,c^2_j)$. Computing the distance between the query vector and all  ($c^1_i$,$c^2_j$) $i \in \{0,\dotsc,K-1\}$, $j \in \{0,\dotsc,K-1\}$ to retrieve the most relevant inverted lists would require $K^2$ distance computations. Multi-D-ADC would therefore bring no benefit over IVFADC and using more than $K^2=2^{16}$ centroids would still be impractical. Instead, the multi-sequence algorithm, introduced in~\cite{Babenko2015} takes advantage of the properties of the index product quantizer $pq_i$ to retrieve relevant inverted lists while requiring $2K$ distance computations only. The multi-sequence algorithm first computes the distance of the first half of the query vector $y$, denoted $y^1$ to all centroids of $\mathcal{C}_1$ and the distance of the query vector $y^2$ to all centroids of $\mathcal{C}_2$. The codebooks $\mathcal{C}_1$ copied and sorted by increasing distance from $y^1$ and $y^2$ respectively. The multi-sequence iterates over the two codebooks by increasing $\operatorname{d}(y^1, c_i^1) + \operatorname{d}(y^2, c_j^2)$ distance, where $\operatorname{d}(y^1, c_i^1)$ is the distance from $y^1$ to the $i$-th centroid of $\mathcal{C}_2$ (respectively for $\operatorname{d}(y^2, c_j^2)$). The inverted lists $(i,j)$ are then marked for scanning, until enough vectors have been gathered.

\paragraph{}
Using a Multi-D-ADC with $K=2^{14}$ (thus $K=2^{28}$ cells in total) and scanning $1/50000^{th}$ of the database gives the same accuracy as using an IVFADC with $K=2^{16}$ cells and scanning $1/1024^{th}$ of the database. Even if the multi-sequence algorithm is more efficient than computing the distance to all ($c^1_i$,$c^2_j$) $i \in \{0,\dotsc,k-1\}$, $j \in \{0,\dotsc,k-1\}$, its cost is not negligible. Consequently, if Multi-D-ADC offers a decrease in response time over IVFADC, this decrease is not proportional to the decrease in code scanned \ie Multi-D-ADC is not 50 times faster in the aforementioned case even if 50 times less short codes are scanned. 

\paragraph{Other approaches} Other approaches have been proposed to partition the database into inverted lists. In~\cite{Babenko2015}, the authors explore the idea of using KD-trees to partition the database. They however show that Multi-D-ADC offers better performance in all cases. The use of a variant of LSH to build the inverted index has also been considered~\cite{Xia2013}. The major flaw of this approach is that it causes a four times increase in memory use, partially offsetting the benefit of product quantization. For this reason, this approach is not very popular.

\subsection{ANN Search}
\label{sec:annsearch}

\paragraph{}
ANN search in a database of vectors encoded as short codes takes three steps: \emph{Index}, which involves retrieving the most relevant inverted lists from the index; \emph{Tables}, which involves computing lookup tables used for fast distance computation and \emph{Scan}, which involves computing the distance between the query vector and short using the previously computed lookup tables. Obviously, the step \emph{Index} is only required for non-exhaustive search, and is skipped in the case of exhaustive search (Algorithm~\ref{alg:ann}). We detail these three steps in the following paragraphs.

\paragraph{Index} In this step, the most relevant lists to the query vector are retrieved from the inverted index. The exact procedure depends on the type of inverted index used; we presented the most common ones in Section~\ref{sec:invindex}. Therefore the implementation of the \textsc{index\_get\_list} function depends on the type of inverted index used (Algorithm~\ref{alg:ann}). If needed, the residual $r(y)$ of the query vector $y$ is also computed. In practice, multiple inverted lists are retrieved from the inverted index. However, for the sake of simplicity, this section describes the search process for a single inverted list. For $\textit{ma}$ inverted lists, each operation should be repeated $\textit{ma}$ times: $\textit{ma}$ inverted lists are retrieved from the inverted index, $\textit{ma}$ residuals are computed, and $\textit{ma}$ inverted lists are scanned. In the remainder of this section, $y'=r(y)$ if an inverted index is used. Otherwise, $y'=y$.

\paragraph{Tables} In this step, a set of $m$ lookup tables are computed $\{D^j\}_{j=0}^m$, where $m$ is the number of sub-quantizers of the product quantizer. The $j$th lookup table $D_j$ is composed of the distances between the $j$th sub-vector of $y'$ and all centroids of the $j$th sub-quantizer:
\begin{equation}
D^j = \left( \left \lVert {y'}^j - \mathcal{C}^j[0] \right\rVert^2,\dots, \left \lVert {y'}^j - \mathcal{C}^j[k-1] \right \rVert^2\right) \label{eqn:tables}
\end{equation}
We do not detail the \textsc{compute\_tables} function in Algorithm~\ref{alg:ann}, but it would correspond to an implementation of Equation~\ref{eqn:tables}.

\begin{algorithm}[t]
  \caption{ANN Search}\label{alg:ann}
  \begin{algorithmic}[1]
    \Function{nns}{$\{\mathcal{C}^j\}_{j=0}^{m}, \mathit{database}, y$}
    \State $\mathit{list}, y' \gets $\Call{index\_get\_list}{$\mathit{database}, y$} \Comment{Index Step}
    \State $\{D^j\}_{j=0}^{m} \gets$ \Call{compute\_tables}{$y', \{\mathcal{C}^j\}_{j=0}^{m}$} \Comment{Tables Step}
    \State \Return{\Call{scan}{$\mathit{list}, \{D^j\}_{j=0}^{m}$}} \Comment{Scan Step}
    \EndFunction
    
    \Function{scan}{$\mathit{list}, \{D^j\}_{j=0}^{m}$}
    \State $\mathit{neighbors} \gets \operatorname{binheap}(r)$\Comment{binary heap of size $r$} \label{alg:line:binheap}
    \For{$i \gets 0$ to $|\mathit{list}|-1$}  \label{alg:line:iter}
    \State $c \gets \mathit{list}[i]$\Comment{$i$th code}
    \State $d \gets$ \Call{adc}{$p, \{D^j\}_{j=0}^{m}$} \label{alg:line:dist}
    \State $\mathit{neighbors}.\operatorname{add}((i, d))$ \label{alg:line:add}
    \EndFor
    \State \Return{$\mathit{neighbors}$}
    \EndFunction
    
    \Function{adc}{$c, \{D^j\}_{j=0}^{m-1}$} \label{alg:line:adc}
    \State $d \gets 0$
    \For{$j \gets 0$ to $m$} \label{alg:line:loop}
    \State $\mathit{index} \gets c[j]$ \label{alg:line:index}
    \State $d \gets d + D^j[index]$ \label{alg:line:ops}
    \EndFor
    \State \Return{$d$}
    \EndFunction
  \end{algorithmic}
\end{algorithm}

\begin{figure}
  \centering
  \input{sota/fig/adc.tex}
  \caption{Asymmetric Distance Computation (ADC)\label{fig:adcexplain}}
\end{figure}
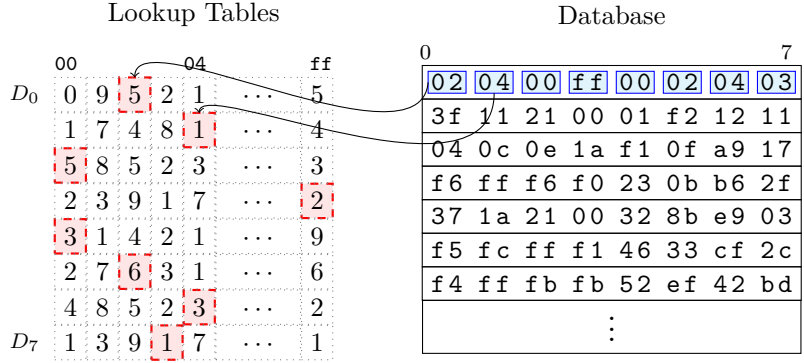

\paragraph{Scan} In this step, the inverted list retrieved from the inverted index is scanned for nearest neighbors (\textsc{scan} function, Algorithm~\ref{alg:ann}). This requires computing the distance between the query vector $y'$ and all short codes in the inverted list (Algorithm~\ref{alg:ann}, line~\ref{alg:line:iter}). Squared Euclidean distances are used for ANN search instead of Euclidean distances because Squared Euclidean distances avoid a costly square root computation and preserve the order. Asymmetric Distance Computation (ADC) uses the lookup tables generated in the previous step to compute the distance between the query vector $y'$ and a short code $c$, as follows:
\begin{equation}
\operatorname{adc}(y',c) = \sum_{j=0}^{m-1} D^{j}[c[j]] \label{eqn:adc1}
\end{equation}
The \textsc{adc} function is a implementation of Equation~\ref{eqn:adc1}. Figure~\ref{fig:adcexplain} also illustrates ADC procedure for short codes codes generated by a $8{\stimes}8$ product quantizer ($m=8$, $b=8$). Each centroids index $c[j]$ constituting the short code $c$ (\texttt{02}, \texttt{04}, \etc) is used to look up a value in the $j$th lookup table ($D_0$, $D_1$, \etc). All looked up values (in red) are then summed to obtain the asymmetric distance.

\paragraph{}
Substituting Equation~\ref{eqn:tables} into Equation~\ref{eqn:adc1}, we have:
\begin{equation}
\operatorname{adc}(y',c) = \sum_{j=0}^{m-1} \left \lVert {y'}^j - \mathcal{C}^j[c[j]] \right \rVert^2 \label{eqn:adc2}
\end{equation}
In Equation~\ref{eqn:adc2}, $\left \lVert {y'}^j - \mathcal{C}^j[c[j]] \right \rVert^2$ is the squared distance between the $j$th sub-vector of $y'$ and the $j$th centroid referenced by code $c$. ADC therefore sums the distances of the sub-vectors of $y'$ to the centroids referenced by the short code $c$ in the $m$ sub-spaces of the product quantizer. Because it splits high-dimensional vectors into $m$ sub-vectors, a product quantizer generates $m$ \emph{orthogonal} sub-spaces. This makes it possible to compute partial distances in each sub-space and to sum them. When the number of codes in cells is large compared to $k$, the number of centroids of sub-quantizers, using lookup tables avoids computing $\lVert {y'}^j - \mathcal{C}^j[i] \rVert^2$ for the same $i$ multiple times. Lookup tables therefore provide a significant speedup.

\paragraph{}
During the scan of an inverted list, nearest neighbors are stored in a binary heap of size $r$ (Algorithm~\ref{alg:ann}, line~\ref{alg:line:binheap}), where $r$ is the number of requested nearest neighbors. Nearest neighbors are stored in the binary heap as $(i,d)$ pairs, where $i$ in the index of the short code and $d$ is the distance to the query vector. The binary heap holds the $r$ pairs with the lowest $d$. If a new pair is added (Algorithm~\ref{alg:ann}, line~\ref{alg:line:add}) and the binary heap is not full, the pair is added. If a new pair is added and the binary heap is full, the pair is added only if its distance $d$ is lower than the distance of the pair with the highest distance in the binary heap. The pair with the largest distance is removed from the heap, and the new pair is added. Adding a pair to the binary heap takes $\operatorname{O}(\log r)$ operations. Checking if a pair has a lower distance than the pair with the highest distance takes $\operatorname{O}(1)$ operations.

\subsection{Product Quantization Derivatives}
\label{sec:pq-derivatives}

\paragraph{Optimized Product Quantization (OPQ)~\cite{Ge2014}}\footnote{Two research teams concurrently discovered a similar generalization of product quantization. One team named this generalization "Optimized Product Quantization"~\cite{Ge2013}, while the other used the term "Cartesian K-Means"~\cite{Norouzi2013}, and both terms are commonly used in the literature. Both papers were originally published at CVPR'13. In this thesis, we refer to this generalization of product quantization as optimized product quantization.} A product quantizer quantizes a $d$-dimensional vector by splitting it into $m$ sub-vectors, and quantizing each sub-vector independently (Section~\ref{sec:vectorencoding}). The $i$th sub-vector comprises the dimensions $i\cdot d/m$ to $(i+1)\cdot d/m$. In this manner, a product quantizer partitions the $d$-dimensional vector spaces into $m$ orthogonal sub-spaces, each sub-space comprising a subset of the dimensions the original vector space. 
Intuitively, we see that this sub-space decomposition may not be optimal, for instance if vectors have different variances across their dimensions. Some sub-spaces may then carry more information than others. Moreover, this sub-space decomposition prevents taking advantage of correlations between dimensions that are not in the same sub-space to decrease quantization error.

\paragraph{}
Optimized product quantization overcomes these issues by multiplying a vector $x \in \mathbb{R}^d$ by an orthogonal matrix $R \in \mathbb{R}^{d \times d}$ before quantizing them with a product quantizer. An optimized product quantizer $\operatorname{opq}$ thus maps a vector $x \in \mathbb{R}^d$ as follows:
\begin{gather*}
\operatorname{opq}(x) = \operatorname{pq}(Rx),\\
\text{such that } R^{T}R = I \text{ and pq is a product quantizer}. 
\end{gather*}
An orthogonal matrix $R$ can represent any permutation of dimensions. Thus, the matrix $R$ determines the decompositions of the $d$-dimensional space in $m$ sub-spaces of equal dimensionality. The matrix $R$ is learned jointly with the codebooks $\{C_j\}_{j=0}^{m-1}$ to minimize quantization error. In~\cite{Ge2014}, the authors train $R$ and $\{C_j\}_{j=0}^{m-1}$ by alternating between two steps (A) and (B) to achieve a locally optimal solution. Step (A) involves fixing $R$ and optimizing $\{C_j\}_{j=0}^{m-1}$, by running a single k-means step in each sub-space. Step (B), involves fixing $\{C_j\}_{j=0}^{m-1}$ and optimizing $R$. For fixed codebooks $\{C_j\}_{j=0}^{m-1}$, $R$ has a closed-form optimal solution obtained through Singular Value Decomposition (SVD). In most cases, the algorithm converges after less than 100 iterations, where one iteration involves running Step (A) followed by Step (B).

\paragraph{}
Optimized product quantizers can be used to encode vectors as short codes, exactly like product quantizers (Section~\ref{sec:vectorencoding}). Like product quantizers, optimized product quantizers divide the vector space in sub-spaces that are orthogonal. Therefore, ADC (Section~\ref{sec:annsearch}) can be used to compute the distance between an high-dimensional vector $y$ and short codes generated by an optimized product quantizer. If short codes were generated by an optimized product quantizer, it is necessary to multiply $y$ by the matrix $R$ before computing tables. Distances to short codes can then be computed by performing $m$ table lookups. Like product quantizers, optimized product quantizers can be combined with inverted indexes. Like IVFADC (Section~\ref{sec:invindex}), IVFOADC~\cite{Babenko2015} uses a vector quantizer $q_i$ to build an inverted index, but uses an optimized product quantizer en encode vectors instead of a product quantizer. Multi-D-ADC uses two product quantizers, one to build the inverted and the other to encode vectors as short codes. OMulti-D-O-ADC replaces both product quantizers by optimized product quantizers~\cite{Babenko2015}.

\paragraph{Additive Quantization (AQ)~\cite{Babenko2014AQ}} Both product quantization and optimized product quantization represent a $d$-dimensional vector $x$ as a \emph{concatenation} of $d/m$-dimensional centroids \ie $\operatorname{pq}(x) = (\mathcal{C}^0[i_0],\dotsc,\mathcal{C}^{m-1}[i_{m-1}])$. On the contrary, additive quantization represents a $d$-dimensional vector $x$ as a sum of $d$-dimensional centroids:
\[
\operatorname{aq}(x) = \mathcal{C}^0[i_0] + \dotsc + \mathcal{C}^{m-1}[i_{m-1}]
\]

Like product quantization, additive quantization can be used to encode vectors as short codes of $m\cdot\log_2(k)$ bits by concatenating centroids indexes, \ie $\operatorname{enc}(x) = (i_0,\dotsc,i_m)$.
Additive quantization therefore uses more information than product quantization to represent a high-dimensional vector: $m$ $d$-dimensional centroids for additive quantization versus $m$ $d/m$-dimensional centroids for product quantization. This allows additive quantization to offer a better accuracy for nearest neighbor search than product quantization. However, additive quantization suffers from two major disadvantages. First, finding the best combination of centroids to sum from the codebooks $\{\mathcal{C}\}_{j=0}^{m-1}$ in order to encode a vector $x$ is a hard problem. In~\cite{Babenko2014AQ}, the authors propose an approximate algorithm based on Beam search but this solution remains multiple orders of magnitude more costly than the encoding process of product quantization. Additive quantization is therefore intractable for billion-size datasets. Second, computing the distance of the query vector to a short code $c$ requires $m^2/2$ table lookups and $m^2/2$ additions, while a distance computation requires $m$ tables lookups and $m$ additions with product quantization. In most cases, nearest neighbor search is more than 3 times slower with additive quantization than with product quantization.

\paragraph{Tree Quantization (TQ)~\cite{Babenko2015TQ}}
Tree quantization is similar to additive quantization in the sense that it represents a high-dimensional vector $x$ by a sum of $m$ $d$-dimensional centroids, $\operatorname{tq}(x) = \mathcal{C}^0[i_0] + \dotsc + \mathcal{C}^{m-1}[i_{m-1}]$. However, it imposes a special tree structure on the codebooks. This tree structure makes the encoding process faster, making tree quantization tractable for larger datasets. Tree quantization encoding however remains much slower than product quantization encoding. In addition, with tree quantization, computing the distance of the query vector to a short code $c$ requires $2m$ table lookups and $2m$ additions. Nearest neigbor search is about 2-2.5 times slower with tree quantization than with product quantization. Tree quantization offers a similar accuracy to additive quantization for nearest neighbor search, and therefore a higher accuracy than product quantization.

\paragraph{Composite Quantization (CQ)~\cite{Zhang2014}}
Like additive quantization and tree quantization, composite quantization represents a high-dimensional vector $x$ by a sum of $m$ $d$-dimensional centroids, $\operatorname{cq}(x) = \mathcal{C}^0[i_0] + \dotsc + \mathcal{C}^{m-1}[i_{m-1}]$. Additive quantization and tree quantization do not enforce any orthogonality constraint on the centroids of the codebooks \ie $\mathcal{C}^{j'}[i_{j'}]\cdot\mathcal{C}^{j}[i_j]$ can be any value $\forall j' \in \{0,\dotsc,m-1\}, \forall j \in \{0,\dotsc,m-1\}, j' \neq j, \forall i_j \in \{0,\dotsc,k-1\}$. Product quantization and optimized product quantization represent a vector $x$ by a concatenation of $m$ centroids, which implies that centroids in different codebooks are orthogonal, $\mathcal{C}^{j'}[i_{j'}]\cdot\mathcal{C}^{j}[i_j]=0$ if $j'\neq j$. The distinctive feature of composite quantization is that it allows some degree of non-orthogonality between centroids, but it enforces that the sum of the pairwise dot products of centroids is a constant for any code $c$, \ie $\sum_{j'=0}^{m-1}\sum_{j=0}^{m-1}\mathcal{C}^{j'}[i_{j'}]\cdot\mathcal{C}^{j}[i_j]=\epsilon$ for $j' \neq j, \forall c = (i_0,\dotsc,i_{m-1})$. Since the sum of pairwise centroid dot products is constant for all codes, it can be be omitted at search time. Therefore computing the distance of the query to a short code requires $m$ table lookups and $m$ additions. Nearest neighbor search with composite quantization is as fast as with product quantization, and therefore much faster than with additive quantization or tree quantization. Composite quantization has be shown to offer a better accuracy than product quantization, but its accuracy has not been compared to additive quantization or tree quantization.

\stopcontents[chapters]

%% file: sota/fig/invertedindex.tex
\newlength\idxCellWidth
\setlength{\idxCellWidth}{\widthof{7}}

\tikzset{
  inverted index cell/.style = {
    rectangle,
    draw,
    text width=\idxCellWidth,
    minimum height=3.7ex,
    align=center,
  },
  inverted index/.style = {
    matrix of math nodes,
    row sep = -\pgflinewidth,
    column sep = -\pgflinewidth,
    nodes={
      inverted index cell
    },
  }
}

\makeatletter
\DeclareRobustCommand{\rvdots}{%
  \vbox{
    \baselineskip4\p@\lineskiplimit\z@
    \kern-\p@
    \hbox{.}\hbox{.}\hbox{.}
  }}
\makeatother

\tikzstyle{list} = [
  rectangle split,
  rectangle split horizontal,
  draw
]

\tikzstyle{slot} = [
  align=center,
  text width=3em
]

\newcommand{\bucketSep}{2.5em}

\begin{tikzpicture}

\node [inverted index] (index)
{%
  0\\
  1\\
  2\\
  3\\
  \rvdots\\
  7\\
};

\node[list,
  rectangle split parts=3,
  right=\bucketSep of index-1-1.east,
  anchor=west] (list0) {
  \nodepart[slot]{one}$v_{5}$
  \nodepart[slot]{two}$v_{9}$
  \nodepart[slot]{three}$v_{13}$
};
\draw[->] (index-1-1.east) -- (list0.west);

\node[list,
  rectangle split parts=2,
  right=\bucketSep of index-2-1.east,
  anchor=west] (list1) {
  \nodepart[slot]{one}$v_{4}$
  \nodepart[slot]{two}$v_{7}$
};
\draw[->] (index-2-1.east) -- (list1.west);

\node[list,
  rectangle split parts=4,
  right=\bucketSep of index-3-1.east,
  anchor=west] (list2) {
  \nodepart[slot]{one}$v_{0}$
  \nodepart[slot]{two}$v_{3}$
  \nodepart[slot]{three}$v_{12}$
  \nodepart[slot]{four}\color{blue}$v_{14}$
};
\draw[->] (index-3-1.east) -- (list2.west);

\node[list,
  rectangle split parts=1,
  right=\bucketSep of index-4-1.east,
  anchor=west] (list3) {
  \nodepart[slot]{one}$v_{1}$
};
\draw[->] (index-4-1.east) -- (list3.west);


\node[list,
  rectangle split parts=2,
  right=\bucketSep of index-6-1.east,
  anchor=west] (list7) {
  \nodepart[slot]{one}$v_{8}$
  \nodepart[slot]{two}$v_{11}$
};
\draw[->] (index-6-1.east) -- (list7.west);

\draw [decoration={brace,raise=5pt},
  decorate] (index-1-1.north west) --  node (indexlbl) [ above=10pt,align=center, text depth=0.25ex] {\emph{Inverted index}\\(partition IDs)} (index-1-1.north east);

\draw [decoration={brace,raise=5pt},
  decorate] (list0.north west) --  node (listlbl) [above=10pt, align=center, text depth=0.25ex] {\emph{Inverted lists} (vector IDs)} (list2.north east |- list0.north east);

\node (boundbox) [
  fit=(indexlbl) (listlbl) (list2) (index)] {};

\begin{axis} [
  xmin=0,
  xmax=1,
  ymin=0,
  ymax=1,
  axis equal image,
  hide axis,
  width=7 cm,
  name=voronoi,
  at={($(boundbox.west)-(0.2cm,0cm)$)},
  anchor=east,
]

\addplot [
  only marks,
  mark=x,
  visualization depends on={\thisrow{label} \as \pointLabel},
  nodes near coords={\pgfmathprintnumber[int trunc]{\pointLabel}},
] table {sota/fig/quantizer_centroids.dat};

\addplot [
  no markers,
  update limits=false,
] table {sota/fig/quantizer_voronoi.dat};

\addplot [
  only marks,
  mark=x,
  color=blue,
  point meta=explicit symbolic,
  nodes near coords
] coordinates {
  (0.42,0.68) [$v_{14}$]
};

\end{axis}

\end{tikzpicture}
\caption{ANN search system based on an inverted index\label{fig:invindex}}

%% file: sota/fig/kdtree.tex
\begin{tikzpicture}
\begin{axis} [
  xmin=0,
  xmax=1,
  ymin=0,
  ymax=1,
  width=7 cm,
  axis equal image,
  hide axis
]

\addplot [no markers] coordinates {
  (0.56, 0)
  (0.56, 1)
};

\addplot [no markers, dashed] coordinates {
  (0.56, 0.73)
  (1, 0.73)

  (0, 0.48)
  (0.56, 0.48)
};

\addplot [no markers, dotted] coordinates {
  (0.19, 0)
  (0.19, 0.48)

  (0.35, 0.48)
  (0.35, 1)

  (0.80, 0)
  (0.80, 0.73)

  (0.77, 0.73)
  (0.77, 1)
};

\addplot [
  only marks,
  mark=x,
  point meta=explicit symbolic,
  nodes near coords
] coordinates {
  (0.4,0.85) [$v_1$]
  (0.62,0.76) [$v_2$]
};

\addplot [
  only marks,
  mark=x,
  point meta=explicit symbolic,
  nodes near coords,
  color=black!60
] coordinates {
  (0.53,0.60) [$q$]
};

\draw[black!50] (axis cs:0.53,0.60) circle [radius=0.282];

\end{axis}
\end{tikzpicture}

%% file: sota/fig/voctree.tex
\begin{tikzpicture}
\begin{axis} [
  xmin=0,
  xmax=1,
  ymin=0,
  ymax=1,
  axis equal image,
  width=7 cm,
  hide axis
]


\addplot [
  no markers,
  update limits=false,
] table {sota/fig/root_voronoi.dat};

\addplot [
  no markers,
  update limits=false,
  dotted
] table {sota/fig/child0_voronoi.dat};

\addplot [
  no markers,
  update limits=false,
  dotted
] table {sota/fig/child1_voronoi.dat};

\addplot [
  no markers,
  update limits=false,
  dotted
] table {sota/fig/child2_voronoi.dat};

\addplot [
  no markers,
  update limits=false,
  dotted
] table {sota/fig/child3_voronoi.dat};

\addplot [
  only marks,
  mark=x,
  point meta=explicit symbolic,
  nodes near coords
] coordinates {
  (0.50,0.65) [$v_1$]
  (0.82,0.70) [$v_2$]
};

\addplot [
  only marks,
  mark=x,
  point meta=explicit symbolic,
  nodes near coords,
  color=black!60
] coordinates {
  (0.73,0.62) [$q$]
};

\draw[black!50] (axis cs:0.73,0.62) circle [radius=0.231];

%
\end{axis}

\end{tikzpicture}

%% file: sota/fig/lsh.tex
\tikzset{cross/.style={cross out, draw=black, minimum   size=2*(#1-\pgflinewidth), inner sep=0pt, outer sep=0pt},
cross/.default={1pt}}

\begin{tikzpicture}

\draw[->] (0,0) -- (0,4cm);
\draw[->] (0,0) -- (4cm,0);

\draw (0,0) -- (2.7cm,3 cm) coordinate (endpoint) 
  coordinate [pos=0.3] (bin1)
  coordinate [pos=0.6] (bin2)
  coordinate [pos=0.69] (shifted)
  coordinate [pos=0.9] (bin3)
  coordinate [pos=0.15] (label1)
  coordinate [pos=0.25] (veca)
  coordinate [pos=0.45] (label2)
  coordinate [pos=0.75] (label3);

\draw (0.6cm,2.35cm) node (point) [cross=2pt] {};

\coordinate (projection) at ($(0,0)!(point)!(endpoint)$);
\draw[dotted] (point) -- (projection);
\node [below=0cm of point.center, anchor=south east] {$v$};

\draw (bin1) circle[radius=1pt];
\draw (bin2) circle[radius=1pt];
\draw (bin3) circle[radius=1pt];

\draw [->, thick] (0,0) -- (veca) node [above left=0.4pt, midway, inner sep=0] {$a$};


\draw (shifted) node [cross=2pt] {};

\draw [decoration={brace,raise=5pt,mirror},
  decorate] (bin1) --  node [below right=5pt] {$w$} (bin2);

\draw [decoration={brace,raise=5pt,},
  decorate] (projection) --  node [above left=5pt] {$bw$} (shifted);

\end{tikzpicture}

%% file: sota/fig/adc.tex
\begin{tikzpicture}[x=20pt, y=20pt, node distance=1pt,outer sep = 0pt]

\begin{scope}[xshift=-1cm]
\node(memtitle) {Database};
\node[draw=none,fill=none,below=0.2cm of memtitle](n){};
\pqvector{a}{n}(0,2  0,4  0,0  f,f  0,0  0,2  0,4  0,3)
\pqvector{b}{a}(3,f  1,1  2,1  0,0  0,1  f,2  1,2  1,1)
\pqvector{c}{b}(0,4  0,c  0,e  1,a  f,1  0,f  a,9  1,7)
\pqvector{d}{c}(f,6  f,f  f,6  f,0  2,3  0,b  b,6  2,f)
\pqvector{e}{d}(3,7  1,a  2,1  0,0  3,2  8,b  e,9  0,3)
\pqvector{f}{e}(f,5  f,c  f,f  f,1  4,6  3,3  c,f  2,c)
\pqvector{g}{f}(f,4  f,f  f,b  f,b  5,2  e,f  4,2  b,d)
\dotsbox{m}{g};
\end{scope}

\def\phantomcell{\phantom{0}\cdots\phantom{0}}

\begin{scope}[xshift=-6.5cm]
\node(title){Lookup Tables};
\node[draw=none,fill=none,below=0.2cm of title](nb){};
\matrix[below = 0cm of nb](dist) [matrix of math nodes, nodes={draw,dotted, color=gray, text=black}]{
0 & 9 & 5 & 2 & 1 & \phantomcell & 5 \\
1 & 7 & 4 & 8 & 1 & \phantomcell & 4 \\
5 & 8 & 5 & 2 & 3 & \phantomcell & 3 \\
2 & 3 & 9 & 1 & 7 & \phantomcell & 2 \\
3 & 1 & 4 & 2 & 1 & \phantomcell & 9 \\
2 & 7 & 6 & 3 & 1 & \phantomcell & 6 \\
4 & 8 & 5 & 2 & 3 & \phantomcell & 2 \\
1 & 3 & 9 & 1 & 7 & \phantomcell & 1 \\};
\end{scope}

\footnotesize
\node[above = 0cm of dist-1-1.north west](lab1){\texttt{\phantom{00}00}};
\node[above = 0cm of dist-1-5.north west](lab1){\texttt{\phantom{00}04}};
\node[above = 0cm of dist-1-7.north east](lab1){\texttt{ff\phantom{00}}};

\node[left = 0.10cm of dist-1-1.west](lab1){$D_0$};
\node[left = 0.10cm of dist-8-1.west](lab1){$D_7$};

\node[above = 0.10cm of a-1-1.north west](lab1){0\phantom{0}};
\node[above = 0.10cm of a-1-16.north east](lab1){7\phantom{0}};

\normalsize

\begin{scope}[on layer=back]
\node[select_dist,fit=(dist-1-3)](d1) {};
\node[select_dist,fit=(dist-2-5)](d2) {};
\node[select_dist,fit=(dist-3-1)](d3) {};
\node[select_dist,fit=(dist-4-7)](d4) {};
\node[select_dist,fit=(dist-5-1)](d5) {};
\node[select_dist,fit=(dist-6-3)](d6) {};
\node[select_dist,fit=(dist-7-5)](d7) {};
\node[select_dist,fit=(dist-8-4)](d8) {};

\node[select_mem,fit=(a-1-1.north west) (a-1-2.south east)](v1b) {};
\node[select_mem,fit=(a-1-3.north west) (a-1-4.south east)](v2b) {};
\node[select_mem,fit=(a-1-5.north west) (a-1-6.south east)](v3b) {};
\node[select_mem,fit=(a-1-7.north west) (a-1-8.south east)](v4b) {};
\node[select_mem,fit=(a-1-9.north west) (a-1-10.south east)](v5b) {};
\node[select_mem,fit=(a-1-11.north west) (a-1-12.south east)](v6b) {};
\node[select_mem,fit=(a-1-13.north west) (a-1-14.south east)](v7b) {};
\node[select_mem,fit=(a-1-15.north west) (a-1-16.south east)](v8b) {};
\end{scope}

\path[->, draw, solid] (v1b.west) ..controls (-5,-3) and +(up:8mm)..(d1.north);
\path[->, draw, solid] (v2b.south) ..controls +(down:16mm) and +(up:5mm)..(d2.north);

\end{tikzpicture}

%% file: perf/perf.tex
\label{sec:perf}

\startcontents[chapters]
\ruledprintcontents

\paragraph{}
All the contributions of this thesis focus on increasing the speed of the \emph{Scan} step of ANN search (Section~\ref{sec:annsearch}). This step consists in computing the asymmetric distance between the query vector $y$ and all short codes in the previously selected inverted list(s). Among the different steps of ANN search (Index, Tables, Scan), the Scan step generally takes the most time. ANN search is trivial to parallelize at the query level by executing different ANN queries on the different cores of a multi-core CPU. The Scan step itself is also trivial to parallelize by scanning a different chunk of the inverted list on each core. Therefore, we focus on the more challenging task of improving the \emph{single-core} performance of the Scan step. In this section, we analyze the factors that limit the performance of the Scan step, and we show that memory accesses are the primary bottlenecks. We also show that the structure of the algorithm prevents an efficient SIMD implementation.

\section{Impact of Memory Accesses}
\label{sec:memory-accesses}
\paragraph{Breakdown of operations} From Algorithm \ref{alg:ann} (Section~\ref{sec:annsearch}) we observe that each Asymmetric Distance Computation (ADC) requires the following operations:
\begin{itemize}
  \item $m$ memory accesses to load centroid indexes $c[j]$ (Algorithm~\ref{alg:ann}, line~\ref{alg:line:index}) [mem1]
  \item $m$ memory accesses to load $D^j[\mathit{index}]$ values from lookup tables (Algorithm~\ref{alg:ann}, line~\ref{alg:line:ops}) [mem2]
  \item $m$ additions (Algorithm~\ref{alg:ann}, line~\ref{alg:line:ops})
\end{itemize}
These operations are the \emph{only} instructions executed by the CPU to perform an ADC. Since the parameter $m$ is a small constant (typical $m=4-32$) known at compile time, compilers are able to unroll the loop (Algorithm~\ref{alg:ann}, line~\ref{alg:line:loop}). The CPU therefore does not have to maintain the loop index and execute jump instructions at runtime. It is crucial for performance that this loop is unrolled: in our experiments, when the loop is not unrolled, the scan is 3 times slower (for typical parameters: $m=8, b=8$). In addition, as the function \textsc{adc} is small and called repeatedly (Algorithm~\ref{alg:ann}, line~\ref{alg:line:dist}), so compilers are able to inline the function call. Therefore, there is no function call overhead at runtime. In this thesis, we take this implementation (Algorithm~\ref{alg:ann}) with compiler optimizations enabled (\ie loop unrolling, function call inlining) as a baseline and seek to improve its performance.

\begin{table}
  \centering
  \caption{Properties of different types of memory (Nehalem-Haswell)\label{tbl:memprop}}
  \begin{tabular}{lll}
    \toprule
    Mem. Type & Size & Latency (CPU cycles)\\
    \midrule
    L1 cache & 32 KiB & 4-5\\
    L2 cache & 256 KiB & 11-13\\
    L3 cache & 2-3 MiB $\times \mathrm{core\_count}$ & 25-40\\
    RAM & 4-512 GiB & 100-300\\
    \bottomrule
  \end{tabular}
\end{table}

\paragraph{Cost of memory accesses} When evaluating the complexity of an algorithm, memory accesses are often overlooked. In the original paper on product quantization~\cite{Jegou2011}, the authors report that "only m additions are required per distance calculation", forgetting about memory accesses. However, memory accesses are costly operations on a CPU. While an additions does not take more than 1 CPU cycle, memory accesses take 4 CPU cycles in the best case. Table~\ref{tbl:memprop} details the latency of memory accesses depending on the cache level, as well as the sizes of the different caches. These figures are valid for Intel CPU architectures from Nehalem (released in 2008) to Haswell (released in 2013). In the list of operations of the previous paragraph, we distinguished between to classes of memory accesses: \emph{mem1} and \emph{mem2}, as they may hit different cache levels. \emph{Mem1} accesses always hit the L1 cache thanks to the \emph{hardware prefetchers} included in CPUs. Hardware prefetchers are able to detect sequential memory accesses, and prefetch data to the L1 cache. We access $c[j]$ values sequentially. We first access $c[0]$, where $c$ is the first code in the database, then $c[1]$ until $c[m-1]$. Next we perform the same accesses on the second code in the database, until we have iterated on all codes in the inverted list. On the contrary, \emph{mem2} accesses may hit the L1 or L3 depending on the $b$ parameter of the product quantizer.

\paragraph{Impact of $m$ and $b$ parameters} In Section~\ref{sec:vectorencoding}, we mentioned that the $m$ and $b$ parameters of a product quantizer impact: (1) the memory use of the database, (2) the accuracy of nearest neighbor search and (3) the speed of nearest neighbor search. An $m{\stimes}b$ product quantizer has a total of $2^{m{\stimes}b}$ centroids, and encodes each high-dimensional vector into a code of $m{\stimes}b$ bits. The higher the $m{\stimes}b$ product, the higher the number of centroids of the product quantizer, and therefore the higher the accuracy of nearest neighbor search. In Table~\ref{tbl:tblsize}, we use Recall@100 as accuracy measure. However, the higher the $m{\stimes}b$ product, the larger the codes, and therefore the higher the memory use of the database. In practice, codes of $m{\stimes}b=64$ bits ($2^{64}$ centroids) or $m{\stimes}b=128$ bits ($2^{128}$ centroids) are used. There has been a recent interest for codes of $m{\stimes}b=32$ bits to achieve extreme compression ratios, although they offer a somewhat lower accuracy (Table~\ref{tbl:tblsize}). For a database of 1 million of 128-dimensional SIFT vectors (SIFT1M), 64-bit codes offer a satisfactory accuracy (Table~\ref{tbl:tblsize}), but 128-bit codes are necessary for larger databases or vectors of higher dimensionality.

\begin{table}
  \centering
  \caption{ANN search speed and accuracy (SIFT1M)\label{tbl:tblsize}}
  \begin{tabular}{llllll}
    \toprule
    $m \stimes b$ & Tables size & Cache & Recall@100 & Tables time & Scan time\\
    \midrule
    \multicolumn{6}{c}{32-bit codes}\\
    \midrule
    $4 \stimes 8$ & 4 KiB  & L1 & 59.0\% & 0.004 ms & 1.3 ms\\
    $2 \stimes 16$ & 512 KiB & L3 & 79.0\% & 0.58 ms & 2.7 ms\\
    \midrule
    \multicolumn{6}{c}{64-bit codes}\\
    \midrule
    $16 \stimes 4$ & 1 KiB & L1 & 83.1\% & 0.001 ms & 6 ms\\
    $8 \stimes 8$  & 8 KiB & L1 & 92.2\% & 0.011 ms & 2.6 ms\\
    $4 \stimes 16$ & 1 MiB & L3 & 96.5\% & 0.82 ms & 7.9 ms\\
    \midrule
    \multicolumn{6}{c}{128-bit codes}\\
    \midrule
    $32 \stimes 4$ & 2 KiB & L1 & 96.5\% & 0.002 ms & 12 ms\\
    $16 \stimes 8$ & 16 KiB & L1 & 99.8\% & 0.009 ms & 5.4 ms\\
    $8 \stimes 16$ & 2 MiB & L3 & 99.8\% & 1.5 ms & 17 ms\\
    \bottomrule
  \end{tabular}
\end{table}

\paragraph{}
For a fixed code size, and thus a fixed $m{\stimes}b$ product (\eg $m{\stimes}b=64$), the parameters $m$ and $b$ impact the accuracy and speed of nearest neighbor search. We observe that a large $b$ and therefore a small $m$, increases accuracy, but also increases scan time (Table~\ref{tbl:tblsize}).  Thus, a $4{\stimes}16$ product quantizer offers a better accuracy than $8{\stimes}8$ product quantizer, but also incurs a higher scan time. This seems counter-intuitive: as small $m$ means less operations per distance computation, and should therefore result in lower scan time. Each distance computation requires $m$ \emph{mem1} accesses, $m$  \emph{mem2} accesses and $m$ additions. A $4{\stimes}16$ product quantizer ($m=4$) requires less operations per distance computation than a $8{\stimes}8$ product quantizer ($m=8$), but still results in a higher scan time. This is because a $4{\stimes}16$ product quantizer leads to larger $\{D^j\}_{j=0}^{m-1}$ lookup tables. The size of the lookup tables is $m\cdot 2^b \cdot \operatorname{sizeof}(\mathrm{float}) = m{\cdot}2^{b}{\cdot}4$ bytes ($m$ lookup tables of $2^b$ floating-point values each). Larger lookup tables need to be stored in slower cache levels, which increases the cost of \emph{mem2} accesses. For a $4{\stimes}16$ product quantizer, lookup tables have to be stored in the L3 cache (slowest cache) while they fit the L1 cache (fastest cache) for a $8{\stimes}8$ product quantizer. Overall, the increase in the cost of \emph{mem2} access outweighs the decrease in the number of operations. This results in a threefold increase in scan time for $4{\stimes}16$ product quantizers compared to  $8{\stimes}8$ product quantizers.

\paragraph{}
Table~\ref{tbl:tblsize} reports the properties of different product quantizers only for a few values of $b$, namely $b=4$, $b=8$ and $b=16$. First, training sub-quantizers (Section~\ref{sec:vectorencoding}) with more than $2^{16}$ centroids is not tractable, thus $b>16$ is not achievable. Theoretically, any $b\leq16$ would be possible, \eg a $5{\stimes}13$ product quantizer ($b=13$, 65-bit codes), $c[j]$ centroid indexes need to be accessed individually (Algorithm~\ref{alg:ann}, line~\ref{alg:line:index}). CPUs can natively address values of 8, 16, 32 or 64 bits in memory. Accessing integers of other sizes, \eg 13-bit integers, requires additional bit shifting and bit masking operations, which increase scan time. Therefore, $b$ values other than $b=8$ or $b=16$ are not suitable. We however added the properties of product quantizers for $b=4$, to illustrate that scan time does not continuously decreases with $b$. Once lookup tables are small enough to fit the L1 cache, it is not useful to further shrink them. This does not make \emph{mem2} accesses less costly, but results in a higher number $m$ of operations.

\paragraph{}
In practice, 8-bit sub-quantizers ($b=8$) are used in almost all publications on product quantization as they offer a good tradeoff between speed and accuracy. As 64-bit codes, and 8-bit sub-quantizers are very common, we focus exclusively on $8{\stimes}8$ product quantizers ($m=8, b=8$, 64-bit codes) in the remainder of this section.

\begin{figure}
  \centering
  \begin{tikzpicture}[baseline, trim axis left]
  \begin{axis}[
    ybar,
    height=5cm,
    width=6.4cm,
    legend pos=north west,
    symbolic x coords={time-all-ms},
    xtick=data,
    xticklabels={time},
    ylabel=(ms),
    ymin=0,
    nodes near coords={\pgfmathprintnumber[fixed, precision=0]{\pgfplotspointmeta}},
    every node near coord/.append style={font=\small},
    enlarge y limits = {abs=3ex,upper},
    legend to name=legend-timeipc,
    legend columns = 4,
  ]
  \addplot+ table [x=counter, y=naive] {perf/plots/scan_time.gnuplot};
  \addplot+ table [x=counter, y=libpq] {perf/plots/scan_time.gnuplot};
  \addplot+ table [x=counter, y=avx] {perf/plots/scan_time.gnuplot};
  \addplot+ table [x=counter, y=gather] {perf/plots/scan_time.gnuplot};
  \legend{baseline,libpq,avx,gather};
  \end{axis}
  \end{tikzpicture}
  \begin{tikzpicture}[baseline, trim axis right]
  \begin{axis}[
    ybar,
    height=5cm,
    width=6.4cm,
    symbolic x coords={ipc},
    xtick=data,
    xticklabels={IPC},
    ymin=0,
    legend pos=north west,
    nodes near coords={\pgfmathprintnumber[fixed, 	precision=1]{\pgfplotspointmeta}},
    every node near coord/.append style={font=\small},
    enlarge y limits = {abs=3ex,upper}
  ]
  \addplot+ table [x=counter, y=naive] {perf/plots/scan_ipc.gnuplot};
  \addplot+ table [x=counter, y=libpq] {perf/plots/scan_ipc.gnuplot};
  \addplot+ table [x=counter, y=avx] {perf/plots/scan_ipc.gnuplot};
  \addplot+ table [x=counter, y=gather] {perf/plots/scan_ipc.gnuplot};
  \end{axis}
  \end{tikzpicture}
  \\
  \pgfplotslegendfromname{legend-timeipc}
  \caption{Scan times (25M vectors) and Intructions Per Cycle (IPC)\label{fig:timeipc}}
\end{figure}
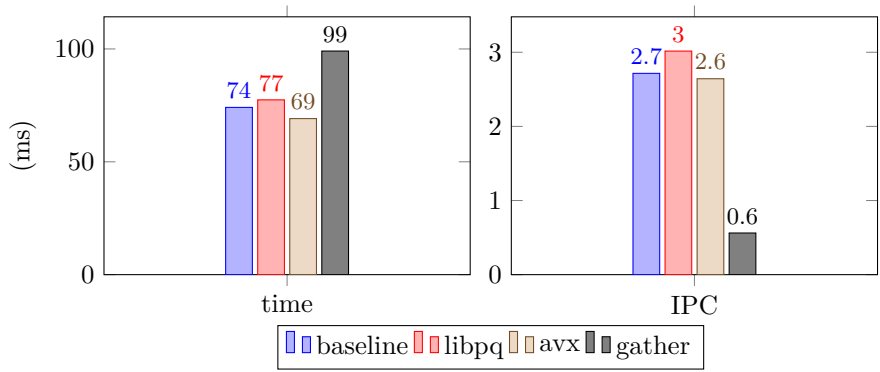

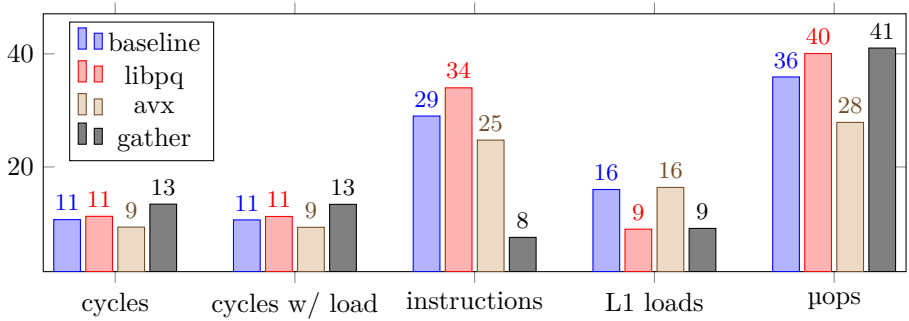
\begin{figure}
  \centering
  \begin{tikzpicture}[trim axis left, trim axis right]
  \begin{axis}[
    ybar,
    legend pos=north west,
    width=\textwidth, height=5cm,
    symbolic x coords={cycles,cycles-pending-load,instructions,L1-dcache-loads,uops},
    xtick=data,
    xticklabels={cycles,cycles w/ load, instructions, L1 loads, {\textmu}ops},
    nodes near coords={\pgfmathprintnumber[fixed, precision=0]{\pgfplotspointmeta}},
    every node near coord/.append style={font=\small},
    enlarge y limits = {upper, abs=3ex}
  ]
    \addplot+ table [x=counter, y=naive] {perf/plots/scan_counters.gnuplot};
    \addplot+ table [x=counter, y=libpq] {perf/plots/scan_counters.gnuplot};
    \addplot+ table [x=counter, y=avx] {perf/plots/scan_counters.gnuplot};
    \addplot+ table [x=counter, y=gather] {perf/plots/scan_counters.gnuplot};
    \legend{baseline,libpq,avx,gather}
  \end{axis}
  \end{tikzpicture}
  \caption{Scan performance counters (per scanned code)\label{fig:perfcounter}}
\end{figure}

\paragraph{Libpq scan implementation} We use hardware performance counters to study the performance of different scan procedure implementations experimentally (Figure~\ref{fig:timeipc}, Figure~\ref{fig:perfcounter}). For all implementations, the number of cycles with pending load operations (cycles w/ load) is almost equal to the number of cycles, which confirms the scan procedure is a memory-intensive. We also measured the number of L1 cache misses (not shown on Figure \ref{fig:perfcounter}). L1 cache misses represent less than 1\% of memory accesses for all implementations, which confirms that both \emph{mem1} and \emph{mem2} accesses hit the L1 cache. The baseline implementation of the scan procedure (Algorithm \ref{alg:ann}) performs 16 L1 loads per scanned vector: $m=8$ \emph{mem1} accesses and $m=8$ \emph{mem2} accesses.
The authors of~\cite{Jegou2011} distribute the libpq
library\footnote{\scriptsize\url{http://people.rennes.inria.fr/Herve.Jegou/projects/ann.html}}, which includes an optimized implementation of the ADC Scan procedure. We obtained a copy of libpq under a commercial licence. Rather than loading $m=8$ centroid indexes of $b=8$ bits each (\emph{mem1} accesses), the libpq implementation of the scan procedure loads a 64-bit word into a register, and performs 8-bit shifts to access individual centroid indexes. This allows reducing the number of \emph{mem1} accesses from $8$ to $1$. Therefore, the libpq implementation performs $9$ L1 loads per scanned vector: $1$ \emph{mem1} access and $8$ \emph{mem2} accesses. However, overall, the libpq implementation is slightly slower than the baseline implementation on our Haswell processor. Indeed, the increase in the number of instructions offsets the increase in IPC (Instructions Per Cycle) and the decrease in L1 loads. It is however likely that this optimization was useful on older processors. Before, the Sandy Bridge architecture, Intel CPUs had a single memory read port, while newer CPUs have two read ports. This means that current CPUs can perform two concurrent cache reads, while older CPUs can only perform one. Therefore, reducing the number of memory reads may bring a higher benefit on older CPUs.

\section{Issues with SIMD Implementation}
\label{sec:simd-issues}

\paragraph{Introduction to SIMD} In addition to memory accesses, each distance computation requires $m$ additions. SIMD instructions are commonly used to increase the performance of algorithms performing arithmetic computations. Therefore, we evaluate the applicability of SIMD instructions to reduce the number of instructions and CPU cycles devoted to additions. SIMD instructions perform the same operation, \eg additions, on multiple data elements in one instruction. To do so, SIMD instructions operate on wide registers. SSE SIMD instructions operate on 128-bit registers, while more recently introduced AVX SIMD instructions operate on 256-bit registers~\cite{IntelArchVol1}. SSE instructions can operate on 4 floating-point \emph{ways} (4$\times$32 bits, 128 bits) while AVX instructions can operate on 8 floating-point \emph{ways} (8$\times$32 bits, 256 bits). In our case, AVX instructions offer only slightly higher performance than SSE instructions. As both instruction sets yield similar results, we only report results for the AVX implementation. We show that the structure of the scan procedure (Algorithm~\ref{alg:ann}) prevents an efficient use of SIMD instructions.

\begin{figure}[b]
  \centering
  \input{perf/fig/simd_add}
  \caption{SIMD vertical add\label{fig:simdadd}}
\end{figure}

\paragraph{Implementation details} To enable the use of fast \emph{vertical} SIMD additions, we compute the asymmetric distance between the query vector and 8 database vectors at a time, designated by the letters $a$ to $h$. We still issue 8 addition instructions, but each instruction involves 8 different vectors, as shown on Figure \ref{fig:simdadd}. Overall, the number of instructions devoted to additions is divided by 8.
However, the gain in cycles brought by the use of SIMD additions is offset by the need to set the ways of SIMD registers one by one.
SIMD processing works best when all values in all ways are \emph{contiguous} in memory and can be loaded in one instruction. Because they were looked up in a table, $D_0[a[0]], D_0[b[0]],\cdots,D_0[h[0]]$ values are not contiguous in memory. We therefore need to insert $D_0[a[0]]$ in the first way of the SIMD register, then $D_0[b[0]]$ in the second way \etc In addition to memory accesses, doing so requires many SIMD instructions, some of which have high latencies.
Overall, this offsets the benefit provided by SIMD additions. This explains why algorithms relying on lookup tables, such as ADC Scan, hardly benefit from SIMD processing. Figure \ref{fig:timeipc} and Figure \ref{fig:timeipc} show that the AVX implementation of ADC Scan requires slightly less instructions than the naive implementation, and is only marginally faster.

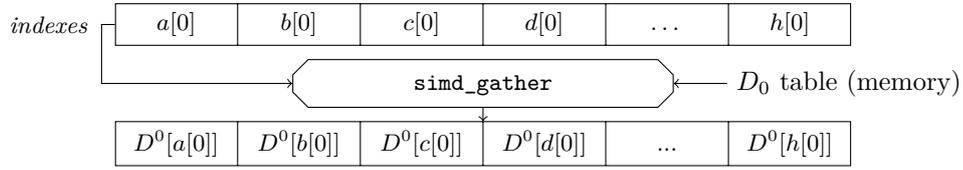
\begin{figure}
  \centering
  \input{perf/fig/simd_gather}
  \caption{SIMD gather\label{fig:simdgather}}
\end{figure}

\begin{figure} [b]
  \centering
  \subfloat[Standard layout]{%
    \centering%
    \input{perf/fig/simd_layout_normal}%
    \label{fig:layoutnorm}%
  }%
  \hfil%
  \subfloat[Transposed layout]{%
    \centering%
    \input{perf/fig/simd_layout_tranposed}%
    \label{fig:layouttrans}%
  }%
  \caption{Standard layout and transposed layout\label{fig:layout}}
\end{figure}
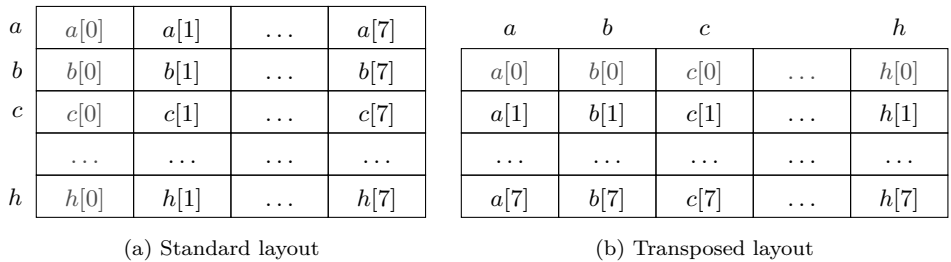

\paragraph{SIMD gather} To tackle this issue, Intel introduced a \texttt{gather} SIMD instruction (AVX2 instruction set) in its latest architecture, Haswell~\cite{IntelArchVol1, IntelArchVol2}. Given an SIMD register containing 8 indexes and a table stored in memory, \texttt{gather} looks up the 8 corresponding elements from the table and stores them in a register, in just one instruction. This avoids having to use many SIMD instructions to set the 8 ways of SIMD registers. Figure \ref{fig:simdgather} shows how \texttt{gather} can be used to look up 8 values in the first distance table ($D_0$). To efficiently use \texttt{gather}, we need $a[0],\cdots,h[0]$ to be stored contiguously in memory, so that they can be loaded in one instruction. To do so, we transpose the memory layout of the inverted list. We divide each inverted list in blocks of 8 codes $(a,\dotsc,h)$, and transpose each block independently. In a transposed block (Figure~\ref{fig:layout}), we store the first components of 8 vectors contiguously ($a[0],\cdots,h[0]$), followed by the second components of the same 8 vectors ($a[1],\cdots,h[1]$) \etc, instead of storing all components of the first vector ($a[0],\cdots,a[7]$), followed by the components of the second vector ($b[0],\cdots,b[7]$). This also allows to reduce the number of \emph{mem1} accesses from $8$ to 1, similarly to the libpq implementation. This transposition is performed online, and does not increase response time. Moreover, this transposition is cheap and does not notable increase the database creation time.

\paragraph{}
However, the gather implementation of ADC Scan is slower than the naive version (Figure \ref{fig:perfcounter}), which can be explained by several factors. First, even if it consists of only one instruction, \texttt{gather} performs 1 memory access for each element it loads, which implies suffering memory latencies. Second, at the hardware level, \texttt{gather} executes 34 {\textmu}ops\footnote{Micro-operations ({\textmu}ops) are the basic operations executed by the processor. Instructions are sequences of {\textmu}ops.} where most instructions execute only 1 {\textmu}op. Figure \ref{fig:perfcounter} shows that the gather implementation has a low instructions count but a high {\textmu}ops count. For other implementations, the number of {\textmu}ops is only slightly higher than the number of instructions. It also has a high latency of 18 cycles and a throughput of 10 cycles, which means it is necessary to wait 10 cycles to pipeline a new \texttt{gather} instruction after one has been issued. This translates into poor pipeline utilization, as shown by the very low IPC of the gather implementation (Figure \ref{fig:perfcounter}). In its documentation, Intel acknowledges that \texttt{gather} instructions may only bring performance benefits in specific cases~\cite{IntelOpt} and other authors reported similar results~\cite{Hofmann2014}.

\section{Lessons Learned}

\paragraph{}
We can draw three conclusions from the experiments conducted in this chapter:
\begin{itemize}
\item Accesses to cache resident lookup tables are the primary bottleneck limiting the performance of the ADC Scan procedure.
\item Using 16-bit sub-quantizers offers a better accuracy than 8-bit sub-quantizers (for a fixed code size), but causes a threefold increase in response time. This is because using 16-bit sub-quantizers causes lookup tables to be stored in the L3 cache, instead of the L1 cache for 8-bit sub-quantizers.
\item The structure of the ADC Scan algorithm prevents an efficient use of SIMD instructions, which are yet commonly used to increase performance. The main issue is that ADC Scan needs to look up values in the cache and insert them one by one in SIMD registers, which is costly.
\end{itemize}
In the remainder of this thesis, we build on these conclusions to design more efficient scan procedures.

\stopcontents[chapters]

%% file: perf/fig/simd_add.tex
\begin{tikzpicture}
\newlength\tmpl
\setlength{\tmpl}{\widthof{$D^0[c[0]]$}}

\node[register] (register0)
{%
\nodepart[elt]{one}$D^0[a[0]]$
\nodepart[elt]{two}$D^0[b[0]]$
\nodepart[elt]{three}$D^0[c[0]]$
\nodepart[elt]{four}$D^0[d[0]]$
\nodepart[elt]{five}...%
\nodepart[elt]{six}$D^0[h[0]]$%
};

\node[rectangle split=true,
  rectangle split horizontal=true,
  rectangle split empty part width=\tmpl,
  below = 0.3em of register0.south west,
  anchor=north west,
  rectangle split parts=6,
] (plus)
{%
  \nodepart[elt]{one}$+$
  \nodepart[elt]{two}$+$
  \nodepart[elt]{three}$+$
  \nodepart[elt]{four}$+$
  \nodepart[elt]{five}
  \nodepart[elt]{six}$+$
};

\node [above=0.1cm of register0.one north, 
  anchor=south,
  inner xsep=0, 
  outer xsep=0] {\small (way 0)};
  
\node [above=0.1cm of register0.six north, 
  anchor=south,
  inner xsep=0, 
  outer xsep=0] {\small (way 7)};

\node[register,
below = 2em of register0.south, anchor=north, 
] (register1)
{%
  \nodepart[elt]{one}$D^1[a[1]]$
  \nodepart[elt]{two}$D^1[b[1]]$
  \nodepart[elt]{three}$D^1[c[1]]$
  \nodepart[elt]{four}$D^1[d[1]]$
  \nodepart[elt]{five}...%
  \nodepart[elt]{six}$D^1[h[1]]$%
};

\node[rectangle split=true,
rectangle split horizontal=true,
rectangle split empty part width=\tmpl,
below = 0.3em of register1.south west,
anchor=north west,
rectangle split parts=6,
] (plus)
{%
  \nodepart[elt]{one}$+$
  \nodepart[elt]{two}$+$
  \nodepart[elt]{three}$+$
  \nodepart[elt]{four}$+$
  \nodepart[elt]{five}
  \nodepart[elt]{six}$+$
};

\node[rectangle split=true,
rectangle split horizontal=true,
rectangle split empty part width=\tmpl,
below = -0.5em of plus.south west,
anchor=north west,
rectangle split parts=6,
] (dots)
{%
  \nodepart[elt]{one}$\vdots$
  \nodepart[elt]{two}$\vdots$
  \nodepart[elt]{three}$\vdots$
  \nodepart[elt]{four}$\vdots$
  \nodepart[elt]{five}
  \nodepart[elt]{six}$\vdots$
};

\node[register,
below = 4em of register1.south, anchor=north, 
] (register2)
{%
  \nodepart[elt]{one}$D^7[a[7]]$
  \nodepart[elt]{two}$D^7[b[7]]$
  \nodepart[elt]{three}$D^7[c[7]]$
  \nodepart[elt]{four}$D^7[d[7]]$
  \nodepart[elt]{five}...%
  \nodepart[elt]{six}$D^7[h[7]]$%
};

\end{tikzpicture}

%% file: perf/fig/simd_gather.tex
\begin{tikzpicture}

\node[register] (indexes)
{%
  \nodepart[elt]{one}$a[0]$
  \nodepart[elt]{two}$b[0]$
  \nodepart[elt]{three}$c[0]$
  \nodepart[elt]{four}$d[0]$
  \nodepart[elt]{five}$\dots$
  \nodepart[elt]{six}$h[0]$
};

\node[operator, below=\vertSep of indexes.south, anchor=north] (gather)
{\texttt{simd\_gather}};

\node[register, below=\vertSep of gather.south, anchor=north] (result)
{%
  \nodepart[elt]{one}$D^0[a[0]]$
  \nodepart[elt]{two}$D^0[b[0]]$
  \nodepart[elt]{three}$D^0[c[0]]$
  \nodepart[elt]{four}$D^0[d[0]]$
  \nodepart[elt]{five}...%
  \nodepart[elt]{six}$D^0[h[0]]$%
};

\draw[->] (indexes.west) -- ++(-\horSep, 0) |- (gather.west);

\node[
  right = 2em of gather.east,
  anchor = west
] (memory) {$D_0$ table (memory)};

\node[label, left=\labelSep of indexes.west, anchor=east] {$\mathit{indexes}$};

\draw[->] (gather.south) -- (result.north);
\draw[->] (memory.west) -- (gather.east);


\end{tikzpicture}

%% file: perf/fig/simd_layout_normal.tex
\begin{tikzpicture}
\node [memory,
  column 1/.style={text=black!70},
] (layout)
{ a[0] & a[1] & \dots & a[7] \\
  b[0] & b[1] & \dots & b[7] \\
  c[0] & c[1] & \dots & c[7] \\
  \dots & \dots & \dots & \dots \\
  h[0] & h[1] &\dots & h[7] \\
};

  \node[label, left=\horSep of layout-1-1.west, anchor=east] {$a$};
  \node[label, left=\horSep of layout-2-1.west, anchor=east] {$b$};
  \node[label, left=\horSep of layout-3-1.west, anchor=east] {$c$};
  \node[label, left=\horSep of layout-5-1.west, anchor=east] {$h$};

\end{tikzpicture}

%% file: perf/fig/simd_layout_tranposed.tex
  \begin{tikzpicture}
  \node [memory, row 1/.style={text=black!70}] (layout)
  { a[0] & b[0] & c[0] &\dots & h[0] \\
    a[1] & b[1] & c[1] &\dots & h[1] \\
    \dots & \dots & \dots & \dots & \dots  \\
    a[7] & b[7] & c[7] &\dots & h[7] \\
  };
  
  \node[label, above=\vertSep of layout-1-1.north, anchor=south] {$a$};
  \node[label, above=\vertSep of layout-1-2.north, anchor=south] {$b$};
  \node[label, above=\vertSep of layout-1-3.north, anchor=south] {$c$};
  \node[label, above=\vertSep of layout-1-5.north, anchor=south] {$h$};
  \end{tikzpicture}

%% file: fastpq/fastpq.tex
\startcontents[chapters]
\ruledprintcontents

\section{Motivation}

\paragraph{}
As it stores large databases in main memory, product quantization makes it possible to scan a large number of nearest neighbor candidates in a small amount of time. However, the ADC Scan procedure remains CPU intensive. In particular, we have shown that the performance of this scan procedure is limited by the large number of accesses to cache-resident lookup tables. Moreover, the structure of this procedure prevents an efficient use of SIMD instructions to boost performance. These limitations call for a modification of ADC Scan.

\paragraph{}
In this chapter, we introduce PQ Fast Scan, a novel scan procedure that achieves 4-6 times better performance than the conventional ADC Scan procedure, while returning the exact same results. PQ Fast Scan focuses exclusively on codes generated by $8{\stimes}8$ product quantizers but this is not a strong limitation, as this type of codes is used in the vast majority of cases. The key idea behind PQ Fast Scan is to replace cache accesses by SIMD in-register shuffles. This change allows PQ Fast Scan to perform less than 2 caches accesses per distance computation, and enables an efficient SIMD implementation of additions. The main design challenge of PQ Fast Scan is that lookup tables used for distance computations are much larger (1 KiB each) than SIMD registers (128 bits). PQ Fast Scan overcomes this challenge by building small tables that fit SIMD registers. These small tables are used to compute lower bounds on distances, and prune unneeded cache accesses. More specifically, this chapter addresses the following points:
\begin{itemize}
  \item We present the design of PQ Fast Scan. We describe the three techniques we use to build small tables: (1) code grouping, (2) minimum tables and (3) quantization of floating-point distances to 8-bit integers. 
  \item We implement PQ Fast Scan on Intel CPUs and evaluate its performance on large datasets of high-dimensional vectors. We determine the parameters that impact its performance, and experimentally show that it achieves a 4-6 times speedup over the conventional scan procedure. 
  \item We discuss the compatibility of PQ Fast Scan with different configurations of inverted indexes. We also review its applicability to other product quantizer configurations than $8{\stimes}8$ as well as its applicability to derivatives of product quantization.
\end{itemize}

\section{Presentation}

\subsection{Overview}

\label{sec:fastpqdesc}

\begin{figure}
  \centering
  \input{fastpq/fig/fastpq_overview.tex}
  \caption{Overview of PQ Fast Scan\label{fig:fastpq-overview}}
\end{figure}
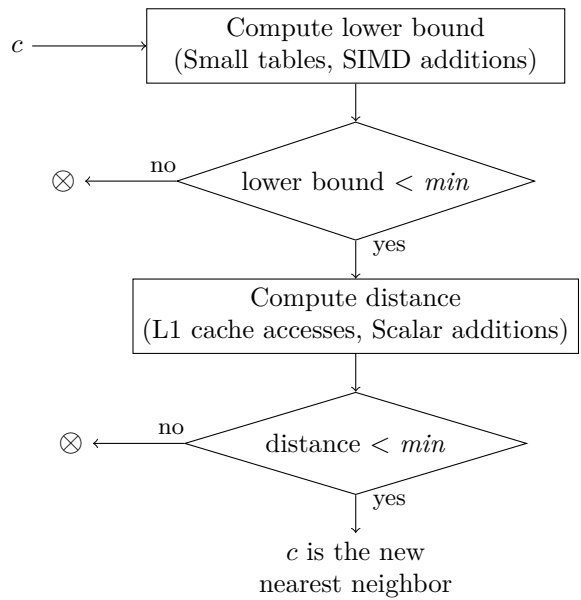

\paragraph{}
The key idea behind PQ Fast Scan is to use \emph{small tables}, sized to fit SIMD registers instead of the cache-resident distance tables. These small tables are used to compute \emph{lower bounds} on distances, without accessing the L1 cache. Therefore, lower bounds computations are fast. In addition, they are implemented using SIMD additions, further improving performance. We use lower bound computations to prune slow distance computations, which access the L1 cache and cannot benefit from SIMD additions. Figure \ref{fig:fastpq-overview} shows the processing steps applied to every database vector $c$. The $\otimes$ symbol means we discard the vector $c$ and move to the next database vector. The $\mathit{min}$ value is the distance of the query vector to the current nearest neighbor. Our experimental results on SIFT data show that PQ Fast Scan is able to prune more than 95\% of distance computations.

\begin{figure}
  \centering
  \input{fastpq/fig/simd_shuffle}
  \caption{SIMD in-register shuffle (\texttt{pshufb})\label{fig:shuffle}}
\end{figure}
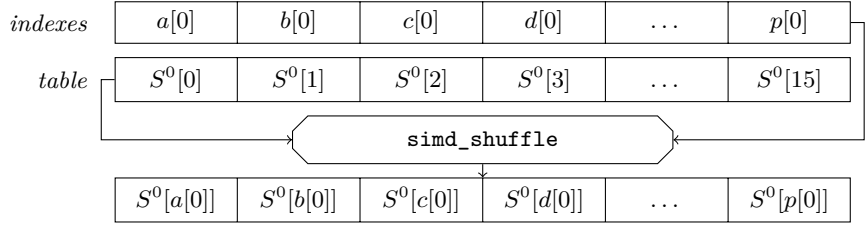

\paragraph{}
Lower bound computations rely on SIMD in-register shuffles (\texttt{pshufb} instruction), which are key to PQ Fast Scan performance. Like cache accesses (\texttt{mov}), or SIMD gather operations (\texttt{gather}), SIMD in-registers shuffles can be used to lookup values in tables. SIMD in-register shuffles however offer much better performance (Table~\ref{tbl:lookup-props}). Using cache accesses allow performing 1 lookup at once, with a latency of 4-5 cycles (lookup tables in L1 cache). SIMD gather has a higher level of parallelism (8 lookups at once), but has a much higher latency (18 cycles). In addition, it has a throughput of 10 cycles, meaning the CPU has to wait 10 cycles before issuing another \texttt{gather} instruction. This makes it difficult to pipeline instructions efficiently, reducing performance. On the other hand, SIMD in-register shuffles offer both a high level of parallelism (16 lookups at once) and a low latency (1 cycle). This comes at the price of strong constraints on lookup tables. When using caches accesses or SIMD gather, lookup tables are stored in memory. Therefore, this is virtually no limit on their size, although it should not exceed the size of the L1 cache for best performance. Moreover, they allow looking up 32-bit floating point values. By contrast, when using SIMD in-register shuffles, lookup tables must be stored in SIMD registers, which limit their size to 128 bits. There are different variants of SIMD in-register shuffles allowing either lookups in tables of 4 element of 32 bits (\texttt{pshufd}) or lookups in tables of 16 elements 8 bits (\texttt{pshufb}). In our case, we need tables with the most possible elements, thus we focus exclusively on \texttt{pshufb} (Figure~\ref{fig:shuffle}).

\begin{table}[b]
  \centering
  \begin{threeparttable}
  \caption{Properties of table lookup techniques\label{tbl:lookup-props}}
  \begin{tabular}{lllllll}
    \toprule
    Lookup technique & Par. \tnote{1} & Lat. \tnote{2} & Throu. \tnote{3} & {\textmu}ops & \multicolumn{2}{c}{Table Elements} \\ \cmidrule(l){6-7}
    & & & & & Count & Size \\

    \midrule
    Cache accesses (\texttt{mov}) & 1 & 4-5 & 0.5 & 1 & No limit & 32 bit \\
    SIMD gather (\texttt{gather}) & 8 & 18 & 10 & 34 & No limit & 32 bit \\
    SIMD shuffle (\texttt{pshufb}) & 16 & 1 & 0.5 & 1 & 16 & 8 bit \\
    \bottomrule
  \end{tabular}%
  \begin{tablenotes}
      \item[1] {\footnotesize Par.: Parallelism. Number of lookups performed by one instruction of this type.}
      \item[2] {\footnotesize Lat.: Latency. Number of cycles to execute one instruction of this type.}
      \item[3] {\footnotesize Throu.: Throughput. Number of cycles to wait before issuing another instruction of this type.}
  \end{tablenotes}
  \end{threeparttable}
\end{table}

\begin{figure}[ht]
  \centering
  \input{fastpq/fig/small_tables.tex}
  \caption{Small tables building process\label{fig:build-small-tables}}
\end{figure}
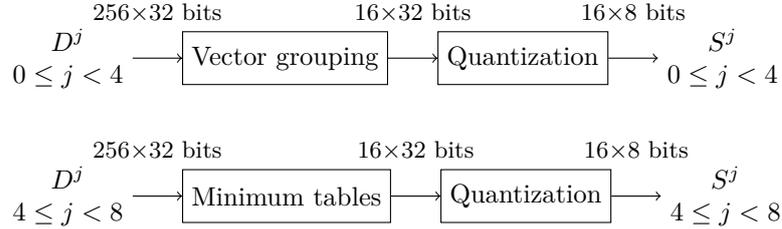

\paragraph{}
To compute asymmetric distances, the conventional ADC Scan algorithm (with a \pq{8}{8} quantizer) uses 8 distance tables $D^j$, $0 \leq j < 8$, and each distance table comprises 256 elements of 32 bits. Hence, one distance table (256$\times$32 bits) does not fit into an SIMD register, which is why we need to build small tables. Just like there are 8 distance tables $D^j$, $0 \leq j < 8$, we build 8 small tables $S^j$, $0 \leq j < 8$. Each small table $S^j$ is stored in a distinct SIMD register and is built by applying transformations to the corresponding $D^j$ table. To build 8 small tables suitable to compute lower bound on distances, we combine three techniques: (1) \emph{code grouping}, (2) computation of \emph{minimum tables} and (3) \emph{quantization of distances}.
The first two techniques, vector grouping and computation of minimum tables, are used to build tables of 16 elements (16$\times$32 bits). The third technique, quantization of distances, is used to shrink each element to 8 bits (16$\times$32 bits $\rightarrow$ 16$\times$8 bits). We group vectors and quantize distances to build the \emph{first four} small tables, $S^0,\dotsc,S^3$. We compute minimum tables and quantize distances to build the \emph{last four} small tables, $S^4,\dotsc,S^7$. Figure \ref{fig:build-small-tables} summarizes this process.

\subsection{Code Grouping}
\label{sec:code-grouping}

\paragraph{}
Database vectors are stored as short codes, which consist of 8 components of 8 bits (Figure \ref{fig:codes-unordered}). When computing between a code and the query vector, each component is used as an index in the corresponding distance table, \eg the 1st component is used as an index in the 1st distance table (Algorithm~\ref{alg:ann}, Section \ref{sec:annsearch}). The key idea behind \emph{code grouping} is to group vectors such that all codes belonging to a \emph{group} hit the same \emph{portion} of 16 elements of a distance table.

\begin{figure}
  \centering
  \input{fastpq/fig/table_portions.tex}
  \caption{Portions of the first distance table\label{fig:table-portions}}
\end{figure}

\begin{figure}[b]
  \centering
  \subfloat[Unordered vectors]{%
    \centering%
    \input{fastpq/fig/unordered_codes}%
    \label{fig:codes-unordered}%
  }\hfil%
  \subfloat[Grouped vectors]{%
    \centering%
    \input{fastpq/fig/grouped_codes}
    \label{fig:codes-grouped}%
  }%
  \caption{Code grouping}
  \label{fig:code-grouping}%
\end{figure}

\paragraph{}
We focus on the first distance table, $D^0$. We group vectors on their first component and divide the $D^0$ table into 16 portions, of 16 elements each (Figure \ref{fig:table-portions}).
All database vectors $c$ having a first component $c[0]$ between \texttt{00} and \texttt{0f} (0 to 15) will trigger lookups in portion 0 of $D^0$ when computing the distance. These vectors form group 0. All vectors having a first component between \texttt{10} and \texttt{1f} (16 to 31) will trigger lookups in portion 1 of $D^0$. These vectors form the group 1. We define 16 groups in this way. Each group is identified by an integer $i$, and contains database vectors $c$ such that:
\[
16(i-1) \leq c[0] < 16i
\]
and only requires the portion $i$ of the first distance table, $D^0$. We apply the same grouping procedure on the \nth{2}, \nth{3} and \nth{4} components. Eventually, each group is identified by four integers $(i_0, i_1, i_2, i_3)$, each belonging to $[0;16[$ and contains vectors such that:
\begin{align*}
& 16(i_0-1) \leq c[0] < 16i_0 \; \operatorname{\land} \; 16(i_1-1) \leq c[1] < 16i_1 \; \operatorname{\land}\\
& 16(i_2-1) \leq c[2] < 16i_2 \; \operatorname{\land} \; 16(i_3-1) \leq c[3] < 16i_3
\end{align*}
Figure \ref{fig:codes-grouped} shows a database where all vectors have been grouped. We can see that all vectors in the group $(3,1,2,0)$ have a first component between \texttt{30} and \texttt{3f}, a second component between \texttt{10} and \texttt{1f}, \etc
To compute the distance of the query vector to any vector of a group $(i_0, i_1, i_2, i_3)$, we only need a portion of $D^0$, $D^1$, $D^2$ and $D^3$. Before scanning a group, we load the relevant portions of $D^0,\cdots,D^3$ into 4 SIMD registers to use them as the small tables  $S^0,\cdots,S^3$. This process is shown by solid arrows on Figure \ref{fig:global-lookup}.

\paragraph{}
We applying grouping on $g=4$ components as it is a good tradeoff between pruning power and constraints on inverted list sizes. Grouping on $g=4$ components allows pruning 95-99\% distance computations, and thus offers a large speedup. Grouping on less that 4 components, \eg 3 components, strongly impacts the tightness of lower bounds, and therefore strongly decreases pruning efficiency. This prevents PQ Fast Scan from offering a significant speedup, as it is not able to prune enough slow distance computations. On the other hand, grouping on more than 4 components, \eg 5 components, does not substantially increases pruning power, while it imposes more constraints on the size of inverted lists. The number of groups created by code grouping is given by $P^g$, where $P$ is the number portions in a lookup table ($P=16$ in our case), and $g$ is the number of components used for grouping. Therefore, the average size of a group is given by $s=n/P^g$. For best performance, $s$ should exceed $s_{min}=50$ codes. Before scanning a group, we load portions of lookup tables in SIMD registers, which is quite costly. If the group comprises less than 50 vectors, a large part of the CPU time is spent loading distance tables. This is detrimental to performance, as shown by our experimental results (Section~\ref{sec:impact-size}). The minimum inverted list size $n_{min}(g)$ to be able to group vectors on $g$ components is therefore given by $n_{min}(g) = s_{min} \cdot P^g$, and increases \emph{exponentially} with the number of components used for grouping. For $g=4$, the minimum inverted list size is 3 million vector ($s_{min}=50, P=16$).

\paragraph{}
Grouping vectors also allows decreasing the amount of memory consumed by the database by approximatively 25\%. In a group, the \nth{1} component of all vectors has the same 4 most significant bits. As we apply grouping on the 4 first components, their \nth{2}, \nth{3} and \nth{4} components also have the same most significant bits. We can therefore avoid storing the 4 most significant bits of the 4 first components each database vector. This saves $4 \times 4 \text{ bits} = 16 \text{ bits}$ on the $8 \times 8 \text{ bits} = 64 \text{ bits}$ of each vector, which leads to a 25\% reduction in memory consumption. Thus, on Figure \ref{fig:codes-grouped}, the grayed out hexadecimal digits (which represent 4 bits) may not be stored.

\begin{figure}
  \hfil
  \subfloat[Arbitrary assignment]{%
    \centering%
    \hspace{0.05\linewidth}%
    \includegraphics[height=4cm]{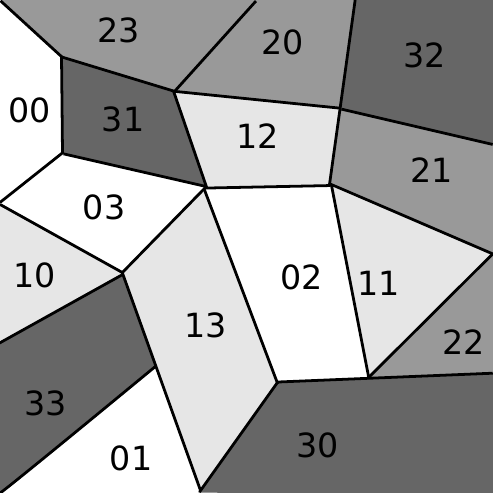}%
    \hspace{0.05\linewidth}%
    \label{fig:aaaa}%
  }\hfil%
  \subfloat[Optimized assignment]{ %
    \centering %
    \hspace{0.05\linewidth}%
    \includegraphics[height=4cm]{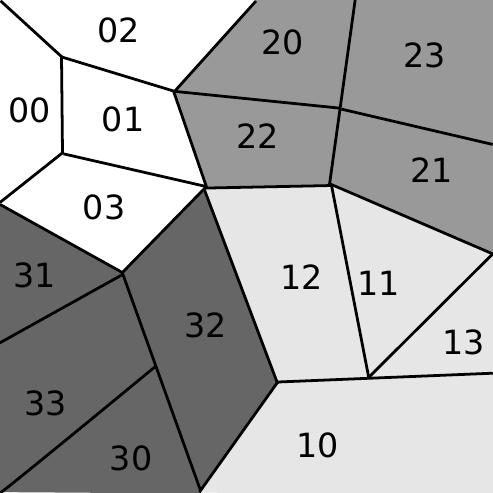} %
    \hspace{0.05\linewidth}%
    \label{fig:cccc} %
  }\hfil%
  \caption{Centroid indexes assignement}
  \label{fig:centroids}%
\end{figure}

\subsection{Minimum Tables}

\begin{figure}
  \centering
  \input{fastpq/fig/minimum_tables}
  \caption{Minimum tables\label{fig:mintables}}
\end{figure}

\paragraph{}
We grouped vectors to build the first four small tables $S^0,\cdots,S^3$. To build the last four small tables, $S^4,\cdots,S^7$ we compute minimum tables. This involves dividing the original distance tables, $D^4,\cdots,D^7$, into 16 portions of 16 elements each. We then keep the minimum of each portion to obtain a table of 16 elements. This process is shown on Figure \ref{fig:mintables}. Using the minimum tables techniques alone results in small tables containing low values, which is detrimental to PQ Fast Scan performance. If these values are too low, the computed lower bound is not tight, \ie far from the actual distance. This limits the ability of PQ Fast Scan to prune costly distance computations.

\paragraph{}
To obtain small tables with higher values, we introduce an \emph{optimized assignment} of sub-quantizer centroids indexes. Each value $D^j[i]$ in a distance table is the distance between the $j$th sub-vector of the query vector and the centroid with index $i$  of the $j$th sub-quantizer (Section \ref{sec:annsearch}). When a sub-quantizer is learned, centroids indexes are assigned arbitrarily. Therefore, there is no specific relation between centroids having indexes corresponding to a portion of a distance table (\eg centroids having indexes between \texttt{00} and \texttt{0f}). On the contrary, our optimized assignment ensures that all indexes corresponding to a given portion (e.g., \texttt{00} to \texttt{0f}) are assigned to centroids close to each other, as shown on Figure~\ref{fig:centroids}. Centroids corresponding to the same portion have the same background color. For the sake of clarity, Figure~\ref{fig:centroids} shows 4 portions of 4 indexes, but in practice we have 16 portions of 16 indexes. This optimized assignment is beneficial because it is likely that a query sub-vector close to a given centroid will also be close to nearby centroids. Therefore, all values in a given portion of a distance table will be close. This allows computing minimum tables with higher values, and thus tighter lower bounds. To obtain this optimized assignment, we group centroids into 16 clusters of 16 elements each using a variant of k-means that forces groups of same sizes~\cite{SSZKmeans}. Centroids in the same cluster are given consecutive indexes, corresponding to one portion of a distance table. This optimized assignment of centroid indexes replaces the arbitrary assignement applied while learning sub-quantizers.

\subsection{Quantization of Distances}
\label{sec:fastpq-floatquant}

\begin{figure}[ht]
  \centering
  \input{fastpq/fig/quantization}
  \caption{Selection of quantization bounds\label{fig:keepscan}}
\end{figure}
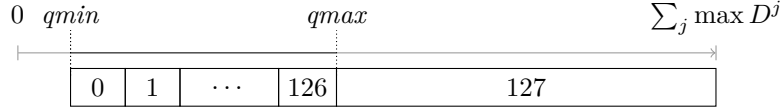

\paragraph{}
The vector grouping and minimum tables techniques are used to build tables of 16 elements of 32 bits each, from the original $D^j$ distance tables (256 $\times$ 32 bits). So that these tables can be used as small tables, we also need to shrink each element to 8 bits. To do so, we quantize floating-point distances to 8-bit integers. As there is no SIMD instruction to compare \emph{unsigned} 8-bit integers, we quantize distances to \emph{signed} 8-bit integers, only utilizing their positive range, \ie 0-127. We quantize floating-point distances between a $\mathit{qmin}$ and a $\mathit{qmax}$ bound into $n=127$ bins. The size of each bin is $(\mathit{qmax}-\mathit{qmin})/n$ and the bin number (0-126) is used as a representation value for the quantized float. All distances above  $\mathit{qmax}$ are quantized to 127 (Figure~\ref{fig:keepscan}).

\paragraph{}
We set $\mathit{qmin}$ to the minimum value across all distance tables, which is the smallest distance we need to represent. Setting $\mathit{qmax}$ to the maximum possible distance, \ie the sum of the maximums of all distance tables, results in a high quantization error. Therefore, to determine $\mathit{qmax}$, we find a \emph{temporary} nearest neighbor of the query vector among the $\mathit{init}\%$ first vectors of the database (usually, $\mathit{init} \approx 1\%$) using the conventional ADC Scan procedure. We then use the distance between the query vector and this temporary nearest neighbor as $\mathit{qmax}$ bound. We do not need to represent distances higher than this distance because all future nearest neighbor candidates will be closer to the query vector than this temporary nearest neighbor. This choice of $\mathit{qmin}$ and $\mathit{qmax}$ bounds allows us to represent a small but relevant range of distances (Figure \ref{fig:keepscan}). Quantization error is therefore minimal, as confirmed by our experimental results (Section~\ref{sec:fastpq-quantonly}). Lastly, to avoid integer overflow issues, we use \emph{saturated} SIMD additions.

\subsection{Lookups in Small tables}
\label{sec:fastpq-lookup}
\paragraph{}
For the sake of clarity, we do not fully describe the SIMD implementation of PQ Fast Scan. Instead, we focus on describing which small tables are used, and how they are indexed to compute lower bounds. The first four small tables, $S^0,\dotsc,S^3$ correspond to quantized portions of $D^0,\dotsc,D^3$. We load these quantized portions into SIMD registers before scanning each group, as shown by the solid arrows on Figure \ref{fig:global-lookup}. Thus, two different groups use different small tables $S^0,\dotsc,S^3$. On the contrary, the last four small tables, $S^4\dotsc,S^7$, built by computing minimum tables do not change and are used to scan the whole database. They are loaded into SIMD registers at the beginning of the scan process. 

\newcommand{\lookuptikzscale}{0.7}
\begin{figure}[b]
  \centering
  \includegraphics[scale=\lookuptikzscale]{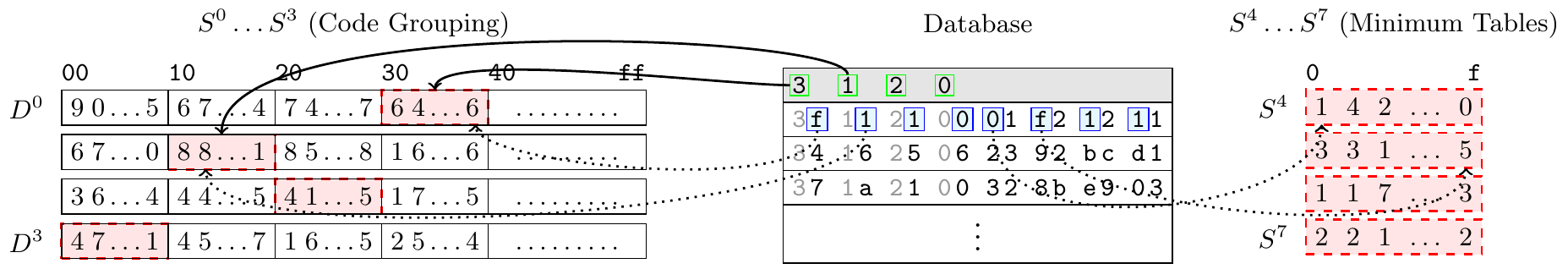}
  \caption{Use of small tables to compute lower bounds\label{fig:global-lookup}}
\end{figure}

\paragraph{}
As small tables contain 16 values, they are indexed by 4 bits. Given a database vector $p$, we use the 4 \emph{least} significant bits of $c[0],\dotsc,c[3]$ to index values in $S^0,\dotsc,S^3$ and the 4 \emph{most} significant bits of $c[4],\dotsc,c[7]$ to index values in $S^4,\dotsc,S^7$. Indexes are circled on Figure \ref{fig:global-lookup} (the 4 most significant bits correspond to the first hexadecimal digit, and the 4 least significant bits to the second hexadecimal digit) and lookups in small tables are depicted by dotted arrows. Lookups depicted by dotted arrows are performed using SIMD in-register shuffles. To efficiently use SIMD in-register shuffles, we need to transpose the vectors in each group, as in the gather implementation of ADC Scan (Section~\ref{sec:simd-issues}). However we do not show this transposition on Figure \ref{fig:global-lookup} for the sake of clarity. To compute the lower bound, we add the 8 looked up values. To decide on pruning distance computations, the lower bound is compared to the \emph{quantized} value of $\mathit{min}$, the distance between the query vector and the current nearest neighbor.

\section{Evaluation}
The aim of this section is twofold: evaluating the performance of PQ Fast Scan and analyzing the parameters that influence it. We show that PQ Fast Scan outperforms ADC Scan by a factor 4-6 in common usage scenarios.

\subsection{Experimental Setup}
\label{sec:expsetup}

\paragraph{}
We implemented PQ Fast Scan in C++ using intrinsics, which allow accessing SIMD instructions from C or C++, without writing assembly code~\cite{IntelCompMan, IntelIntrinsicsGuide}. Our implementation uses 128-bit SIMD instructions from the SSSE3, SSE3 and SSE2 instruction sets. We compared our implementation of PQ Fast Scan with the libpq implementation of the ADC Scan procedure, introduced in Section \ref{sec:memory-accesses}.
On all our test platforms, we used the gcc and g++ compilers version 4.9.2, with the following compilation options: \texttt{-O3 -m64 -march=native -ffast-math}. We released our source code\footnote{\small\url{https://github.com/technicolor-research/pq-fast-scan}} under the Clear BSD license.

\begin{table}[ht]
  \centering
  \caption{Size of inverted lists used for experiments\label{tbl:dbsizes}}
  \begin{tabular}{lllllllll}
    \toprule
    Inverted list ID & 0 & 1 & 2 & 3 & 4 & 5 & 6 & 7 \\
    \midrule
    \# vectors & 25M & 3.4M & 11M & 11M & 11M & 11M & 4M & 23M \\
    \# queries & 2595 & 307 & 1184 & 1032 & 1139 & 1036 & 390 & 2317 \\
    \bottomrule
  \end{tabular}%
\end{table}

\paragraph{}
We evaluate PQ Fast Scan on the largest public dataset of high-dimensional vectors, ANN\_SIFT1B\footnote{\small\url{http://corpus-texmex.irisa.fr/}}. It consists of 3 parts: a learning set of 100 million vectors, a base set of 1 billion vectors and a query set of 10000 vectors. We restricted the learning set for the product quantizer to 10 million vectors. Vectors of this dataset are SIFT descriptors of dimensionality 128. We use two subsets of ANN\_SIFT1B for experiments:
\begin{itemize}
  \item ANN\_SIFT100M1, a subset of 100 million vectors of the base set. We build an index with 8 inverted lists; each query is directed to the most relevant inverted list which is then scanned with PQ Fast Scan and ADC Scan. Table \ref{tbl:dbsizes} summarizes the sizes of the different inverted lists.
  
  \item ANN\_SIFT1B, the full base set of 1 billion vectors to test our algorithm on a larger scale.
\end{itemize}

\paragraph{}
We study the following parameters impacting the performance of PQ Fast Scan:
\begin{itemize}
  \item $\mathit{init}$, the percentage of vectors kept at the beginning of the database (Section \ref{sec:fastpq-floatquant}). Even when using PQ Fast Scan, these vectors are scanned using the conventional ADC Scan procedure to find a temporary nearest neighbor. The distance of the query vector to the temporary nearest neighbor is then used as the $\mathit{qmax}$ value for quantization of distance tables.
  \item $\mathit{r}$, the number of nearest neighbors returned by the search process. For the sake of simplicity, we described ADC Scan and PQ Fast Scan as if they returned a single nearest neighbor. In practice, they return multiple nearest neighbors \eg $\mathit{r}=100$ for information retrieval in multimedia databases. 
  \item \emph{inverted list size}, the number of vectors in the scanned inverted list size.
\end{itemize}

\paragraph{}
We compare the performance of PQ Fast Scan, and the libpq implementation of ADC Scan. We compare their respective \emph{scan speed}, expressed in millions of codes scanned per second (M codes/s). Scan speed is obtained by dividing response times by the size of inverted lists. We do not evaluate PQ Fast Scan accuracy, recall or precision because PQ Fast Scan returns the exact same results as ADC Scan and PQ accuracy has already been extensively studied~\cite{Jegou2011}. In addition to theoretical guarantees, we checked that PQ Fast Scan returned the same results as the libpq implementation of ADC Scan for every experiment. Lastly, we run PQ Fast Scan across a variety of different platforms (Table \ref{tbl:fastpq-config}) and demonstrate it consistently outperforms ADC Scan by a factor of 4-6. All experiments were run on a \emph{single processor core}. Unless otherwise noted, experiments were run on laptop (A) (Table \ref{tbl:fastpq-config}).

\subsection{Distribution of Response Times}

\begin{table}
  \centering
  \caption{Response time distribution\label{tbl:runtimedist}}
  \begin{tabular}{llllll}
    \toprule
    & Mean & 25\% & Median & 75\% & 95\% \\
    \midrule
    ADC Scan (libpq) & 73.9 & 73.6 & 73.8 & 74.0 & 74.5 \\
    PQ Fast Scan & 13.7 & 12.3 & 12.9 & 14.1 & 18.0 \\
    \textbf{Speedup} & \textbf{5.4} & \textbf{6.0} & \textbf{5.7} & \textbf{5.2} & \textbf{4.1} \\
    \bottomrule
  \end{tabular}
\end{table}

\paragraph{}
We study the distribution of PQ Fast Scan response times. Contrary to ADC Scan, PQ Fast Scan response time varies with the query vector. Indeed, PQ Fast Scan performance depends on the amount of asymmetric distance computations that can be pruned, which depends on the query vector.
Figure \ref{fig:fastpq-runtimedist} shows the distribution of response times of 2595 nearest neighbor queries executed on the inverted list 0. As expected, ADC Scan response time is almost constant across different query vectors. PQ Fast Scan response time is more dispersed, but it responds to the bulk of queries 4-6 times faster than ADC Scan, as shown in Table \ref{tbl:runtimedist}.
In the remainder of this section, when studying the impact of different parameters on PQ Fast Scan performance, we plot \emph{median} response times or \emph{median} scan speeds. We use the \emph{\nth{1} quartile} (\nth{25} percentile) and \emph{\nth{3} quartile} (\nth{75} percentile) to draw error bars. Because it directly impacts performance, we also plot the percentage of pruned distance computations.

\subsection{Performance Counters}
\label{sec:perfcounters}

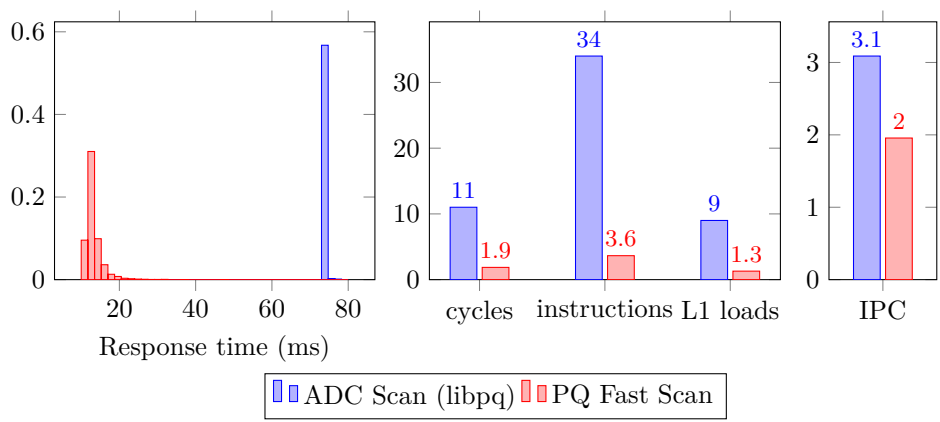
\begin{figure}
  \centering
  \begin{tikzpicture}[baseline, trim axis left]
  \begin{axis}[
  ybar,
  height=5 cm,
  xlabel=Response time (ms),
  ymin=0,
  ]
  
  \addplot+[hist={bins=40, data min=10, data max=80, density}] table [y=pq_ms] {fastpq/plots/time_distribution.gnuplot};
  
  \addplot+[hist={bins=40, data min=10, data max=80, density}] table [y=fast_pq_ms] {fastpq/plots/time_distribution.gnuplot};
  
  \end{axis}
  \end{tikzpicture}
  \begin{tikzpicture}[baseline]
  \begin{axis}[
  ybar,
  height=5cm,
  width=6.2cm,
  ymin= 0,
  enlarge x limits=0.2,
  symbolic x coords={cycles, instructions, l1_loads},
  xtick=data,
  xticklabels={cycles, instructions, L1 loads},
  nodes near coords={\pgfmathprintnumber[fixed, precision=1]{\pgfplotspointmeta}},
  every node near coord/.append style={font=\small},
  enlarge y limits = {abs=3ex, upper},
  ]
  
  \addplot+ table[x=counter, y=pq_median] {fastpq/plots/perf_counters.gnuplot};
  \addplot+ table[x=counter, y=fast_pq_median] {fastpq/plots/perf_counters.gnuplot};
  \end{axis}
  \end{tikzpicture}
  \begin{tikzpicture}[baseline, trim axis right]
  \begin{axis}[
  ybar,
  height=5 cm,
  width=3cm,
  symbolic x coords={ipc},
  xtick=data,
  xticklabels={IPC},
  ymin = 0,
  nodes near coords={\pgfmathprintnumber[fixed, precision=1]{\pgfplotspointmeta}},
  every node near coord/.append style={font=\small},
  enlarge y limits = {abs=3ex, upper},
  legend to name = legend-timeperf,
  legend columns = 4,
  ]
  \addplot+ table[x=counter, y=pq_median] {fastpq/plots/ipc.gnuplot};
  \addplot+ table[x=counter, y=fast_pq_median] {fastpq/plots/ipc.gnuplot};
  \legend{ADC Scan (libpq), PQ Fast Scan}
  \end{axis}
  \end{tikzpicture}
  \\
  \pgfplotslegendfromname{legend-timeperf}
  \caption{Distribution of response times, performance counters and IPC\label{fig:fastpq-runtimedist}}
\end{figure}

\paragraph{}
We use performance counters to measure the usage of CPU resources of PQ Fast Scan and ADC Scan when scanning the inverted list 0 (Figure~\ref{fig:fastpq-runtimedist}). Thanks to the use of register-resident small tables, PQ Fast Scan only performs 1.3 L1 loads per scanned vector, where the libpq implementation of ADC Scan requires 9 L1 loads. PQ Fast Scan requires 89\% less instructions than ADC Scan thanks to the use of SIMD instructions instead of scalar ones (respectively 3.7 and 34 instructions per scanned vector). PQ Fast Scan uses 83\% less cycles than ADC Scan (respectively 1.9 and 11 cycles per vector).
The decrease in cycles is slightly less than the decrease in instructions because PQ Fast Scan has a lower IPC than ADC Scan. This is because SIMD instructions can be less easily pipelined than scalar instructions.

\newcommand{\keep}{$\mathit{init}$\xspace}
\newcommand{\topk}{$\mathit{r}$\xspace}

\subsection{Impact of init and r Parameters}

\begin{figure}
  \centering
  \begin{tikzpicture}[baseline, trim axis left]
  \begin{semilogxaxis} [
    height=5cm,
    ylabel=Pruned (\%),
    xlabel=init (\%),
    error bars/y dir=both,
    error bars/y explicit,
  ]
  \addplot+[smooth] table[
    x=keep_percent,
    y=pruned_bh100_median,
    y error plus expr=\thisrow{pruned_bh100_quart3}-\thisrow{pruned_bh100_median},
    y error minus expr=\thisrow{pruned_bh100_median}-\thisrow{pruned_bh100_quart1}
  ] {fastpq/plots/keep_pruned_speed.gnuplot};
  \addplot+[smooth] table[
    x=keep_percent,
    y=pruned_bh1000_median,
    y error plus expr=\thisrow{pruned_bh1000_quart3}-\thisrow{pruned_bh1000_median},
    y error minus expr=\thisrow{pruned_bh1000_median}-\thisrow{pruned_bh1000_quart1}
  ] {fastpq/plots/keep_pruned_speed.gnuplot};
  \end{semilogxaxis}
  \end{tikzpicture}
  \begin{tikzpicture}[baseline, trim axis right]
  \begin{semilogxaxis} [
    height=5cm,
    ylabel=Scan speed (M codes/s),
    xlabel=init (\%),
    legend entries={PQ Fast Scan (r=100), PQ Fast Scan (r=1000),
      ADC Scan (libpq r=100), ADC Scan (libpq r=1000)},
    legend to name=legend-keep,
    legend columns=2,
    error bars/y dir=both,
    error bars/y explicit,
  ]
  \addplot+[smooth] table[
    x=keep_percent,
    y=speed_fast_pq_bh100_median,
    y error plus expr=\thisrow{speed_fast_pq_bh100_quart3} -\thisrow{speed_fast_pq_bh100_median},
    y error minus expr=\thisrow{speed_fast_pq_bh100_median}-\thisrow{speed_fast_pq_bh100_quart1}
  ] {fastpq/plots/keep_pruned_speed.gnuplot};
  \addplot+[smooth] table[
    x=keep_percent,
    y=speed_fast_pq_bh1000_median,
    y error plus expr=\thisrow{speed_fast_pq_bh1000_quart3}-\thisrow{speed_fast_pq_bh1000_median},
    y error minus expr=\thisrow{speed_fast_pq_bh1000_median}-\thisrow{speed_fast_pq_bh1000_quart1}
  ] {fastpq/plots/keep_pruned_speed.gnuplot};
  \addplot+[smooth] table[
    x=keep_percent,
    y=speed_pq_bh100_median,
    y error plus expr=\thisrow{speed_pq_bh100_quart3}-\thisrow{speed_pq_bh100_median},
    y error minus expr=\thisrow{speed_pq_bh100_median}-\thisrow{speed_pq_bh100_quart1}
  ] {fastpq/plots/keep_pruned_speed.gnuplot};
  \addplot+[smooth] table[
    x=keep_percent,
    y=speed_pq_bh1000_median,
    y error plus expr=\thisrow{speed_pq_bh1000_quart3}-\thisrow{speed_pq_bh1000_median},
    y error minus expr=\thisrow{speed_pq_bh1000_median}-\thisrow{speed_pq_bh1000_quart1}
  ] {fastpq/plots/keep_pruned_speed.gnuplot};
  \end{semilogxaxis}
  \end{tikzpicture}
  \\
  \pgfplotslegendfromname{legend-keep}
  \caption{Impact of init Parameter (all inverted lists)\label{fig:fastpq-keep}}
\end{figure}
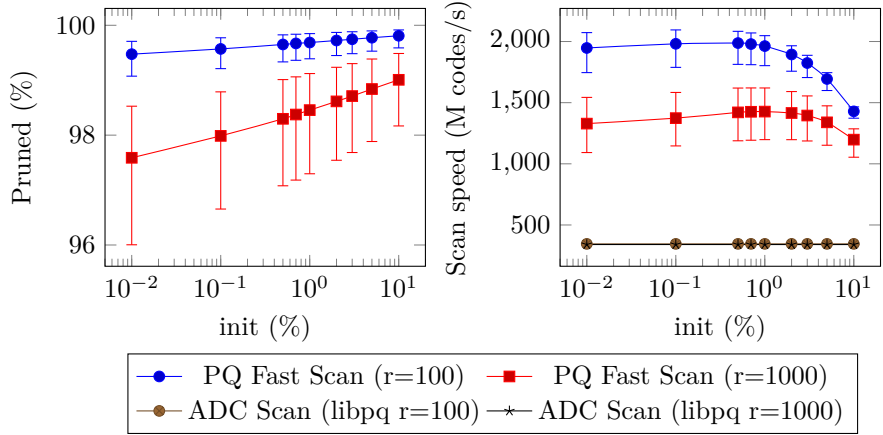

\paragraph{}
Both \keep and \topk impact the amount of pruned distance computations, and thus PQ Fast Scan performance. For information retrieval in multimedia databases, \topk is often set between 100 and 1000. Therefore, we start by studying the impact of \keep for $\mathit{r}=100$ and $\mathit{r}=1000$. The \keep parameter impacts the tightness of the $\mathit{qmax}$ bound used for quantization. A higher \keep value means more vectors are scanned using the conventional ADC Scan procedure to find a temporary nearest neighbor (Section~\ref{sec:fastpq-floatquant}).
This makes the $\mathit{qmax}$ bound tighter and decreases the distance quantization error. Figure~\ref{fig:fastpq-keep} shows that the pruning power increases with \keep; however this increase is moderate.
For $\mathit{r}=1000$, the pruning power is lower than for $\mathit{r}=100$ and more sensitive to \keep. PQ Fast Scan can prune a distance computation if the lower bound of the currently scanned vector is higher than the distance between the query vector and the current $r$th nearest neighbor. A higher \topk value implies a higher distance between the query vector and the $r$th nearest neighbor. Therefore, less distance computations can be pruned.

\begin{figure}[b]
  \centering
  \begin{tikzpicture}[baseline, trim axis left]
  \begin{semilogxaxis} [
    height=5cm,
    ylabel=Pruned (\%),
    xlabel=$r$,
    error bars/y dir=both,
    error bars/y explicit,
    cycle list shift=1
  ]
  \addplot+[smooth] table[
    x=bh_size,
    y=pruned_median,
    y error plus expr=\thisrow{pruned_quart3}-\thisrow{pruned_median},
    y error minus expr=\thisrow{pruned_median}-\thisrow{pruned_quart1}
  ] {fastpq/plots/topk_pruned_speed.gnuplot};
  \end{semilogxaxis}
  \end{tikzpicture}
  \begin{tikzpicture}[baseline, trim axis right]
  \begin{semilogxaxis} [
    height=5cm,
    ylabel=Scan speed (M codes/s),
    xlabel=$r$,
    legend to name=legend-topk,
    legend columns=2,
    error bars/y dir=both,
    error bars/y explicit,
  ]
  \addplot+[smooth] table[
    x=bh_size,
    y=pq_speed_median,
    y error plus expr=\thisrow{pq_speed_quart3}-\thisrow{pq_speed_median},
    y error minus expr=\thisrow{pq_speed_median}-\thisrow{pq_speed_quart1}
  ] {fastpq/plots/topk_pruned_speed.gnuplot};
  \addplot+[smooth] table[
    x=bh_size,
    y=fast_pq_speed_median,
    y error plus expr=\thisrow{fast_pq_speed_quart3}-\thisrow{fast_pq_speed_median},
    y error minus expr=\thisrow{fast_pq_speed_median}-\thisrow{fast_pq_speed_quart1}]
  {fastpq/plots/topk_pruned_speed.gnuplot};
  \legend{ADC Scan (libpq), PQ Fast Scan}
  \end{semilogxaxis}
  \end{tikzpicture}
  \\
  \pgfplotslegendfromname{legend-topk}
  \caption{Impact of r Parameter (all inverted lists, init=0.5\%)\label{fig:fastpq-topk}}
\end{figure}
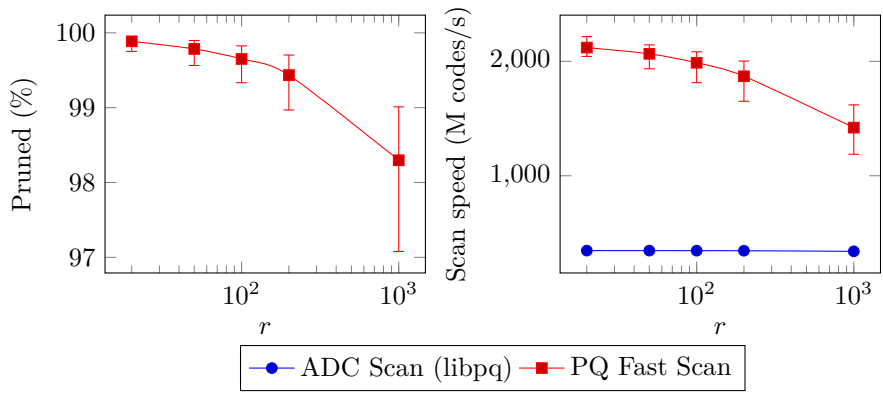

\paragraph{}
The scan speed increases slightly with \keep as more distance computations get pruned, up to a  threshold where it starts to collapse. After this threshold, the increase in pruned distance computations provided by the tighter $\mathit{qmax}$ bound is outweighed by the increased time spent scanning the first $\mathit{init}\%$ vectors using the slow ADC Scan procedure. Overall, PQ Fast Scan is not very sensitive to \keep, and a decent $\mathit{qmax}$ bound is found quickly. Any \keep value between 0.1\% and 1\% is suitable. We set $\mathit{init}=0.5$\% for the remainder of experiments. Lastly, we evaluate PQ Fast Scan performance for more \topk values. Figure~\ref{fig:fastpq-topk} confirms that PQ Fast Scan performance decreases with \topk.

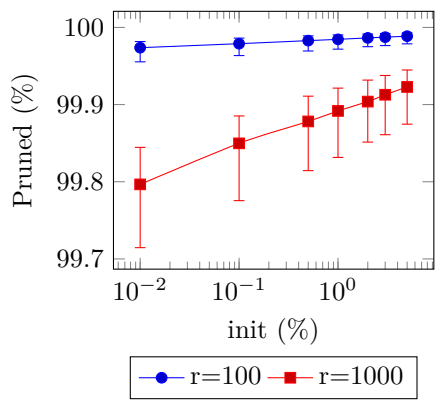
\begin{figure}
  \centering
  \begin{tikzpicture}[baseline, trim axis left, trim axis right]
  \begin{semilogxaxis} [
    height=5cm,
    legend to name=legend-quantonly,
    legend columns=2,
    xlabel=init (\%),
    ylabel=Pruned (\%),
    error bars/y dir=both,
    error bars/y explicit,
  ]
  \addplot+[smooth] table[
    x=keep_percent,
    y=pruned_bh100_median,
    y error plus expr=\thisrow{pruned_bh100_quart3}-\thisrow{pruned_bh100_median},
    y error minus expr=\thisrow{pruned_bh100_median}-\thisrow{pruned_bh100_quart1}
  ] {fastpq/plots/quant_only.gnuplot};
  \addplot+[smooth] table[
    x=keep_percent,
    y=pruned_bh1000_median,
    y error plus expr=\thisrow{pruned_bh1000_quart3}-\thisrow{pruned_bh1000_median},
    y error minus expr=\thisrow{pruned_bh1000_median}-\thisrow{pruned_bh1000_quart1}
  ] {fastpq/plots/quant_only.gnuplot};
  \legend{r=100, r=1000}
  \end{semilogxaxis}
  \end{tikzpicture}
  \\
  \pgfplotslegendfromname{legend-quantonly}
  \caption{Pruning power using quantization only (all inverted lists)\label{fig:fastpq-quantonly}}
\end{figure}

\subsection{Impact of Distance Quantization}
\label{sec:fastpq-quantonly}

\paragraph{}
PQ Fast Scan uses three techniques to build small tables: (1) vector grouping, (2) minimum tables and (3) quantization of distances. Among these three techniques, minimum tables and quantization of distances impact the tightness of lower bounds, and therefore pruning power. To assess the respective impact on pruning power of these two techniques, we implement a quantization-only version of PQ Fast Scan which relies \emph{only} on quantization of distances (Figure~\ref{fig:fastpq-quantonly}). This version uses tables of 256 8-bit integers, while the full version of PQ Fast Scan uses tables of 16 8-bit integers. Therefore, the quantization-only version cannot use SIMD and offers no speedup. Hence, Figure~\ref{fig:fastpq-quantonly} shows only the pruning power and does not show the scan speed. The quantization-only version of PQ Fast Scan achieves 99.9\% to 99.97\% pruning power. This is higher than the pruning power of the full version of PQ Fast Scan (\ie using the three techniques) which is 98\% to 99.7\% (Figure~\ref{fig:fastpq-keep}). This demonstrates that our quantization scheme is highly efficient and that \emph{most of the loss of pruning power comes from minimum tables}.

\begin{figure}[b]
  \centering
  \begin{tikzpicture}[baseline, trim axis left]
  \begin{axis} [
    height=5cm,
    ybar,
    symbolic x coords={0,7,2,4,5,3,6,1},
    xtick=data,
    xlabel=Inverted list ID,
    ylabel=Pruned (\%),
    bar width=4pt,
    ymin=90,
    error bars/y dir=both,
    error bars/y explicit,
    cycle list shift=1
  ]
  \addplot+ table[
    x=cluster_id,
    y=pruned_median,
    y error plus expr=\thisrow{pruned_quart3}-\thisrow{pruned_median},
    y error minus expr=\thisrow{pruned_median}-\thisrow{pruned_quart1}
  ] {fastpq/plots/database_pruned_speed.gnuplot};
  \end{axis}
  \end{tikzpicture}
  \begin{tikzpicture}[baseline, trim axis right]
  \begin{axis} [
    height=5cm,
    ybar,
    symbolic x coords={0,7,2,4,5,3,6,1},
    xtick=data,
    xlabel=Inverted list ID,
    ylabel=Scan speed (M codes/s),
    bar width=4 pt,
    legend to name=legend-invsize,
    legend columns=2,
    error bars/y dir=both,
    error bars/y explicit,
  ]
  \addplot+ table[
    x=cluster_id,
    y=pq_speed_median,
    y error plus expr=\thisrow{pq_speed_quart3}-\thisrow{pq_speed_median},
    y error minus expr=\thisrow{pq_speed_median}-\thisrow{pq_speed_quart1}
  ] {fastpq/plots/database_pruned_speed.gnuplot};
  \addplot+ table[
    x=cluster_id,
    y=fast_pq_speed_median,
    y error plus expr=\thisrow{fast_pq_speed_quart3}-\thisrow{fast_pq_speed_median},
    y error minus expr=\thisrow{fast_pq_speed_median}-\thisrow{fast_pq_speed_quart1}
  ] {fastpq/plots/database_pruned_speed.gnuplot};
  \legend{ADC Scan (libpq), PQ Fast Scan}
  \end{axis}
  \end{tikzpicture}
  \\
  \pgfplotslegendfromname{legend-invsize}
  \caption{Impact of inverted list size (init=0.5\%, r=100)\label{fig:fastpq-size}}
\end{figure}
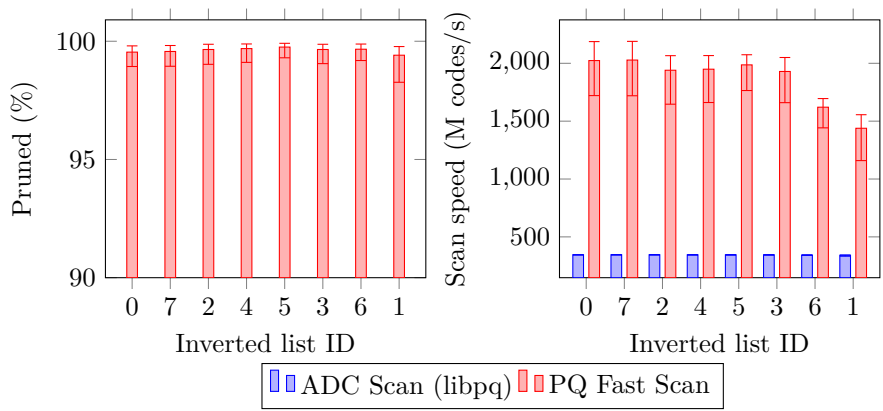

\subsection{Impact of the Size of Inverted Lists}
\label{sec:impact-size}

\paragraph{}
The size of inverted lists impacts scan speed without impacting pruning power (Figure~\ref{fig:fastpq-size}). Inverted lists 0, 7, 2, 4, 5 and 3 have sizes comprised between 10 million vectors and 25 million vectors, and PQ Fast Scan speed is almost constant across all theses inverted lists. Smaller inverted lists, \eg inverted lists 6 and 1, exhibit lower scan speeds. PQ Fast Scan groups vectors on 4 components for inverted lists exceeding 3 million vectors,  (Section~\ref{sec:code-grouping}). As inverted lists sizes approaches this threshold, the scan speed decreases. This is because the inverted list size $s$ approaches the minimum inverted list size of $s_{min} = 50$ codes. Small tables are loaded too frequently in comparison with the number of codes scanned, which impacts performance.

\subsection{Large Scale Experiment}
\label{sec:large-scale}

\paragraph{}
We test PQ Fast Scan on the full database of 1 billion vectors (ANN\_SIFT1B). For this database, we build an index which divides the database into 128 inverted lists. Inverted lists therefore have an average size of about 8 million vectors. We run 10000 NN queries. The most appropriate inverted list for each query is selected using the index, and scanned to find nearest neighbors. We scan inverted lists using both ADC Scan and PQ Fast Scan, and we compare mean response times to queries (Figure \ref{fig:fastpq-otherarch}, SIFT1B). In addition to its lower response time, PQ Fast Scan also allows decreasing the amount of memory consumed by the database thanks to vector grouping (Section \ref{sec:code-grouping}). Unlike previous experiments, this experiment was run on workstation (B) instead of laptop (A) (Table \ref{tbl:fastpq-config}). The parameters $\mathit{init}=1\%$, $\mathit{r}=100$ were chosen.

\begin{table}
  \centering
  \caption{Systems\label{tbl:fastpq-config}}
  \begin{tabular}{lllll}
    \toprule
    & laptop (A) & workstation (B) & server (C) & server (D) \\
    \midrule
    CPU Model & Core  & Xeon  & Xeon & Xeon \\
    & i7-4810MQ & E5-2609v2 & E5-2640 & X5570 \\
    CPU Arch. & Haswell & Ivy Bridge & Sandy Bridge & Nehalem \\
    CPU Freq. & 2.8-3.8 Ghz & 2.5-2.5 Ghz & 2.5-3.0 Ghz & 2.9-3.33 Ghz \\
    \midrule
    Mem. & 8 GB & 16 GB & 64 GB & 24 GB \\
    & ($2\times4$ GB) & ($4\times4$ GB) & ($4\times16$ GB) & ($6\times4$ GB) \\
    Mem. Type & DDR3 & DDR3 & DDR3 & DDR3 \\
    Mem. Freq. & 1600 Mhz & 1333 Mhz & 1333 Mhz & 1066 Mhz \\
    \bottomrule
  \end{tabular}
\end{table}

\subsection{Impact of CPU Architecture}

\begin{figure}
  \centering
  \begin{tikzpicture}[baseline, trim axis left]
  \begin{axis} [
  height=5cm,
  width=3cm,
  ybar,
  symbolic x coords={workstation},
  xtick=data,
  xticklabels={(B)},
  ylabel=Memory use (GiB),
  ymin=0
  ]
  \addplot+ table[
  x=architecture,
  y=pq_memory,
  ] {fastpq/plots/large_scale.gnuplot};
  \addplot+ table[
  x=architecture,
  y=fast_pq_memory,
  ] {fastpq/plots/large_scale.gnuplot};
  \end{axis}
  \end{tikzpicture}
  \begin{tikzpicture}[baseline]
  \begin{axis} [
  height=5cm,
  width=3cm,
  ybar,
  symbolic x coords={workstation},
  xtick=data,
  xticklabels={(B)},
  ylabel=Mean response time (ms),
  ymin=0,
  ]
  \addplot+ table[
  x=architecture,
  y=pq_time,
  ] {fastpq/plots/large_scale.gnuplot};
  \addplot+ table[
  x=architecture,
  y=fast_pq_time,
  ] {fastpq/plots/large_scale.gnuplot};
  \end{axis}
  \end{tikzpicture}
  \begin{tikzpicture}[baseline, trim axis right]
  \begin{axis} [
  height=5cm,
  ybar,
  ymin=0,
  symbolic x coords={laptop, workstation, rennstorm, parapide},
  xtick=data,
  xticklabels={(A), (B), (C), (D)},
  enlarge x limits=0.25,
  ylabel=Scan speed (M codes/s),
  legend to name=legend-arch,
  legend columns=2,
  error bars/y dir=both,
  error bars/y explicit,
  ]
  \addplot+ table[
  x=architecture,
  y=pq_speed_median,
  y error plus expr=\thisrow{pq_speed_quart3}-\thisrow{pq_speed_median},
  y error minus expr=\thisrow{pq_speed_median}-\thisrow{pq_speed_quart1}
  ] {fastpq/plots/architectures.gnuplot};
  \addplot+ table[
  x=architecture,
  y=fast_pq_speed_median,
  y error plus expr=\thisrow{fast_pq_speed_quart3}-\thisrow{fast_pq_speed_median},
  y error minus expr=\thisrow{fast_pq_speed_median}-\thisrow{fast_pq_speed_quart1}
  ] {fastpq/plots/architectures.gnuplot};
  \legend{ADC Scan (libpq), PQ Fast Scan}
  \end{axis}
  \end{tikzpicture}
  \\
  \pgfplotslegendfromname{legend-arch}
  \caption{Experiments on other CPU architectures (see Table~\ref{tbl:fastpq-config})\label{fig:fastpq-otherarch}}
\end{figure}
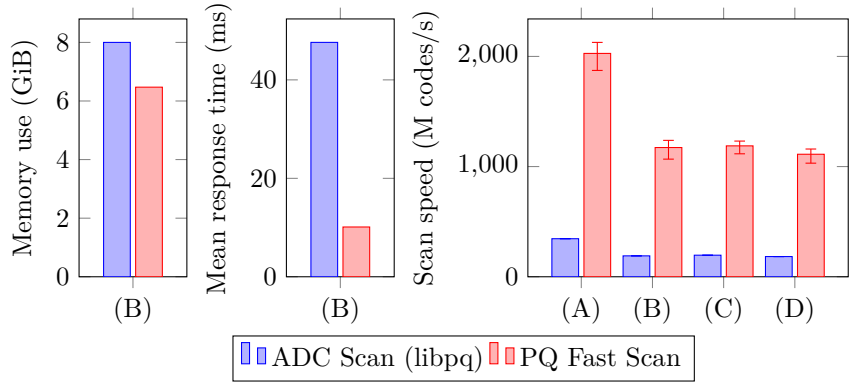

\paragraph{}
To conclude our evaluation section, we compare PQ Fast Scan and ADC Scan over a wide range of using processors released between 2009 and 2014 (Table \ref{tbl:fastpq-config}). On all these systems, PQ Fast Scan median speed exceeds ADC Scan median speed by a factor of 4-6, thus validating our performance analysis and design hypotheses (Figure~\ref{fig:fastpq-otherarch}, Scan speed). In addition, PQ Fast Scan performance is not sensitive to processor architecture. PQ Fast Scan loads 6 bytes from memory for each lower bound computation. Thus, a scan speed of 2000 M codes/s correspond to a bandwidth use of 16 GB/s. The memory bandwidth of Intel server processors ranges from 40 GB/s to 70 GB/s. When answering 8 queries concurrently on an 8-core server processor, PQ Fast Scan is bound by the memory bandwidth, thus demonstrating its highly efficient use of CPU resources.

\section{Discussion}

\subsection{Compatibility with Inverted Indexes}
\label{sec:pq-comp-invindex}

\paragraph{}
The main limitation of PQ Fast Scan is that it is not compatible with the most advantageous configurations of inverted indexes. Inverted indexes are commonly combined with product quantization to decrease response time and increase accuracy.  It has been shown that using an inverted index that partitions the database into a large number of small inverted lists (fine partitioning) offers better results than an inverted index that partitions the database into a small number of inverted lists (Section~\ref{sec:invindex}). For large datasets (\eg 1 billion vectors), it is common to use inverted indexes that partitions the database into $K=8192-65536$ inverted lists. Inverted lists therefore have a size of $s=15000-200000$ codes. On the other hand, PQ Fast Scan requires inverted lists of at least 3 million vectors (Section~\ref{sec:code-grouping}), and works better with larger inverted lists (Section~\ref{sec:impact-size}). In practice, this makes it difficult to combine the speedup provided by fine inverted indexes and the speedup provided by PQ Fast Scan. In our experiments  on a 1 billion vector dataset (Section~\ref{sec:large-scale}), we used an inverted index with only $K=128$ inverted lists. The mean size of inverted lists is therefore of 8 million codes, ideal for PQ Fast Scan. Combining a finer inverted index (\eg, with $K=8192-65536$  inverted lists) with PQ Fast Scan would have offered an even better speedup, but this is currently not possible.

\paragraph{}
Two factors mitigate this issue. First, in very large databases \ie exceeding hundreds of billions of vectors, inverted lists may exceed 3 million codes, even if a fine inverted index is used. Second, the upcoming AVX-512 SIMD instruction set will allow PQ Fast Scan to work with smaller inverted lists. The constraints on inverted lists sizes imposed by PQ Fast Scan come from code grouping (Section~\ref{sec:code-grouping}). We showed that the minimum inverted list size is given by $n_{min}(g)=s_{min}{\cdot}P^g$, where $s_{min}$ is the minimum number of codes per group (in our case $s_{min}=50$), and $P$ is the number of portions in lookup tables. The current version of PQ Fast Scan relies on 128-bit in-register shuffles, which allows for small tables of 16 elements ($16{\stimes}8$ bits). Lookup tables therefore have to be divided into $P=256/16=16$ portions, which leads to a minimum inverted list size $n_{min}(4) = 50 \cdot 16^4 $ of 3 million codes. The AVX-512 version of PQ Fast Scan would use 512-bit in-register shuffles, and allow for small tables of 64 elements ($64{\stimes}8$ bits). Lookup tables would therefore be divided in $P=256/64=4$ portions, which would lead to a minimum inverted list size $n_{min}(4) = 50 \cdot 4^4$ of 12000 codes. Moreover, using minimum tables of 64 elements instead of 16 would allow grouping on 3 components instead of 4 components without impacting pruning power too strongly. The minimum inverted list size $n_{min}(3)$ of the AVX-512 version of PQ Fast Scan may be as low as 3000 codes. Therefore, the AVX-512 version of PQ Fast Scan would be compatible with almost all configurations of inverted indexes.

\subsection{Applicability to Product Quantization Derivatives}
\label{sec:fastpq-deriv-discuss}

\paragraph{}
As currently designed, PQ Fast Scan is only compatible with $m{\stimes}b=8{\stimes}8$ product quantizers (64-bit codes). Minor adjustments may however make it compatible with other $m{\stimes}8$ product quantizers, \eg $16{\stimes}8$ product quantizers (128-bit codes). In all cases, it is only possible to use code grouping for $g=4$ components (Section~\ref{sec:code-grouping}). For $16{\stimes}8$ product quantizers, this means that we would need to use the minimum tables technique on $m-g=16-4=12$ tables. Experiments would be required to determine if PQ Fast Scan still has a high enough pruning power to offer a significant speedup in this scenario. Making PQ Fast Scan compatible with $m{\stimes}8$ product quantizers with $m < 8$, \eg $4{\stimes}8$ product quantizers (32-bit codes) would be easy. These cases are indeed less challenging than $8{\stimes}8$ product quantizers. Code grouping can still be used to build $g=4$ (or less) small tables. This means that minimum tables have to be computed for only $m-g < 4$ tables and therefore guarantees that PQ Fast Scan will have a high enough pruning power. However, this type of codes is still rarely used. Making PQ Fast Scan compatible with $b$ parameters other than $b=8$ would not make much sense, as it would require changing the whole algorithm. PQ Fast Scan has been designed to work with lookup tables of 256 floating-point values, resulting from the use of 8-bit sub-quantizers.

\paragraph{}
In Section~\ref{sec:pq-derivatives}, we presented derivatives of product quantization, that have been recently introduced: Optimized Product Quantization (OPQ), Additive Quantization (AQ), Tree quantization (TQ) or composite quantization (CQ). Like product quantization, these approaches represent high-dimensional vectors by a combination of $m$ centroids taken from $m$ codebooks of $2^b$ centroids each. All these approaches offer a lower quantization error than product quantization. These approaches use (1) a different process to learn quantizer codebooks, and (2) a different ADC procedure from the one of product quantization. We discuss the applicability of PQ Fast Scan to these derivatives of product quantization.

\paragraph{}
The codebook learning process of OPQ, AQ, TQ and CQ has only been tested with 8-bit sub-quantizers ($m{\stimes}8$ codes). PQ Fast Scan also relies on the use of 8-bit sub-quantizers, thus it requires no changes to the codebook learning process of these product quantization derivatives. Besides, OPQ and CQ use the same ADC procedure as product quantization. Therefore, the adaptation of PQ Fast Scan to OPQ or CQ is straightforward. The case of AQ and TQ is more challenging. The ADC process of AQ requires about $m+m^2/2$ table lookups and additions, while the ADC process of TQ requires about $2{\cdot}m$ tables lookups and additions. In all cases, PQ Fast Scan groups codes on $g = 4$ components and computes minimum tables for the remaining tables. For AQ, this means that $g + g^2/2$ small tables would be built using code grouping, and $(m - g) + (m - g)^2/2$ small tables would be built computing minimum tables. Likewise, for TQ, $2g$ tables would be built using code grouping and $2(m-g)$ small tables would be built computing minimum tables. In both cases, experiments are required to determine if this combination of code grouping and minimum tables provides a high enough pruning power. If this is the case, PQ Fast Scan would be able to provide a higher speedup for AQ and TQ than for PQ. As these derivatives require more additions ($m+m^2/2$ and $2m$) than product quantization, they would benefit more of the speedup offered by SIMD additions.

\stopcontents[chapters]

%% file: fastpq/fig/fastpq_overview.tex
\begin{tikzpicture}

\def\vertsep{0.5cm}
\def\discardsep{1.2cm}

\tikzstyle{computation} = [
  shape=rectangle,
  minimum width=5.5cm,
  align=center,
  draw
]

\tikzstyle{check} = [
  shape=diamond,
  minimum width=4.5cm,
  aspect=3,
  align=center,
  draw
]

\node (input) {$c$};

\node[computation,
  right=1.5cm of input.east,
  anchor=west
] (lbcomp) {Compute lower bound\\(Small tables, SIMD additions)};

\draw[->] (input.east) -- (lbcomp.west);

\node[check,
  below=\vertsep of lbcomp.south,
  anchor=north,
] (check1) {lower bound $<$ $\mathit{min}$};
  
\node[
  left=\discardsep of check1.west,
  anchor=east
] (discard1) {\Large $\otimes$};

\draw[->] (lbcomp.south) -- (check1.north);
\draw[->] (check1.west) -- (discard1.east);

\node[
  right=0.1cm of check1.south,
  inner ysep=0,
  outer ysep=0,
  anchor=north west
] {\small yes};

\node[
  inner xsep=0,
  outer xsep=0,
  right=0cm of check1.west,
  anchor=south east
] {\small no};

\node[shape=rectangle,
draw,
below=\vertsep of check1.south,
anchor=north,
align=center] (pqcomp) {Compute distance\\(L1 cache accesses, Scalar additions)};

\draw[->] (check1.south) -- (pqcomp.north);

\node[check,
  below=\vertsep of pqcomp.south,
  anchor=north] (check2) {distance $<$ $\mathit{min}$};
  
\node[
  left=\discardsep of check2.west,
  anchor=east
] (discard2) {\Large $\otimes$};

\node[
  right=0.1cm of check2.south,
  inner ysep=0,
  outer ysep=0,
  anchor=north west
] {\small yes};

\node[
inner xsep=0,
outer xsep=0,
right=0cm of check2.west,
anchor=south east
] {\small no};

\draw[->] (pqcomp.south) -- (check2.north);
\draw[->] (check2.west) -- (discard2.east);

\node[shape=rectangle,
  below=\vertsep of check2.south,
  anchor=north,
  align=center] (result) {$c$ is the new\\nearest neighbor};

\draw[->] (check2.south) -- (result.north);

\end{tikzpicture}

%% file: fastpq/fig/simd_shuffle.tex
\begin{tikzpicture}
\node[register] (indexes)
{%
  \nodepart[elt]{one}$a[0]$
  \nodepart[elt]{two}$b[0]$
  \nodepart[elt]{three}$c[0]$
  \nodepart[elt]{four}$d[0]$
  \nodepart[elt]{five}$\dots$%
  \nodepart[elt]{six}$p[0]$%
};

\node[register, below=\vertSep of indexes.south, anchor=north] (table)
{%
  \nodepart[elt]{one}$S^0[0]$
  \nodepart[elt]{two}$S^0[1]$
  \nodepart[elt]{three}$S^0[2]$
  \nodepart[elt]{four}$S^0[3]$
  \nodepart[elt]{five}$\dots$%
  \nodepart[elt]{six}$S^0[15]$
};

\node[operator, below=\vertSep of table.south, anchor=north] (shuffle)
{\texttt{simd\_shuffle}};

\node[register, below=\vertSep of shuffle.south, anchor=north] (result)
{%
  \nodepart[elt]{one}$S^0[a[0]]$
  \nodepart[elt]{two}$S^0[b[0]]$
  \nodepart[elt]{three}$S^0[c[0]]$
  \nodepart[elt]{four}$S^0[d[0]]$
  \nodepart[elt]{five}$\dots$
  \nodepart[elt]{six}$S^0[p[0]]$
};

\draw[->] (table.west) -- ++(-\horSep,0) |- (shuffle.west);
\draw[->] (indexes.east) -- ++(\horSep, 0) |- (shuffle.east);
\draw[->] (shuffle.south) -- (result.north);
\node[label, left=\labelSep of indexes.west, anchor=east] {$\mathit{indexes}$};
\node[label, left=\labelSep of table.west, anchor=east] {$\mathit{table}$};
\end{tikzpicture}

%% file: fastpq/fig/small_tables.tex
\begin{tikzpicture}

\def\sep{0.65cm}
\def\vertsep{0.8cm}

\node[
  align=center,
] (dj1) {$D^j$\\$\mathbf{0} \leq j < \mathbf{4}$};

\node[
  draw,
  minimum height=0.7cm,
  align=center,
  right=\sep of dj1
](vecgroup) {Vector grouping};

\node[
  draw,
  minimum height=0.7cm,
  align=center,
  right=\sep of vecgroup
](quant1) {Quantization};

\node[
  align=center,
 right=\sep of quant1
] (sj1) {$S^j$\\$0 \leq j < 4$};

\draw[->] (dj1.east) -- (vecgroup.west) node[above=10pt, midway] {\small 256$\times$32 bits};
\draw[->] (vecgroup.east) -- (quant1.west) node[above=10pt, midway] {\small 16$\times$32 bits};
\draw[->] (quant1.east) -- (sj1.west) node[above=10pt, midway] {\small 16$\times$8 bits};

\node[
  align=center,
  below=\vertsep of dj1,
] (dj2) {$D^j$\\$\mathbf{4} \leq j < \mathbf{8}$};

\node[
  draw,
  minimum height=0.7cm,
  align=center,
  right=\sep of dj2
](mintbl) {Minimum tables};

\node[
  draw,
  minimum height=0.7cm,
  align=center,
  right=\sep of mintbl
](quant2) {Quantization};

\node[
  align=center,
  right=\sep of quant2
] (sj2) {$S^j$\\$4 \leq j < 8$};

\draw[->] (dj2.east) -- (mintbl.west) node[above=10pt, midway] {\small 256$\times$32 bits};
\draw[->] (mintbl.east) -- (quant2.west) node[above=10pt, midway] {\small 16$\times$32 bits};
\draw[->] (quant2.east) -- (sj2.west) node[above=10pt, midway] {\small 16$\times$8 bits};

\end{tikzpicture}

%% file: fastpq/fig/table_portions.tex
\begin{tikzpicture}

\def\z#1{\textcolor{black}{\textbf{#1}}}
\def\longdots{\phantom{0} \dots \dots \dots \phantom{0}}
\def\phantomcell{\phantom{3\; 1} \dots \phantom{8}}

\def\innersepdists{2mm}

\tikzset{
  disttable/.style={
    matrix of math nodes,
    row sep=-\pgflinewidth,
    column sep=-\pgflinewidth,
    nodes={
      text width=width("00"),
      text height=1.5ex,
      minimum height=4mm,
      inner xsep=0.8mm
    },
    execute at empty cell={\node[draw=none]{};}
  }
}

\tikzstyle{index} = [
  rectangle,
  inner sep=0,
  outer ysep=0.7mm,
  anchor=south west
]

\matrix[disttable] (dist) {
  7 & 2 & \cdots & 1 & 8 & 3 & \cdots & 5 & \phantom{0} & \phantom{0} & \cdots & \phantom{0} & 6 & 9 & \cdots & 5 \\
};

\tikzstyle{portion} = [
  rectangle,
  draw,
  inner sep=-\pgflinewidth,
  outer sep=0
]

\node[left = 0.10cm of dist-1-1.west](lab1){$D^0$};

\node[portion,
  fit=(dist-1-1.north west) (dist-1-4.south east)](v1b) {};
\node[portion,
  fit=(dist-1-5.north west) (dist-1-8.south east)](v1b) {};
\node[portion,
  fit=(dist-1-9.north west) (dist-1-12.south east)](v1b) {};
\node[portion,
  fit=(dist-1-13.north west) (dist-1-16.south east)](v1b) {};

\node[index,
  above = 0cm of dist-1-1.north west,
  anchor=south west](p0s){\footnotesize \texttt{00}};

\node[index,
  above = 0cm of dist-1-4.north west,
  anchor=south west](p0e){\footnotesize \texttt{0f}};

\node[index,
  above = 0cm of dist-1-5.north west,
  anchor=south west](p1s){\footnotesize \texttt{10}};

\node[index,
  above = 0cm of dist-1-8.north west,
  anchor=south west](p1e){\footnotesize \texttt{1f}};

\node[index,
  above = 0cm of dist-1-13.north west,
  anchor=south west](pfs){\footnotesize \texttt{f0}};
  
\node[index,
  above = 0cm of dist-1-16.north west,
  anchor=south west](pfe){\footnotesize \texttt{ff}};
  
 \draw [decoration={brace, raise=1.5pt}, decorate] 
  (p0s.north west) --  node[above=3pt]{portion 0}(p0e.north east);
 \draw [decoration={brace, raise=1.5pt}, decorate] 
 (p1s.north west) --  node[above=3pt]{portion 1}(p1e.north east);
 \draw [decoration={brace, raise=1.5pt}, decorate] 
 (pfs.north west) --  node[above=3pt]{portion 15}(pfe.north east);

\end{tikzpicture}

%% file: fastpq/fig/unordered_codes.tex
\begin{tikzpicture}[x=20pt, y=20pt, node distance=1pt,outer sep = 0pt]
\node[draw=none,fill=none](n){};
\pqvector{a}{n}(0,1  0,3  0,2  0,5  0,6  0,9  0,4  0,8)
\pqvector{b}{a}(3,f  1,1  2,1  0,0  0,1  f,2  1,2  1,1)
\pqvector{d}{b}(f,6  f,f  f,6  f,0  2,3  0,b  b,6  2,f)
\pqvector{f}{d}(f,5  f,c  f,f  f,1  4,6  3,3  c,f  2,c)
\pqvector{h}{f}(0,8  0,a  0,b  0,1  3,d  b,c  8,2  d,6)
\pqvector{i}{h}(0,e  0,6  0,2  1,9  b,0  8,e  c,9  1,3)
\pqvector{j}{i}(0,1  0,2  0,8  0,4  a,1  9,7  6,d  a,f)
\pqvector{k}{j}(3,4  1,6  2,5  0,6  2,3  9,2  b,c  d,1)
\largedotsbox{m}{k};

\end{tikzpicture}

%% file: fastpq/fig/grouped_codes.tex
\begin{tikzpicture}[x=20pt, y=20pt, node distance=1pt,outer sep = 0pt]

\def\pqshadecolor{gray!80}
\node[draw=none,fill=none](n){};
\pqhead{z1}{n}(0)(0)(0)(0)
\pqvector{a1}{z1}(0,1  0,3  0,2  0,5  0,6  0,9  0,4  0,8)
\pqvector{c1}{a1}(0,8  0,a  0,b  0,1  3,d  b,c  8,2  d,6)
\smediumdotsbox{d1}{c1}


\pqhead{z3}{d1}(3)(1)(2)(0)
\pqvector{a3}{z3}(3,f  1,1  2,1  0,0  0,1  f,2  1,2  1,1)
\pqvector{b3}{a3}(3,4  1,6  2,5  0,6  2,3  9,2  b,c  d,1)
\smediumdotsbox{d3}{b3}

\pqhead{z4}{d3}(f)(f)(f)(f)
\pqvector{a4}{z4}(f,5  f,c  f,f  f,1  4,6  3,3  c,f  2,c)
\pqvector{b4}{a4}(f,6  f,f  f,6  f,0  2,3  0,b  b,6  2,f)
\ssmalldotsbox{d4}{b4}

%
%
%
\end{tikzpicture}

%% file: fastpq/fig/minimum_tables.tex
\begin{tikzpicture}[node distance=1pt,outer sep = 0pt]

\tikzstyle{box}=[rectangle,draw,text centered, font=\ttfamily]
\tikzstyle{info}=[ellipse,text centered] 

\tikzstyle{box}=[rectangle,draw,text centered, font=\ttfamily]
\tikzstyle{normalbox}=[box,minimum height=12.5pt,text width=6.48em]

\def\longdots{\phantom{0} \dots \dots \phantom{0}}
\def\phantomcell{\phantom{3\; 1} \dots \phantom{8}}

\def\z#1{\textcolor{red}{\textbf{#1}}}

\begin{scope}[xshift=-5cm]
\node(title){Distance Tables};
\matrix[below = 0.4cm of title](dist) [
  matrix of math nodes,
  column sep=-\pgflinewidth,
  row sep=1mm,
  nodes={draw}]{
	2 \; 5 \dots \z{1} &  \z{4}\;6 \dots 8 & \longdots &  6\;1 \dots \z{1}\\
	4 \; 6 \dots \z{3} &  8\;7 \dots \z{3} & \longdots &  7\;8 \dots \z{7} \\
	1 \; \z{1} \dots 4 &  \z{1}\;3 \dots 2 & \longdots &  \z{0}\;2 \dots 6 \\
	7 \; 5 \dots \z{2} &  5\;\z{2} \dots 4 & \longdots &  2\;6 \dots \z{0} \\};
\end{scope}

\def\phdots{\phantom{0}\dots\phantom{0}}

\begin{scope}[xshift=0cm]
\node(title){Minimum Tables};
\matrix[below = 0.4cm of title] (mindist) [
  matrix of math nodes,
  column sep=-\pgflinewidth,
  row sep=1mm,
  nodes = {draw}]{
	1 & 4 & \phdots &  1 \\
	3 & 3 & \phdots &  7 \\
	1 & 1 & \phdots &  0 \\
	2 & 2 & \phdots &  0 \\};
\end{scope}

\footnotesize
\node[above = 0cm of dist-1-1.north west](lab1){\texttt{\phantom{00}00}};
\node[above = 0cm of dist-1-2.north west](lab1){\texttt{\phantom{00}10}};
\node[above = 0cm of dist-1-4.north west](lab1){\texttt{\phantom{00}f0}};

\node[above = 0cm of mindist-1-1.north west](lab1){\texttt{\phantom{0}0}};
\node[above = 0cm of mindist-1-4.north east](lab1){\texttt{f\phantom{0}}};

\node[left = 0.10cm of dist-1-1.west](lab1){$D^4$};
\node[left = 0.10cm of dist-4-1.west](lab1){$D^7$};

\normalsize

\path (dist)--(mindist) node[midway] {$\Longrightarrow$};

\end{tikzpicture}

%% file: fastpq/fig/quantization.tex
\def\qminDist{0.7cm}
\def\qmaxDist{3.5cm}
\def\totMaxDist{5cm}
  
\def\binWidth{0.72cm}

\def\boundSize{0.6cm}
\def\labelSep{0.4cm}
\def\axesSep{\boundSize*0.5-0.1cm}
  
\tikzstyle{bound} = [
  densely dotted
]

\tikzstyle{bin} = [
  draw, minimum width=\binWidth, minimum height=0.5cm
]

\begin{tikzpicture}

\draw [|-, color=black!40] (0,0) -- +(\qminDist, 0)
  coordinate (qminPos);
\node[above=\labelSep of qminPos, anchor=base] {$\mathit{qmin}$};
\path (0,0) -- +(0, \labelSep) node[anchor=base] {$0$};
\draw[bound] ($(qminPos) - (0,\boundSize*0.5)$) -- +(0,\boundSize);


\draw [-] (qminPos) -- +(\qmaxDist, 0)
  coordinate (qmaxPos);
\node[above=\labelSep of qmaxPos, anchor=base] {$\mathit{qmax}$};
\draw[bound] ($(qminPos) - (0,\boundSize*0.5)$) -- +(0,\boundSize);

\draw[color=black!40, ->|] (\qminDist+\qmaxDist,0) -- +(\totMaxDist, 0) 
  coordinate (endPos);
\node[above=\labelSep of endPos, anchor=base] {$\sum_j \max D^j$};
\draw[bound] ($(qmaxPos) - (0,\boundSize*0.5)$) -- +(0,\boundSize);

\coordinate (binStart) at ($(qminPos) - (0,\axesSep)$);
\node[bin, right=-\pgflinewidth*0.5 of binStart, anchor=north west] (bin0) {$0$};
\node[bin, right=-\pgflinewidth of bin0, anchor=west] (bin1) {$1$};
\node[bin, right=-\pgflinewidth of bin1, anchor=west,
  minimum width=\qmaxDist-3*\binWidth-3*\pgflinewidth,
  align=center] (bindots) {$\dots$};
\node[bin, right=-\pgflinewidth of bindots, anchor=west] (bin126) {$126$};
\node[bin, right=-\pgflinewidth of bin126, anchor=west, 
  minimum width=\totMaxDist-\pgflinewidth*0.5] {$127$};
\end{tikzpicture}

%% file: quickadc/quickadc.tex
\startcontents[chapters]
\ruledprintcontents

\section{Motivation}

\paragraph{}
Even if product quantization is among the fastest nearest neighbor search approaches, we have shown that the procedure used to scan inverted lists of short codes has a low computational efficiency. We have proposed PQ Fast Scan, a highly efficient scan procedure that replaces slow cache accesses by fast SIMD in-register shuffles. Replacing cache accesses by SIMD in-register shuffles requires building small tables that fit SIMD registers. PQ Fast Scan achieves this result by (1) grouping codes, (2) computing minimum tables and (3) quantizing floating-point distances to 8-bit integers. We have shown that grouping codes imposes a minimum size on inverted lists. This makes it impossible to combine PQ Fast Scan with inverted indexes, another widespread search acceleration technique. 

\paragraph{}
In this chapter, we introduce Quick ADC, a fast scan procedure that can be combined with inverted indexes. Like PQ Fast Scan, Quick ADC offers a 4-6 speedup over the conventional scan procedure. Quick ADC also builds on the idea of replacing costly cache accesses by cheap SIMD in-register shuffles. Unlike PQ Fast Scan, Quick ADC achieves this results by (1) using 4-bit sub-quantizers and (2) quantizing floating-point distances to 8-bit integers. Using 4-bit sub-quantizers makes it possible to build small tables that fit SIMD registers, without imposing size constraints on inverted lists. On the flip side, using 4-bit sub-quantizers ($m{\stimes}4$ codes) instead if the commonly used 8-bit sub-quantizers ($m{\stimes}8$ codes) causes a slight decrease in recall. We however show that this decreases is generally small to negligible and unlikely to have a practical impact. More specifically, this chapter addresses the following points:

\begin{itemize}
  \item We briefly present the key points of the design of Quick ADC. We show that using 4-bit sub-quantizers allows fast distance computations relying on SIMD in-register shuffles.
  \item We evaluate Quick ADC in a wide range range of scenarios. We evaluate the recall of Quick ADC not only on SIFT descriptors (128 dimensions), but also on more challenging types of vectors such as deep neural codes (256 dimensions) or GIST descriptors (960 dimensions). We show that Quick ADC only causes a slight decrease in recall, especially when combined with inverted indexes and optimized product quantization. In all cases, Quick ADC yields 4-6 times better performance than the conventional scan procedure.
  \item We discuss the differences between PQ Fast Scan and Quick ADC. We review the applicability of Quick ADC to derivatives of product quantization. 
\end{itemize}

\section{Presentation}

\subsection{Overview}
\label{sec:quickadc-overview}

\paragraph{4-bit sub-quantizers}
Quick ADC relies on SIMD in-register shuffles for distance computations. We have demonstrated that SIMD in-register shuffles offer much better performance than cache accesses or SIMD gather operations. Thus, SIMD in-register shuffles perform 16 table lookups in 1 cycle while caches accesses only perform 1 tables lookup in 1 cycle. However, SIMD in-register shuffles require lookup tables to be stored in SIMD registers, and limit their size to 16 elements of 8 bit each (Section~\ref{sec:fastpqdesc}). Quick ADC achieves this result by (1) imposing the use of 4-bit sub-quantizers and (2) quantizing floating point distances to 32-bit integers. Using 4-bit sub-quantizers inherently lead to lookup tables of 16 elements, without any additional modifications. An $m{\stimes}b$ product quantizer generates $m$ lookup tables of $2^b$ elements each. An $m{\stimes}8$ product quantizer (product quantizer with 8-bit sub-quantizers) generates lookup tables of $2^{8}=256$ floating-point values. As PQ Fast Scan uses 8-bit sub-quantizers, it has to perform additional modifications to shrink lookup tables of 256 values to 16 values (grouping codes, and computing tables of minimums). By contrast, an $m{\stimes}4$ product quantizer (product quantizer with 4-bit sub-quantizers) generates lookup tables of $2^{4}=16$ floating-point values. Therefore, Quick ADC does not need to perform additional modifications to obtain lookup tables of 16 elements. 

\begin{figure} [b]
  \centering%
  \input{quickadc/fig/simd_layout_tranposed}%
  \caption{Transposed block\label{fig:blocktrans}}
\end{figure}

\paragraph{Floating-point quantization}
Although product quantizers that use 4-bit sub-quantizers inherently generate lookup tables of 16 elements, these elements are 32-bit floating-point distances. SIMD in-register shuffles require lookup tables of 16 elements of 8-bit each ($16{\stimes}8$ bit). Therefore, like in PQ Fast Scan we quantize floating-point distances to 8-bit integers (Section~\ref{sec:fastpq-floatquant}). As there is no SIMD instruction to compare 8-bit unsigned integers, we quantize distances to 8-bit integers, only using their positive range. Like in PQ Fast Scan, we quantize dsitances between a $\mathit{qmin}$ and $\mathit{qmax}$ bound intro $n=127$ bins (0-126) uniformly. Values above $\mathit{qmax}$ are quantized to 127. We compute $\mathit{qmin}$ and $\mathit{qmax}$ as in PQ Fast Scan. Thus, $\mathit{qmin}$ is the minimum value across all lookup tables. To determine $\mathit{qmax}$ we scan $\mathit{init}$ vectors to find a temporary nearest neighbor. We use the distance of the query vector to this temporary nearest neighbor as the $\mathit{qmax}$ bound.

\subsection{SIMD Distance Computations}

\begin{algorithm}[t]
  \caption{ANN Search with Quick ADC}\label{alg:quickadc}
  \begin{algorithmic}[1]
    \Function{lookup\_add}{$\mathit{comps}, D^j, \mathit{acc}$}  \Comment{Fig. \ref{fig:lookup_add}} \label{alg:line:lookadd}
    \State \texttt{r128} masked $\gets$ \texttt{simd\_and}($\mathit{comps}, \texttt{0x0f}$)
    \State \texttt{r128} partial $\gets$ \texttt{simd\_shuffle}($\mathit{comps}, D^j$)
    \State  \Return{\texttt{simd\_add\_saturated}($\mathit{acc}, \mathit{partial}$)}
    \EndFunction
    
    \Function{quick\_adc\_block}{$\mathit{blk}, \{D^j\}_{j=0}^{m-1}$}
    \State \texttt{r128} $\mathit{acc} \gets \{0\}$
    \For{$j \gets 0$ to $m/2 - 1$} \label{alg:line:rowiter}
    \State \texttt{r128} $\mathit{comps} \gets$ \texttt{simd\_load}($\mathit{blk} + j\cdot16$) \label{alg:line:rowload}
    \State $\mathit{acc} \gets$ \Call{lookup\_add}{$\mathit{comps}, D^{2j}, \mathit{acc}$} \label{alg:line:lookadd1}
    \State $\mathit{comps} \gets$ \texttt{simd\_right\_shift}($\mathit{comps}, 4$) \Comment{Fig. \ref{fig:shift}} \label{alg:line:shift}
    \State $\mathit{acc} \gets$ \Call{lookup\_add}{$\mathit{comps}, D^{2j+1}, \mathit{acc}$} \label{alg:line:lookadd2}
    \EndFor
    \Return{$\mathit{acc}$}
    \EndFunction
    
    \Function{quick\_adc\_scan}{$\mathit{tlist}, \{D^j\}_{j=0}^{m-1}, R$} \label{alg:line:qadc}
    \State $\mathit{neighbors} \gets \operatorname{binheap}(R)$
    \For{$\mathit{blk}$ in $\mathit{tlist}$} \label{alg:line:blockiter}
    \State \texttt{r128} $\mathit{acc} \gets$ \Call{quick\_adc\_block}{$\mathit{blk}, \{D^j\}_{j=0}^{m-1}$}
    \State \Call{extract\_matches}{$\mathit{acc}, \mathit{neighbors}$} \label{alg:line:extract}
    \EndFor
    \State \Return{$\mathit{neighbors}$}
    \EndFunction
  \end{algorithmic}
\end{algorithm}

\paragraph{}
As the Quick ADC algorithm is simpler than PQ Fast Scan, we are able to detail how we utilize SIMD instructions to implement it. In the evaluation section, we use a 256-bit version of Quick ADC implemented with the new AVX 256-bit SIMD instruction set. Yet, for the sake of simplicity, we describe a 128-bit version of Quick ADC, and show how to generalize it to 256-bit SIMD at the end of the section. The 128-bit version of Quick ADC has also the advantage of being compatible with older Intel CPUs and ARM CPUs. The two versions are similar, as 256-bit SIMD in-register shuffles do not perform full-length shuffles on 256-bit ($32{\stimes}8$ bit) lookup tables. Instead, 256-bit SIMD in-registers perform shuffles in two independent 128-bit tables, stored in two independent 128-bit lanes. The speedup provided by the 256-bit version in comparison with the 128-bit is small: it is generally comprised between 10\% and 15\%.

\paragraph{}
Quick ADC operates on a block-transposed inverted list of codes, like PQ Fast Scan (Section~\ref{sec:fastpq-lookup}) or the SIMD gather implementation of the conventional ADC Scan algorithm (Section~\ref{sec:simd-issues}). This is required as SIMD instructions operate on 16 codes at once. SIMD in-register shuffles are able to perform 16 tables lookups at once, but in the \emph{same} lookup table. Therefore, to perform the first SIMD shuffle, we need to load the first component of 16 codes $(a_0,\dotsc,p_0)$ in an SIMD register. This can be done efficiently only if these 16 components are contiguous in memory, which is why we need to block-transpose the inverted list. To block-transpose the inverted list, we first divide it into blocks of 16 codes $(a,\dotsc,p)$. We then transpose each block independently. In a transposed block (Figure~\ref{fig:layout}), we store the first components of 16 codes contiguously ($a_0,\cdots,p_0$), followed by the second components of the same 8 code ($a_1,\cdots,p_1$) \etc, instead of storing all components of the first vector ($a_0,\cdots,a_7$), followed by the components of the second vector ($b_0,\cdots,b_7$) \etc 

\paragraph{}
The \textsc{quick\_adc\_scan} function (Algorithm~\ref{alg:quickadc}, line~\ref{alg:line:qadc}) scans a block-transposed inverted list $\mathit{tlist}$ using $m$ \emph{quantized} lookup tables $\{D^j\}_{j=0}^{m-1}$, where $m$ is the number of sub-quantizers of the product quantizer. Each lookup table is stored in a distinct SIMD register. In Algortihm~\ref{alg:quickadc}, SIMD instructions are denoted by the prefix \texttt{simd\_}. SIMD instructions use 128-bit variables, denoted by \texttt{r128}. The function named \textsc{quick\_adc\_scan}  iterates over blocks $\mathit{blk}$ of 16 codes (Algorithm~\ref{alg:quickadc}, line~\ref{alg:line:blockiter}). The \textsc{quick\_adc\_block} function computes the distance between the query vector and the 16 codes ($a,\dotsc,p$) of the block $\mathit{blk}$.

\begin{figure}
  \centering
  \input{quickadc/fig/simd_lookup_add}
  \caption{SIMD Lookup-add\label{fig:lookup_add} ($j=0$)}
\end{figure}
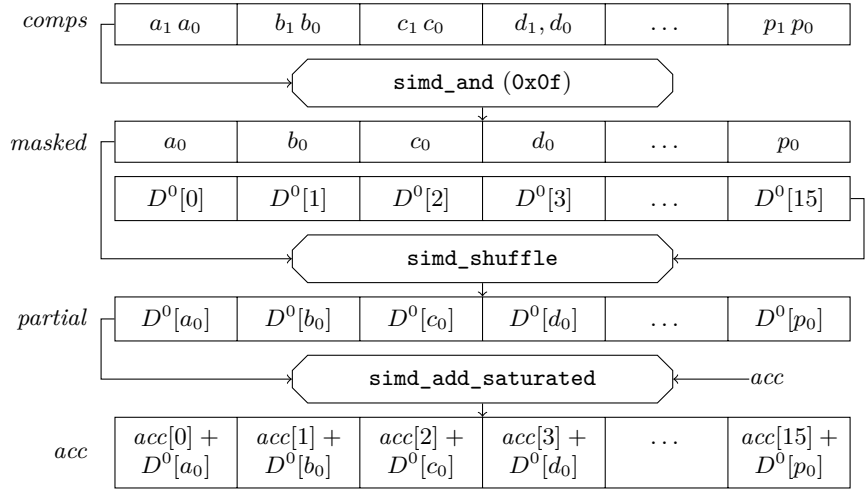

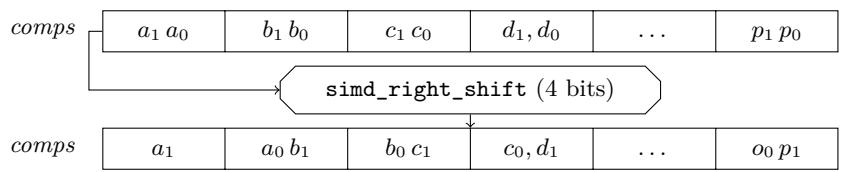
\begin{figure}
  \centering
  \input{quickadc/fig/simd_shift}
  \caption{SIMD 4-bit Right Shift\label{fig:shift} ($j=0$)}
\end{figure}

\paragraph{}
Each block comprises $m/2$ rows of 16 bytes (128 bits). Each row stores the $j$th and $(j+1)$th components of 16 codes (Figure~\ref{fig:blocktrans}). The \textsc{quick\_adc\_block} function iterates over each row (Alorithm~\ref{alg:quickadc}, line~\ref{alg:line:rowiter}), and loads it in the $\mathit{comps}$ register sequentially (Algorithm~\ref{alg:quickadc}, line~\ref{alg:line:rowload}). Two lookup-add operations are performed on each row (Algorithm~\ref{alg:quickadc}, line~\ref{alg:line:lookadd1} and line~\ref{alg:line:lookadd2}): one for the $(2j)$th components, and one for $(2j+1)$th components of the codes. Figure~\ref{fig:lookup_add} describes the succession of operations performed by the \textsc{lookup\_add} function for the first row ($j=0$). As each byte of the first row stores two components, \eg the first byte of the first row stores $a_1$ and $a_0$ (Figure~\ref{fig:lookup_add}), we start by masking the lower 4 bits of each byte (\texttt{and} with \texttt{0x0f}), to obtain the first components ($a_0,\dotsc,p_0$) only. The remainder of the function looks up values in the $D^0$ table and accumulates distances in $\mathit{acc}$ variable. Before the \textsc{lookup\_add} function can be used to process the second components ($a_1,\dotsc,p_1$), it is necessary that ($a_1,\dotsc,p_1$) are in the lowest 4 bits of each byte of the register. We therefore right shift the $\mathit{comps}$ register by 4 bits (Figure~\ref{fig:shift}) before calling \textsc{lookup\_add} (Algorithm~\ref{alg:quickadc}, line~\ref{alg:line:shift}). The \textsc{extract\_matches} function (Algorithm~\ref{alg:quickadc}, line~\ref{alg:line:extract}), the implementation of which is not shown, extracts distances from the $\mathit{acc}$ register and inserts them in the binary heap $\mathit{neighbors}$.

\paragraph{}
Among 256-bit SIMD instructions (AVX and AVX2 instruction sets) supported on recent CPUs, some, like in-register shuffles, operate concurrently on two independent 128-bit lanes. This prevents use of 256-bit lookup tables (32 8-bit integers) but allows an easy generalization of the 128-bit version of Quick ADC. While the 128-bit version of Quick ADC iterates on block rows one by one (Algorithm~\ref{alg:quickadc}, line~\ref{alg:line:rowiter}), the 256-bit version processes two rows at once: one row in each 128-bit lane. The number of iterations is thus reduced from $m/2$ to $m/4$. Lastly, instead of storing each $D^j$ table in a distinct 128-bit register, the tables $D^j$ and $D^{2j}$, $j \in \{0,\dotsc,m/2-1\}$, are stored in each of the two lanes of a 256-bit register.

\section{Evaluation}
\label{sec:quickadc-eval}

\subsection{Experimental Setup}

\paragraph{}
We implemented Quick ADC in C++, using intrinsics to access SIMD instructions~\cite{IntelCompMan,IntelIntrinsicsGuide}. While PQ Fast Scan is implemented using 128-bit SIMD (Section~\ref{sec:expsetup}), we implemented Quick using 256-bit SIMD, which allows and additional performance boost of 10\%-15\%. Our implementation therefore uses the AVX and AVX2 instruction sets. We used the g++ compiler version 5.3, with the options \texttt{-03 -ffast-math -m64 -march=native}. Exhaustive search and non-exhaustive search (inverted indexes, IVF) were implemented as described in~\cite{Jegou2011}. We use the yael library and the ATLAS library version 3.10.2. We compiled an optimized version of ATLAS on our system. To learn product quantizers and optimized product quantizers, we used the implementation \footnote{\url{https://github.com/arbabenko/Quantizations}} of the authors of~\cite{Babenko2015TQ, Babenko2014AQ}.

\begin{table}
  \centering
  \caption{Systems\label{tbl:platforms}}
  \begin{tabular}{lll@{ }l}
    \toprule
    & CPU &  \multicolumn{2}{l}{RAM} \\
    \midrule
    workstation & Xeon E5-1650v3 & 16GB & DDR4 2133Mhz\\
    server & Xeon E5-2630v3 & 128GB & DDR4 1866Mhz\\
    \bottomrule
  \end{tabular}
\end{table}

\paragraph{}
Unless otherwise noted, experiments were performed on our workstation (Table \ref{tbl:platforms}). To get accurate timings, we processed queries sequentially on a single core. We evaluate our approach on two publicly available\footnote{\url{http://corpus-texmex.irisa.fr/}} datasets of SIFT descriptors, one dataset of GIST descriptors, and one dataset of deep features\footnote{\url{http://sites.skoltech.ru/compvision/projects/aqtq/}} (Table~\ref{tbl:datasets}). The Deep1M dataset consists of deep neural codes that were $L_2$-normalized and PCA-compressed to 256 dimensions~\cite{Babenko2015TQ}. For SIFT1B, the learning set is needlessly large to train product quantizers, so we used the first 2 million vectors. We used the first 1000 queries from the query sets of SIFT1M and SIFT1B.

\begin{table}
  \centering
  \caption{Datasets\label{tbl:datasets}}
  \begin{tabular}{lllll}
    \toprule
    & Base set & Learning set & Query set & Dim. \\
    \midrule
    SIFT1M & 1M & 100K & 10K & 128\\
    SIFT1B & 1000M & 100M (2M) & 10K & 128 \\
    GIST1M & 1M & 500K & 1K & 960 \\
    Deep1M & 1M & 300K & 1K & 256\\
    \bottomrule
  \end{tabular}
\end{table}

\subsection{Exhaustive Search in SIFT1M}

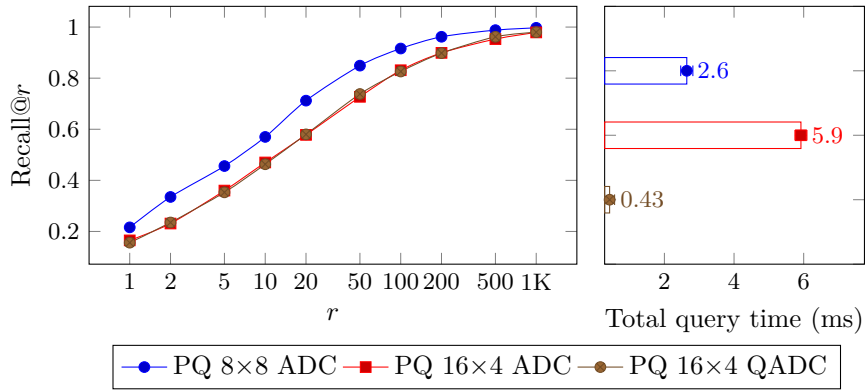
\begin{figure}[b]
  \centering
   \begin{tikzpicture}[baseline, trim axis left]
   \begin{semilogxaxis}[
     recall plot,
     height=5 cm,
     width=8 cm,
     ticklabel style={font=\small},
     legend to name=legend-qadcpq,
     legend columns=3
   ]
   \addplot+[smooth] 
   table[x=r,y=pq8x8adc] {quickadc/plots/SIFT1M.flat.recall.gnuplot};
   \addplot+[smooth] 
   table[x=r,y=pq16x4adc] {quickadc/plots/SIFT1M.flat.recall.gnuplot};
   \addplot+[smooth] 
   table[x=r,y=pq16x4qadc] {quickadc/plots/SIFT1M.flat.recall.gnuplot};
   \legend{PQ $8{\stimes}8$ ADC, PQ $16{\stimes}4$ ADC, PQ $16{\stimes}4$ QADC}
   \end{semilogxaxis}
   \end{tikzpicture}
   \begin{tikzpicture}[baseline, trim axis right]
   \begin{axis}[
     time plot,
     height=5 cm,
     width=5 cm,
     enlarge x limits={abs={5ex}, upper},
     enlarge y limits={rel=0.5},
     ticklabel style={font=\small},
   ]
   \addplot+[time bar]
   table [x=pq8x8adcmean,
   x error=pq8x8adcstd,
   y expr=1,
   meta=pq8x8adcmean]
   {quickadc/plots/SIFT1M.flat.time.transposed.gnuplot};
   \addplot+[time bar]
   table[x=pq16x4adcmean,
   x error=pq16x4adcstd,
   y expr=0,
   meta=pq16x4adcmean]
   {quickadc/plots/SIFT1M.flat.time.transposed.gnuplot};
   \addplot+[time bar]
   table[x=pq16x4qadcmean,
   x error=pq16x4qadcstd,
   y expr=-1,
   meta=pq16x4qadcmean]
   {quickadc/plots/SIFT1M.flat.time.transposed.gnuplot};
   \end{axis}
   \end{tikzpicture}
  \\
  \pgfplotslegendfromname{legend-qadcpq}
  \caption{ADC and QADC response time and recall (PQ, SIFT1M, Exhaustive search)\label{fig:exhaust}}
\end{figure}

\paragraph{}
Using $16{\stimes}4$ Quick ADC (QADC) instead of $8{\stimes}8$ ADC causes a decrease in recall which is due to two factors: (1) use of $16{\stimes}4$ quantizers instead of $8{\stimes}8$ quantizers and (2) use of quantized lookup tables (Section~\ref{sec:quickadc-overview}). We evaluate the global decrease in recall caused by the use of $16{\stimes}4$ QADC instead of $8{\stimes}8$ ADC, but also the relative impact of factors (1) and (2). To do so, we use the SIFT1M dataset and follow an exhaustive search strategy. Because we do not use an inverted index, we encode the original vectors into short codes, and not residuals. This maximizes quantization error and thus represents a worst-case scenario for QADC. We scan $\mathit{init}=200$ vectors to set the $\mathit{qmax}$ bound for quantization of lookup tables (Section ~\ref{sec:quickadc-overview}).

\begin{figure}
  \centering
   \begin{tikzpicture}[baseline, trim axis left]
   \begin{semilogxaxis}[
     recall plot,
     height=5 cm,
     width=8 cm,
     ticklabel style={font=\small},
     legend to name=legend-qadcopq,
     legend columns=3
   ]
   \addplot+[smooth] 
   table[x=r,y=opq8x8adc] {quickadc/plots/SIFT1M.flat.recall.gnuplot};
   \addplot+[smooth] 
   table[x=r,y=opq16x4adc] {quickadc/plots/SIFT1M.flat.recall.gnuplot};
   \addplot+[smooth] 
   table[x=r,y=opq16x4qadc] {quickadc/plots/SIFT1M.flat.recall.gnuplot};
   \legend{OPQ $8{\stimes}8$ ADC, OPQ $16{\stimes}4$ ADC, OPQ $16{\stimes}4$ QADC}
   \end{semilogxaxis}
   \end{tikzpicture}
   \begin{tikzpicture}[baseline, trim axis right]
   \begin{axis}[
     time plot,
     height=5 cm,
     width=5 cm,
     enlarge x limits={abs={5ex}, upper},
     enlarge y limits={rel=0.5},
     ticklabel style={font=\small},
   ]
   \addplot+[time bar]
   table [x=opq8x8adcmean,
   x error=opq8x8adcstd,
   y expr=1,
   meta=opq8x8adcmean]
   {quickadc/plots/SIFT1M.flat.time.transposed.gnuplot};
   \addplot+[time bar]
   table[x=opq16x4adcmean,
   x error=opq16x4adcstd,
   y expr=0,
   meta=opq16x4adcmean]
   {quickadc/plots/SIFT1M.flat.time.transposed.gnuplot};
   \addplot+[time bar]
   table[x=opq16x4qadcmean,
   x error=opq16x4qadcstd,
   y expr=-1,
   meta=pq16x4qadcmean]
   {quickadc/plots/SIFT1M.flat.time.transposed.gnuplot};
   \end{axis}
   \end{tikzpicture}
  \\
  \pgfplotslegendfromname{legend-qadcopq}
  \caption{ADC and QADC response time and recall (OPQ, SIFT1M, Exhaustive search)\label{fig:exhaustopq}}
\end{figure}
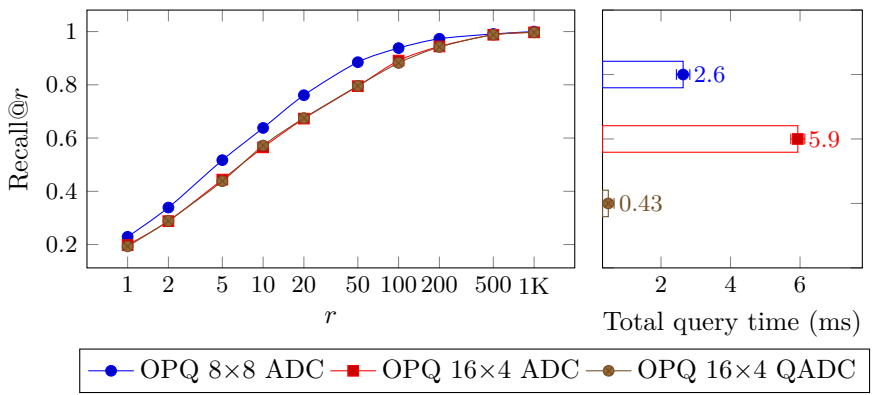

\paragraph{}
Using $16{\stimes}4$ ADC causes a small decrease in recall (Figure~\ref{fig:exhaust}). However, $16{\stimes}4$ QADC, which uses quantized lookup tables, does not further decrease recall in comparison with $16{\stimes}4$ ADC. Interestingly, the recall of  $16{\stimes}4$ QADC is even higher than the recall of $16{\stimes}4$ ADC for some points \eg the Recall@50 of $16{\stimes}4$ QADC is 0.738 against 0.727 for $16{\stimes}4$ ADC. This is because ADC already approximates real distances, as ADC computes distances between the query vector and \emph{quantized} database vectors. QADC adds another layer of approximation, which can either compensate or worsen the error introduced by PQ and ADC.  OPQ yiels better results than PQ in all cases (Figure~\ref{fig:exhaustopq}), which is consistent with~\cite{Norouzi2013, Ge2014}. Moreover, the difference in recall between $8{\stimes}8$ ADC and $16{\stimes}4$ QADC is lower in the case of OPQ than in the case of PQ. OPQ optimizes the decomposition of the input vector space into $m$ sub-spaces, which are used by the optimized product quantizer (Section \ref{sec:pq-derivatives}). For $m=16$, OPQ has more degrees of freedom than for $m=8$ and is therefore able to bring a greater level of optimization.

\paragraph{}
For an exhaustive search in 1 million vectors, $16{\stimes}4$ QADC is ${\sim}14$ times faster than $16{\stimes}4$ ADC and $6$ times faster than $8{\stimes}8$ ADC (Figure~\ref{fig:exhaust}, Figure~\ref{fig:exhaustopq}) (85\% decrease in response time). Response times for PQ and OPQ are similar (Figure~\ref{fig:exhaustopq}).
In practice, $8{\stimes}8$ ADC is much more common than $16{\stimes}4$ ADC~\cite{Babenko2014AQ, Babenko2015TQ, Babenko2015, Norouzi2013, Zhang2014}, thus we only compare $16{\stimes}4$ QADC with $8{\stimes}8$ ADC in the remainder of this section. Overall, QADC therefore proposes trading a small decrease in recall, and almost negligible in the case of OPQ, for a large improvement in response time.

\paragraph{}
Non-exhaustive search offers both lower response time and higher recall than exhaustive search (Section~\ref{sec:invindex}) and is often preferred in practice. Therefore, in the remainder of this section, we evaluate QADC in the context of non-exhaustive search, for a wide range of scenarios: SIFT and GIST descriptors, PQ and OPQ, 64 and 128 bit codes, and varying index parameters. We show that in most cases, when combined with OPQ and inverted indexes, QADC offers a speedup of 3.3-4 (70-75\% decrease in reponse time) for a small or negligible loss in recall.

\subsection{Non-exhaustive Search in SIFT1M}

\begin{table}
\centering
\begin{threeparttable}
  \caption{Non-exhaustive search, SIFT1M, 64 bit\label{tbl:sift1m}}
  \begin{tabular}{lllllll}
    \toprule
    PQ & ADC \tnote{*} & R@100 & Index & Tables & Scan & Total \\
    \midrule
    \multicolumn{7}{c}{\textbf{SIFT1M, IVF, K=1024, ma=48}} \\
    \midrule
    \input{quickadc/tables/SIFT1M.1024.48.booktabs.tex}
    \midrule
    \multicolumn{7}{c}{\textbf{SIFT1M, IVF, K=256, ma=24}} \\
    \midrule
    \input{quickadc/tables/SIFT1M.256.24.booktabs.tex}
    \bottomrule
  \end{tabular}
  \begin{tablenotes}
    \item[*] ADC: $8{\stimes}8$ ADC, QADC: $16{\stimes}4$ QADC
  \end{tablenotes}
\end{threeparttable}
\end{table}

\paragraph{}
Some approaches like PQ Fast Scan~\cite{Andre2015} or the inverted multi-index~\cite{Babenko2015} have only been evaluated on large datasets. To show the versatility of QADC, we evaluate it on both large and small datasets, such as SIFT1M. Table~\ref{tbl:sift1m} compares ADC and QADC in terms of Recall@100 (R@100), total ANN search time (Total), and the time spent in each of the search steps (Index, Tables, and Scan) detailed in Section~\ref{sec:annsearch}. All times are in milliseconds (ms). OPQ requires a rotation of the input vector before computing lookup tables (Section~\ref{sec:pq-derivatives}). We include the time to perform this rotation in the Tables column.

\paragraph{}
When using inverted indexes, the parameters $K$, the total number of cells of the inverted index, and $\mathit{ma}$, the number of cells scanned to answer a query, impact response time and recall (Section~\ref{sec:annsearch}). To answer a query $S \cdot \mathit{ma}/K$ vectors are scanned on average, where $S$ is the size of the database, and $\mathit{ma}$ tables are computed (Section~\ref{sec:annsearch}). Recall increases with $\mathit{ma}/K$ but so does response time. For a constant $\mathit{ma}/K$, a higher $K$ provides better recall.
We use an inverted index with $K=1024$ cells, as in~\cite{Jegou2011}, and $\mathit{ma}=48$. In this case, QADC offers a 66\% decrease in scan time (Table~\ref{tbl:sift1m}), which less than the 85\% decrease for exhaustive search (Figure~\ref{fig:exhaust}c). This is because when using inverted indexes, scanned codes are scattered across $\mathit{ma}$ cells of relatively small size (on average, cells comprise $S/K\approx976$ vectors). A significant amount of time is therefore spent switching cells.
Using larger cells solves this issue: for $K=256$, $\mathit{ma}=24$, QADC offers a 75\% decrease in scan time. The scan time is not significantly increased (0.072ms versus 0.067ms) for $K=256$, $\mathit{ma}=24$ despite twice more vectors are scanned compared to $K=1024$, $\mathit{ma}=48$. However, for $\mathit{ma}=24$, two times less lookup tables are computed, which significantly decreases response time. Therefore, in the context of QADC, the configuration $K=256$, $ma=24$ is more interesting.

\paragraph{}
Thanks to the use of 4-bit quantizers, which results in smaller and faster to compute lookup tables, QADC offers a 50-70\% decrease in tables computation time. This decrease is lower with OPQ due to the time spent rotating the query vector.
The loss in recall is significantly lower with OPQ (-1.5\% to 2.1\%) than with PQ (-3.6\% to 4.4\%).
In most cases, QADC offers a close to 70\% decrease in total response time.
\subsection{Non-exhaustive Search in GIST1M}

\begin{table}
  \centering
  \begin{threeparttable}
    \caption{Non-exhaustive search, GIST1M, 128 bit\label{tbl:gist1m128}}
    \begin{tabular}{lllllll}
      \toprule
      PQ & ADC \tnote{*} & R@100 & Index & Tables & Scan & Total \\
      \midrule
      \multicolumn{7}{c}{\textbf{GIST1M, IVF, K=256, ma=24}} \\
      \midrule
      \input{quickadc/tables/GIST1M.256.24.booktabs.tex}
      \bottomrule
    \end{tabular}
    \begin{tablenotes}
      \item[*] ADC: $16{\stimes}8$ ADC, QADC: $32{\stimes}4$ QADC
    \end{tablenotes}
  \end{threeparttable}
\end{table}

\paragraph{}
Due to their higher dimensionality GIST descriptors (960 dimensions) suffer a higher quantization error than SIFT descriptors (128 dimensions). To mitigate this issue, 128-bit codes can be used instead of 64-bit codes~\cite{Norouzi2013, Babenko2015}. Thus, here, we compare $16{\stimes}8$ ADC and $32{\stimes}4$ QADC. As for SIFT1M, we keep the parameters $K=256$, $\mathit{ma}=24$. Like for 64-bit codes, QADC provides a 75-80\% decrease in scan time with 128-bit codes (Table~\ref{tbl:gist1m128}). However, the decrease in recall when using QADC with PQ is higher for GIST descriptors (-24\%) than for SIFT descriptors (-3.5\% to 4.5\%). However, using OPQ settles this issue and reduces the recall loss to 5\%. OPQ should therefore be preferred for GIST descriptors. The decrease in reponse time is however less important for OPQ (-45\%). This is because a large part of response time is spent computing tables in the case of OPQ. OPQ requires rotating the query vector before computing tables, the computational cost of which depends on the dimensionality of the query vector. For 960-dimensional GIST descriptors this cost is much higher than for 128-dimensional SIFT descriptors. Even in the less favourable case of GIST descriptors, Quick ADC still achieves a close to 50\% decrease in response time.

\subsection{Non-exhaustive Search in Deep1M}

\begin{table}
  \centering
  \begin{threeparttable}
    \caption{Non-exhaustive search, Deep1M, 64 bit\label{tbl:deep1m}}
    \begin{tabular}{lllllll}
      \toprule
      PQ & ADC \tnote{*} & R@100 & Index & Tables & Scan & Total \\
      \midrule
      \multicolumn{7}{c}{\textbf{Deep1M, IVF, K=256, ma=24}} \\
      \midrule
      \input{quickadc/tables/DEEP1M.256.24.booktabs.tex}
      \bottomrule
    \end{tabular}
    \begin{tablenotes}
      \item[*] ADC: $8{\stimes}8$ ADC, QADC: $16{\stimes}4$ QADC
    \end{tablenotes}
  \end{threeparttable}
\end{table}

\begin{table}
  \centering
  \begin{threeparttable}
    \caption{Non-exhaustive search, Deep1M, 128 bit\label{tbl:deep1m128}}
    \begin{tabular}{lllllll}
      \toprule
      PQ & ADC \tnote{*} & R@100 & Index & Tables & Scan & Total \\
      \midrule
      \multicolumn{7}{c}{\textbf{Deep1M, IVF, K=256, ma=24}} \\
      \midrule
      \input{quickadc/tables/DEEP1M128.256.24.booktabs.tex}
      \bottomrule
    \end{tabular}
    \begin{tablenotes}
      \item[*] ADC: $16{\stimes}8$ ADC, QADC: $32{\stimes}4$ QADC
    \end{tablenotes}
  \end{threeparttable}
\end{table}

\paragraph{}
For the vectors of Deep1M, we evaluate QADC with 64-bit codes (Table~\ref{tbl:deep1m}) and 128-bit codes (Table~\ref{tbl:deep1m128}). We keep the parameters $K = 256$ and $\mathit{ma}=24$ used for SIFT1M and GIST1M. With 64-bit codes, QADC suffers higher loss of recall (13\%) with deep features than with SIFT descriptors (3.6\%). The increase in quantization error caused by the use of 4-bit quantizers (QADC) instead of 8-bit quantizers (ADC) is stronger for the PCA-compressed deep features than for SIFT descriptors. This is due to the higher dimensionality of the PCA-compressed deep features (256 dimensions, versurs 128 dimensions for SIFT descriptors). As for GIST descriptors, using OPQ instead of PQ mitigates this issue. With OPQ, QADC incurs a 2.2\% loss in recall, compared to a 13\% loss with OPQ (Table~\ref{tbl:deep1m}). With OPQ, the decrease in response time offered by QADC is a slightly lower for the PCA-compressed deep features than for SIFT descriptors (-62\% versus -67\%). This is because the rotation matrix for PCA-compressed deep features is larger that the rotation matrix for SIFT descriptor. Thus, the time spent to compute lookup tables is increased. With 64-bit codes and OPQ, QADC offers a 60\% decrease in response time for a 2.2\% decrease in recall (Table~\ref{tbl:deep1m}). With 128-bit codes, the loss of recall caused by the use of QADC is lower than for than 64-bit codes, even for PQ (-6.8\% versus -13\%). For 128-bit codes and OPQ, QADC also offers a 60\% decrease in response time. The loss of recall becomes negligible at 0.4\%.

\subsection{Non-exhaustive Search in SIFT1B}

\begin{table}
  \centering
  \begin{threeparttable}
    \caption{Non-exhaustive search, SIFT1B, 64 bit\label{tbl:sift1b}}
    \begin{tabular}{lllllll}
      \toprule
      PQ & ADC \tnote{*} & R@100 & Index & Tables & Scan & Total \\
      \midrule
      \multicolumn{7}{c}{\textbf{SIFT1B, IVF, K=8192, ma=64}} \\
      \midrule
      \input{quickadc/tables/SIFT1B.8192.64.booktabs.tex}
      \midrule
      \multicolumn{7}{c}{\textbf{SIFT1B, IVF, K=65536, ma=64}} \\
      \midrule
      \input{quickadc/tables/SIFT1B.65536.64.booktabs.tex}
      \bottomrule
    \end{tabular}
    \begin{tablenotes}
      \item[*] ADC: $8{\stimes}8$ ADC, QADC: $16{\stimes}4$ QADC
    \end{tablenotes}
  \end{threeparttable}
\end{table}

\begin{table}
  \centering
  \begin{threeparttable}
    \caption{Non-exhaustive search, SIFT1B, 128 bits\label{tbl:sift1b128}}
    \begin{tabular}{lllllll}
      \toprule
      PQ & ADC \tnote{*} & R@100 & Index & Tables & Scan & Total \\
      \midrule
      \multicolumn{7}{c}{\textbf{SIFT1B, IVF, K=65536, ma=64}} \\
      \midrule
      \input{quickadc/tables/SIFT1B128.65536.64.booktabs.tex}
      \bottomrule
    \end{tabular}
    \begin{tablenotes}
      \item[*] ADC: $16{\stimes}8$ ADC, QADC: $32{\stimes}4$ QADC
    \end{tablenotes}
  \end{threeparttable}
\end{table}

\paragraph{}
For non-exhaustive search in 1 billion SIFT vectors, we test two inverted indexes configurations $K=8192$, like in~\cite{Jegou2011R} and $K=65536$, like in~\cite{Babenko2015}. For this larger dataset, we scan $\mathit{init}=1000$ vectors before quantizing lookup tables (Section~\ref{sec:quickadc-overview}). In all cases, QADC offers a close to 78-79\% decrease in scan time (Table~\ref{tbl:sift1b}). For $K=8192$, QADC also offers a 78\% decrease in total query time but the loss in recall is relatively high (-15\% for PQ, and -10\% for OPQ). As in all other experiments, OPQ offers a much higher recall for a small to negligible increase in response time. Thus, from now on, we only report figures for OPQ only. The configuration $K=65536$ is more interesting than $K=8192$ as it allows both a lower response time and a higher recall, both for ADC and QADC. For $K=65536$ and OPQ, QADC achieves a recall 0.747 in 1.7ms (68\% decrease in response time). In comparison, in~\cite{Babenko2015}, the state-of-the-art OMulti-D-OADC system achieves the same recall in 2 ms.

\paragraph{}
With 64-bit codes (Table~\ref{tbl:sift1b}), the recall of ANN search in 1 billion vectors is relatively low, even for ADC. To offer better recall, it is possible to use 128-bit codes (Table~\ref{tbl:sift1b128}). With 128-bit codes, the database uses $20$GB of RAM, which exceeds the RAM capacity of our workstation. We therefore ran this experiment on a server (Table~\ref{tbl:platforms}). This configuration allows QADC to achieve a recall of 0.94 in 3.4 ms. In comparison, in~\cite{Babenko2015}, the OMulti-D-OADC system achieves a recall of 0.901 in 5 ms and a recall of 0.969 in 16 ms.

\section{Discussion}

\subsection{Compatibility with Inverted Indexes}

\paragraph{}
Contrary to PQ Fast Scan, Quick ADC imposes no constraints on the size of inverted lists. Quick ADC is therefore compatible with any type of inverted index. It is not only compatible with fine inverted indexes, but also with multi-indexes (Section~\ref{sec:invindex}). Besides, Quick ADC is compatible with $m{\stimes}4$ product quantizers, for any value of $m$. In this regard, in Section~\ref{sec:quickadc-eval}, we tested PQ Fast Scan with $16{\stimes}4$ (64-bit codes) product quantizers and $32{\stimes}4$ (128-bit codes) product quantizers. Clearly, Quick ADC is not compatible with $b$ parameters other than $b=4$, as using 4-bit sub-quantizers is one of the key ideas of Quick ADC.

\subsection{Applicability to Product Quantization Derivatives}
\label{sec:quickadc-deriv-discuss}

\paragraph{}
We evaluated Quick ADC with Product Quantization (PQ) and Optimized Product Quantization (OPQ), but we did not evaluate it in the context of other product quantization derivatives presented in Section~\ref{sec:pq-derivatives}, namely Additive Quantization (AQ), Tree Quantization (TQ) and Composite Quantization (CQ). These approaches offer a higher accuracy than PQ or OPQ. They however use (1) a different process to learn quantizer codebooks, and (2) a different ADC procedure. We discuss the compatibility of Quick ADC with these approaches. Using 4-bit sub-quantizers instead of 8-bit sub-quantizers ($b=4$ instead of $b=8$) requires doubling the $m$ parameter (\ie $m'=2m$) to maintain a similar accuracy (accuracy mainly depends on the $m{\stimes}b$ product, Section~\ref{sec:vectorencoding}). AQ, TQ and CQ codebook learning processes have complexities in $m\cdot 2^b$ or $m^2 \cdot 2^{2b}$. Therefore, doubling $m$ is unlikely to cause issues for practical values ($m \in 8,16$, $m' \in 16,32$). Still, experiments are required to determine if the codebook learning process remains tractable, especially for 128-bit codes ($m'=32$). CQ uses the same ADC procedure as PQ or OPQ. Therefore, Quick ADC is fully compatible with CQ. The ADC procedure of AQ and TQ have complexities in $m + m^2/2$ and $2m$. For these two approaches, doubling $m$ strongly increases the number of operations per distance computation. Even if Quick ADC makes table lookups and additions fast thanks to SIMD, such an increase in the number of operations would lessen its benefit. In conclusion, combining Quick ADC with CQ is more interesting than combining Quick ADC with TQ or AQ. The main drawback of Quick ADC is that it incurs a small decrease in accuracy. As CQ offers a better accuracy than OPQ, it would contribute to eliminate this drawback. The resulting solution would therefore be very competitive, as it would offer both superior performance, thanks to Quick ADC, and a high accuracy, thanks to CQ.

\stopcontents[chapters]

%% file: quickadc/fig/simd_layout_tranposed.tex
    \newlength\longMemCellHeight
    \setlength{\longMemCellHeight}{\heightof{\small$p_1\,p_0$}+0.5mm}
    \newlength\longMemCellWidth
    \setlength{\longMemCellWidth}{\widthof{\small$a_{m-1}\,a_{m-2}$}+1ex}

    \tikzset{
      long memory cell/.style={
        rectangle,
        draw,
        font=\small,
        text depth=0.25ex,
        text height=\longMemCellHeight,
        text width=\longMemCellWidth,
        align=center,
      },      
      long memory/.style={
        matrix of math nodes,
        row sep=-\pgflinewidth,
        column sep=-\pgflinewidth,
        nodes={
            long memory cell
        },
        execute at empty cell={\node[draw=none]{};}
      },
      long memory multicell/.style={
        draw,
        inner ysep=\pgflinewidth/2,
        inner xsep=-\pgflinewidth/2,
        font=\small,
        align=center,
        text height=\longMemCellHeight,
        text depth=0.1ex,
      }
    }

    \tikzstyle{memory label} = [
        font=\small
    ]
  \begin{tikzpicture}
  \node [long memory, row 1/.style={text=black!70}] (layout)
  { a_1\,a_0 & b_1\,b_0 & c_1\,c_0 &\dots & p_1\,p_0 \\
    \dots & \dots & \dots & \dots & \dots  \\
    a_{m-1}\,a_{m-2} & b_{m-1}\,b_{m-2} & c_{m-1}\,c_{m-2} & \dots & p_{m-1}\,p_{m-2} \\
  };
  
  \node[label, above=\vertSep of layout-1-1.north, anchor=south] {$a$};
  \node[label, above=\vertSep of layout-1-2.north, anchor=south] {$b$};
  \node[label, above=\vertSep of layout-1-3.north, anchor=south] {$c$};
  \node[label, above=\vertSep of layout-1-5.north, anchor=south] {$p$};
  \end{tikzpicture}

%% file: quickadc/fig/simd_lookup_add.tex
\begin{tikzpicture}

\node[register] (indexes01)
{%
  \nodepart[elt]{one}$a_1\,a_0$
  \nodepart[elt]{two}$b_1\,b_0$
  \nodepart[elt]{three}$c_1\,c_0$
  \nodepart[elt]{four}$d_1,d_0$%
  \nodepart[elt]{five}$\dots$%
  \nodepart[elt]{six}$p_1\,p_0$%
};

\node[operator, below=\vertSep of indexes01.south, anchor=north] (mask0) {\texttt{simd\_and} (\texttt{0x0f})};

\draw[->] (indexes01.west) -- ++(-\horSep,0) |- (mask0.west);

\node[register, below=\vertSep of mask0.south, anchor=north] (indexes0)
{%
  \nodepart[elt]{one}$a_0$
  \nodepart[elt]{two}$b_0$
  \nodepart[elt]{three}$c_0$
  \nodepart[elt]{four}$d_0$
  \nodepart[elt]{five}$\dots$%
  \nodepart[elt]{six}$p_0$%
};

\draw[->] (mask0.south) -- (indexes0.north);

\node[label, left=\labelSep of indexes01.west, anchor=east] {$\mathit{comps}$};
\node[label, left=\labelSep of indexes0.west, anchor=east] {$\mathit{masked}$};
\node[register, below=\vertSep of indexes0.south, anchor=north] (tables0)
{%
  \nodepart[elt]{one}$D^0[0]$
  \nodepart[elt]{two}$D^0[1]$
  \nodepart[elt]{three}$D^0[2]$
  \nodepart[elt]{four}$D^0[3]$
  \nodepart[elt]{five}$\dots$%
  \nodepart[elt]{six}$D^0[15]$%
};

\node[operator, below=\vertSep of tables0.south, anchor=north] (shuffle0) {\texttt{simd\_shuffle}};

\draw[->] (indexes0.west) -- ++(-\horSep,0) |- (shuffle0.west);
\draw[->] (tables0.east) -- ++(\horSep,0) |- (shuffle0.east);

\node[register, below=\vertSep of shuffle0.south, anchor=north] (look0)
{%
  \nodepart[elt]{one}$D^0[a_0]$
  \nodepart[elt]{two}$D^0[b_0]$
  \nodepart[elt]{three}$D^0[c_0]$
  \nodepart[elt]{four}$D^0[d_0]$
  \nodepart[elt]{five}$\dots$%
  \nodepart[elt]{six}$D^0[p_0]$%
};

\draw[->] (shuffle0.south) -- (look0.north);
\node[label, left=\labelSep of look0.west, anchor=east] {$\mathit{partial}$};

\node[operator, below=\vertSep of look0.south, anchor=north] (add0) {
  \texttt{simd\_add\_saturated}};

\draw[->] (look0.west) -- ++(-\horSep,0) |- (add0.west);

\node[register, below=\vertSep of add0.south, anchor=north] (res0)
{%
  \nodepart[elt]{one}$\mathit{acc[0]} + D^0[a_0]$
  \nodepart[elt]{two}$\mathit{acc[1]} + D^0[b_0]$
  \nodepart[elt]{three}$\mathit{acc[2]} + D^0[c_0]$
  \nodepart[elt]{four}$\mathit{acc[3]} + D^0[d_0]$
  \nodepart[elt]{five}$\dots$%
  \nodepart[elt]{six}$\mathit{acc[15]} + D^0[p_0]$%
};

\node[label, left=\labelSep of res0.west, anchor=east] {$\mathit{acc}$};

\draw[->] (add0.south) -- (res0.north);
\node[label, anchor=west, right=1cm of add0.east] (prev) {$\mathit{acc}$};
\draw [->] (prev.west) -- (add0.east);

\end{tikzpicture}

%% file: quickadc/fig/simd_shift.tex
\begin{tikzpicture}
\node[register] (indexes01)
{%
  \nodepart[elt]{one}$a_1\,a_0$
  \nodepart[elt]{two}$b_1\,b_0$
  \nodepart[elt]{three}$c_1\,c_0$
  \nodepart[elt]{four}$d_1,d_0$
  \nodepart[elt]{five}$\dots$%
  \nodepart[elt]{six}$p_1\,p_0$%
};

\node[operator, below=\vertSep of indexes01.south, anchor=north] (shift) {\texttt{simd\_right\_shift} (4 bits)};

\draw[->] (indexes01.west) -- ++(-\horSep,0) |- (shift.west);

\node[register, below=\vertSep of shift.south] (shiftindexes01)
{%
  \nodepart[elt]{one}$a_1$
  \nodepart[elt]{two}$a_0\,b_1$
  \nodepart[elt]{three}$b_0\,c_1$
  \nodepart[elt]{four}$c_0,d_1$
  \nodepart[elt]{five}$\dots$%
  \nodepart[elt]{six}$o_0\,p_1$%
};

\draw[->] (shift.south) -- (shiftindexes01.north);

\node[label, left=\labelSep of shiftindexes01.west, anchor=east] {$\mathit{comps}$};
\node[label, left=\labelSep of indexes01.west, anchor=east] {$\mathit{comps}$};

\node[label, anchor=west, right=1cm of shift.east] (prev) {\phantom{$\mathit{acc}$}};
\node[label, left=\labelSep of indexes01.west, anchor=east] {\phantom{$\mathit{masked}$}};

\end{tikzpicture}

%% file: quickadc/tables/SIFT1M.1024.48.booktabs.tex
PQ & ADC & 0.951 & 0.023 & 0.35 & 0.2 & 0.57 \\ 
    & QADC & 0.917 & 0.024 & 0.1 & 0.067 & 0.19 \\ 
    &      & \emph{-3.6\%} &    & \emph{-70\%} & \emph{-66\%} & \emph{-66\%} \\ 
\midrule
OPQ & ADC & 0.977 & 0.023 & 0.39 & 0.19 & 0.6 \\ 
    & QADC & 0.956 & 0.023 & 0.18 & 0.068 & 0.28 \\ 
    &      & \emph{-2.1\%} &    & \emph{-53\%} & \emph{-65\%} & \emph{-54\%} \\ 

%% file: quickadc/tables/GIST1M.256.24.booktabs.tex
PQ & ADC & 0.675 & 0.038 & 0.77 & 0.71 & 1.5 \\ 
    & QADC & 0.515 & 0.038 & 0.26 & 0.15 & 0.45 \\ 
    &      & \emph{-24\%} &    & \emph{-67\%} & \emph{-79\%} & \emph{-71\%} \\ 
\midrule
OPQ & ADC & 0.918 & 0.039 & 1.7 & 0.73 & 2.5 \\ 
    & QADC & 0.872 & 0.038 & 1.2 & 0.16 & 1.4 \\ 
    &      & \textbf{\emph{-5\%}} &    & \emph{-32\%} & \emph{-78\%} & \textbf{\emph{-45\%}} \\ 

%% file: quickadc/tables/DEEP1M.256.24.booktabs.tex
PQ & ADC & 0.772 & 0.015 & 0.24 & 0.33 & 0.58 \\ 
    & QADC & 0.669 & 0.013 & 0.082 & 0.076 & 0.17 \\ 
    &      & \emph{-13\%} &    & \emph{-66\%} & \emph{-77\%} & \emph{-71\%} \\ 
\midrule
OPQ & ADC & 0.922 & 0.013 & 0.34 & 0.32 & 0.67 \\ 
    & QADC & 0.902 & 0.013 & 0.16 & 0.08 & 0.25 \\ 
    &      & \emph{-2.2\%} &    & \emph{-53\%} & \emph{-75\%} & \emph{-62\%} \\ 

%% file: quickadc/tables/DEEP1M128.256.24.booktabs.tex
PQ & ADC & 0.922 & 0.015 & 0.33 & 0.58 & 0.93 \\ 
    & QADC & 0.859 & 0.013 & 0.13 & 0.14 & 0.28 \\ 
    &      & \emph{-6.8\%} &    & \emph{-62\%} & \emph{-77\%} & \emph{-70\%} \\ 
\midrule
OPQ & ADC & 0.988 & 0.013 & 0.39 & 0.58 & 0.99 \\ 
    & QADC & 0.984 & 0.013 & 0.24 & 0.14 & 0.39 \\ 
    &      & \emph{-0.4\%} &    & \emph{-39\%} & \emph{-76\%} & \emph{-60\%} \\ 

%% file: quickadc/tables/SIFT1B.8192.64.booktabs.tex
PQ & ADC & 0.746 & 0.09 & 0.42 & 23 & 23 \\ 
    & QADC & 0.635 & 0.088 & 0.13 & 5 & 5.2 \\ 
    &      & \emph{-15\%} &    & \emph{-69\%} & \emph{-78\%} & \emph{-78\%} \\ 
\midrule
OPQ & ADC & 0.792 & 0.09 & 0.53 & 23 & 24 \\ 
    & QADC & 0.712 & 0.087 & 0.22 & 4.9 & 5.2 \\ 
    &      & \emph{-10\%} &    & \emph{-58\%} & \emph{-79\%} & \emph{-78\%} \\ 

%% file: quickadc/tables/SIFT1B128.65536.64.booktabs.tex
OPQ & ADC & 0.95 & 0.73 & 1.1 & 10 & 12 \\ 
    & QADC & 0.94 & 0.72 & 0.49 & 2.2 & 3.4 \\ 
    &      & \textbf{\emph{-1.1\%}} &    & \emph{-57\%} & \emph{-78\%} & \textbf{\emph{-72\%}} \\ 

%% file: derivedquant/derivedquant.tex
\startcontents[chapters]
\ruledprintcontents

\section{Motivation}

\paragraph{}
For most current ANN search use cases, product quantization with 8-bit sub-quantizers offers a sufficient accuracy. With Quick ADC, we have shown that even 4-bit sub-quantizers can be used. We have proposed two highly efficient scan procedures for these use cases, PQ Fast Scan and Quick ADC. However, some emerging uses cases require a higher accuracy. Thus, descriptors generated by deep neural networks are increasingly popular for multimedia similarity search. When they are not PCA-compressed, such descriptors have several thousand dimensions (usually 4096). For these very high-dimensional vectors, a higher quantization accuracy is required. Therefore, 16-bit quantizers are used to increase product quantization accuracy. In addition, there has been a recent interest in using 32-bit codes instead of the usual 64-bit codes~\cite{Babenko2014AQ}. For 32-bit codes, 16-bit quantizers are required as 8-bit quantizers have a too low accuracy. However, using 16-bit quantizer causes a threefold increase in search time, which often outweighs their accuracy advantage. We have shown that this threefold increase in response time is caused by the fact that lookup tables used for distance computations are stored in the L3 cache when using 16-bit sub-quantizers. On the contrary, when using 8-bit sub-quantizers, lookup tables are stored in the much faster L1 cache (Section~\ref{sec:memory-accesses}).

\paragraph{}
In this chapter, we introduce a novel approach, derived quantizers, that makes 16-bit quantizers as fast as 8-bit quantizers, while retaining their accuracy. The key idea behind our approach is to derive 8-bit quantizers, named derived quantizers, from the 16-bit quantizers so that they share the same short codes. Therefore, our approach does not incur any increase in memory usage. The codebook of derived quantizers can be seen as an approximate version of the codebook of the high-resolution 16-bit quantizers. This allows us to design a two-pass nearest neighbor search procedure that provides both a low response time and a high-accuracy. More specifically, this chapter addresses the following points:
\begin{itemize}
  \item We present in detail the algorithm we designed to derive fast 8-bit quantizers from high-resolution 16-bit quantizers. We also describe how our two-pass nearest neighbor search procedure exploits derived quantizers to offer both a low response time and a high accuracy.
  \item We evaluate derived quantizers in the context of product quantization and optimized product quantization. We show that our approach achieves close response time to 8-bit quantizers, while having the same accuracy as 16-bit quantizers. 
  \item We discuss the differences between our approach and other approaches that provide a high accuracy, \eg additive quantization or tree quantization. We show that because it provides a large increase in accuracy, without significantly impacting response time, our approach compares favorably to the state of the art.
\end{itemize}
\section{Presentation}

\subsection{Overview}

\paragraph{}
The key idea behind our approach is to use product quantizers with 16-bit sub-quantizers, and associate an 8-bit \emph{derived quantizer} with each 16-bit sub-quantizer. \emph{Derived quantizers} are used to compute \emph{compact lookup tables}, which fit the L1 cache. Compact lookup tables are used to compute approximate distances between the query vectors and short codes. A candidate set of the $r2$ closest vectors is built using the computed approximate distances. A precise distance evaluation is then performed for all vectors of the candidate set. This precise distance evaluation relies on large lookup tables, stored in the L3 cache and computed from the codebooks of the 16-bit sub-quantizers. Performing a precise distance evaluation is slow, as it requires accessing the L3 cache. However, as precise distance evaluation is only performed for the vectors of the candidate set, the overall ANN search speed is high.

\paragraph{}
We denote \pq{m}{\overline{b},b} a product quantizer using $m$ sub-quantizers of $b$ bits each, associated with $m$ derived quantizers of $\overline{b}$ bits each. We denote O\pq{m}{\overline{b},b} an optimized product quantizer with the similar properties. In this paper, we focus on \pq{m}{8,16} and O\pq{m}{8,16}, \ie product quantizers and optimized product quantizers with 16-bit sub-quantizers and 8-bit derived quantizers.

\paragraph{}
A naive way of training the 8-bit quantizers would be to trains the codebooks $\overline{\mathcal{C}_j}, j \in \{0,\dots,m-1\}$ of 8-bit quantizers in the same way that the codebooks $\mathcal{C}_j$ of sub-quantizers are trained (Section~\ref{sec:vectorencoding}). We would therefore need to encode high-dimensional vectors twice: once with the 16-bit sub-quantizers, and once with the 8-bit sub-quantizers. We would therefore obtain two codes per vector. This would double the memory use of the database, which would outweigh the benefit of our solution. Instead we \emph{derive} the codebook of the 8-bit quantizers from the codebook of the 16-bit quantizers so that they share the same codes. Sub-quantizers are used to encode vectors and during ANN search, while the derived 8-bit quantizers are used exclusively for ANN search.

\subsection{Training Derived Quantizers}

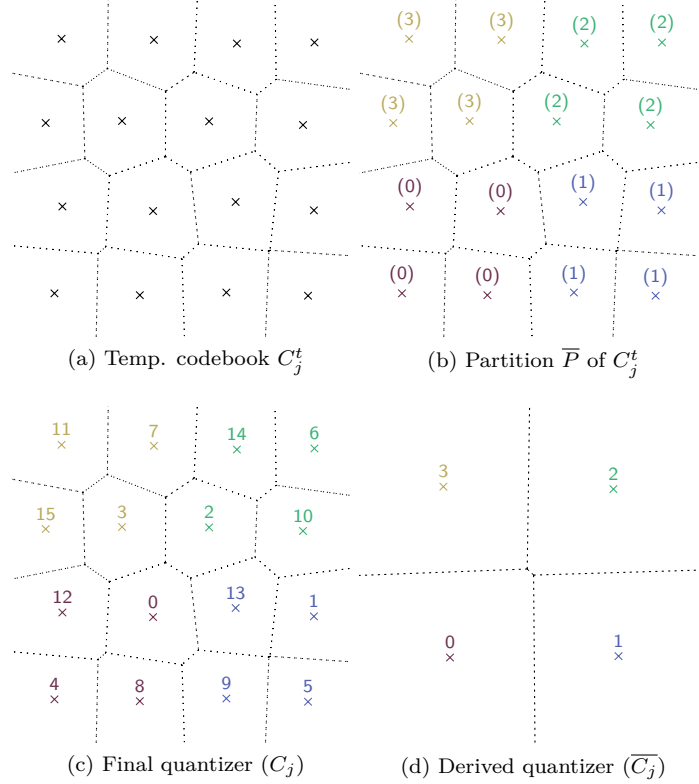
\begin{figure}[t]%
  \centering
  \subfloat[][Temp. codebook $C_j^t$]{%
    \input{derivedquant/fig/quantizer0}
    \label{fig:quant0}
  }
  \subfloat[][Partition $\overline{P}$ of $C_j^t$]{%
    \input{derivedquant/fig/quantizer1}
    \label{fig:quant1}
  }\hfill

  \subfloat[][Final quantizer ($C_j$)]{%
    \input{derivedquant/fig/quantizer2}
    \label{fig:quant2}
  }
  \subfloat[][Derived quantizer ($\overline{C_j}$)]{%
    \input{derivedquant/fig/quantizer3}
    \label{fig:quant3}
  }\hfill%
  \caption{Derived quantizer training process}
  \label{fig:vecgroup}%
\end{figure}

A product quantizer uses $m$ sub-quantizers, each having a different codebook $\mathcal{C}_j, j \in \{0,\dots,m-1\}$. We derive the codebook of each derived quantizer $\overline{\mathcal{C}_j}, j \in \{0,\dots,m-1\}$ from the codebook of the corresponding sub-quantizer. Thus, $\mathcal{C}_0$ is derived from  $\overline{\mathcal{C}_0}$ \etc

\paragraph{}
The training process of the codebook of the $j$th sub-quantizer, with $k=2^{b}$ centroids, and of the codebook of the $j$th derived quantizer, with $\overline{k}=2^{\overline{b}}$ centroids is described in Algorithm~\ref{alg:jointbuild}. The \textsc{kmeans} function is a standard implementation of the k-means algorithm. It takes a training set $V_t$ and a parameter $k$, the number of clusters desired. It returns a codebook $\mathcal{C}$ and a partition $P$ of the training set into $k$ clusters. We denote $P[i]$ the $i$th cluster of $P$. The \textsc{kmeans\_same\_size} function is a k-means variant which produces clusters $G$ of identical sizes, \ie $\forall G_l \in P, |G_l| = |G_0|$ \cite{SSZKmeans}. For the sake of simplicity, Figure~\ref{fig:vecgroup} illustrates the training process for $k=16$ and $\overline{k} = 4$, although we use $k=2^{16}$ and $\overline{k}=2^8$ in practice. The training process takes three steps, described in the three following paragraphs.

\begin{algorithm}[t]
  \caption{Training derived quantizers\label{alg:jointbuild}}
  \begin{algorithmic}[1]
    \Function{build\_quantizers}{$V_t,  \overline{k} , k$}
    \State $\mathcal{C}_j^t, P\gets$ \Call{kmeans}{$V_t, k$}
    \Comment{Step 1}
    \State $\overline{\mathcal{C}_j}, \overline{P} \gets$ \Call{kmeans\_same\_size}{$\mathcal{C}_j^t, \overline{k}$}
    \Comment{Step 2}
    \State $\mathcal{C}_j \gets$ \Call{build\_final\_codebook}{$\overline{P}$, $\overline{b}$}
    \Comment{Step 3}
    \State \Return{$\mathcal{C}_j, \mathcal{C}_j'$}
    \EndFunction
    \Function{build\_final\_codebook}{$\overline{P}$, $\overline{k}$}
    \State $\overline{b} = \log_2(k)$
    \For{$l \gets 0$ to $\overline{k}$}
    \State $\overline{G} \gets \overline{P[i]}$
    \For{$\overline{i} \gets 0$ to $|\overline{G}|-1$}
    \State $\mathcal{C}_j[\overline{i} \ll \overline{b} \mathrel{|} l] = \overline{G}[\overline{i}]$ \Comment{\scriptsize{$|$ is binary or}}
    \Statex \Comment{\scriptsize{$\ll$ is bitwise left shift}} \label{alg:line:order}
    \normalsize
    \EndFor
    \EndFor
    \State \Return{$\mathcal{C}_j$}
    \EndFunction
  \end{algorithmic}
\end{algorithm}

\paragraph{Step 1.} Train a temporary codebook $\mathcal{C}_j^t$ using the \textsc{kmeans} function. Figure~\ref{fig:quant0} shows the result of this step. Each point represents a centroid of $\mathcal{C}_j^t$. Vectors of the training set $V_t$ are not shown. Implicitly, centroids of $\mathcal{C}_j^t$ have an index associated with them, which is their position in the list $\mathcal{C}_j^t$, \ie the index of $\mathcal{C}_j^t[i]$ is i. As the indexes of centroids are not used in the remainder of the training process, they are not shown. 

\paragraph{Step 2.} Partition $\mathcal{C}_j^t$ into $\overline{k}$ clusters using the \textsc{kmeans\_}\textsc{same\_} \textsc{size} function. To do so, $\mathcal{C}_j^t$ is used as the training set argument for \textsc{kmeans\_same\_size}. Figure~\ref{fig:quant1} shows the partition $\overline{P}=(\overline{G_l}), l \in \{0,\dots,\overline{k}-1\}$ of $\mathcal{C}_j^t$. The number in parentheses above each centroid in the index of the cluster it has been assigned to \ie the number $l$ such that the centroid belongs to $\overline{G_l}$. The codebook $\overline{\mathcal{C}_j}$ returned by \textsc{kmeans\_same\_size} is the codebook of the derived quantizer, shown on Figure~\ref{fig:quant3}. Each centroids $\mathcal{C}_j[l]$ of the derived quantizer is the centroid of the cluster $\overline{G_l}$. We obtain "centroids of centroids" because we used as codebook of centroids as the training set.

\paragraph{Step 3.} Build the final codebook $\mathcal{C}_j$ by reordering the centroids of the temporary codebook $\mathcal{C}_j^t$. This reordering, or re-assignement of centroids indexes, is what allows sub-quantizers and derived quantizers to share the same code. The order of centroids in $\mathcal{C}_j$ must be such that the lowest $\overline{b}$ bits of the index assigned to each centroid of $\mathcal{C}_j$ corresponds to the cluster $\overline{G_l}$ it has been assigned to in step 2. If we denote $\operatorname{low_{\overline{b}}}(i)$, the lower $b$ bits of the index $i$, the order of centroids must obey the property:
\begin{gather*}
  \forall {i \in \{0..k-1\}}, \forall l \in \{0..\overline{k}-1\},\\
  \operatorname{low_{\overline{b}}}(i) = l \Leftrightarrow \mathcal{C}_j[i] \in G_l \tag{P1}
\end{gather*}
The \textsc{build\_final\_codebook} function produces an assignment of centroid indexes which obeys property P1 (Algorithm~\ref{alg:jointbuild}, line~\ref{alg:line:order}). Figure~\ref{fig:quant2} shows the final assignment of centroid indexes. In this example, $k=16$ and $\overline{k}=4$ ($b=4$ and $\overline{b}=2$). The centroids belonging to cluster 1 (\texttt{01} in binary) have been assigned the indexes 9 (\texttt{1001}), 13 (\texttt{1101}), 1 (\texttt{0001}), and 5 (\texttt{0101}). The lowest $\overline{b}=2$ bits of 9,13,1 and 5 are \texttt{01}, which matches the partition number (1, or \texttt{01}). This property similarly holds for all partitions and all centroids.

\paragraph{}
This joint training process allows the derived quantizer $\overline{\mathcal{C}_j}$ to be used as an approximate version of $\mathcal{C}_j$ during the NN search process. To encode a vector $x \in \mathbb{R}^d$ into a short code, the codebooks $\mathcal{C}_j$ are used. The code $c$ resulting from the encoding of vector $x$ is such that for all $j \in \{0,\dots,m-1\}$, $\mathcal{C}_j[c[j]]$ is the closest centroid of $x^j$ in $\mathcal{C}_j$. Our training process ensures that the centroid $\overline{\mathcal{C}_j}[\operatorname{low_{\overline{b}}}(c[j])]$ is close to $x^j$. In other words, the centroid index assigned by the quantizer $\mathcal{C}_j$ remains meaningful in the derived quantizer $\overline{\mathcal{C}_j}$.

\subsection{ANN Search with Derived Quantizers}
\label{sec:budsearch}

\paragraph{}
ANN Search with derived quantizers takes two passes. The first pass builds a candidate set of $r2$ vectors, while the second pass reranks the candidate set to produce a final result set of $r$ vectors. The first pass takes three steps: first, lookup tables are computed (Step 1.1) and quantized (Step 1.2). The full database is then scanned to build the candidate set (Step 1.3). The second pass takes a single step (Step 2.), which consists in performing a precise distance evaluation for all vectors of the candidate set, using the quantizers $\{\mathcal{C}_j\}_{j=0}^{m-1}$. Each step is detailed in the following paragraphs

\paragraph{Step 1.1} Compute a set a of $m$ \emph{short lookup tables}, $\{\overline{D_j}\}_{j=0}^{m}$ from the derived quantizers $\{\overline{\mathcal{C}_j}\}_{j=0}^{m}$. Compact loookup tables are computed using the same \textsc{compute\_tables} function used in the conventional ANN search procedure (Section~\ref{sec:annsearch}, Algorithm~\ref{alg:ann}). The compact lookup table $\overline{D_j}$ consists of the distances between the sub-vectors $y^j$ and all centroids of the codebook $\overline{\mathcal{C}_j}$.

\paragraph{Step 1.2} Quantize floating-point distances in compact lookup tables $\{\overline{D}_j\}_{j=0}^{m-1}$ to 8-bit integers in order to build quantized lookup tables $\{Q_j\}_{j=0}^{m-1}$. This quantization step is necessary because our scan procedure (Step 1.3) uses a data structure  optimized for fast insertion, capped\_buckets, which requires distances to be quantized to 8-bit integers. We follow a quantization procedure similar to the one used in~\cite{Andre2015}. We quantize floating-point distances uniformly into 255 (0-254) bins between a $\mathit{qmin}$ and $\mathit{qmax}$ bound. All distances above $\mathit{qmax}$ are quantized to 255. We use the minimum distance across all lookup tables as the $\mathit{qmin}$ bound. To determine $\mathit{qmax}$, we compute the distance between the query vector and the $r2$ first vectors of the database. The greatest distance is used as the $\mathit{qmax}$ bound. Once $\mathit{qmin}$ and $\mathit{qmax}$ have been set, quantized lookup tables are computed as follows:
\begin{gather*}
\forall j \in \{0,\dots,m-1\}, \forall i \in \{0, \dots, \overline{k}-1\}, \\
Q_j[i] = \left \lfloor \frac{\overline{D_j}[i] - \mathit{qmin}}{\mathit{qmax} - \mathit{qmin}} \right \rfloor \cdot 255
\end{gather*}

\begin{algorithm}[t]
  \caption{ANN Search with derived quantizers}\label{alg:annbud}
  \begin{algorithmic}[1]
    \Function{nns\_derived}{$\{\mathcal{C}^j\}_{j=0}^{m}, \{\overline{\mathcal{C}^j}\}_{j=0}^{m} ,\mathit{db}, y, r, r2$}
    \State $\{\overline{D^j}\}_{j=0}^{m} \gets$ \Call{compute\_tables}{$y, \{\overline{\mathcal{C}^j}\}_{j=0}^{m}$} \Comment{Step 1.1} \label{alg:line:bstep1}
    \State $\{Q_j\}_{j=0}^m \gets$ \Call{quantize}{$\{\overline{D^j}\}_{j=0}^{m}, db, r2$} \Comment{Step 1.2} \label{alg:line:bstep2}
    \State $\mathit{cand} \gets $ \Call{scan}{$db, \{Q_j\}_{j=0}^m, r2$} \Comment{Step 1.3} \label{alg:line:bstep3}
    \State \Return{\Call{rerank}{$\mathit{db}, \mathit{cand},\{\mathcal{C}^j\}_{j=0}^{m},y,r,r2$}}
    \Comment{Step 2} \label{alg:line:bstep4}
    \EndFunction
    
    \Function{scan}{$\mathit{db}, \{Q_j\}_{j=0}^m, r2$} \label{alg:line:bscan}
    \State $\mathit{cand} \gets \operatorname{capped\_buckets}(r2)$ \label{alg:line:bucketinst}
    \For{$i \gets 0$ to $|\mathit{db}|-1$}
    \State $c \gets \mathit{db}[i]$\Comment{$i$th short code}
    \State $d \gets$ \Call{adc\_low\_bits}{$c, \{Q^j\}_{j=0}^{m}$}
    \State $\mathit{cand}.\operatorname{put}(d, i)$ \label{alg:line:bucketput}
    \EndFor
    \State \Return{$\mathit{cand}$}
    \EndFunction
    
    \Function{adc\_low\_bits}{$c, \{Q_j\}_{j=0}^m$}
    \State $d \gets 0$
    \For{$j \gets 0$ to $m-1$}
    \State $d \gets d + Q_j[\operatorname{low_{\overline{b}}}(c[j])]$ \label{alg:line:qaccess}
    \EndFor
    \State \Return $d$
    \EndFunction
    
    \Function{rerank}{$\mathit{db},\mathit{cand},\{\mathcal{C}^j\}_{j=0}^{m},y,r,r2$}
    \State $i_{bucket} \gets 0$
    \State $\mathit{count} \gets 0$
    \State $\mathit{neighbors} \gets \operatorname{binheap}(r)$ \label{alg:line:finalset}
    \State $\{D_j\}_{j=0}^{m-1} \gets \{\{-1\}\}$ \label{alg:line:tblfill}
    \While{$count < r2$}
    \State $\mathit{bucket} \gets \mathit{cand}.\operatorname{get\_bucket}(i_{bucket})$ \label{alg:line:bucketiter}
    \ForAll{$i \in \mathit{bucket}$} \label{alg:line:bucketproc}
    \State $d \gets$ \Call{adc\_rerank}{$db[i], \{D^j\}_{j=0}^{m-1}, \{\mathcal{C}^j\}_{j=0}^{m}$}
    \State $\mathit{neighbors}.\operatorname{add}((i, d))$
    \EndFor
    \State $\mathit{count} \gets \mathit{count} + \mathit{bucket}$.size
    \State $i_{bucket} \gets i_{bucket} + 1$ \label{alg:line:bucketiter2}
    \EndWhile
    \EndFunction
    
    \Function{adc\_rerank}{$c, \{D^j\}_{j=0}^{m-1}, \{\mathcal{C}_j\}_{j=0}^{m-1}$}
    \State $d \gets 0$
    \For{$j \gets 0$ to $m-1$}
    \If{$D^j[c[j]] = -1$} \label{alg:line:notinit}
    \State $D^j[c[j]] \gets \left \lVert y^j - \mathcal{C}^j[c[j]] \right \rVert^2$ \label{alg:line:calctbl}
    \EndIf
    \State $d \gets d + D^j[c[j]]$
    \EndFor
    \Return{$d$}
    \EndFunction
  \end{algorithmic}
\end{algorithm}
 
\paragraph{Step 1.3} Scan the full database to build a candidate set of $r2$ vectors using the quantized lookup tables (Algorithm~\ref{alg:annbud}, line~\ref{alg:line:bscan}). Our scan procedure is similar to the scan procedure of the conventional ANN search algorithm (Algorithm~\ref{alg:ann}), apart from two differences. The first difference is that the \textsc{adc\_low\_bits} function is used to compute distances in place of the \textsc{adc} function. The \textsc{adc\_low\_bits} function masks the lowest $\overline{b}$ bits of centroids indexes $c[j]$ to perform lookups in quantized tables (Algorithm~\ref{alg:annbud}, line~\ref{alg:line:qaccess}), instead of using the full centroids indexes to access the full lookup tables. The second difference is that candidates are stored in a data structure optimized for fast insertion, capped buckets (Algorithm~\ref{alg:annbud}, line~\ref{alg:line:bucketinst}), instead of a binary heap. Capped buckets consists of an array of buckets, one for each possible distance (0-254). Each bucket is a list of vector IDs. The put operation (Algorithm~\ref{alg:annbud}, line~\ref{alg:line:bucketput}) involves retrieving the bucket $d$, and appending the vector ID $i$ to the list.
Because adding a vector to a capped buckets data structure requires much less operations than adding a vector to a binary heap, it is much faster. It however requires distances to be quantized to 8-bit integers, which are used as bucket IDs. This not an issue, as quantizing distances to 8-bit integers has been shown not to impact recall significantly~\cite{Andre2015}. On the contrary, a fast insertion is highly beneficial because the candidate set is relatively large (typical $r2$=10K-200K) in comparison with the final result set (typical $r$=10-100). Therefore, many insertions are performed in the candidate set.
To avoid the capped\_buckets structure to grow indefinitely, we maintain an upper bound on distances (\ie bucket IDs). Vectors having distances higher than the upper bound are discarded instead of being inserted in the capped\_bucket. We used the distance (\ie bucket ID)  of the $r2$-th farthest vector in the capped\_buckets as upper bound. 

\begin{figure}[t]
  \centering
  \input{derivedquant/fig/buckets.tex}
  \caption{Capped buckets data structure}
\end{figure}
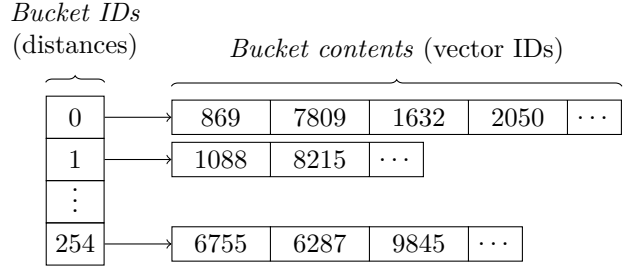

\paragraph{Step 2.} Extract $r2$ vectors from the capped buckets, and perform a precise distance evaluation using the full quantizers $\{\mathcal{C}_j\}_{j=0}^{m-1}$. This precise distance evaluation is used to build the result set of nearest neighbors, named $\mathit{neighbors}$ (Algorithm~\ref{alg:annbud}, line~\ref{alg:line:finalset}). We iterate over capped buckets by increasing bucket ID $i_{bucket}$ (Algorithm~\ref{alg:annbud}, line~\ref{alg:line:bucketiter} and line~\ref{alg:line:bucketiter2}). We process bucket 0, bucket 1, \etc until $r2$ vectors have been processed. When processing a bucket, we iterate over all vectors IDs stored in this bucket (Algorithm~\ref{alg:annbud}, line~\ref{alg:line:bucketproc}), and compute precise distances using the \textsc{adc\_rerank} function. This function is similar to the \textsc{adc} function used in the conventional search process (Algorithm~\ref{alg:ann}). However, here, we do not pre-compute the full lookup tables $\{D_j\}_{j=0}^{m-1}$ but rely on a dynamic programming technique. We fill all tables with the value -1 (Algorithm~\ref{alg:annbud}, line~\ref{alg:line:tblfill}), and compute table elements on demand. Whenever, the value -1 is encountered during a distance computation (Algorithm~\ref{alg:annbud}, line~\ref{alg:line:notinit}), it means that this table element has not yet be computed. The appropriate centroid to sub-vector distance is therefore computed and stored in the appropriate lookup table (Algorithm~\ref{alg:annbud}, line~\ref{alg:line:calctbl}). This strategy is beneficial because it avoids computing the full $\{D_j\}_{j=0}^{m-1}$ tables, which is costly (Table~\ref{tbl:tblsize}). In the case of \pq{m}{8,16}, a full lookup table $D_j$ comprises a large number of elements $k=2^{16}=65536$, but only a small number are accessed (usually 5\%-20\%). This is because vectors of the candidate set are relatively close to the query vector, and thus their short codes tend to have similar indexes. 

\section{Evaluation}
\label{sec:derived-eval}
\subsection{Experimental Setup}

\paragraph{}
All ANN search methods evaluated in this section were implemented in C++. We use gcc version 5.3, with the options \texttt{-O3 -ffast-math -m64} \texttt{-march=native}. For linear algebra primitives, we use the ATLAS library version 3.10.2, of which we compiled an optimized version for our system. We trained the full codebooks $\mathcal{C}_j$ of product quantizers and optimized product quantizers using the implementation\footnote{\url{https://github.com/arbabenko/Quantizations}} of the authors of~\cite{Babenko2014AQ, Babenko2015TQ}. We use two datasets\footnote{\url{http://corpus-texmex.irisa.fr/}}: a small dataset of 1 million vectors (SIFT1M), and a large dataset of 100 million vectors (SIFT100M), consisting of the first 100 million vectors of the BIGANN dataset. We experiment with 32-bit codes and 64-bit codes. All experiments were run on workstation equipped with an Intel Xeon E5-1650v3 CPU and 16 GiB of RAM (DDR4 2133Mhz).

\subsection{Small Datasets, 64-bit Codes}

\pgfplotsset{sfit1mr2 plot/.style={
    xtick={2000,5000,10000,20000,50000},
    xticklabels={2K,5K,10K,20K,50K},
}}

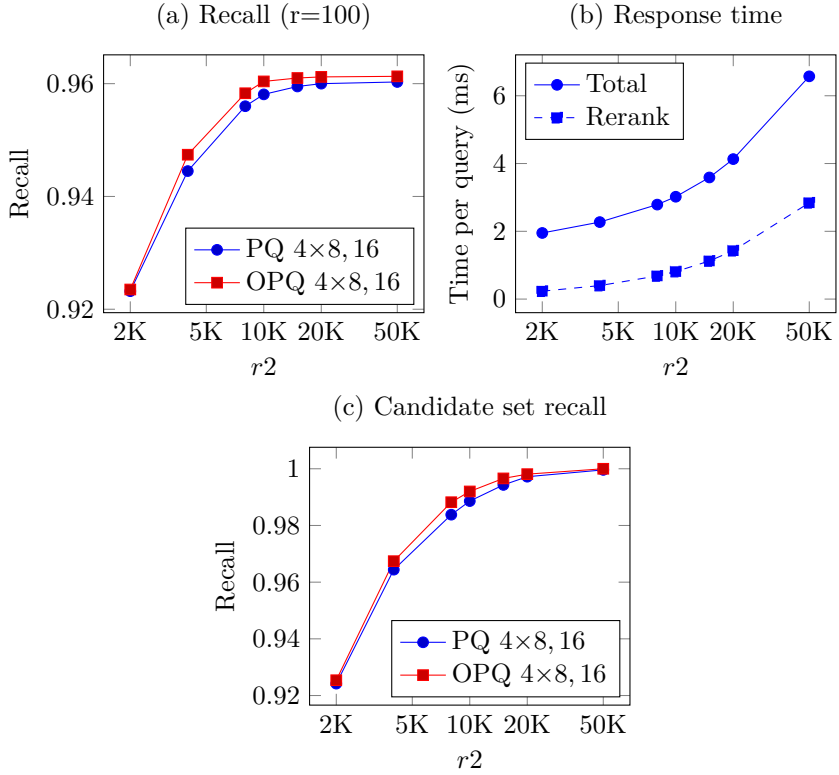
\begin{figure}
  \centering
  \begin{tikzpicture}[trim axis left]
  \begin{semilogxaxis} [
    sfit1mr2 plot,
    height=5 cm,
    ylabel=Recall,
    xlabel=$r2$,
    name=recall graph,
    legend pos=south east,
    legend cell align=left,
    title={(a) Recall (r=100)},
  ]
  \addplot+
  table[x=r2, y=recall] {derivedquant/plots/sift1m_r2_graphs.dat};
  \addplot+
  table[x=r2, y=opq_recall] {derivedquant/plots/sift1m_r2_graphs.dat};
  \legend{PQ $4{\stimes}8,16$\\ OPQ $4{\stimes}8,16$\\}
  \end{semilogxaxis}
  \end{tikzpicture}
  \begin{tikzpicture}[trim axis right]
  \begin{semilogxaxis} [
    sfit1mr2 plot,
    name=time graph,
    height=5 cm,
    anchor=west,
    xlabel=$r2$,
    ylabel=Time per query (ms),
    cycle list={blue},
    legend pos=north west,
    legend cell align=left,
    title=(b) Response time,
  ]
  \addplot+[blue, mark=*]
  table[x=r2, y=query_us] {derivedquant/plots/sift1m_r2_graphs.dat};
  \addplot+[blue, mark=square*, dashed]
  table[x=r2, y=rerank_us] {derivedquant/plots/sift1m_r2_graphs.dat};
  \legend{Total, Rerank}
  \end{semilogxaxis}
  \end{tikzpicture}

  \begin{tikzpicture}[trim axis left, trim axis right]
  \begin{semilogxaxis} [
    sfit1mr2 plot,
    height=5 cm,
    anchor=west,
    xlabel=$r2$,
    ylabel=Recall,
    legend pos=south east,
    legend cell align=left,
    title=(c) Candidate set recall,
  ]
  \addplot+
  table[x=r2, y=recall] {derivedquant/plots/sift1m_r2first_graphs.dat};
  \addplot+
  table[x=r2, y=opq_recall] {derivedquant/plots/sift1m_r2first_graphs.dat};
  \legend{PQ $4{\stimes}8,16$\\ OPQ $4{\stimes}8,16$\\}
  \end{semilogxaxis}
  \end{tikzpicture}
  \caption{Impact of $r2$ on recall and response time (SIFT1M, 64-bit codes)  \label{fig:sift1mr2}}
\end{figure}

\label{sec:r2impact}
\paragraph{}
Our ANN search procedure utilizing derived quantizers takes place in two passes: (1) a candidate set of $r2$ vectors is built by scanning the full database, and (2) a final result set of $r$ nearest neighbors is built by reranking the candidate set using high-accuracy quantizers. Both the parameters $r$ and $r2$ impact response time and recall. By contrast, the conventional ANN search procedure depends on a single parameter: $r$, the number of nearest neighbors in the final result set. In the remainder of this section we denote our approach (16-bit quantizers with derived 8-bit quantizers) PQ$m{\stimes}8,16$ for product quantization, and OPQ$m{\stimes}8,16$ for optimized product quantization.

\paragraph{Impact of $r2$} We first study the impact of $r2$ on recall and accuracy. To do so, we set $r=100$ and vary $r2$ between 2K and 50K. The recall increases with $r2$, until a point where it stabilizes (Figure~\ref{fig:sift1mr2}a). The total response time is also very sensitive to $r2$; it suffers a threefold increase between $r2=$2K and $r2$=50K (Figure~\ref{fig:sift1mr2}b).
This increase in response time comes from two factors: (1) an increased time spent building the candidate set (first pass) and (2) an increased time spent reranking vectors of the candidate set (second pass). Reranking the vectors of the candidate set requires performing lookups in the full lookup tables $D_j$, which is costly as they are stored in slow cache levels (Section~\ref{sec:budsearch}). Moreover, as the elements of the tables $D_j$ are computed on dynamically, a large candidate set means more costly centroid-to-subvector distances will need to be computed. Thus, the reranking time strongly increases with $r2$ (Figure~\ref{fig:sift1mr2}b) and plays a major role in the increase of the response time. The other cause of the increase in response time is the increased time spent building the candidate set (first pass). The number of distance computations performed in the first pass is independent of $r2$ (Algortihm~\ref{alg:annbud}). However, the time spent in the first pass still increases with $r2$ because a large $r2$ means a large candidate set. Thus, more insertions in the candidate set are performed, which increases response time.

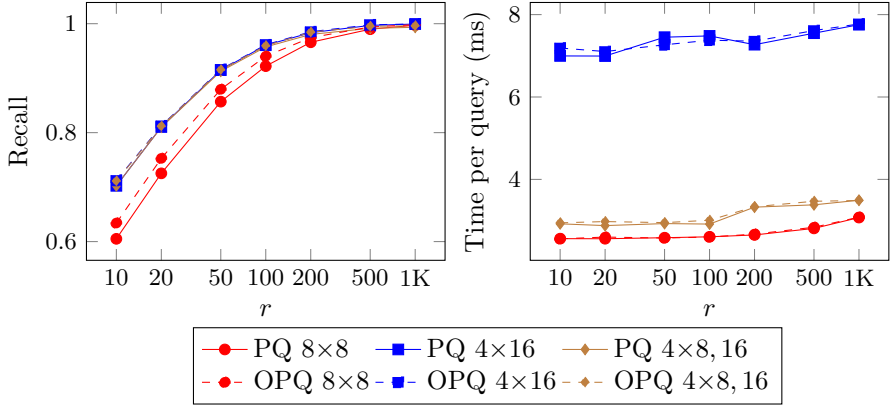
\begin{figure}
  \centering
  \begin{tikzpicture} [baseline, trim axis left]
  \begin{semilogxaxis} [
    r plot,
    cycle list={black},
    ylabel=Recall,
    legend to name=legend-derived-64-1m,
    legend columns=3,
    legend cell align=left,
  ]
  \addplot+ [solid, color=red, mark=*]
  table[x=r, y=pq8_recall] {derivedquant/plots/sift1m_r_recall.dat};
  \addplot+ [solid, color=blue, mark=square*]
  table[x=r, y=pq16_recall] {derivedquant/plots/sift1m_r_recall.dat};
  \addplot+ [solid, color=brown, mark=diamond*]
  table[x=r, y=pq816_recall] {derivedquant/plots/sift1m_r_recall.dat};
  
  \addplot+ [dashed, color=red, mark=*]
  table[x=r, y=opq8_recall] {derivedquant/plots/sift1m_r_recall.dat};
  \addplot+ [dashed, color=blue, mark=square*]
  table[x=r, y=opq16_recall] {derivedquant/plots/sift1m_r_recall.dat};
  \addplot+ [dashed, color=brown, mark=diamond*]
  table[x=r, y=opq816_recall] {derivedquant/plots/sift1m_r_recall.dat};
  \legend{PQ $8{\stimes}8$\\
    PQ $4{\stimes}16$\\
    PQ $4{\stimes}8,16$\\
    OPQ $8{\stimes}8$\\
    OPQ $4{\stimes}16$\\
    OPQ $4{\stimes}8,16$\\}
  \end{semilogxaxis}
  \end{tikzpicture}
  \begin{tikzpicture} [baseline, trim axis right]
  \begin{semilogxaxis} [
    r plot,
    cycle list={black},
    legend pos=north west,
    ylabel=Time per query (ms)
  ]
  \addplot+ [solid, color=red, mark=*]
  table[x=r, y=pq8_query_us] {derivedquant/plots/sift1m_r_time.dat};
  \addplot+ [solid, color=blue, mark=square*]
  table[x=r, y=pq16_query_us] {derivedquant/plots/sift1m_r_time.dat};
  \addplot+ [solid, color=brown, mark=diamond*]
  table[x=r, y=pq816_query_us] {derivedquant/plots/sift1m_r_time.dat};
  
  \addplot+ [dashed, color=red, mark=*]
  table[x=r, y=opq8_query_us] {derivedquant/plots/sift1m_r_time.dat};
  \addplot+ [dashed, color=blue, mark=square*]
  table[x=r, y=opq16_query_us] {derivedquant/plots/sift1m_r_time.dat};
  \addplot+ [dashed, color=brown, mark=diamond*]
  table[x=r, y=opq816_query_us] {derivedquant/plots/sift1m_r_time.dat};
  \end{semilogxaxis}
  \end{tikzpicture}
  \\
  \pgfplotslegendfromname{legend-derived-64-1m}
  \caption{Recall and response time (SIFT1M, 64-bit codes)\label{fig:derived-sift1m}}
\end{figure}

\paragraph{}
Due to its significant impact on both response time and recall, it is essential to fine-tune $r2$ to obtain both a a high recall and low response time. To determine $r2$, we measure the recall of the first pass of our ANN search procedure depending on $r2$ (Figure~\ref{fig:sift1mr2}c). We only execute the first pass of our ANN search procedure \ie we build the candidate set but we do not rerank it. We use this candidate set as result set, and measure the recall. From Figure~\ref{fig:sift1mr2}a and Figure~\ref{fig:sift1mr2}c, it may seem that the first pass of our ANN search procedure (Figure~\ref{fig:sift1mr2}c) achieves a higher recall than the full procedure (Figure~\ref{fig:sift1mr2}). However, this is not the case. In the case of the full procedure (Figure~\ref{fig:sift1mr2}a), the result set comprises $r=100$ vectors, while in the case of the first pass (Figure~\ref{fig:sift1mr2}c), the candidate set is used as result set, and thus comprises $r2=$2K-50K vectors. Our goal is to select $r2$ values as small as possible but which preserve the recall of PQ$4{\stimes}16$. For a given $r$ value, we set $r2$ such that the recall of the the first pass (Figure~\ref{fig:sift1mr2}c) is 5-10\% higher than the recall@r of PQ$4{\stimes}16$ (Figure~\ref{fig:derived-sift1m}). For $r \leq 100$, PQ$4{\stimes}16$ achieves a recall@r $\leq 0.9$. Therefore, we set $r2=9000$ (recall of first step is $0.98$) for $r \leq 100$. For $r > 100$, PQ$4{\stimes}16$ achieves very high recalls. Thus, for these cases we set $r2=120000$ (recall of first step is $0.99$).

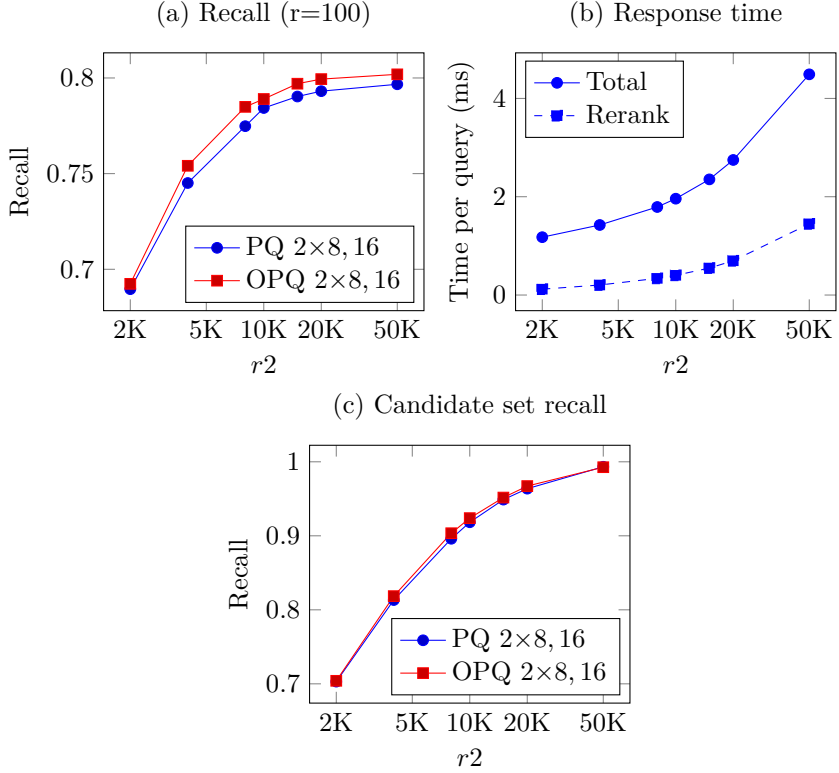
\begin{figure}[t]
  \centering
  \begin{tikzpicture}[trim axis left]
  \begin{semilogxaxis} [
    sfit1mr2 plot,
    height=5 cm,
    ylabel=Recall,
    xlabel=$r2$,
    name=recall graph,
    legend pos=south east,
    legend cell align=left,
    title={(a) Recall (r=100)}
  ]
  \addplot+
  table[x=r2, y=recall] {derivedquant/plots/sift1m32_r2_graphs.dat};
  \addplot+
  table[x=r2, y=opq_recall] {derivedquant/plots/sift1m32_r2_graphs.dat};
  \legend{PQ $2{\stimes}8,16$\\ OPQ $2{\stimes}8,16$\\}
  \end{semilogxaxis}
  \end{tikzpicture}
  \begin{tikzpicture}[trim axis right]
  \begin{semilogxaxis} [
    sfit1mr2 plot,
    name=time graph,
    height=5 cm,
    anchor=west,
    xlabel=$r2$,
    ylabel=Time per query (ms),
    cycle list={blue},
    legend pos=north west,
    legend cell align=left,
    title=(b) Response time
  ]
  \addplot+[blue, mark=*]
  table[x=r2, y=query_us] {derivedquant/plots/sift1m32_r2_graphs.dat};
  \addplot+[blue, mark=square*, dashed]
  table[x=r2, y=rerank_us] {derivedquant/plots/sift1m32_r2_graphs.dat};
  \legend{Total, Rerank}
  \end{semilogxaxis}
  \end{tikzpicture}

  \begin{tikzpicture}[trim axis left, trim axis right]
  \begin{semilogxaxis} [
    sfit1mr2 plot,
    height=5 cm,
    anchor=west,
    xlabel=$r2$,
    ylabel=Recall,
    legend pos=south east,
    legend cell align=left,
    title=(c) Candidate set recall
  ]
  \addplot+
  table[x=r2, y=recall] {derivedquant/plots/sift1m32_r2first_graphs.dat};
  \addplot+
  table[x=r2, y=opq_recall] {derivedquant/plots/sift1m32_r2first_graphs.dat};
  \legend{PQ $2{\stimes}8,16$\\ OPQ $2{\stimes}8,16$\\}
  \end{semilogxaxis}
  \end{tikzpicture}
  \caption{Impact of $r2$ on recall and response time (SIFT1M, 32-bit codes)  \label{fig:sift1m32r2}}
\end{figure}

\paragraph{Comparison with (O)PQ$4{\stimes}16$ and (O)PQ$8{\stimes}8$} PQ$4{\stimes}8,16$ and OPQ$4{\stimes}8,16$ offer the same recall as PQ$4{\stimes}16$ and OPQ$4{\stimes}16$ (Figure~\ref{fig:derived-sift1m}), while maintaining a response time close to PQ$8{\stimes}8$ and OPQ$8{\stimes}8$ (Figure~\ref{fig:derived-sift1m}), thus demonstrating the effectiveness of derived quantizers. Compared to (O)PQ$4{\stimes}8,16$, (O)PQ$4{\stimes}16$ is 2.5 times faster, while offering the same recall. At $r=50$, PQ$4{\stimes}16$ offers a 7\% increase in recall for a 196\% increase in response time, compared to PQ$8{\stimes}8$. By contrast, PQ$4{\stimes}8,16$ also offers a 7\% increase in recall at $r=50$, but the increase in response time is limited to 13\%. The response time of (O)PQ$4{\stimes}8,16$ increases at $r=200$ because we increased $r2$ from 9000 to 12000 at this point. This increase is necessary for (O)PQ$4{\stimes}8,16$ to offer the same recall as (O)PQ$4{\stimes}16$ beyond $r=200$ (Section \ref{sec:r2impact}). Lastly, we observe that OPQ$8{\stimes}8$ offers a higher recall than PQ$4{\stimes}16$, but OPQ$4{\stimes}16$ does not offer a higher recall than PQ$4{\stimes}16$ (Figure~\ref{fig:derived-sift1m}). This because OPQ optimizes the distribution of information between the $m$ sub-spaces of a product quantizers. The greater the number $m$ of sub-spaces, the larger is the gain provided by OPQ. For PQ$8{\stimes}8$ ($m$=8 sub-spaces), OPQ is able to provide a noticeable gain, but for PQ$4{\stimes}16$ ($m$=8 sub-spaces), the gain is too small to be visible.

\subsection{Small Dataset, 32-bit Codes}

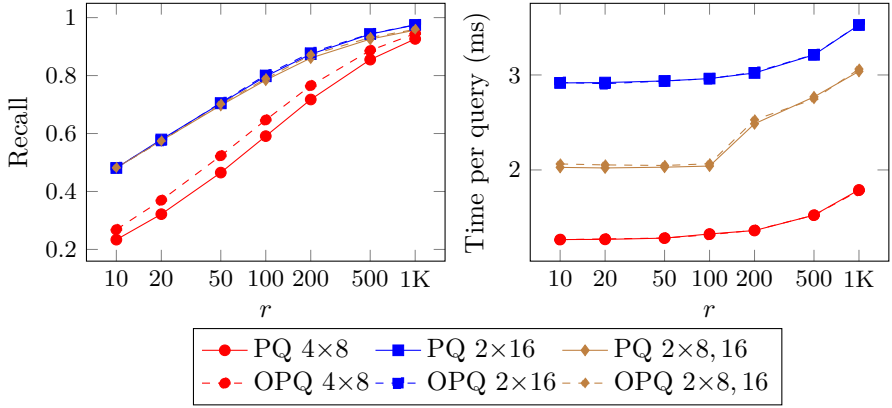
\begin{figure}
  \centering
  \begin{tikzpicture}[baseline, trim axis left]
  \begin{semilogxaxis} [
    r plot,
    cycle list={black},
    legend pos=south east,
    ylabel=Recall,
    legend to name=legend-derived-32-1m,
    legend columns=3,
    legend cell align=left,
  ]
  \addplot+ [solid, color=red, mark=*]
  table[x=r, y=pq8_recall] {derivedquant/plots/sift1m32_r_recall.dat};
  \addplot+ [solid, color=blue, mark=square*]
  table[x=r, y=pq16_recall] {derivedquant/plots/sift1m32_r_recall.dat};
  \addplot+ [solid, color=brown, mark=diamond*]
  table[x=r, y=pq816_recall] {derivedquant/plots/sift1m32_r_recall.dat};
  
  \addplot+ [dashed, color=red, mark=*]
  table[x=r, y=opq8_recall] {derivedquant/plots/sift1m32_r_recall.dat};
  \addplot+ [dashed, color=blue, mark=square*]
  table[x=r, y=opq16_recall] {derivedquant/plots/sift1m32_r_recall.dat};
  \addplot+ [dashed, color=brown, mark=diamond*]
  table[x=r, y=opq816_recall] {derivedquant/plots/sift1m32_r_recall.dat};
  \legend{PQ $4{\stimes}8$\\
    PQ $2{\stimes}16$\\
    PQ $2{\stimes}8,16$\\
    OPQ $4{\stimes}8$\\
    OPQ $2{\stimes}16$\\
    OPQ $2{\stimes}8,16$\\}
  \end{semilogxaxis}
  \end{tikzpicture}
  \centering
  \begin{tikzpicture}[baseline, trim axis right]
  \begin{semilogxaxis} [
  r plot,
  cycle list={black},
  legend pos=north west,
  ylabel=Time per query (ms)
  ]
  \addplot+ [solid, color=red, mark=*]
  table[x=r, y=pq8_query_us] {derivedquant/plots/sift1m32_r_time.dat};
  \addplot+ [solid, color=blue, mark=square*]
  table[x=r, y=pq16_query_us] {derivedquant/plots/sift1m32_r_time.dat};
  \addplot+ [solid, color=brown, mark=diamond*]
  table[x=r, y=pq816_query_us] {derivedquant/plots/sift1m32_r_time.dat};
  
  \addplot+ [dashed, color=red, mark=*]
  table[x=r, y=opq8_query_us] {derivedquant/plots/sift1m32_r_time.dat};
  \addplot+ [dashed, color=blue, mark=square*]
  table[x=r, y=opq16_query_us] {derivedquant/plots/sift1m32_r_time.dat};
  \addplot+ [dashed, color=brown, mark=diamond*]
  table[x=r, y=opq816_query_us] {derivedquant/plots/sift1m32_r_time.dat};
  \end{semilogxaxis}
  \end{tikzpicture}
  \\
  \pgfplotslegendfromname{legend-derived-32-1m}
  \caption{Recall and response time (SIFT1M, 32-bit codes)\label{fig:derived-sift1m32}}
\end{figure}

\paragraph{}
Like with 64-bit codes, both recall and response time increase with $r2$ (Figure~\ref{fig:sift1m32r2}a and Figure~\ref{fig:sift1m32r2}b). However, the second pass (reranking pass) represents a greater part of the total time for 32-bit codes than for 64-bit codes (Figure~\ref{fig:sift1m32r2}b). This is because the first pass (building the candidate set) is faster for 32-bit than for 64-bit codes, while the second pass has the same cost for the two code sizes. The first pass is faster for 32-bit codes (PQ$2{\stimes}8,16$) because each distance computation requires $m=2$ memory accesses instead of the $m=4$ cache accesses required for 64-bit codes (PQ$4{\stimes}8,16$). Even with the conventional search process, PQ$2{\stimes}16$ (Figure~\ref{fig:derived-sift1m32}) is faster than PQ$4{\stimes}16$ (Figure~\ref{fig:derived-sift1m}). This is partly due the lower number of cache accesses, but also to the fact that for PQ$2{\stimes}16$, lookup tables fit the L2 cache, which is faster than the L3 cache used by PQ$4{\stimes}16$ (Table~\ref{tbl:tblsize}, Section~\ref{sec:memory-accesses}). Overall, the higher cost of the reranking step and the relatively low cost of PQ$2{\stimes}16$ make the difference in response time between PQ$2{\stimes}16$ and PQ$4{\stimes}16$ lower. We determine $r2$ values using the rule describe in Section~\ref{sec:r2impact}. We set $r2=10000$ for $r \leq 100$, $r2=15000$ for $r = 200$, $r2=17000$ for $r = 500$ and $r2=20000$ for $r = 1000$.

\paragraph{}
Like for 64-bit codes, (O)PQ$2{\stimes}8,16$ offers the same recall as (O)PQ$2{\stimes}16$ (Figure~\ref{fig:derived-sift1m32}). However, the relative difference in response time is higher for (O)PQ$2{\stimes}8,16$ than for (O)PQ$2{\stimes}8,16$ (Figure~\ref{fig:derived-sift1m32}). At $r=50$, PQ$2{\stimes}16$ offers a 52\% increase in recall for a 123\% increase in response time, compared to PQ$4{\stimes}8$. PQ$2{\stimes}8,16$ also offers a 50\% increase in recall at $r=50$, for a 53\% increase in response time. If our approach leads to a higher increase in response time for 32-bit codes than for 64-bit codes (50\% versus 13\%), it also leads to a higher increase in recall (50\% versurs 7\%). This is because 32-bit codes lead to a higher quantization error than 64-bit codes. Thus, any technique to reduce the quantization error has more impact on 32-bit codes than on 64-bit codes.

\subsection{Large Dataset, 64-bit Codes}

\pgfplotsset{sfit100mr2 plot/.style={
  xtick={20000,50000,100000,500000,1000000},
  xticklabels={20K,50K,100K,500K,1M},
  ticklabel style={font=\small}
}}

\begin{figure}[t]
  \centering
  \begin{tikzpicture}[trim axis left]
  \begin{semilogxaxis} [
    sfit100mr2 plot,
    height=5 cm,
    ylabel=Recall,
    xlabel=$r2$,
    name=recall graph,
    legend pos=south east,
    legend cell align=left,
    title={(a) Recall (r=100)}
  ]
  \addplot+
  table[x=r2, y=recall] {derivedquant/plots/sift100m_r2_graphs.dat};
  \addplot+
  table[x=r2, y=opq_recall] {derivedquant/plots/sift100m_r2_graphs.dat};
  \legend{PQ $4{\stimes}8,16$\\ OPQ $4{\stimes}8,16$\\}
  \end{semilogxaxis}
  \end{tikzpicture}
  \begin{tikzpicture}[trim axis right]
  \begin{semilogxaxis} [
    sfit100mr2 plot,
    name=time graph,
    height=5 cm,
    anchor=west,
    xlabel=$r2$,
    ylabel=Time per query (ms),
    cycle list={blue},
    legend pos=north west,
    legend cell align=left,
    title=(b) Response time
  ]
  \addplot+[blue, mark=*]
  table[x=r2, y=query_us] {derivedquant/plots/sift100m_r2_graphs.dat};
  \addplot+[blue, mark=square*, dashed]
  table[x=r2, y=rerank_us] {derivedquant/plots/sift100m_r2_graphs.dat};
  \legend{Total, Rerank}
  \end{semilogxaxis}
  \end{tikzpicture}

  \begin{tikzpicture}[trim axis left, trim axis right]
  \begin{semilogxaxis} [
    sfit100mr2 plot,
    height=5 cm,
    anchor=west,
    xlabel=$r2$,
    ylabel=Recall,
    legend pos=south east,
    legend cell align=left,
    title=(c) Candidate set recall
  ]
  \addplot+
  table[x=r2, y=recall] {derivedquant/plots/sift100m_r2first_graphs.dat};
  \addplot+
  table[x=r2, y=opq_recall] {derivedquant/plots/sift100m_r2first_graphs.dat};
  \legend{PQ $4{\stimes}8,16$\\ OPQ $4{\stimes}8,16$\\}
  \end{semilogxaxis}
  \end{tikzpicture}
  \caption{Impact of $r2$ on recall and response time (SIFT100M, 64-bit codes) \label{fig:sift100mr2}}
\end{figure}

\begin{figure}
  \centering
  \begin{tikzpicture} [baseline, trim axis left]
  \begin{semilogxaxis} [
    r plot,
    cycle list={black},
    legend pos=south east,
    ylabel=Recall,
    legend to name=legend-derived-64-100m,
    legend columns=3,
    legend cell align=left,
  ]
  \addplot+ [solid, color=red, mark=*]
    table[x=r, y=pq8_recall] {derivedquant/plots/sift100m_r_recall.dat};
  \addplot+ [solid, color=blue, mark=square*]
    table[x=r, y=pq16_recall] {derivedquant/plots/sift100m_r_recall.dat};
  \addplot+ [solid, color=brown, mark=diamond*]
    table[x=r, y=pq816_recall] {derivedquant/plots/sift100m_r_recall.dat};
  
  \addplot+ [dashed, color=red, mark=*]
    table[x=r, y=opq8_recall] {derivedquant/plots/sift100m_r_recall.dat};
  \addplot+ [dashed, color=blue, mark=square*]
    table[x=r, y=opq16_recall] {derivedquant/plots/sift100m_r_recall.dat};
  \addplot+ [dashed, color=brown, mark=diamond*]
    table[x=r, y=opq816_recall] {derivedquant/plots/sift100m_r_recall.dat};
  \legend{PQ $8{\stimes}8$\\
    PQ $4{\stimes}16$\\
    PQ $4{\stimes}8,16$\\
    OPQ $8{\stimes}8$\\
    OPQ $4{\stimes}16$\\
    OPQ $4{\stimes}8,16$\\}
  \end{semilogxaxis}
  \end{tikzpicture}
  \begin{tikzpicture}[baseline, trim axis right]
  \begin{semilogxaxis} [
    r plot,
    cycle list={black},
    legend pos=north west,
    ylabel=Time per query (ms),
  ]
  \addplot+ [solid, color=red, mark=*]
    table[x=r, y=pq8_query_us] {derivedquant/plots/sift100m_r_time.dat};
  \addplot+ [solid, color=blue, mark=square*]
    table[x=r, y=pq16_query_us] {derivedquant/plots/sift100m_r_time.dat};
  \addplot+ [solid, color=brown, mark=diamond*]
    table[x=r, y=pq816_query_us] {derivedquant/plots/sift100m_r_time.dat};
  
  \addplot+ [dashed, color=red, mark=*]
    table[x=r, y=opq8_query_us] {derivedquant/plots/sift100m_r_time.dat};
  \addplot+ [dashed, color=blue, mark=square*]
    table[x=r, y=opq16_query_us] {derivedquant/plots/sift100m_r_time.dat};
  \addplot+ [dashed, color=brown, mark=diamond*]
    table[x=r, y=opq816_query_us] {derivedquant/plots/sift100m_r_time.dat};
  \end{semilogxaxis}
  \end{tikzpicture}
  \\
  \pgfplotslegendfromname{legend-derived-64-100m}
  \caption{Recall and response time (SIFT100M, 64-bit codes)\label{fig:derived-sift100m}}
\end{figure}
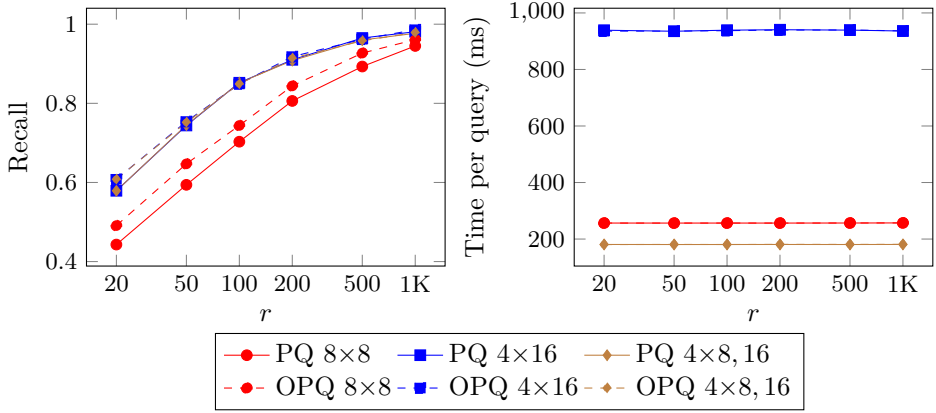

\paragraph{}
To conclude our evaluation section, we evaluate our approach on a large dataset of 100 million vectors. Like for the small dataset, recall and response time increase with $r2$ (Figure~\ref{fig:sift100mr2}a and Figure~\ref{fig:sift100mr2}b). However, contrary to the small dataset, the reranking time (second pass) only represents a small fraction of the response time in this case. Building the candidate set (first pass) requires performing a distance evaluation for all codes stored in the database. As the large is 100 times larger than the small database, the cost of this step in multiplied by 100. By constrast, the size of the candidate set is only multiplied by 10 ($r2=$20K-10M for the large dataset versus $r2=$2K-50K). As a consequence, the second pass has less impact on response time for the large dataset than for the small dataset. In addition, response time increases more slowly with $r2$ for the large dataset than for the small dataset. For the small dataset, the response time is multiplied by 3.4 between $r2=$2K and $r2=$50K, while for the large dataset, the response time is multiplied by 1.8 between $r2=$20K and $r2=$1M. Thus, it is less necessary to fine-tune $r2$. We set $r2$=300K for all $r$ values.

\paragraph{}
Like for the small dataset, (O)PQ$4{\stimes}8,16$ provides the same recall as (O)PQ$4{\stimes}16$ (Figure~\ref{fig:derived-sift100m}). However, the response time of (O)PQ$4{\stimes}8,16$ is lower than the response time of (O)PQ$8{\stimes}8$ instead of being slightly higher (Figure~\ref{fig:derived-sift100m}). At $r=50$, PQ$4{\stimes}16$ offers a 25\% increase in recall for a 270\% increase in response time. PQ$4{\stimes}8,16$ offers a 25\% increase in recall \emph{and} a 42\% decrease in response time. With PQ$4{\stimes}8,16$, the first pass of our ANN search procedure consists in performing distance computations between the query vector and $4{\stimes}8$ codes. Therefore, the first pass of our ANN search procedure is about two times faster than the conventional search procedure with $8{\stimes}8$ codes. For the small dataset, the time gain is outweighed by the large amount of time spent in the second pass. For the large dataset, the second pass only represents a fraction of the total response time (Figure~\ref{fig:sift100mr2}b). The overall response in thus lower than the response time of the conventional ANN search procedure in the case of larger datasets.

\section{Discussion}

\paragraph{}
In this section, we discuss the strengths and weaknesses of our approach and compare it with the state of the art. 

\subsection{Limitations}

\paragraph{} As shown in Section~\ref{sec:derived-eval}, the use of 8-bit derived quantizers combined with 16-bit quantizers does not significantly increase response time while providing a substantial increase in recall, especially on large datasets. However, our approach still requires encoding vectors with 16-bit quantizers, and training 16-bit quantizers. This impacts both the vector encoding time and the codebook training time. For instance, on our workstation, encoding 1 million vectors with a $4{\stimes}16$ product quantizer takes 150 seconds (0.15 ms/vector), while encoding 1 million vectors with $8{\stimes}8$ product quantizer takes 0.91 seconds (0.00091 ms/vector). Similarly, training the codebooks of a $4{\stimes}16$ product quantizer takes 4 minutes (50 k-means iterations, 100K SIFT descriptors), while training the codebooks of a $8{\stimes}8$ product quantizer takes 5 seconds. Nonetheless, it is unlikely that any of this disadvantages would have a practical impact. Codebooks are trained once and for all, therefore training time does not have much impact, as long as it remains tractable. As regards encoding time, even with $4{\stimes}16$ product quantizers, encoding a vectors into its short code takes much less time than an ANN query (0.15ms versus 2-200ms). Moreover, many real-world workloads are read-mostly: ANN queries are performed more often than vectors are added to the database.

\subsection{Comparison with Other Quantization Approaches}
\label{sec:derivedquant-deriv-discuss}

\paragraph{}
In this chapter, we combined derived quantizers with Product Quantization (PQ) and Optimized Product Quantization (OPQ). In Section~\ref{sec:pq-derivatives}, we presented derivatives of product quantization that have been recently introduced: Optimized Product Quantization (OPQ), Additive Quantization (AQ), Tree Quantization (TQ) or Composite Quantization (CQ). Like derived quantizers, these approaches decrease quantization error to offer a higher ANN search accuracy. 

\paragraph{} Our approach compares favorably to these new quantization approaches as it does not significantly increase response time and only moderately increases encoding time. On the contrary, AQ and TQ result in a more than twofold increase in ANN search time~\cite{Babenko2015TQ}. Moreover, AQ and TQ increase vector encoding time by multiple orders of magnitude. Thus encoding a vector into a 64-bit code takes more than 200ms (1333 times more than our approach) for AQ, and 6ms for TQ (40 times more than our approach)~\cite{Babenko2015}. Even if encoding time is not as important as query response time, such an increase makes AQ and TQ hardly tractable for large datasets. Thus, AQ and TQ have not been evaluated on datasets of more than 1 million vectors~\cite{Babenko2014AQ, Babenko2015TQ}. More importantly, our approach is essentially orthogonal to the approach taken by AQ, TQ or CQ. Therefore, derived quantizers may be combined with AQ, TQ or CQ, but only if the codebook learning processes remain tractable.

\paragraph{}
 The codebooks learning process of AQ has a time complexity in $m^2\cdot 2^{2b}$, thus this process is likely to be intractable for 16-bit quantizers ($b=16$). For TQ, the codebook learning process has a complexity in $m{\cdot}2^b$ and is thus likely to be tractable, but this needs to be confirmed by experiments. For its part, CQ requires data structures with a size in $m^2\cdot2^{2b}$. However, these data structures are mainly used to speed up computations, so it is possible to eliminate these data structures at the expense of a moderate increase in computation time. Adapting CQ to 16-bit quantizers is therefore a possible research direction. Combining the ADC process of AQ, TQ or CQ with derived quantizers is likely to be feasible. Using 16-bit quantizers enables halving the $m$ parameter ($m' = m/2$) for a constant code size. This makes it easy to adapt the ADC procedures, even if they have complexities in $m+m^2/2$ or $2m$.

\subsection{Comparison with ADC+R}

\paragraph{}
The idea of building a relatively large candidate set and then reranking it to obtain a final result set was inspired by the ADC+R approach~\cite{Jegou2011R}. The first major difference between our approach and ADC+R is that ADC+R uses 8-bit sub-quantizers, while our approach uses 16-bit sub-quantizers combined with 8-bit derived quantizers. This allows our approach to offer a significantly higher accuracy. ADC+R mainly aims at making ANN search with 8-bit sub-quantizers faster. On the contrary, derived quantizers aim at providing a higher accuracy (16-bit sub-quantizers), while maintaining a constant speed (8-bit derived quantizers). Besides, ADC+R has been superseded by other approaches such as multi-indexes (Section~\ref{sec:invindex}). The second major difference between our approach is that the techniques they use to build the candidate set and rerank it are completely unrelated. The key idea of ADC+R is to use two product quantizers: a first product quantizer $\operatorname{pq_1}$ to encode vectors into short codes, and a second product quantizer $\operatorname{pq_2}$ to encode the error vector $r(x) = x - \operatorname{pq_1(x)}$ resulting from the quantization by $\operatorname{pq_1}$. A vector $x$ is encoded into two
codes. To maintain memory use constant, ADC+R uses $m/2$ sub-quantizers for $\operatorname{pq_1}$ and $\operatorname{pq_2}$ Thus, a vector $x \in \mathbb{R}^d$ is encoded into two $4{\stimes}8$ codes, instead of a single $8{\stimes}8$ code. The candidate set is built using the conventional ANN search  (Section~\ref{sec:annsearch}) algorithm and the codes produced by $\operatorname{pq_1}$. To rerank the candidate set, approximations $\hat{x} = \operatorname{pq_1}(x) + \operatorname{pq_2}(r(x))$ of the input vectors $x$ need to be rebuilt. The distance between the query vector $y$ and approximations $\hat{x}$ are then computed in $\mathbb{R}^d$. These two operations are costly, therefore making reranking slow. Therefore, reranking a candidate set of 20000 vectors takes about 40ms with ADC+R. On the contrary, our approach encodes the vectors with a single product quantizer $\operatorname{pq}$, which is built in such a way that codes can be either scanned with 16-bit sub-quantizers, or 8-bit sub-quantizers. This allows our approach to rerank a candidate set of 20000 vectors in less than 1.5ms.

\stopcontents[chapters]

%% file: derivedquant/fig/quantizer0.tex
\begin{tikzpicture}

\begin{axis}[voronoiaxis]
\addplot [
    only marks, mark=x]
    table {derivedquant/fig/quantizer_centroids.dat};
\addplot [no markers, update limits=false, dotted] table {derivedquant/fig/quantizer_voronoi.dat};
\end{axis}

\end{tikzpicture}

%% file: derivedquant/fig/quantizer1.tex
\begin{tikzpicture}

\begin{axis}[voronoiaxis]
\addplot [
    only marks, mark=x, 
    scatter, scatter src=explicit,
    colormap name=rainbow16,
    visualization depends on={\thisrow{group} \as \group},
    nodes near coords*={(\pgfmathprintnumber[int trunc]{\pgfplotspointmeta})},
    every node near coord/.append style={font=\nodeFont, adjust near node color},
    point meta=\thisrow{group},
] table {derivedquant/fig/quantizer_centroids.dat};
\addplot [no markers, update limits=false, dotted] table {derivedquant/fig/quantizer_voronoi.dat};
\end{axis}

\end{tikzpicture}

%% file: derivedquant/fig/quantizer2.tex
\begin{tikzpicture}
\begin{axis}[voronoiaxis]
\addplot [
    only marks, mark=x, 
    scatter, scatter src=explicit,
    colormap name=rainbow16,
    visualization depends on={\thisrow{label} \as \pointLabel},
    nodes near coords*={$\pgfmathprintnumber[int trunc]{\pointLabel}$},
    every node near coord/.append style={font=\nodeFont, adjust near node color},
    point meta=\thisrow{group},
    ]
    table {derivedquant/fig/quantizer_centroids.dat};
\addplot [no markers, update limits=false, dotted] table {derivedquant/fig/quantizer_voronoi.dat};

\end{axis}
\end{tikzpicture}

%% file: derivedquant/fig/quantizer3.tex
\begin{tikzpicture}
\begin{axis}[voronoiaxis]
\addplot [
    only marks, mark=x, 
    scatter, scatter src=explicit,
    colormap name=rainbow16,
    nodes near coords*={\pgfmathprintnumber[int trunc]{\pgfplotspointmeta}},
    every node near coord/.append style={font=\nodeFont, adjust near node color},
    point meta=\thisrow{label},
    ]
    table {derivedquant/fig/quantizer_cheap_centroids.dat};
\addplot [no markers, update limits=false, dotted] table {derivedquant/fig/quantizer_cheap_voronoi.dat};

\end{axis}
\end{tikzpicture}

%% file: derivedquant/fig/buckets.tex
\newcommand{\idHeight}{2eX}

\newlength\cellWidth
\setlength{\cellWidth}{\widthof{254}}

\tikzset{
  array cell/.style = {
    rectangle,
    draw,
    text width=\cellWidth,
    minimum height=3.7ex,
    align=center,
  },
  array/.style = {
    matrix of math nodes,
    row sep = -\pgflinewidth,
    column sep = -\pgflinewidth,
    nodes={
      array cell
    },
  }
}


\tikzstyle{bucket} = [
  rectangle split,
  rectangle split horizontal,
  draw
]

\newcommand{\slotWidth}{3em}

\tikzstyle{slot} = [
  align=center,
  text width=\slotWidth
]

\newcommand{\bucketSep}{2.5em}

\makeatletter
\DeclareRobustCommand{\rvdots}{%
  \vbox{
    \baselineskip4\p@\lineskiplimit\z@
    \kern-\p@
    \hbox{.}\hbox{.}\hbox{.}
  }}
\makeatother

\begin{tikzpicture}

\node [array] (distances)
{%
  0\\
  1\\
  \rvdots\\
  254\\
};

\node [bucket,
rectangle split parts=5,
right=\bucketSep of distances-1-1.east,
anchor=west] (bucket0) {%
  \nodepart[slot]{one}869
  \nodepart[slot]{two}7809
  \nodepart[slot]{three}1632
  \nodepart[slot]{four}2050
  \nodepart{five}\dots
};

\draw[->] (distances-1-1.east) -- (bucket0.west);

\node [bucket,
  rectangle split parts=3,
  right=\bucketSep of distances-2-1.east,
  anchor=west] (bucket1) {%
  \nodepart[slot]{one}1088
  \nodepart[slot]{two}8215
  \nodepart{three}\dots
};

\draw[->] (distances-2-1.east) -- (bucket1.west);

\node [bucket,
rectangle split parts=4,
right=\bucketSep of distances-4-1.east,
anchor=west] (bucket254) {%
  \nodepart[slot]{one}6755
  \nodepart[slot]{two}6287
  \nodepart[slot]{three}9845
  \nodepart{four}\dots
};

\draw[->] (distances-4-1.east) -- (bucket254.west);

\draw [decoration={brace,raise=5pt},
  decorate] (distances-1-1.north west) --  node[above=10pt,align=center] {\emph{Bucket IDs}\\(distances)} (distances-1-1.north east);
  
\draw [decoration={brace,raise=5pt},
decorate] (bucket0.north west) --  node[above=10pt] {\emph{Bucket contents} (vector IDs)} (bucket0.north east);

\end{tikzpicture}

%% file: conclusion/conclusion.tex
\startcontents[chapters]
\ruledprintcontents

\section{Summary of the Thesis}

\subsection{Context}

\paragraph{}
In this thesis, we have proposed several contributions on the topic of nearest neighbor search in high dimensionality and at large scale. This problem is of particular importance in the current context of mass use of online services. The recent development of mass internet access, online social networks, and smartphone applications has enabled corporations to collect data produced by millions or even billions of users. This trend of large scale data collection, known as \emph{big data} is often characterized by three adjectives: \emph{volume}, \emph{velocity} and \emph{variety}. Distributed computing plaforms like Hadoop and Spark, have been recently introduced to address the high volume of datasets and the high velocity at which new data is produced. This thesis focuses on the \emph{variety} aspect of big data: users not only upload textual data (messages, chats, blog posts) but also multimedia data (image and video files).

\paragraph{}
High-dimensional nearest neighbor search is of particular importance in multimedia databases. Multimedia objects (images, videos, audio files) can be represented as high-dimensional feature vectors that capture their contents. Finding similar multimedia objects then comes down to finding multimedia objects that have similar feature vectors. Many other applications, such as recommender systems or scientific data processing require efficient high-dimensional nearest neighbor search techniques. However, nearest neighbor search remains a challenging problem, especially at large scale and in high-dimensional spaces. Thus, the notorious curse of dimensionality makes \emph{exact} nearest neighbor search intractable. The current research focus is therefore on \emph{approximate} nearest neighbor search solutions.

\paragraph{}
Product quantization is one the most efficient approximate nearest neighbor search approaches, and is at the basis of all contributions of this thesis. The key advantage of product quantization is that it compresses high-dimensional vectors into short codes, which enables large databases to be stored entirely in RAM. Product Quantization can therefore answer nearest neighbor queries without accessing the SSD or HDD, unlike other approaches such as LSH. This approach makes it possible to achieve low response times, because RAM typically has a 100-1000 times lower latency than SSDs. The contributions of this thesis bet on exploiting the capabilities of modern CPUs to further decrease response times offered by product quantization.

\subsection{Contributions}

\paragraph{}
Product quantization compresses high-dimensional vectors into short codes, and stores the entire database in RAM in the form of lists of short codes. To answer nearest neighbor queries, product quantization scans these lists for nearest neighbors. Product quantization therefore computes the distance between the query vector and every code in the lists. This process, named Asymmetric Distance Computation (ADC), remains CPU intensive, and does not exploit the capabilities of modern CPUs. On a single-core, ADC is able to scan lists of short codes at 2-3 GB/s while the theoretical single-core memory bandwidth is 12-16 GB/s. This demonstrates that ADC is strongly CPU bound. In this thesis, we first analyzed the factors that limit the performance of ADC (Performance Analysis). Based on this analysis, we subsequently proposed three contributions (PQ Fast Scan, Quick ADC, Derived Quantizers) that increase the performance of ADC in different scenarios.

\paragraph{Performance Analysis} To compute distances, ADC relies on a set of cache-resident lookup tables. Each distance computation requires a small number of table lookups and additions. In the first part of our analysis, we have shown that even when lookup tables fit the fastest cache levels, cache accesses limit the performance of ADC. Moreover, our study demonstrates that the performance of ADC strongly depends on the cache level in which lookup tables are stored. The product quantization parameter $b$, the number of bits per quantizer (typical $b=8-16$ bits), affects both the accuracy of nearest neighbor search and the cache level in which lookup tables are stored. For instance, we show that using $b=16$ bits per quantizer yields a higher accuracy than using $b=8$ bits per quantizer, but also causes lookup tables to be stored in the L3 cache instead of the L1 cache. This in turn translates into a threefold increase in response time. In the second part of our analysis, we show that ADC cannot be efficiently implemented in SIMD, including using \texttt{gather} instructions introduced in recent Intel Haswell CPUs. Our study concludes that both the large number of cache accesses it performs, and the impossibility of an SIMD implementation limit the performance of ADC. These limitations call for a modification of the ADC procedure to increase performance.

\paragraph{PQ Fast Scan} The first solution we proposed to overcome the limitations of the conventional ADC algorithm is PQ Fast Scan. PQ Fast Scan achieves a 4-6 times speedup over ADC, while returning the exact same results. The key idea behind PQ Fast Scan is to replace costly cache accesses by much faster SIMD in-register shuffles. Using SIMD in-register shuffles requires lookup tables to be stored in SIMD registers, which are much smaller than cache. PQ Fast Scan therefore builds small tables that fit SIMD registers. These small tables are used to compute lower bounds on distances. Lower bounds are then used to prune unneeded distance computations, thus limiting the number of cache accesses. To build small tables, PQ Fast Scan modifies the layout of the database in memory. This makes it incompatible with inverted indexes, a search acceleration technique commonly used in combination with product quantization.

\paragraph{Quick ADC} The second solution we proposed is Quick ADC. Like PQ Fast Scan, Quick ADC achieves a 4-6 times speedup over the conventional ADC algorithm. Quick ADC also builds on the idea of replacing cache accesses by SIMD in-register shuffles. However, contrary to PQ Fast Scan, Quick ADC is compatible with inverted indexes. Instead of building small tables that can be used to compute lower bounds, Quick ADC uses 4-bit quantizers ($b=4$). Using 4-bit quantizers ($b=4$) instead of the commonly used 8-bit quantizers makes it easy to build lookup tables that fit SIMD registers, but causes a slight decrease in recall. While this decrease is generally small, it may still have a practical impact in very large databases. To eliminate this issue, we combine Quick ADC  with optimized product quantization, a derivative of product quantization that yields a better accuracy. We show that when combined with optimized product quantization, the decrease of accuracy caused by Quick ADC is always small, including in the case of very large databases.

\paragraph{Derived Quantizers} For most current use cases of nearest neighbor search, product quantization with 8-bit quantizers, or even 4-bit quantizers, offers a sufficient accuracy. Newer use cases may however require 16-bit quantizers. For instance, there has been a recent interest in using 32-bit codes, instead of 64-bit codes, to achieve higher compression ratios. In this case, the accuracy offered by 8-bit quantizers is too low and 16-bit quantizers become necessary. The main drawback of this approach is that 16-bit incur a threefold increase in response time. To tackle this issue, we have proposed derived quantizers. Derived quantizers make 16-bit quantizers as fast as 8-bit quantizers, while retaining their accuracy. The key idea behind our approach is to compute 8-bit derived quantizers that approximate the 16-bit high-resolution quantizers. To answer nearest neighbor queries, a small candidate set is built by scanning the database using the fast 8-bit derived quantizers. This candidate set is then reranked using the high-resolution 16-bit quantizers.

\section{Perspectives}

\paragraph{}
The contributions presented in this thesis open both short term and more long-term perspectives. Among short-term perspectives is the application of the fast scan techniques we developed to derivatives of product quantization, such as additive quantization or composite quantization. After evaluating the applicability of our techniques to these recent derivatives of product quantization, we discuss two long-term perspectives. We first review how our techniques could be adapted to exploit future hardware. We then examine how they can be generalized to other algorithms that rely heavily on lookup tables or other database algorithms.

\subsection{Application to Product Quantization Derivatives}

\paragraph{}
In this thesis, we proposed three contributions (PQ Fast Scan, Quick ADC, Derived Quantizers) that increase the speed of the ADC procedure in different scenarios. We tested these solutions in the context of product quantization and optimized product quantization. Product Quantization (PQ) and Optimized Product Quantization (OPQ) use the same ADC procedure: $m$ table lookups and $m$ additions. Recently, compositional quantization models inspired by product quantization have been introduced by the research community. Additive Quantization (AQ), Tree Quantization (TQ) and Composite Quantization (CQ) are among the most promising approaches (Section~\ref{sec:pq-derivatives}). All these approaches offer a lower quantization error than product quantization or optimized product quantization. They however sometimes use a different ADC procedure. For instance, AQ requires about $m+m^2/2$ table lookups and additions, and TQ requires $2{\cdot}m$ table lookups and additions. Moreover, all these new approaches use a different process to learn quantizer codebooks from the one of PQ. So far, these approaches have only been tested with 8-bit quantizers ($b=8$). Some of our solutions use 4-bit quantizers (Quick ADC), or 16-bit quantizers (Derived Quantizers). Adapting the codebook learning processes of these new approaches to other quantizer sizes may therefore open challenges. In this section, we recap the research opportunities opened by the application of our solutions to these new compositional quantization models.

\paragraph{PQ Fast Scan} PQ Fast Scan relies on the use of 8-bit sub-quantizers, and all these new quantization approaches have also been tested with 8-bit sub-quantizers. Therefore, there is no need to adapt the codebook learning process of AQ, TQ or CQ. In addition, CQ uses a similar ADC procedure to the one of PQ and OPQ. Therefore, adapting PQ Fast Scan to CQ is straightforward. We have shown that the case of TQ or AQ is more challenging (Section~\ref{sec:fastpq-deriv-discuss}). The ADC procedure of TQ and AQ requires more table lookups ($m+m^2/2$ or $2{\cdot}m$), than the ADC procedure of PQ or OPQ. This means that a different combination of code grouping and minimum tables has to be used. Determining if PQ Fast Scan has a high enough pruning power under these conditions is therefore a possible research direction. If this is the case, PQ Fast Scan would offer a high speedup for AQ and TQ. The SIMD implementation of additions allowed by PQ Fast Scan would highly benefit AQ and TQ, as they require a high number of additions ($m+m^2/2$ or $2{\cdot}m$).

\paragraph{Quick ADC} Quick ADC uses 4-bit sub-quantizers while the recently introduced compositional quantization approaches have been tested with 8-bit sub-quantizers only. This means that it is necessary to adjust the codebook learning process of AQ, TQ or CQ. Using 4-bit sub-quantizers instead of 8-bit sub-quantizers requires doubling the $m$ parameter to maintain a comparable accuracy. This impacts the complexity of the codebook learning process of AQ, TQ and CQ. We have shown that this process should however remain tractable (Section~\ref{sec:quickadc-deriv-discuss}), but experiments are required to determine if it is the case in practice. We have also shown that Quick ADC is unlikely to strongly benefit the ADC procedure of AQ and TQ. For these two quantization approaches, doubling the $m$ parameter causes a large increase in the number of additions ($m+m^2/2$ or $2{\cdot}m$), lessening the benefit of Quick ADC (Section~\ref{sec:quickadc-deriv-discuss}). In conclusion, the most interesting research opportunity consists in combining Quick ADC and CQ, as its ADC procedure requires only $m$ additions. The resulting solution would offer both superior performance, thanks to Quick ADC, and a high accuracy, thanks to CQ.

\paragraph{Derived Quantizers} The key idea of derived quantizers is to use 16-bit quantizers instead of the common 8-bit quantizers to increase accuracy. Therefore, derived quantizers have the same goal as AQ, TQ or CQ. Because they increase accuracy without impacting response time, derived quantizers compare favorably to these approaches. More importantly, the approach of derived quantizers is orthogonal to these other approaches. Derived quantizers may therefore be combined with AQ, TQ or CQ, but only if the codebook learning processes remain tractable for 16-bit sub-quantizers. We have shown that the codebook learning process of TQ and CQ is likely to remain tractable, while the codebook learning process of AQ is likely to be intractable (Section~\ref{sec:derivedquant-deriv-discuss}). Therefore, the most interesting research opportunity is to combine derived quantizers with TQ or CQ. Lastly, using 16-bit quantizers makes it possible to divide the $m$ parameter by 2 ($m'=m/2$), while maintaining a greater accuracy. Thus the ADC procedures can be easily adapted, as they have complexities in $m+m^2/2$ or $2m$.

\subsection{Exploitation of Future Hardware}

\paragraph{Future SIMD instruction sets} The upcoming Intel Xeon Skylake Purley, expected in 2017, will include support for the new AVX-512 SIMD instruction set~\cite{IntelAVX512,IntelArchExt}. The AVX-512 instruction set will provide twice as wide SIMD registers (512 bits) as the current AVX2 SIMD instruction set (256 bits). In addition, AVX-512 will offer new SIMD instructions that do not exist yet in AVX2. For instance, AVX-512 will offer a full-width SIMD in-register shuffle allowing 512-bit small tables ($64{\stimes}8$ bits). In comparison, the current AVX2 instruction set only offers half-width SIMD in-register shuffles, allowing lookups in 256-bit small tables ($16{\stimes}8$ bits). AVX-512 will increase the applicability of PQ Fast Scan, and the accuracy of Quick ADC. PQ Fast Scan requires inverted lists of at least 3 million codes to be efficient, hindering the compatibility with inverted indexes (Section~\ref{sec:code-grouping}). This size constraint stems from the use of code grouping. More specifically, we have shown that the minimum size of inverted lists depends on the size of small tables stored in SIMD registers (Section~\ref{sec:pq-comp-invindex}). For 128-bit small tables ($16{\stimes}8$), the minimum size is 3 million codes, while for 512-bit small tables ($64{\stimes}8$), the minimum size would be 3000-12000 codes. Thus, AVX-512 would make PQ Fast Scan compatible with most inverted indexes configurations. Quick ADC causes a slight drop in accuracy because it uses 4-bit quantizers ($b=4$) instead of the more common and more accurate 8-bit quantizers ($b=8$). Relying on 512-bit small tables would allow Quick ADC to use 6-bit quantizers ($b=6$), thus reducing the drop in accuracy. In addition to improving applicability or accuracy, AVX-512 (512-bit SIMD) will allow process two times more data per cycles than AVX2 (256-bit SIMD), or four times more data per cycles than SSE (128-bit SIMD). The benefit of AVX-512 is therefore twofold: increase in accuracy or applicability and improvement in performance.

\paragraph{Other Architectures} So far, we have implemented and evaluated PQ Fast Scan on current Intel CPUs. However, other architectures, such as ARM CPUs are used on a massive scale in consumer devices such as smartphones or tablets~\cite{ARMProcessors}. ARM CPUs might also gain traction in the datacenter market in the future. PQ Fast Scan and Quick ADC require support for SIMD in-register shuffles, and SIMD saturated adds. In addition, they require an SIMD width of at least 128 bit. The ARM NEON instruction set is a 128-bit SIMD instruction set included in ARM Cortex-A CPUs, ARM's range of CPUs for smartphones and tablets \cite{ARMNEON, ARMNEONInst}. ARM NEON supports in-register shuffles (\texttt{VTBL} and \texttt{VTBX} instructions) as well as saturated adds (\texttt{VQADD} instruction). Therefore, implementing PQ Fast Scan and Quick ADC with ARM NEON instructions is a possible research direction. Moreover, experiments would allows assessing the speedup obtained on ARM processors. In August 2016, ARM has announced its intent to provide very wide SIMD engines (up to 2048 bits) in its future generation of CPUs via the Scalable Vector Extensions (SVE) instruction set \cite{ARMSVE}. By allowing larger small tables, SVE would bring similar benefits as AVX-512 to ARM CPUs.

\paragraph{} In 2013, IBM announced its intent to offer the POWER technology for licensing, and to allow third-parties to include POWER CPUs in their products~\cite{PowerFoundation}. So far, POWER CPUs were almost exclusively used in mainframes manufactured by IBM itself. This has triggered a surge of interest for the POWER architecture, and the OpenPOWER Foundation~\cite{PowerFoundationWeb} was created to coordinate the efforts of all parties interested in POWER CPUs. The OpenPOWER Foundation now includes major players such as Google or NVIDIA, and the first POWER servers manufactured by third-parties are appearing. POWER servers might therefore enter the market in the near future. Like ARM, the POWER architecture offers 128-bit SIMD. In particular, SIMD in-register shuffles (\texttt{vperm}) and 8-bit saturated additions (\texttt{vaddubs}) are supported~\cite{PowerArch}. PQ Fast Scan and Quick ADC may therefore be implemented on POWER CPUs.

\subsection{Generalization to Other Algorithms}

\paragraph{} When designing PQ Fast Scan and Quick ADC, we developed different techniques that exploit SIMD to accelerate nearest neighbor search. As a medium to long-term research perspective, these techniques may be used beyond the context of nearest neighbor search. More specifically, these techniques can be useful to algorithms that rely on lookup tables, or database workloads that process large amounts of data. In this section, we discuss how the techniques we developed in the context of PQ Fast Scan and Quick ADC can be generalized to other workloads.

\paragraph{SIMD in-registers shuffles} The main idea behind PQ Fast Scan and Quick ADC is to store lookup tables in SIMD registers, while storing them in the L1 cache is generally considered as best practice for efficiency. Therefore, any algorithm relying on lookup tables is a candidate for applying this idea. Among practical uses of lookup tables is query execution in compressed databases. Compression schemes, either generic~\cite{Roth1993,Graefe1991,Abadi2006,Stonebraker2005} or specific (\eg SAX for time series~\cite{Lin2007}), have been widely adopted in database systems. In the case of dictionary-based compression (or quantization), the database stores short codes. A dictionary (or codebook) holds the actual values corresponding to the short codes. Query execution then relies on lookup tables, derived from the dictionary. In this case, storing lookup tables in SIMD registers allows for better performance. If lookup tables are small enough (16 entries), they may be stored directly in SIMD registers, after quantization of their elements to 8-bit integers. Otherwise, it possible to build small tables for different types of queries.
For top-k queries, it is possible to build small tables enabling computation of lower or upper bounds. Like in PQ Fast Scan, lower bounds can then be used to limit L1-cache accesses. To compute upper bounds instead of lower bounds, maximum tables can be used instead of minimum tables. For approximate aggregate queries (\eg approximate mean), tables of aggregates (\eg tables of means) can be used instead of minimum tables.

\paragraph{Saturated Integer Arithmetic} Another idea behind PQ Fast Scan is to use 8-bit saturated arithmetic. This idea can be applied for queries which do not use lookup tables, such as queries executed on uncompressed data. Top-k queries require exact score evaluation for a small number of items, so 8-bit arithmetic can be used to discard candidates. Similarly, 8-bit arithmetic may provide enough precision for approximate queries. In the context of SIMD processing, 8-bit arithmetic allows processing 4 times more data per instruction than 32-bit floating-point arithmetic and thus provides a significant speedup.

\stopcontents[chapters]

%% file: thesis.bbl
\begin{thebibliography}{10}

\bibitem{Abadi2006}
D.~J. Abadi, S.~R. Madden, and M.~C. Ferreira.
\newblock {Integrating Compression and Execution in Column-Oriented Database
  Systems}.
\newblock In {\em SIGMOD}, 2006.

\bibitem{Andre2015}
F.~Andr\'{e}, A.-M. Kermarrec, and N.~Le~Scouarnec.
\newblock {Cache locality is not enough: High-Performance Nearest Neighbor
  Search with Product Quantization Fast Scan}.
\newblock {\em PVLDB}, 9(4), 2015.

\bibitem{HadoopWebsite}
{Apache Software Foundation}.
\newblock {Apache Hadoop}.
\newblock \url{https://hadoop.apache.org/}, 2016.
\newblock Accessed 30 June 2016.

\bibitem{SparkWebsite}
{Apache Software Foundation}.
\newblock {Apache Spark}.
\newblock \url{https://spark.apache.org/}, 2016.
\newblock Accessed 30 June 2016.

\bibitem{ARMProcessors}
ARM.
\newblock {ARM Processors}.
\newblock \url{https://www.arm.com/products/processors}.
\newblock Accessed 02 October 2016.

\bibitem{ARMNEON}
ARM.
\newblock {NEON}.
\newblock \url{http://www.arm.com/products/processors/technologies/neon.php}.
\newblock Accessed 02 October 2016.

\bibitem{ARMNEONInst}
ARM.
\newblock {NEON Instructions}.
\newblock
  \url{http://infocenter.arm.com/help/index.jsp?topic=/com.arm.doc.dui0489e/CJAJIIGG.html}.
\newblock Accessed 02 October 2016.

\bibitem{Babenko2014AQ}
A.~Babenko and V.~Lempitsky.
\newblock {Additive Quantization for Extreme Vector Compression}.
\newblock In {\em CVPR}, 2014.

\bibitem{Babenko2015}
A.~Babenko and V.~Lempitsky.
\newblock {The Inverted Multi-Index}.
\newblock {\em TPAMI}, 37(6), 2015.

\bibitem{Babenko2015TQ}
A.~Babenko and V.~Lempitsky.
\newblock {Tree quantization for large-scale similarity search and
  classification}.
\newblock In {\em CVPR}, 2015.

\bibitem{Bay2006}
H.~Bay, T.~Tuytelaars, and L.~Van~Gool.
\newblock {SURF: Speeded Up Robust Features}.
\newblock In {\em ECCV}, 2006.

\bibitem{Beis1997}
J.~S. Beis and D.~G. Lowe.
\newblock {Shape indexing using approximate nearest-neighbour search in
  high-dimensional spaces}.
\newblock In {\em CVPR}, 1997.

\bibitem{Bentley1975}
J.~L. Bentley.
\newblock {Multidimensional Binary Search Trees Used for Associative
  Searching}.
\newblock {\em CACM}, 18(9), 1975.

\bibitem{CERNCompute}
{CERN}.
\newblock {Computing}.
\newblock \url{https://home.cern/about/computing}.
\newblock Accessed 27 June 2016.

\bibitem{ARMSVE}
I.~Cutress.
\newblock {ARM Announces ARM v8-A with Scalable Vector Extensions: Aiming for
  HPC and Data Center}.
\newblock {\em Anandtech}, August 2016.

\bibitem{Datar2004}
M.~Datar, N.~Immorlica, P.~Indyk, and V.~S. Mirrokni.
\newblock {Locality-sensitive hashing scheme based on p-stable distributions}.
\newblock {\em SCG}, 2004.

\bibitem{Dean2004}
J.~Dean and S.~Ghemawat.
\newblock {MapReduce: Simplified Data Processing on Large Clusters}.
\newblock In {\em OSDI}, 2004.

\bibitem{FacebookQ1}
Facebook.
\newblock {Facebook Q1 2016 Results}.
\newblock \url{https://investor.fb.com/financials/?section=annualreports},
  2016.
\newblock Accessed 27 June 2016.

\bibitem{FortuneAmazon}
{Fortune}.
\newblock {Amazon's recommendation secret}.
\newblock \url{http://fortune.com/2012/07/30/amazons-recommendation-secret/},
  2016.
\newblock Accessed 27 June 2016.

\bibitem{Friedman1977}
{Friedman, Jerome H. and Bentley, Jon Louis and Finkel, Raphael Ari}.
\newblock {An Algorithm for Finding Best Matches in Logarithmic Expected Time}.
\newblock {\em TOMS}, 3(3), 1977.

\bibitem{Fukunaga1975}
K.~Fukunaga and P.~M. Narendra.
\newblock {A Branch and Bound Algorithm for Computing k-Nearest Neighbors}.
\newblock {\em TC}, C-24(7), 1975.

\bibitem{Gan2012}
J.~Gan, J.~Feng, Q.~Fang, and W.~Ng.
\newblock {Locality-sensitive Hashing Scheme Based on Dynamic Collision
  Counting}.
\newblock In {\em SIGMOD}, 2012.

\bibitem{Ge2013}
T.~Ge, K.~He, Q.~Ke, and J.~Sun.
\newblock {Optimized Product Quantization for Approximate Nearest Neighbor
  Search}.
\newblock In {\em CVPR}, 2013.

\bibitem{Ge2014}
T.~Ge, K.~He, Q.~Ke, and J.~Sun.
\newblock {Optimized Product Quantization}.
\newblock {\em TPAMI}, 36(4), 2014.

\bibitem{Gionis1999}
A.~Gionis, P.~Indyk, and R.~Motwani.
\newblock {Similarity Search in High Dimensions via Hashing}.
\newblock In {\em VLDB}, 1999.

\bibitem{Graefe1991}
G.~Graefe and L.~D. Shapiro.
\newblock {Data Compression and Database Performance}.
\newblock In {\em SAC}, 1991.

\bibitem{Gray1984}
R.~Gray.
\newblock {Vector Quantization}.
\newblock {\em IEEE ASSP Magazine}, 1(2), 1984.

\bibitem{FourthParadigm}
T.~Hey, S.~Tansley, and K.~Tolle, editors.
\newblock {\em {The Fourth Paradigm: Data-Intensive Scientific Discovery}}.
\newblock Microsoft Research, 2009.

\bibitem{Hofmann2014}
J.~Hofmann, J.~Treibig, G.~Hager, and G.~Wellein.
\newblock {Comparing the Performance of Different x86 SIMD Instruction Sets for
  a Medical Imaging Application on Modern Multi- and Manycore Chips}.
\newblock In {\em WPMVP}, 2014.

\bibitem{PowerArch}
IBM.
\newblock {\em {Power ISA Version 2.07}}, May 2013.

\bibitem{Instagram400M}
{Instagram Blog}.
\newblock {Celebrating a Community of 400 Million}.
\newblock \url{http://blog.instagram.com/post/129662501137/150922-400million},
  2015.
\newblock Accessed 27 June 2016.

\bibitem{IntelIntrinsicsGuide}
Intel.
\newblock {Intel Intrinsics Guide}.
\newblock \url{https://software.intel.com/sites/landingpage/IntrinsicsGuide/}.

\bibitem{IntelCompMan}
Intel.
\newblock {\em {User and Reference Guide for the Intel C++ Compiler 15.0}}.

\bibitem{IntelAVX512}
Intel.
\newblock {Intel AVX-512 instructions}.
\newblock
  \url{https://software.intel.com/en-us/blogs/2013/avx-512-instructions}, 2013.
\newblock Accessed 02 October 2016.

\bibitem{IntelOpt}
Intel.
\newblock {\em {Intel 64 and IA-32 Architectures Optimization Reference
  Manual}}, June 2016.

\bibitem{IntelArchVol1}
Intel.
\newblock {\em {Intel 64 and IA-32 Architectures Software Developer's Manual,
  Volume 1: Basic Architecture}}, September 2016.

\bibitem{IntelArchVol2}
Intel.
\newblock {\em {Intel 64 and IA-32 Architectures Software Developer's Manual,
  Volume 2 (2A, 2B, 2C \& 2D): Instruction Set Reference, A-Z}}, September
  2016.

\bibitem{IntelArchExt}
Intel.
\newblock {\em {Intel Architecture Instruction Set Extensions Programming
  Reference}}, September 2016.

\bibitem{Jegou2011}
H.~J\'{e}gou, M.~Douze, and C.~Schmid.
\newblock {Product quantization for nearest neighbor search.}
\newblock {\em TPAMI}, 33(1), 2011.

\bibitem{Lin2007}
J.~Lin, E.~Keogh, L.~Wei, and S.~Lonardi.
\newblock {Experiencing SAX: a novel symbolic representation of time series}.
\newblock {\em Data Mining and Knowledge Discovery}, 15(2), 2007.

\bibitem{Liu2014}
Y.~Liu, J.~Cui, Z.~Huang, H.~Li, and H.~Shen.
\newblock {SK-LSH: An Efficient Index Structure for Approximate Nearest
  Neighbor Search}.
\newblock {\em PVLDB}, 7(9), 2014.

\bibitem{Lloyd1982}
S.~Lloyd.
\newblock {Least squares quantization in PCM}.
\newblock {\em Transactions on Information Theory}, 28(2), 1982.

\bibitem{Lowe1999}
D.~Lowe.
\newblock {Object recognition from local scale-invariant features}.
\newblock In {\em ICCV}, 1999.

\bibitem{Muja2014}
M.~Muja and D.~G. Lowe.
\newblock {Scalable Nearest Neighbour Algorithms for High Dimensional Data}.
\newblock {\em TPAMI}, 36(X), 2014.

\bibitem{Nister2006}
D.~Nister, H.~Stewenius, D.~Nist, and H.~Stew.
\newblock {Scalable Recognition with a Vocabulary Tree}.
\newblock In {\em CVPR}, 2006.

\bibitem{Norouzi2013}
M.~Norouzi and D.~J. Fleet.
\newblock {Cartesian K-Means}.
\newblock In {\em CVPR}, 2013.

\bibitem{Oliva2001}
A.~Oliva and A.~Torralba.
\newblock {Modeling the Shape of the Scene: A Holistic Representation of the
  Spatial Envelope}.
\newblock {\em IJCV}, 42(3), 2001.

\bibitem{PowerFoundationWeb}
{OpenPOWER Foundation}.
\newblock \url{http://openpowerfoundation.org/}.

\bibitem{Roth1993}
M.~A. Roth and S.~J. Van~Horn.
\newblock Database compression.
\newblock {\em {ACM SGIMOD Record}}, 22(3), 1993.

\bibitem{SSZKmeans}
E.~Schubert.
\newblock Same-size k-means variation, 2012.
\newblock \url{http://elki.dbs.ifi.lmu.de/wiki/Tutorial/SameSizeKMeans}.

\bibitem{Silpa2008}
C.~Silpa-Anan and R.~Hartley.
\newblock {Optimised KD-trees for fast image descriptor matching}.
\newblock In {\em CVPR}, 2008.

\bibitem{Sivic2003}
J.~Sivic and A.~Zisserman.
\newblock {Video Google: a text retrieval approach to object matching in
  videos}.
\newblock In {\em ICCV}, 2003.

\bibitem{Stonebraker2005}
M.~Stonebraker, D.~J. Abadi, A.~Batkin, X.~Chen, M.~Cherniack, M.~Ferreira,
  E.~Lau, A.~Lin, S.~Madden, E.~O'Neil, et~al.
\newblock {C-store: a column-oriented DBMS}.
\newblock In {\em VLDB}, 2005.

\bibitem{Tao2009}
Y.~Tao, K.~Yi, C.~Sheng, and P.~Kalnis.
\newblock {Quality and Efficiency in High Dimensional Nearest Neighbor Search}.
\newblock In {\em SIGMOD}, 2009.

\bibitem{Jegou2011R}
R.~Tavenard, H.~Jegou, M.~Douze, and L.~Amsaleg.
\newblock {Searching in one billion vectors: Re-rank with source coding}.
\newblock In {\em ICASSP}, 2011.

\bibitem{PowerFoundation}
J.~Walton.
\newblock {{IBM Offers POWER Technology for Licensing, Forms OpenPOWER
  Consortium}}.
\newblock {\em Anandtech}, August 2013.

\bibitem{Weber1998}
R.~Weber, H.-J. Schek, and S.~Blott.
\newblock {A Quantitative Analysis and Performance Study for Similarity-Search
  Methods in High-Dimensional Spaces}.
\newblock In {\em VLDB}, 1998.

\bibitem{Xia2013}
Y.~Xia, K.~He, F.~Wen, and J.~Sun.
\newblock {Joint Inverted Indexing}.
\newblock In {\em ICCV}, 2013.

\bibitem{Youtube400hours}
{YoutTube Engineering and Developers Blog}.
\newblock {Machine learning for video transcoding}.
\newblock
  \url{https://youtube-eng.blogspot.fr/2016/05/machine-learning-for-video-transcoding.html},
  2016.
\newblock Accessed 27 June 2016.

\bibitem{Zaharia2012}
M.~Zaharia, M.~Chowdhury, T.~Das, A.~Dave, J.~Ma, M.~McCauley, M.~J. Franklin,
  S.~Shenker, and I.~Stoica.
\newblock {Resilient Distributed Datasets: A Fault-tolerant Abstraction for
  In-memory Cluster Computing}.
\newblock In {\em NSDI}, 2012.

\bibitem{Zhang2014}
T.~Zhang, C.~Du, and J.~Wang.
\newblock {Composite Quantization for Approximate Nearest Neighbor Search}.
\newblock In {\em ICML}, 2014.

\end{thebibliography}
